\documentclass[10pt,twocolumn,letterpaper]{article}

\usepackage[pagenumbers]{iccv} 

\pdfoutput=1

\definecolor{iccvblue}{rgb}{0.21,0.49,0.74}
\usepackage[pagebackref,breaklinks,colorlinks,allcolors=iccvblue]{hyperref}

\usepackage{stfloats}
\usepackage[inkscapearea=page]{svg}
\usepackage{adjustbox}
\usepackage{afterpage}
\usepackage{multirow}
\usepackage{makecell} 
\usepackage{float}    
\usepackage{placeins} 
\usepackage{caption} 
\usepackage{tikz}
\usepackage{graphicx}
\usepackage{overpic}
\usepackage[dvipsnames]{xcolor}
\usepackage{amsmath}
\usepackage[symbol, hang, flushmargin]{footmisc}
\usepackage[accsupp]{axessibility} 

\def\paperID{_} 
\def\confName{ICCV}
\def\confYear{2025}

\title{SPIE: Semantic and Structural Post-Training of Image Editing Diffusion Models with AI feedback}

\author{Elior Benarous$^{1,2,*}$\\
\and
Yilun Du$^{1}$\\
\and
Heng Yang$^{1}$\\
}

\begin{document}
\newcommand{\raisedrule}[2][0em]{\leaders\hbox{\rule[#1]{1pt}{#2}}\hfill}
\twocolumn[{%
\renewcommand\twocolumn[1][]{#1}%
\maketitle

\vspace{-3.75em}
\centering{$^1$Harvard University\ \ \ $^2$ETH Zürich}
\vspace{3.75em}

\vspace{-3em}
\begin{center}
    \captionsetup{type=figure, justification=centering}
    \subfloat[Input Image]{%
    \includegraphics[width=0.15\textwidth]{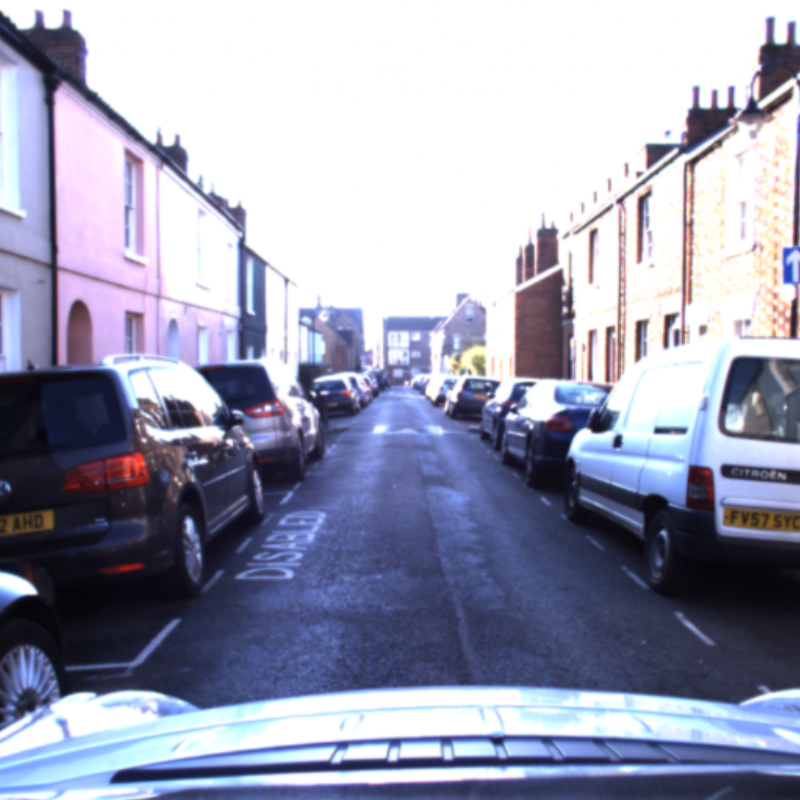}
    }\hspace{1em} 
    \subfloat[\textit{``Add snow \\ on the road"}]{
    \includegraphics[width=0.15\textwidth]{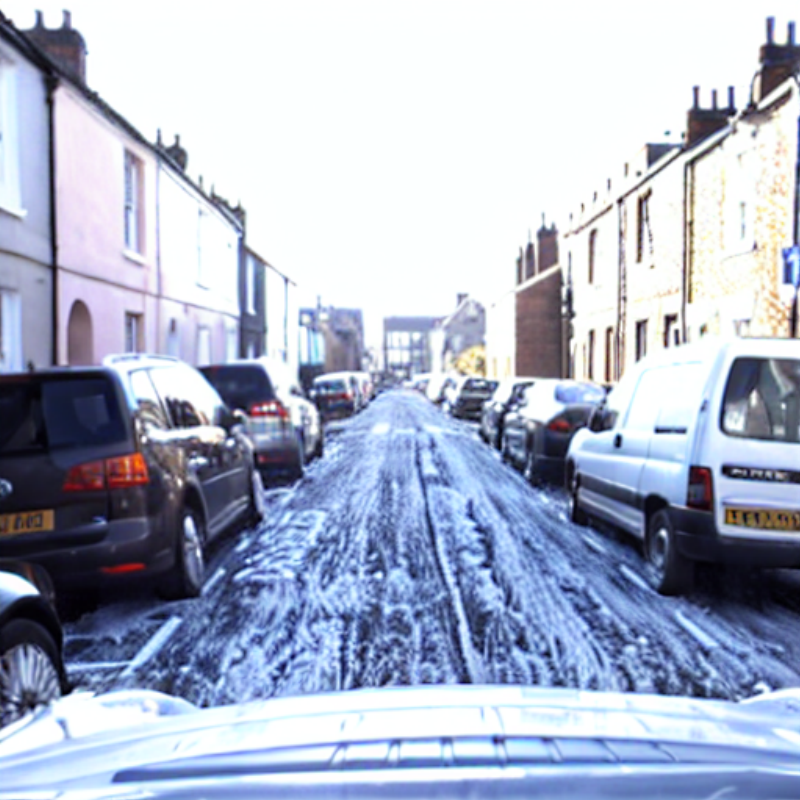}
    }\hspace{1em} 
    \subfloat[\textit{``Change the time \\ to nighttime"}]{%
    \includegraphics[width=0.15\textwidth]{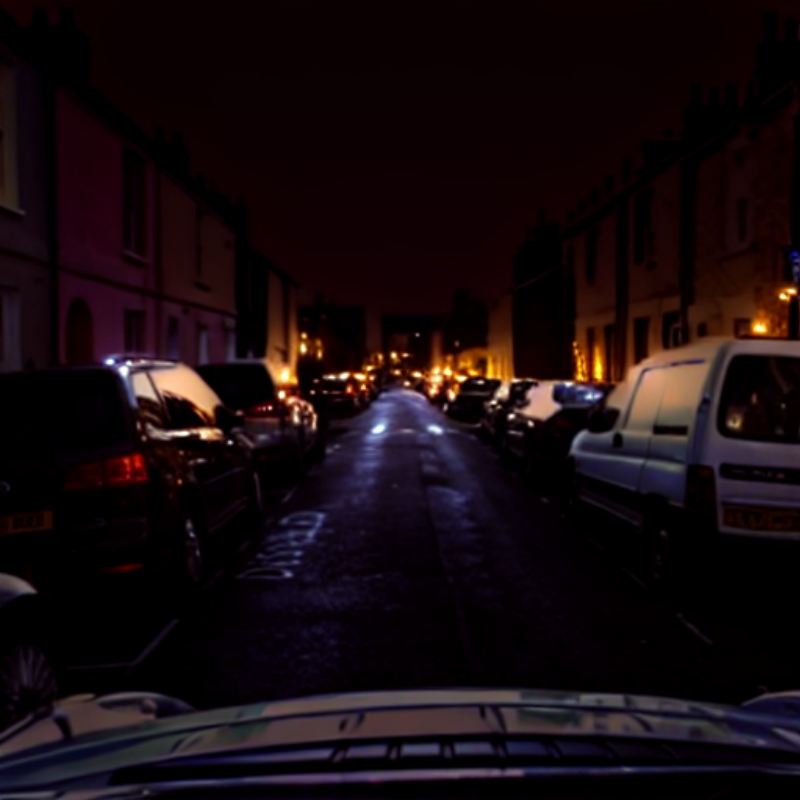}
    }\hspace{1em} 
    \subfloat[\textit{``Turn the road \\ into wood"}]{%
    \includegraphics[width=0.15\textwidth]{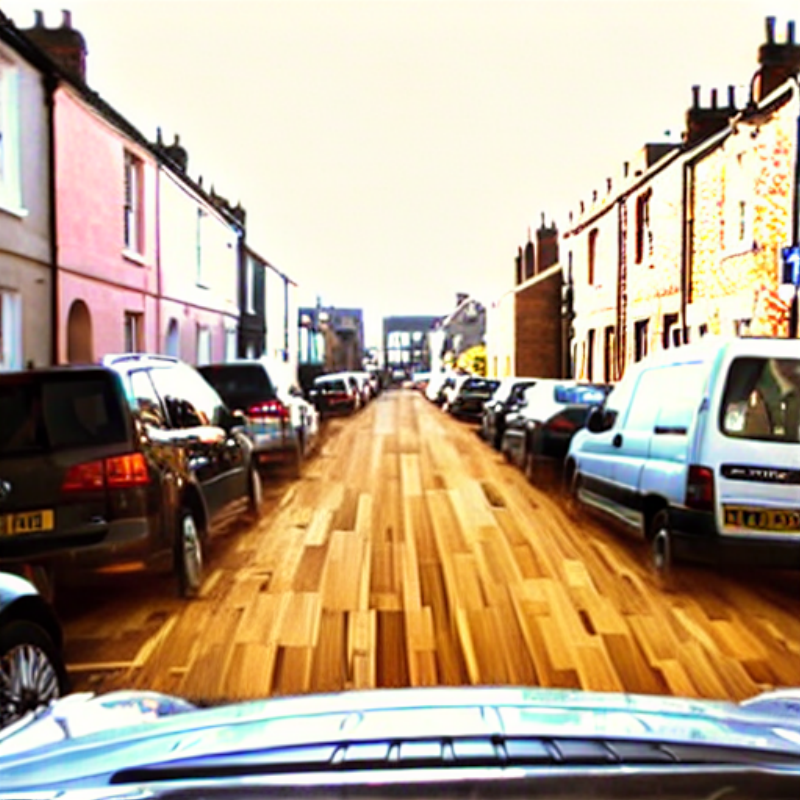}
    }\hspace{1em} 
    \subfloat[\textit{``Add rain \\ on the road"}]{%
    \includegraphics[width=0.15\textwidth]{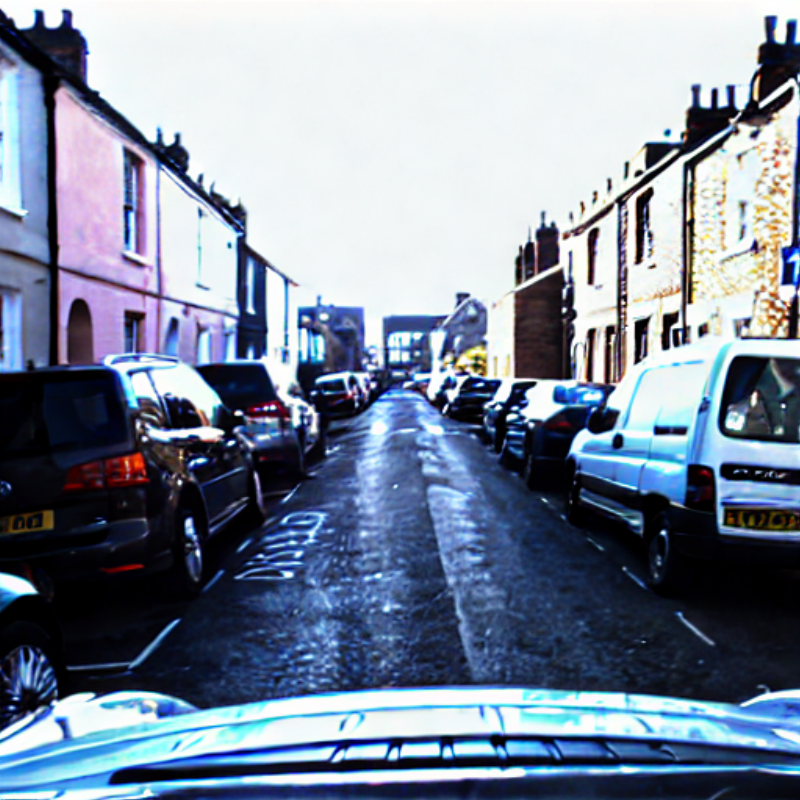}
    } 
    \vspace{-0.5em}
    {\captionsetup{justification=justified}%
    \caption{Representative results showcasing our method's ability to perform precise and realistic edits. The input image is displayed alongside four diverse edits, highlighting SPIE's capacity to align with user intentions while preserving structural coherence.}}
\end{center}
\vspace{0.5em}

}]

\begin{abstract}
This paper presents SPIE: a novel approach for semantic and structural post-training of instruction-based image editing diffusion models, addressing key challenges in alignment with user prompts and consistency with input images.
We introduce an online reinforcement learning framework that aligns the diffusion model with human preferences without relying on extensive human annotations or curating a large dataset.
Our method significantly improves the alignment with instructions and realism in two ways.
First, SPIE captures fine nuances in the desired edit by leveraging a visual prompt, enabling detailed control over visual edits without lengthy textual prompts. 
Second, it achieves precise and structurally coherent modifications in complex scenes while maintaining high fidelity in instruction-irrelevant areas.
This approach simplifies users' efforts to achieve highly specific edits, requiring only 5 reference images depicting a certain concept for training.
Experimental results demonstrate that SPIE can perform intricate edits in complex scenes, after just 10 training steps.
Finally, we showcase the versatility of our method by applying it to robotics, where targeted image edits enhance the visual realism of simulated environments, which improves their utility as proxy for real-world settings.
\end{abstract}    
\section{Introduction}
\label{sec:intro}
Text-to-image (T2I) generative models have achieved remarkable success in creating visually compelling images from text prompts \cite{ho2020denoisingdiffusionprobabilisticmodels, stablediffusion, ramesh2022hierarchicaltextconditionalimagegeneration}, driven by advancements in aligning captions with images \cite{radford2021learningtransferablevisualmodels, alayrac2022flamingovisuallanguagemodel}.
Leveraging these impressive generative capabilities, T2I models have facilitated the development of instructional image editing, offering a highly practical approach for semantic modifications \cite{ip2p, emuedit, mgie, smartedit, ultraedit, hqedit, magicbrush, hive}.
Unlike conventional image editing techniques \cite{prompt2prompt, masactrl, plugandplay, nulltextinversion, imagic, liew2022magicmixsemanticmixingdiffusion} that necessitate detailed descriptive captions for both the input and modified images, instruction-based image editing relies only on natural-language directives to specify changes while leaving unrelated attributes intact.

\footnotetext[1]{Corresponding author: \texttt{ebenarous@ethz.ch}. Code is available at: \url{https://github.com/ebenarous/SPIE}}

While instructional editing has gained popularity for creative applications, its potential remains underexplored in domains requiring precision and consistency.
Simplistic synthetic data has proven effective for pretraining \cite{baradad2022learninglookingnoise, benarous2023harnessingsyntheticdatasetsrole, kataoka2021pretrainingnaturalimages}.
Yet, current data augmentation methods via semantic editing \cite{DA_fusion, zhang2023expandingsmallscaledatasetsguided, zhou2023trainingairimproveimage, DiffAug_recent, bansal2023leavingrealityimaginationrobust, shipard2023diversitydefinitelyneededimproving, dunlap2023diversifyvisiondatasetsautomatic, sariyildiz2023faketillmakeit, shin2023fillupbalancinglongtaileddata, yuan2024justprettypicturesinterventional, azizi2023syntheticdatadiffusionmodels} fail to leverage instructional guidance for generating samples closely aligned with user intentions. 
To address this gap, we identify two key criteria for effective image editing models.

\noindent\textbf{Semantic Alignment.} 
Models should enable fine-grained control over desired modifications by leveraging both textual and visual prompts.
Visual prompts capture stylistic nuances that are difficult to articulate in text alone, alleviating user effort while ensuring edits align closely with expectations.

\noindent\textbf{Structural Alignment.}
Modifications should be confined to specified regions while preserving high fidelity elsewhere.
Current state-of-the-art methods like InstructPix2Pix \cite{ip2p} often struggle with precise edits due to limitations inherited from their training data \cite{prompt2prompt}.
These limitations can result in edits that inadvertently affect background elements or fail to maintain global coherence—an essential factor for achieving realism and preserving the overall structure of the original image.

To meet these criteria, we propose aligning diffusion models with human preferences through online reinforcement learning from AI feedback (RLAIF). 
Unlike traditional RLHF methods, which rely on human annotations, RLAIF enforces semantic and structural alignment from AI-generated preference feedback, effectively emulating human judgments.
This approach also circumvents the limitations of the standard denoising objective, which requires an oracle to generate precise input-output pairs for training. 
Such oracles are often biased, lack scalability, or produce low-quality samples that necessitate extensive pruning \cite{ip2p, emuedit, magicbrush, suti}. 
By moving away from the denoising objective, our method focuses on preserving high-level structural coherence and capturing nuanced semantic features, ensuring edits align closely with user expectations without compromising realism or precision.

In this work, we introduce \textbf{SPIE}: a novel framework for semantic and structural post-training of image editing diffusion models to produce edits highly aligned with visual prompts while preserving original structures in non-pertinent areas. 
Our method builds on InstructPix2Pix \cite{ip2p} and surpasses state-of-the-art baselines in both structural preservation, instruction adherence and predicted human preference. 
Beyond creative applications, we showcase its utility in robotics by enhancing simulated environments with realistic edits that improve alignment with real-world settings.
Our contributions are summarized as follows:
\vspace{0.5em}
\begin{enumerate}
  \item We propose a novel self-play RLAIF-based framework addressing semantic and structural alignment challenges in image editing.
  \vspace{0.5em}
  \item We adapt T2I diffusion models to capture nuanced visual styles from exemplars while adhering to simple textual instructions.
  \vspace{0.5em}
  \item We conduct comprehensive quantitative and qualitative evaluations, demonstrating enhanced precision in intricate edits, stronger alignment with instruction prompts, and practical utility like improving the realism of simulation environments in robotics. 
\end{enumerate}
    
\section{Related Works}
\label{sec:related_works}

\paragraph{Text-guided Image Editing.}
Prior approaches to text-guided image editing can be categorized into three distinct groups: architectural modifications, per-sample optimization, and large scale finetuning.

In the first category, methods like Prompt-to-Prompt (P2P) \cite{prompt2prompt} manipulate attention maps in the diffusion model's U-Net \cite{unet} to control the layout and content of the editing.
Plug-and-Play \cite{plugandplay} injects self-attention maps and spatial features to improve structural coherence.
While these approaches are effective for specific tasks, their reliance on architectural tweaks often limits their ability to handle complex scenes with intricate details.
Other works leverage segmentation masks \cite{ling2021editganhighprecisionsemanticimage, masactrl, xie2022smartbrushtextshapeguided, shi2022semanticstyleganlearningcompositionalgenerative, wang2023imageneditoreditbenchadvancing, Avrahami_2022, glide} or semantic masks \cite{sdedit, controlnet, t2iadapter, ipadapter} during the forward pass to guide edits, ensuring modifications are both targeted and structurally consistent. 
However, these methods impose additional burdens on users by requiring them to provide these masks.
MGIE \cite{mgie} and SmartEdit \cite{smartedit} address this limitation by integrating multimodal large language models (LLMs) to enhance instruction comprehension and reasoning capabilities.
Despite their advancements, these methods introduce significant architectural overhead by incorporating large additional components, greatly increasing computational demands.

In the second category, optimization-based methods like Null-Text Inversion \cite{nulltextinversion} adjust the null-text embedding during the inversion for each input image.
Imagic \cite{imagic} finetunes model weights and embeddings to align with both the input image and the edit prompt.
RB-Modulation \cite{rbmodulation} uses a stochastic optimal controller to align content and style with visual prompts.
While effective, these methods are time-intensive as they require optimization for each individual sample during inference, resulting in slower generation speeds.

The third category includes methods that adopt standard denoising training on large synthetic datasets. 
InstructPix2Pix \cite{ip2p} trains on a dataset generated using P2P and instruction-based prompts. 
Emu Edit \cite{emuedit} expands this dataset with semantic and structural filters and employs multi-task training for improved generalization. 
SuTI \cite{suti}, MagicBrush \cite{magicbrush}, HQ-Edit \cite{hqedit}, and UltraEdit \cite{ultraedit} rely on curated datasets synthesized using models like Imagen \cite{imagen}, DALL-E 2 and 3 \cite{ramesh2022hierarchicaltextconditionalimagegeneration}, or LLMs.
These datasets are often manually pruned or filtered to ensure quality. 
Alchemist \cite{alchemist}, on the other hand, uses a rendering tool specifically designed for material attribute modifications. 
Nevertheless, all of these methods depend on an oracle to generate training data, which introduces biases, requires curation efforts, and may still carry limitations from the oracle itself.

Our method distinguishes itself from these approaches in several ways.
Unlike architectural modification methods, we do not rely on large additional components in the forward pass architecture to improve performance, ensuring simplicity and efficiency.
Unlike per-sample optimization techniques, our method does not require computationally expensive optimization steps during inference.
Unlike large-scale finetuning approaches that depend on curated synthetic datasets generated by external oracles, we leverage the diffusion model’s own samples for training without additional data generation pipelines or curation efforts.
Building upon InstructPix2Pix, our approach addresses misalignments in structural preservation and prompt adherence through targeted finetuning steps while maintaining generalization to unseen input images. By focusing on simplicity and efficiency in both training and inference stages, SPIE achieves state-of-the-art performance without relying on external datasets or complex architectural modifications.

\paragraph{Visual Prompting.}
Most works on visual prompting for image generation have focused on style transfer, where the style of an image is modified across the entire frame \cite{ruiz_styledrop, hertz2024stylealignedimagegeneration, wang2024instantstylefreelunchstylepreserving}. 
Recent studies have also explored subject-driven editing by finetuning pre-trained T2I models using a set of reference images \cite{dreambooth, gal2022imageworthwordpersonalizing}. 
However, these methods often require unique identifiers to encode the concept from the prompt into the edit, limiting their flexibility and usability.
In contrast, our work focuses on local edits conditioned on visual prompts, paired with simple text instructions.
By leveraging visual prompts, we reduce the user's burden to articulate complex edits in detail while maintaining precise control over the desired modifications. 
This combination of visual and textual guidance ensures alignment with user intentions without requiring lengthy or intricate text prompts.
Moreover, prior approaches often rely on the diffusion denoising objective for style alignment, which can lead to reproductions of reference styles that fail to meet human expectations. 
Instead, we enforce alignment in the latent space of an encoder trained to match human judgments, capturing both high-level semantic features and subtle stylistic nuances while preserving structural fidelity. 
Our method enables localized edits with minimal user effort and moves beyond the limitations of traditional denoising objectives, ensuring results that are visually appealing and closely aligned with user preferences.

\paragraph{Reinforcement Learning for Diffusion.} 
Aligning model outputs with human preferences has been widely successful in language modeling. 
For objectives that are difficult to define explicitly, reinforcement learning with human feedback (RLHF) \cite{rlhf_og, rlhf_openaiog, openaiRLHF, anthropicrlhf} has emerged as a popular strategy. 
RLHF involves training a reward function to mimic human preferences and using reinforcement learning algorithms like proximal policy optimization \cite{ppo} to finetune models based on these rewards.

In the context of diffusion models, several works have explored using human feedback for T2I generation. \citet{lee2023aligningtexttoimagemodelsusing} collect human annotations and perform maximum likelihood training where the reward is applied as a naive weight. 
Further, \citet{hps} design a reward model that captures fine-grained human preferences more effectively. 
DDPO \cite{ddpo} and DPOK \cite{dpok} demonstrate that diffusion models can be trained with RL using a reward model emulating human preferences, such as ImageReward \cite{imagereward}. 
For instructional image editing specifically, HIVE \cite{hive} extends large-dataset supervised training by collecting human feedback on edits and performing offline RLHF training.

These methods rely heavily on reward models trained on large-scale human annotations, which introduce significant limitations. 
First, the annotation process is cumbersome and costly, and the resulting supervision often lacks consistency. 
Human evaluators' ability to detect structural preservation inconsistencies diminishes over time due to fatigue and attention variability. 
Second, semantic alignment remains vague as it is only compared to short instruction prompts, leaving room for subjective interpretation and disagreement among annotators.

Our method addresses these challenges by leveraging RLAIF \cite{rlaif_og, rlaifvsrlhf}, eliminating the need for human-in-the-loop supervision. 
Instead of relying on human annotations, we use AI models to provide preference supervision tailored to address semantic and structural alignment issues. 
Additionally, unlike offline RL methods such as HIVE, we adopt an online training framework inspired by D3PO \cite{d3po}, which uses samples generated throughout training to ensure the learning process remains adaptable and unrestricted by the fixed distribution of pre-collected datasets.
With such, SPIE only needs a few steps of post-training to produce edits that are semantically precise, structurally coherent, and aligned with user expectations, without relying on large-scale human annotations.
    
\section{Method}
\label{sec:method}
In this section, we describe the custom objective designed to obtain parallel supervision for the semantic and structural alignment.
In Sec. \ref{method_rlframework}, we describe how to alleviate the need for a reward model.
Then, we explain in Sec. \ref{method_objective} how we design our two separate objectives.
Finally, in Sec. \ref{method_archi}, we present the modified architecture to intake the additional visual prompt conditioning and its modified score estimate formulation for classifier-free guidance with three conditionings.

\subsection{Reinforcement Learning for Diffusion Models}
\label{method_rlframework}
Most RLHF methods train a reward model to then train a downstream model.
However, Direct Preference Optimization (DPO) \cite{dpo} showed that preference ranking can be used to train language models and circumvent reward models, which \citet{diffusion_dpo} extended to diffusion models.
In our work, we leverage the framework introduced by D3PO \cite{d3po}, which expands that of DPO into a multi-step Markov Decision Process (MDP).

Given a pair of outputs $(y_1, y_2) \sim \pi_{\text{ref}}(y \vert x)$ generated from a reference pre-trained model $\pi_{\text{ref}}$, we denote the preference as $y_w \succ y_l \vert x$ and store the ranking tuple $(x, y_w, y_l)$ in dataset $\mathcal{D}$, where $y_w$ and $y_l$ are the preferred and unpreferred samples respectively. Following the Bradley-Terry model \cite{braddleyterry}, the human preference distribution $p^*$ can be expressed by using a reward function $r^*$ as:
\begin{equation}
p^*(y_w \succ y_l \mid x) = \frac{\exp(r^*(x, y_w))}{\exp(r^*(x, y_w)) + \exp(r^*(x, y_l))}
\end{equation}

\noindent A parametrized reward model $r_\phi$ can then be trained via maximum likelihood estimation to approximate $r^*$ with:
\begin{equation}
\label{eq:reward_MLE}
\scalebox{0.85}{$
\mathcal{L}_R(r_{\phi}, \mathcal{D}) = -\mathbb{E}_{(x, y_w, y_l) \sim \mathcal{D}} \left[\log \rho(r_{\phi}(x, y_w) - r_{\phi}(x, y_l))\right]
$}
\end{equation}

\noindent where $\rho$ is the logistic function.
Prior works in RL have for objective to optimize a distribution such that its associated reward is maximized, while regularizing this distribution with the KL divergence to remain similar to its initial reference distribution:
\begin{equation}
\scalebox{0.87}{$
\underset{\pi_\theta}{\max} \;\mathbb{E}_{x \sim \mathcal{D}, y \sim \pi_\theta(y \mid x)} \left[ r_{\phi}(x, y) \right] - \beta \mathbb{D}_{\text{KL}}\left[\pi_\theta(y \vert x) \parallel \pi_{\text{ref}}(y \vert x)\right]
$}
\end{equation}

\noindent where $\beta$ controls the deviation between $\pi_\theta$ and $\pi_{\text{ref}}$. 
This distribution takes the following for optimal solution:
\begin{equation}
\label{eq:opti_rl}
\pi_r(y \mid x) = \frac{1}{Z(x)} \pi_{\text{ref}}(y \mid x) \exp\left(\frac{1}{\beta} r(x, y)\right)
\end{equation}

\noindent where $Z(x) = \sum_y \pi_{\text{ref}}(y \vert x) \exp\left(\frac{1}{\beta} r(x, y)\right)$ is the partition function.
Reorganizing Eq. \ref{eq:opti_rl}, we obtain the expression for the reward as a function of its associated optimal policy.
\begin{equation}
r(x, y) = \beta \log \frac{\pi_r(y \mid x)}{\pi_{\text{ref}}(y \mid x)} + \beta \log Z(x)
\end{equation}

\noindent Substituting the parametrized reward function and policy for their optimal counterparts, we reintegrate that expression into Eq. \ref{eq:reward_MLE}. 
With the change of variables, the loss function is now expressed over policies rather than over reward functions. 
This closed form avoids having to train a reward model, but rather allows us to directly optimize the model.
\begin{equation}
\scalebox{0.79}{$
\mathcal{L}_{\text{DPO}}(\pi_\theta; \pi_{\text{ref}}) = -\mathbb{E}_{(x, y_w, y_l) \sim \mathcal{D}} \left[ \log \rho \left( \beta \log \frac{\pi_\theta(y_w \vert x)}{\pi_{\text{ref}}(y_w \vert x)} - \beta \log \frac{\pi_\theta(y_l \vert x)}{\pi_{\text{ref}}(y_l \vert x)} \right) \right]
$}
\end{equation}

\noindent Extending this to diffusion models, we note a key difference from the derived framework.
The output is not generated from a single forward pass, but rather a sequential process. 
To address this, we pose the $T$-horizon MDP formulation, adapted from \cite{ddpo}, for the $T$-timesteps denoising process.
\begin{alignat*}{2}
    &\boldsymbol{s}_t = (\mathbf{x}_{T-t}, \boldsymbol{c}, t) \quad \hspace{0.3em}P_0(s_0) = \left(\mathcal{N}(\boldsymbol{0}, \boldsymbol{I}), p(\boldsymbol{c}), \delta_0\right)& \\
    &\boldsymbol{a}_t = \mathbf{x}_{T-t-1} \qquad\hspace{0.7em} P(\boldsymbol{s}_{t+1} \mid \boldsymbol{s}_t, \boldsymbol{a}_t) = (\delta_{\mathbf{x}_{T-t-1}}, \delta_c, \delta_{t+1})& \\
    &r(\boldsymbol{s}_t, \boldsymbol{a}_t) = r((\mathbf{x}_{T-t}, \boldsymbol{c}, t), \mathbf{x}_{T-t-1}) \\
    &\pi(\boldsymbol{a}_t \mid s_t) = p_\theta(\mathbf{x}_{T-t-1} \mid \mathbf{x}_{T-t}, \boldsymbol{c}, t)
\end{alignat*}
\noindent where $p_{\theta}(\mathbf{x}_{0:T} \vert \cdot)$ is a T2I diffusion model, $\delta$ is the Dirac delta distribution, and $\boldsymbol{c}$ is the conditioning distributed according to $p(\boldsymbol{c})$.
Note that we disregard $r$ as our method circumvents it.
With such, we treat the denoising process as a sequence of observations and actions: $\sigma = \{s_0, a_0, s_{1}, a_{1}, ..., s_{T-1}, a_{T-1}\}$. 
Since we can only judge the denoised output, we would need to update $\pi_\theta(\sigma) = \prod_{t}^{T} \pi_\theta(s_t, a_t)$, which is intractable.
Following \cite{d3po}, we assume that if the final output of a sequence is preferred over that of another sequence, then any state-action pair of the winning sequence is preferred over that of the losing sequence.
Hence, we determine the preferred sequence by sampling an initial state $s_0=s_0^w=s_0^l$, generating two independent sequences, and ranking their output.
Accordingly, we express the objective at a certain timestep as: 
\begin{equation}
\scalebox{0.79}{$
    \mathcal{L}_t(\pi_\theta) = -\mathbb{E}_{(s_t, \sigma_w, \sigma_l)}\left[\log \rho \left(\beta \log \frac{\pi_\theta(a^w_t \vert s^w_t)}{\pi_{\text{ref}}(a^w_t \vert s^w_t)} - \beta \log \frac{\pi_\theta(a^l_t \vert s^l_t)}{\pi_{\text{ref}}(a^l_t \vert s^l_t)}\right)\right]
    $}
\end{equation}

\subsection{Multi-objective Joint Training}
\label{method_objective}

Having established the mathematical framework for converting preference rankings into a trainable loss, we now detail how we determine the relative ranking between two generated samples, $I_\text{gen}$.
SPIE optimizes a composite objective that enforces both structural alignment with the input image $I_\text{in}$, and semantic alignment with the text instruction and visual style prompts, $c_T$ and $I_\text{sty}$ respectively.
We achieve this by decoupling the overall objective into two separate scores.
\vspace{-1em}

\paragraph{Structural Score.}
To measure how well the structure of the input image is preserved, we employ a monocular depth estimation model \cite{depthanythingv2}.
Given a pair of input and edited images, we compute their respective depth map, and define the structural score as the $L_1$ distance between the two maps.
\vspace{-0.75em}
\begin{equation}
\mathcal{L}_{struct} = \frac{1}{h\cdot w}\sum^{h, w}_{i, j} \left| f_\phi(I_\text{in})_{i,j} - f_\phi(I_\text{gen})_{i,j}\right|
\label{eq:loss_struct}
\vspace{-0.75em}
\end{equation}
\noindent where $f_\phi$ is the depth model and $h \times w$ is the image resolution. 
This metric effectively captures any missing, additional, or deformed elements in the edit relative to the original input.
\vspace{-1em}

\paragraph{Semantic Score.}
The semantic alignment is evaluated within the region of interest, where the edit is expected.
To identify this region, we use a text-conditioned segmentation model, grounded-SAM2 \cite{ren2024grounded, liu2023grounding, ravi2024sam2segmentimages}, which locates the element to be edited based on the instruction. 
Similarly, the visual prompt’s style may not span the entire frame, so its relevant region is also segmented.
For global edits that are intended to cover the whole frame, the mask can be defined to encompass the entire image.
The semantic alignment score is computed by measuring the distance between embeddings of instruction-relevant patches in the generated image and those in the style prompt image. 
Additionally, we incorporate a pixel-space reconstruction objective for instruction-irrelevant regions that should remain unchanged. 
This acts as a regularizer to enforce sharper boundaries and prevent the style from spreading into areas outside the intended edit region.
Accordingly, our semantic score is:
\begin{equation}
\begin{aligned}
    \mathcal{L}_{sem} = & \;D(m_\text{in} \odot I_\text{in}, \; m_\text{sty} \odot I_\text{sty}, \; f_\theta) \\ 
    & + \lambda \cdot (1 - m_\text{in}) \odot \| I_\text{in} - I_\text{gen} \|_2^2
\end{aligned}
\label{eq:loss_semantic}
\end{equation}
where $D(\cdot)$ is a distance metric, here the cosine distance, $m_\text{in}$ and $m_\text{sty}$ are binary segmentation masks for the input and style images respectively, $\odot$ defines the element-wise multiplication, and $f_{\theta}$ is an encoder.
The hyperparameter $\lambda$ balances the influence of the pixel reconstruction term relative to semantic alignment; empirically, we set $\lambda=0.5$ for optimal performance.
Among various encoders tested, DreamSim \cite{dreamsim} most effectively captures task-relevant features (see Sec. \ref{exp_ablation} for ablations).

Similar to \cite{yiluncomposition}, we obtain \textit{advantages} \cite{NIPS1999_464d828b} by normalizing the scores on a per-batch basis using the mean and variance of each training batch.
We then combine the distinct structural and semantic advantages into a unique score, with their relative contribution weighed by a hyperparameter $\alpha$.

\begin{center}
\scalebox{1}{$
\mathcal{L}_{total} = \hat{A}_{struct} + \alpha \cdot \hat{A}_{sem} \hspace{0.4em} \text{, where} \hspace{0.4em}
\hat{A}= \frac{\mathcal{L} - \mu_{\mathcal{L}}}{\sqrt{\sigma_{\mathcal{L}}^2 + \epsilon}}
$}
\end{center}

\noindent Our experiments indicate that $\alpha=1$ is optimal; however $\alpha$, like $\lambda$, can be tuned based on the model's zero-shot performance to accelerate learning convergence. Finally, we rank generated sequences according to their total score.

\subsection{Architecture for Multiple Conditionings}
\label{method_archi}

Our architecture builds upon InstructPix2Pix \cite{ip2p}, which itself adapts Stable Diffusion \cite{stablediffusion} with a key architectural modification: additional input channels in the first convolutional layer of the U-Net to incorporate the encoded input image.
We extend this approach by adding further input channels to intake both the input and style images simultaneously. 
The weights for these newly added channels are initialized to zero to ensure stable training.
We enhance performance by incorporating a cross-attention layer before feeding the visual prompt into the network. This mechanism helps better localize regions in both images that are relevant to the editing directive. Specifically, the query is formed from a linear projection of the concatenated VAE encodings of the input and style images, $\mathcal{E}(I_\text{in})$ and $\mathcal{E}(I_\text{sty})$, while the key and value are derived from projections of the CLIP-encoded instruction prompt. Importantly, the cross-attention output maintains the same spatial dimensions as the VAE-encoded images, preserving compatibility with the pre-trained U-Net architecture.

For sampling, we leverage classifier-free guidance (CFG) \cite{cfg_og}, which shifts probability mass toward regions where an implicit classifier assigns high likelihood to the conditioning, thereby improving sample quality and faithfulness. 
In our case, we compose the CFG estimate with respect to both the input image, textual prompt, and visual prompt \cite{liu2022compositional}, allowing for more precise control over the editing process (see Appendix \ref{appendix_cfg} for the complete derivation).

\section{Experiments}
\label{sec:experiment}

This section presents a comprehensive analysis of our experimental results, including baseline comparisons, ablation studies, and applications in robotics. 
We demonstrate our method's effectiveness in performing precise and realistic edits across diverse scenarios. For consistency, we use the default guidance scale parameters from InstructPix2Pix and set our visual conditioning score to $s_{I_\text{sty}}=3$ (discussed in Sec. \ref{exp_ablation}).

Our evaluations focus on localized edits in complex scenes using images from the Oxford RobotCar \cite{RobotCarDatasetIJRR} and Places \cite{places} datasets, covering various edit types such as weather and material changes (see full list of edits in Sec. \ref{appendix_trainingdetails}). 
We evaluate 29,500 images at $512 \times 512$ resolution, showing that our model outperforms baselines in realism and prompt alignment.

\subsection{Baseline Comparisons}
\label{exp_baseline}

We evaluate SPIE against state-of-the-art baselines, including InstructPix2Pix (IP2P) \cite{ip2p}, HIVE \cite{hive}, MagicBrush (MBrush) \cite{magicbrush}, and HQ-Edit \cite{hqedit}, all built on the stable diffusion v1.5 backbone.

\begin{figure}[t]
\centering
\newcommand{\imgwidth}{0.075}
\setlength{\tabcolsep}{1pt} 
\renewcommand{\arraystretch}{0.25} 
\begin{tabular}{cccccc} 
    Input & IP2P & MBrush & HQ-Edit & HIVE & \textbf{SPIE} \vspace{0.2em} \\

    \includegraphics[width=\imgwidth\textwidth]{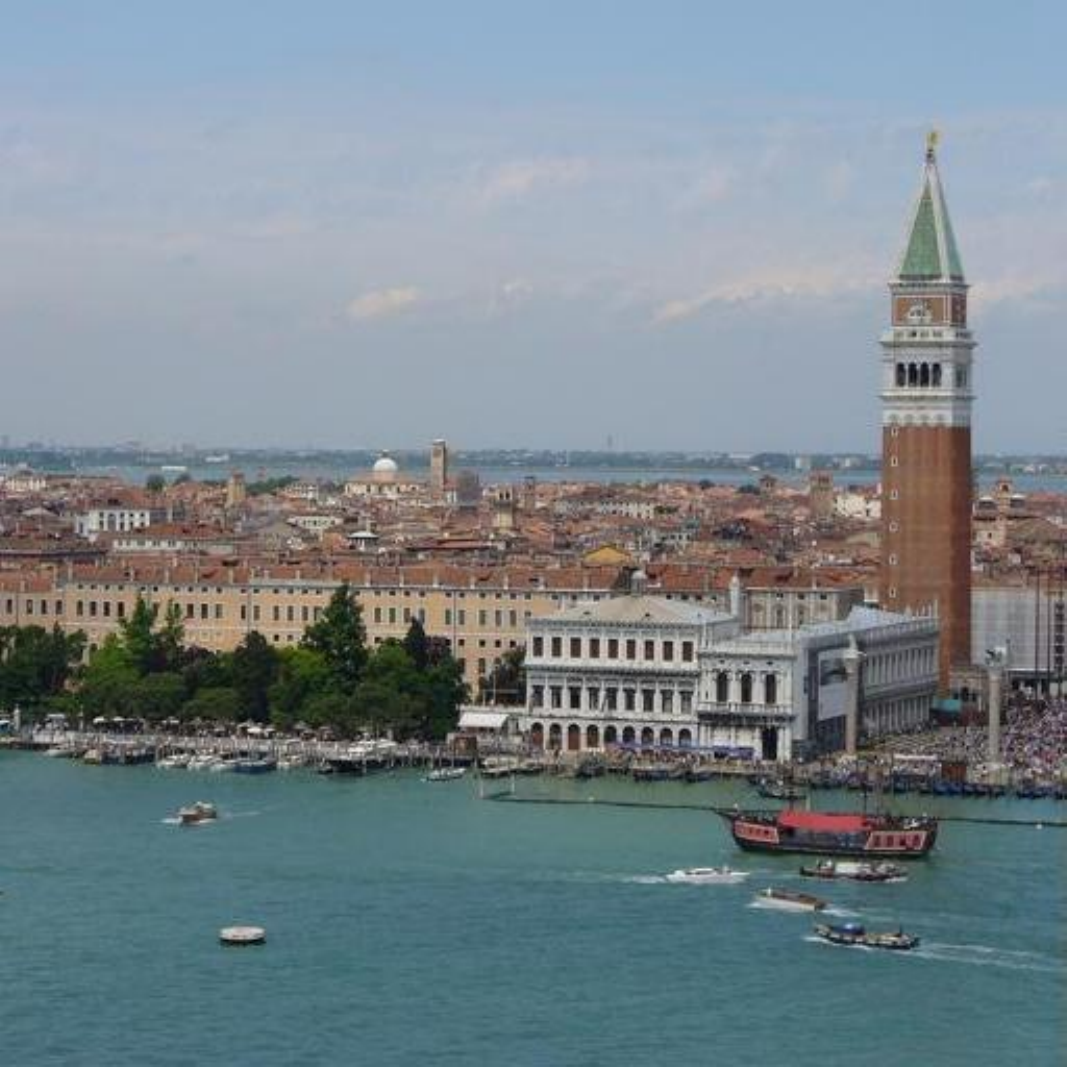}&
    \includegraphics[width=\imgwidth\textwidth]{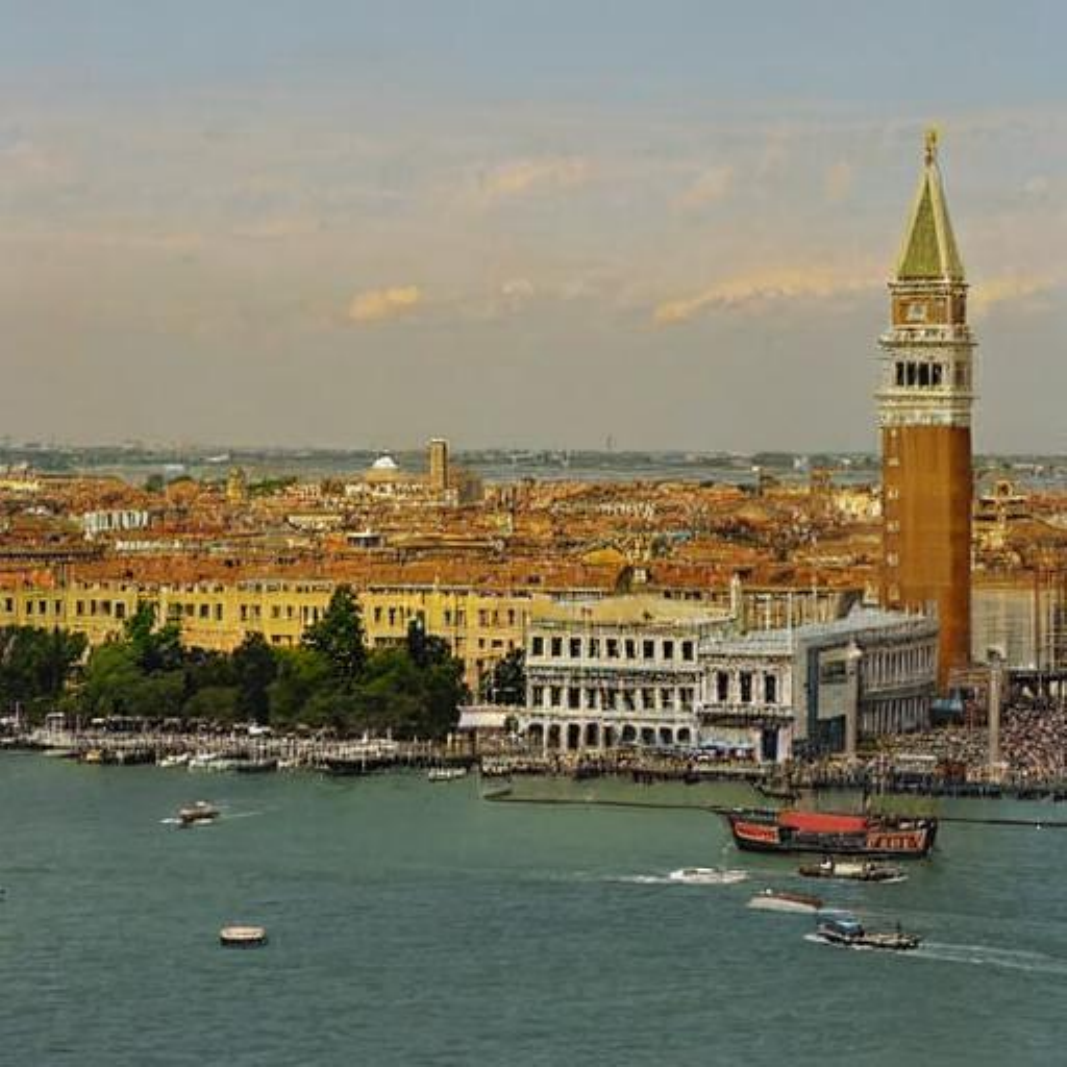} &
    \includegraphics[width=\imgwidth\textwidth]{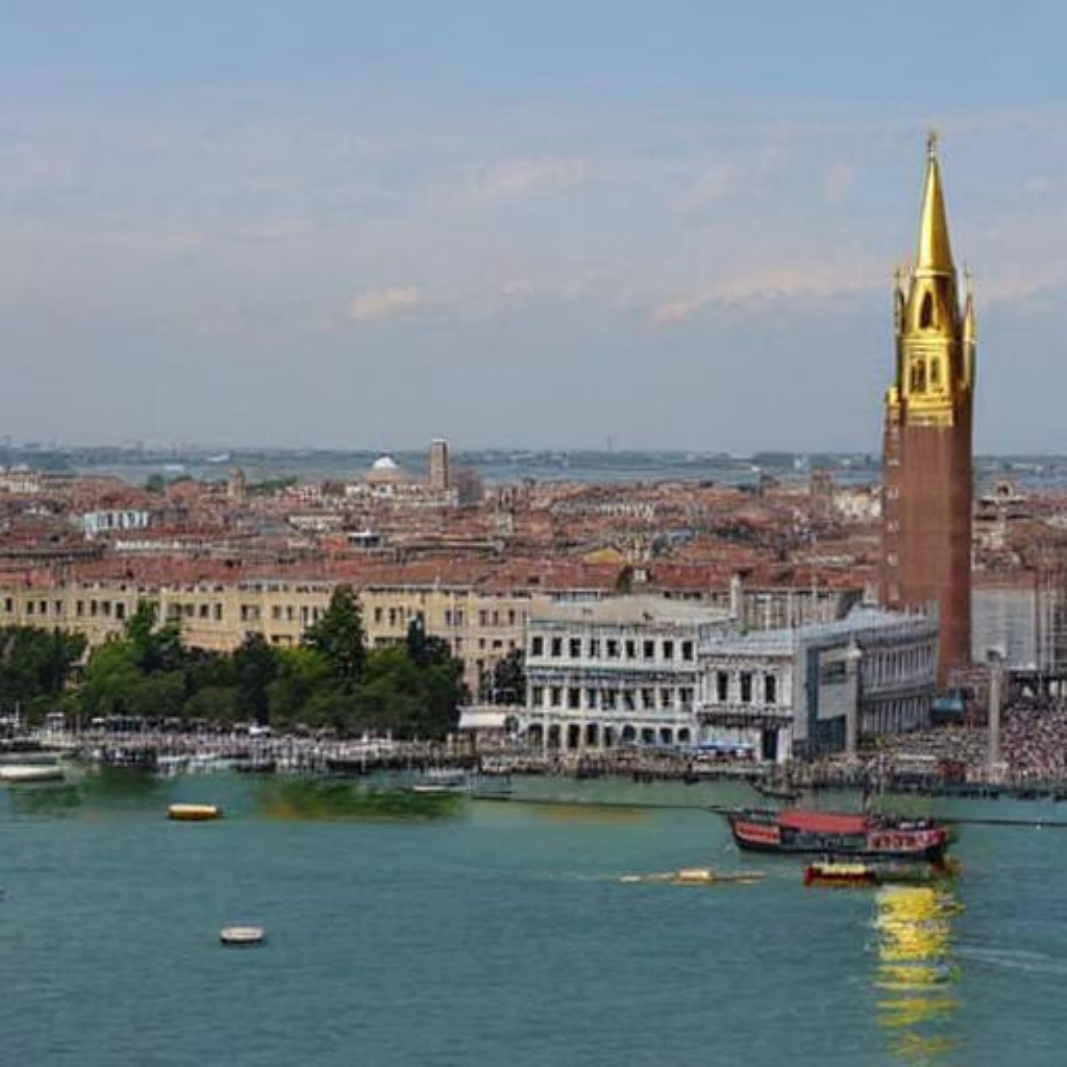} &
    \includegraphics[width=\imgwidth\textwidth]{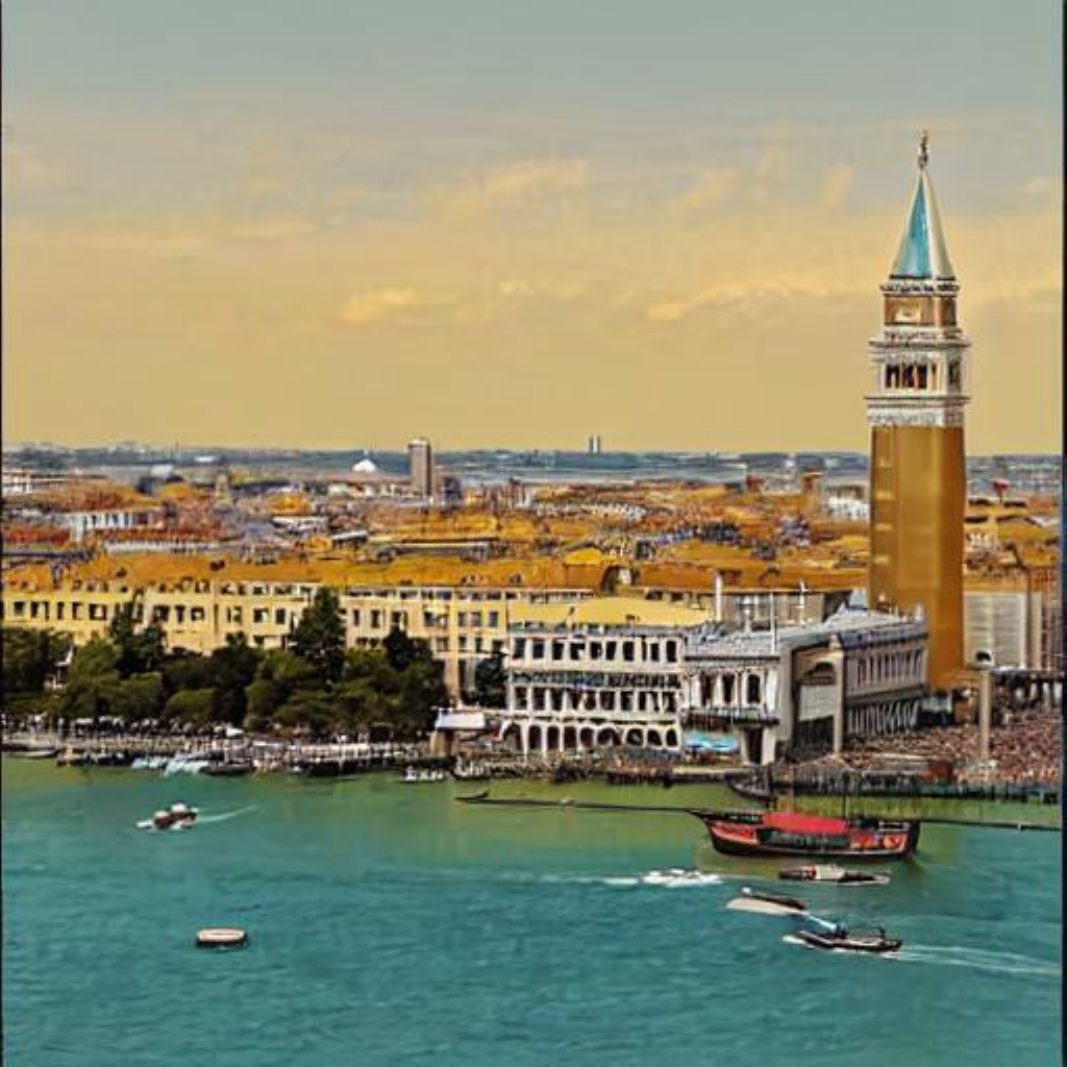} &
    \includegraphics[width=\imgwidth\textwidth]{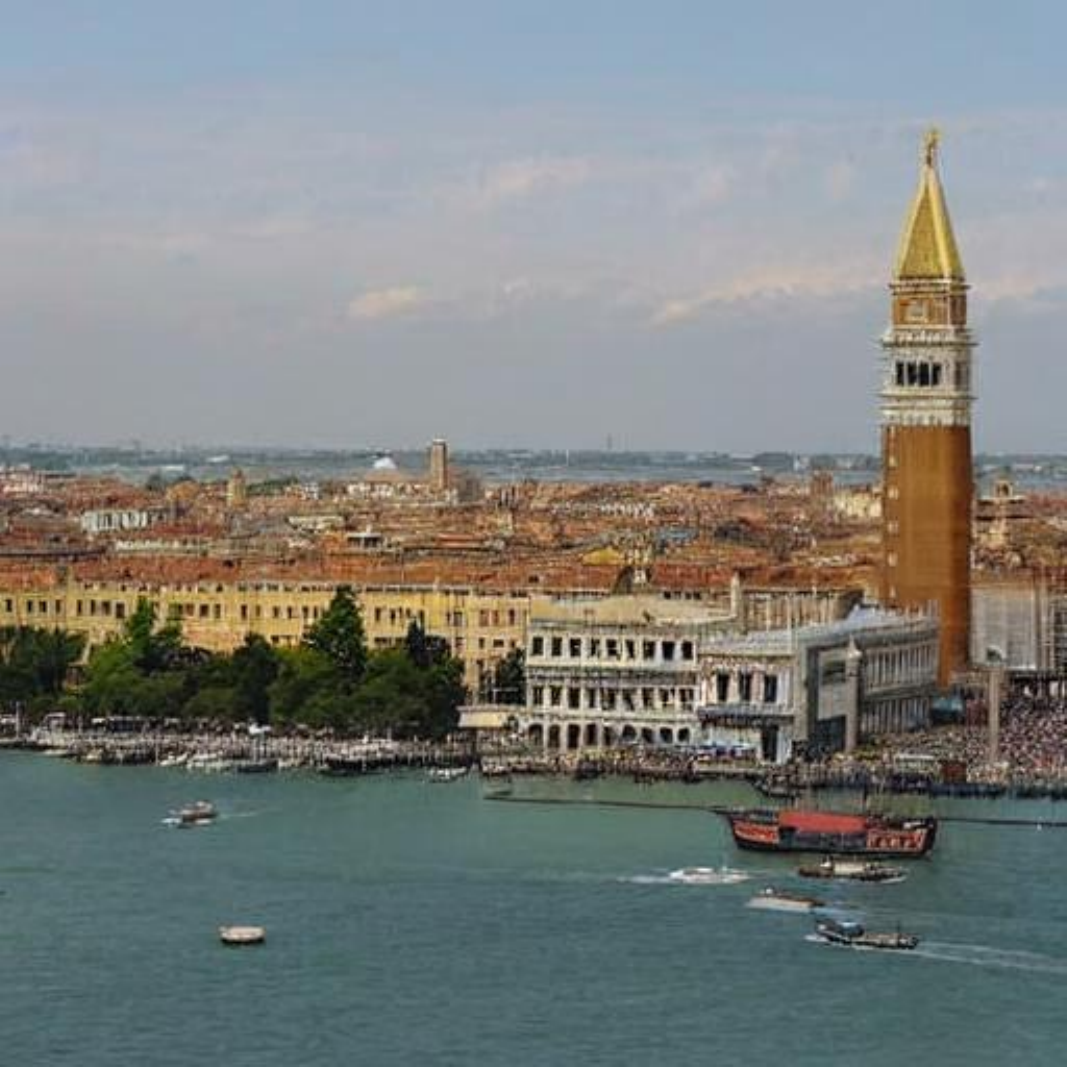} &
    \includegraphics[width=\imgwidth\textwidth]{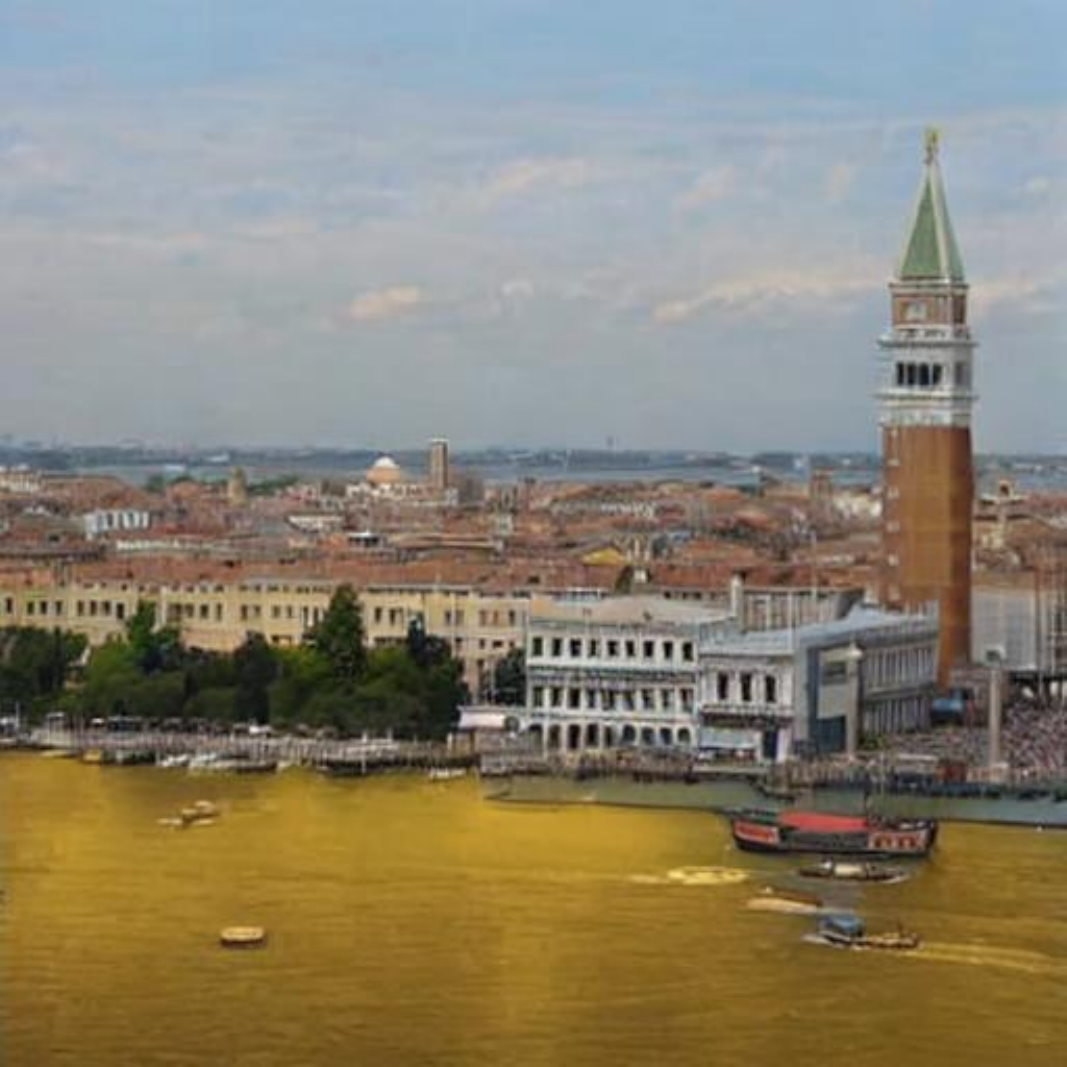} \vspace{0.1em}\\
    \multicolumn{6}{c}{\scriptsize ``change the water for gold"} \vspace{0.6em}\\ 

    \includegraphics[width=\imgwidth\textwidth]{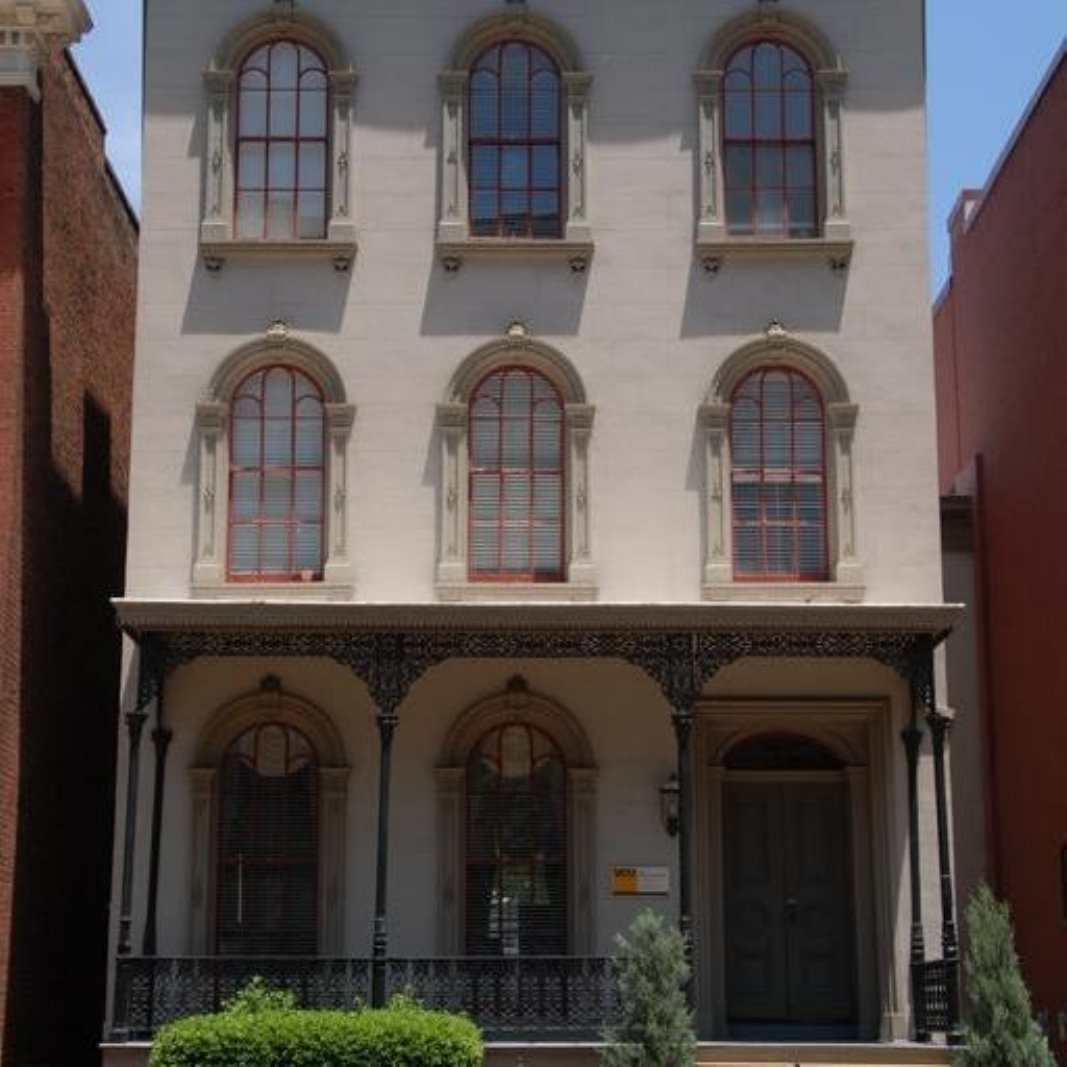} &
    \includegraphics[width=\imgwidth\textwidth]{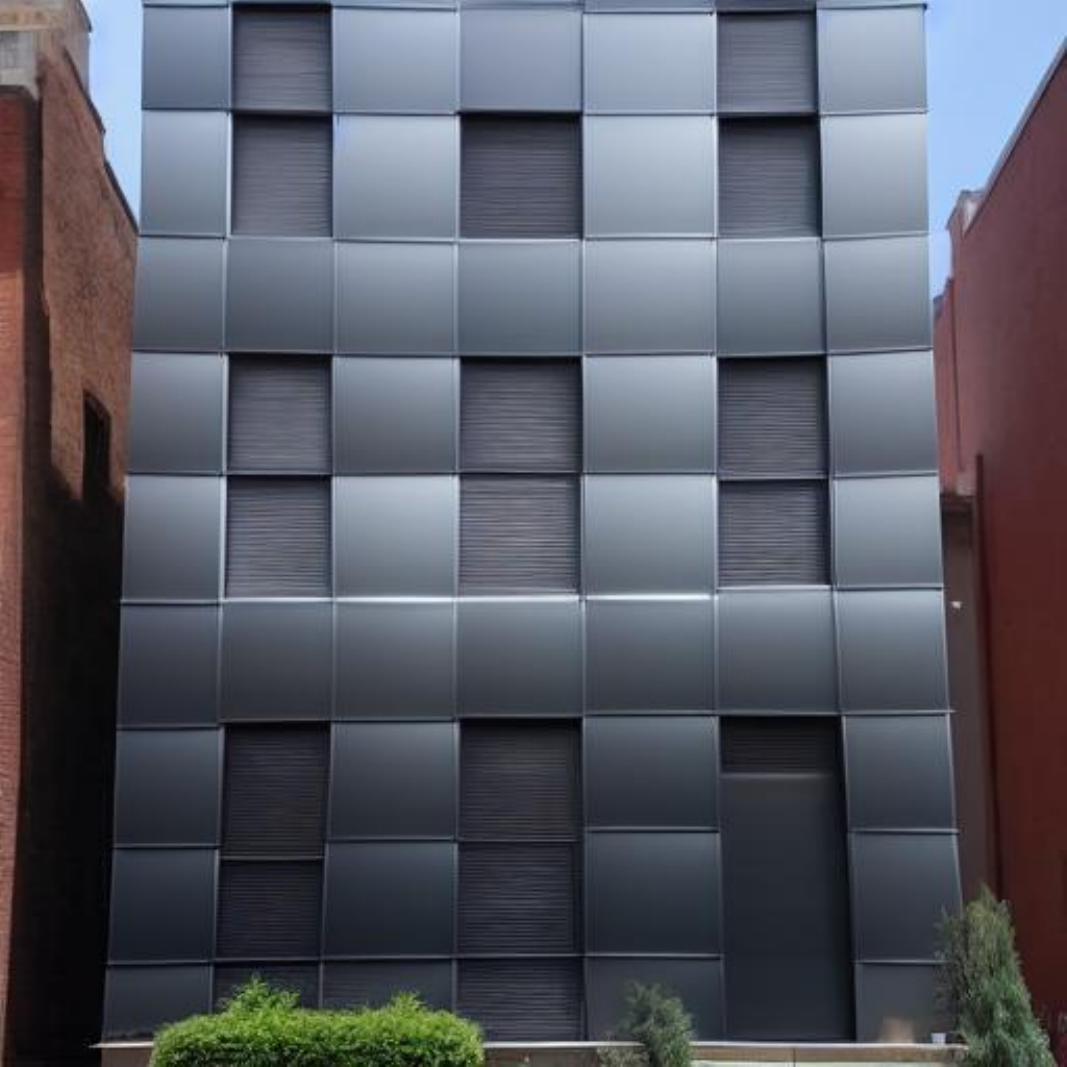} &
    \includegraphics[width=\imgwidth\textwidth]{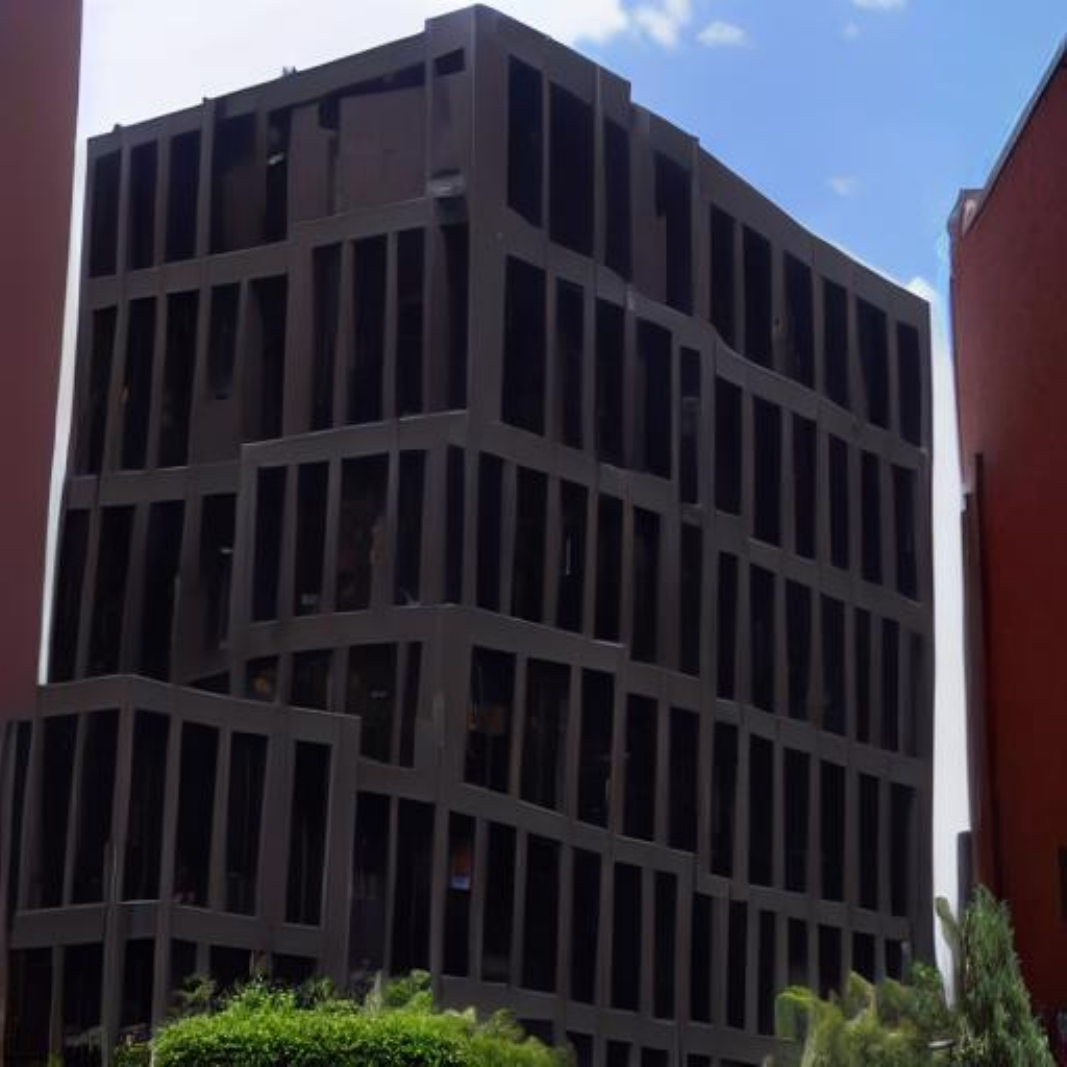} &
    \includegraphics[width=\imgwidth\textwidth]{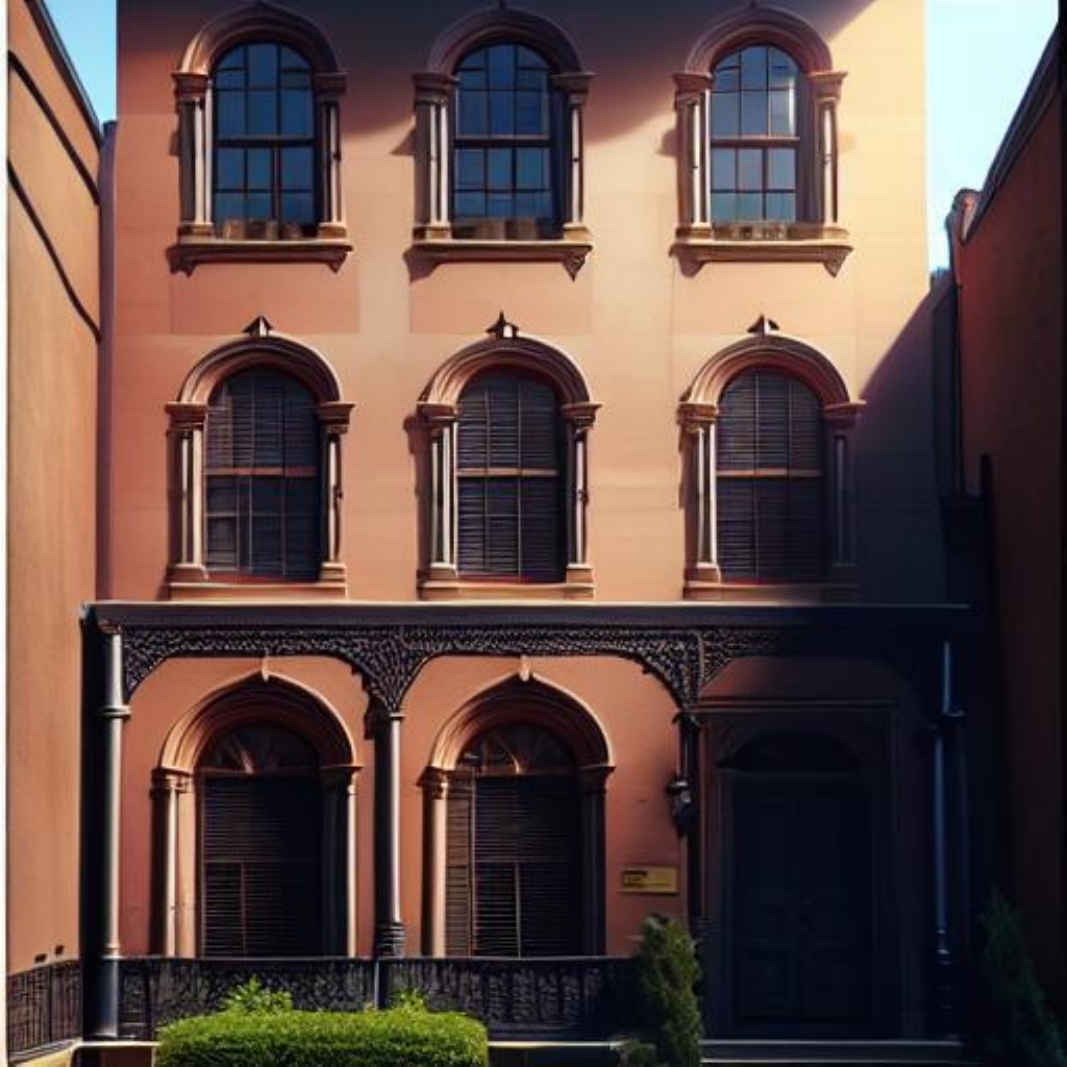} &
    \includegraphics[width=\imgwidth\textwidth]{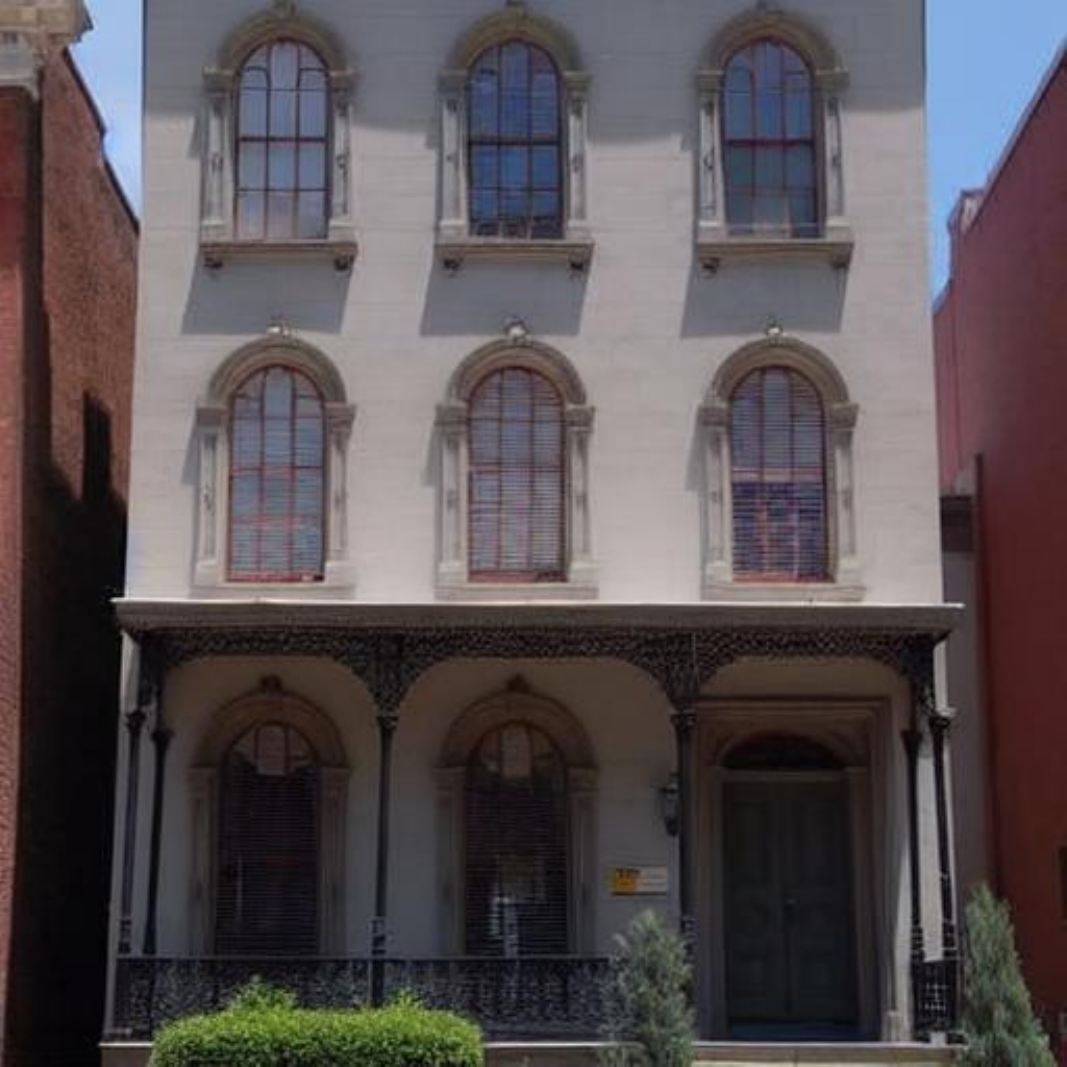} &
    \includegraphics[width=\imgwidth\textwidth]{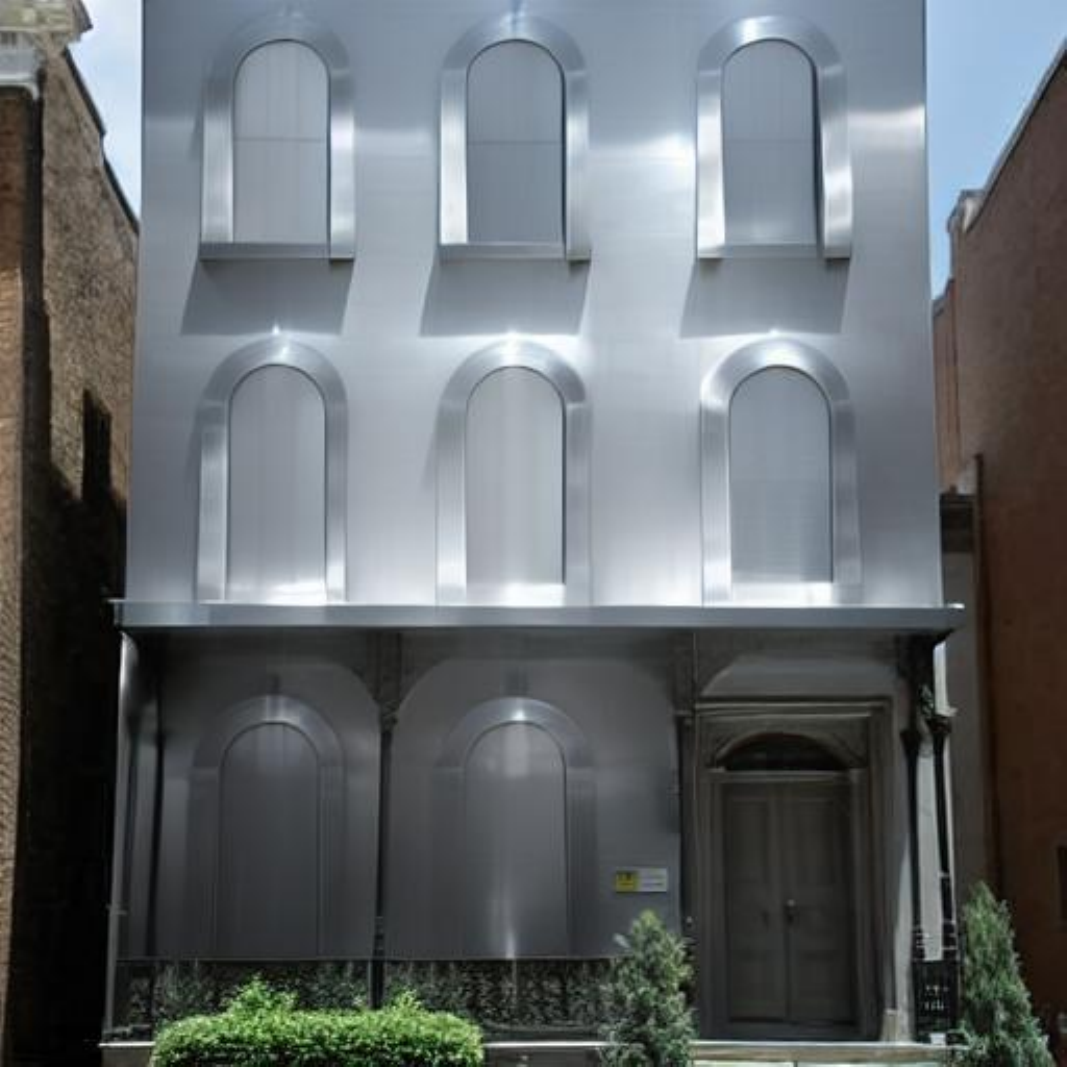} \vspace{0.1em}\\
    \multicolumn{6}{c}{\scriptsize ``turn the building into steel"} \vspace{0.6em}\\ 

    \includegraphics[width=\imgwidth\textwidth]{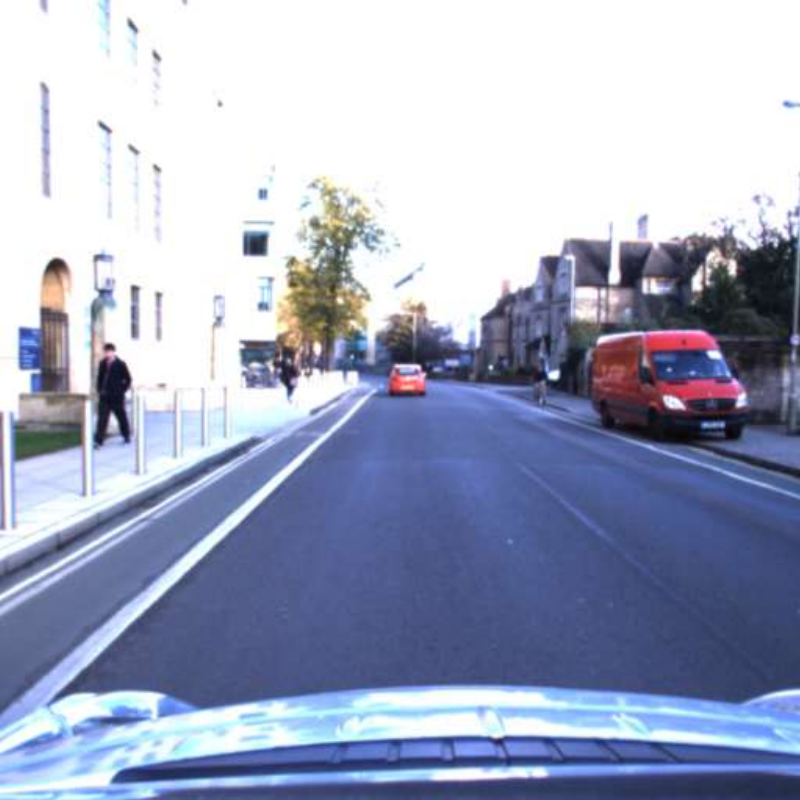} &
    \includegraphics[width=\imgwidth\textwidth]{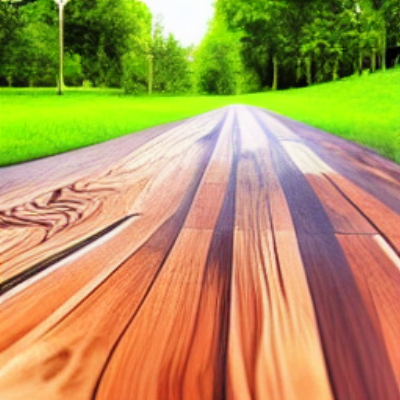} &
    \includegraphics[width=\imgwidth\textwidth]{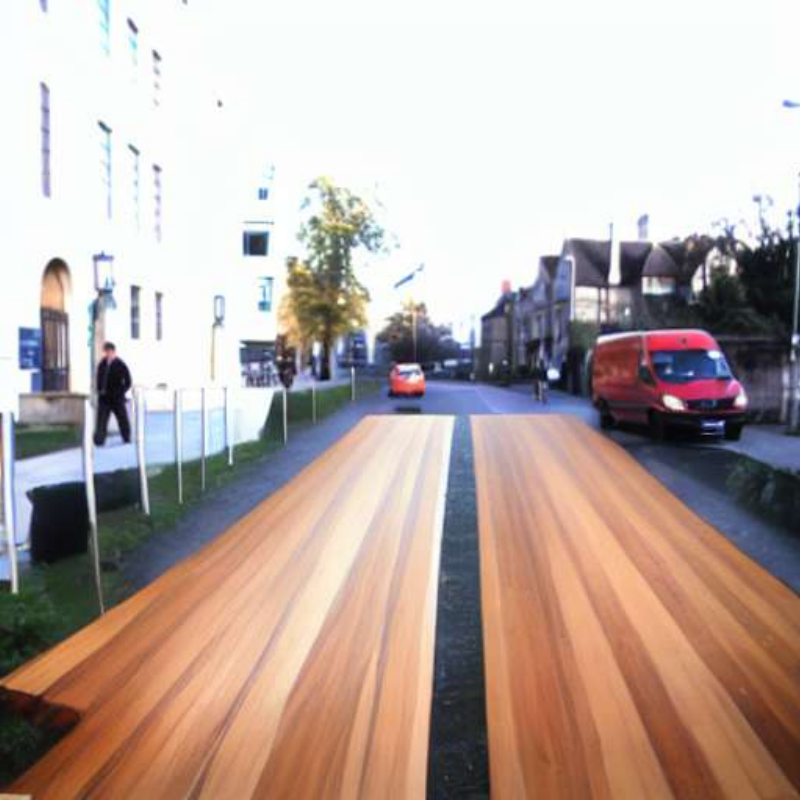} &
    \includegraphics[width=\imgwidth\textwidth]{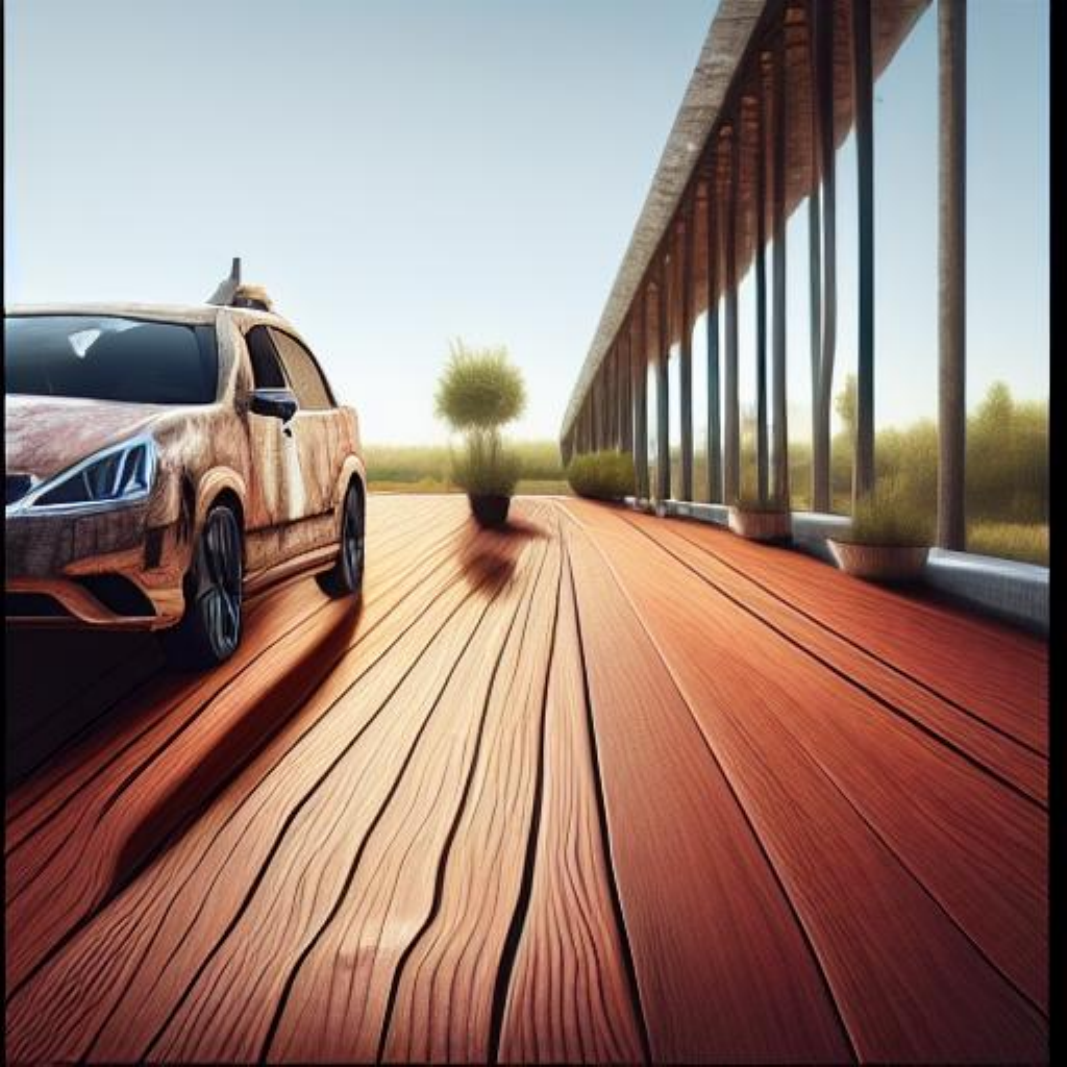} &
    \includegraphics[width=\imgwidth\textwidth]{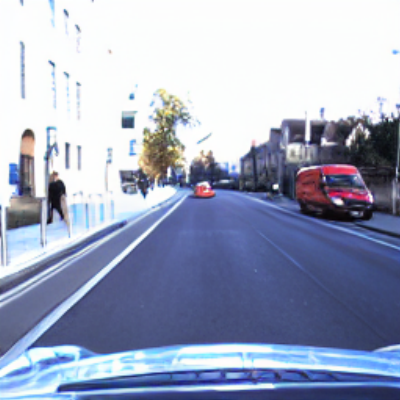} &
    \includegraphics[width=\imgwidth\textwidth]{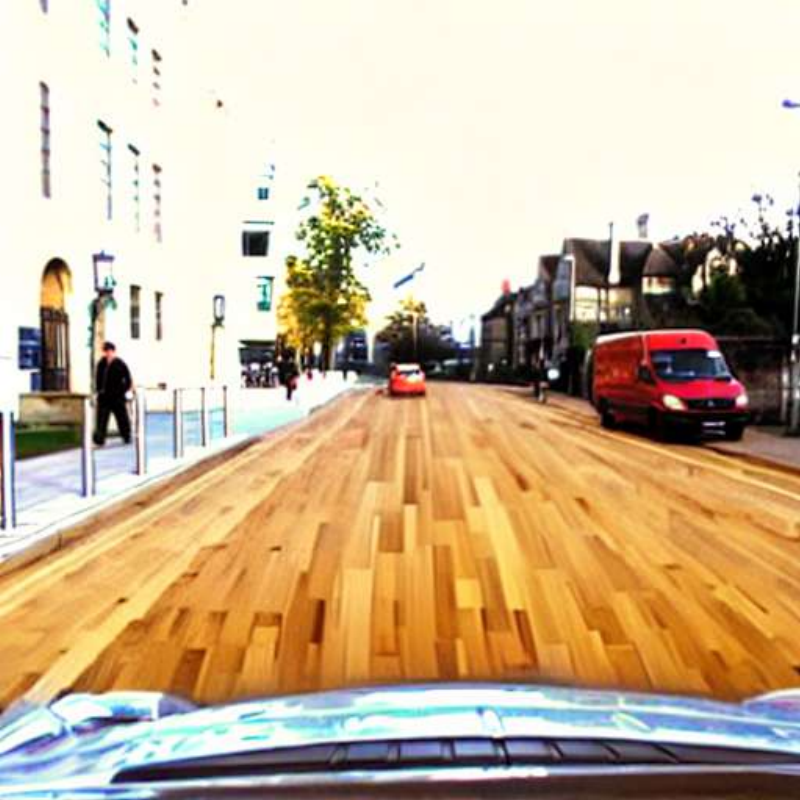} \vspace{0.1em}\\
    \multicolumn{6}{c}{\scriptsize ``turn the road into wood"} \vspace{0.6em}\\

\end{tabular}
\caption{Qualitative comparison. SPIE outperforms its counterparts by significantly editing the image while sharply preserving the structure of regions unrelated to the instruction. We exclude the conditioning style image from the visualization since it is not applicable to the other methods. Additional samples are shown in Appendix \ref{appendix}.}
\label{real_qualitative}
\vspace{-0.75em}
\end{figure}

We begin by comparing the qualitative results in Figure \ref{real_qualitative}. InstructPix2Pix often struggles to precisely locate the edit region, while MagicBrush and HQ-Edit generate unnatural edits with unrealistic prompts. 
HIVE tends to prioritize input fidelity over executing edits, sometimes resulting in insufficient modifications. 
In contrast, our method strikes a better balance between editing strength and input fidelity, producing edits that are both prompt-aligned and faithful to the original image. 
Additionally, SPIE mitigates text prompt-induced biases by suppressing irrelevant hallucinations, like mistaking ``wood" for a forest.

\begin{figure}[t]
\centering
\adjustbox{trim=0 0 0 0, clip}{
\includegraphics[width=0.47\textwidth]{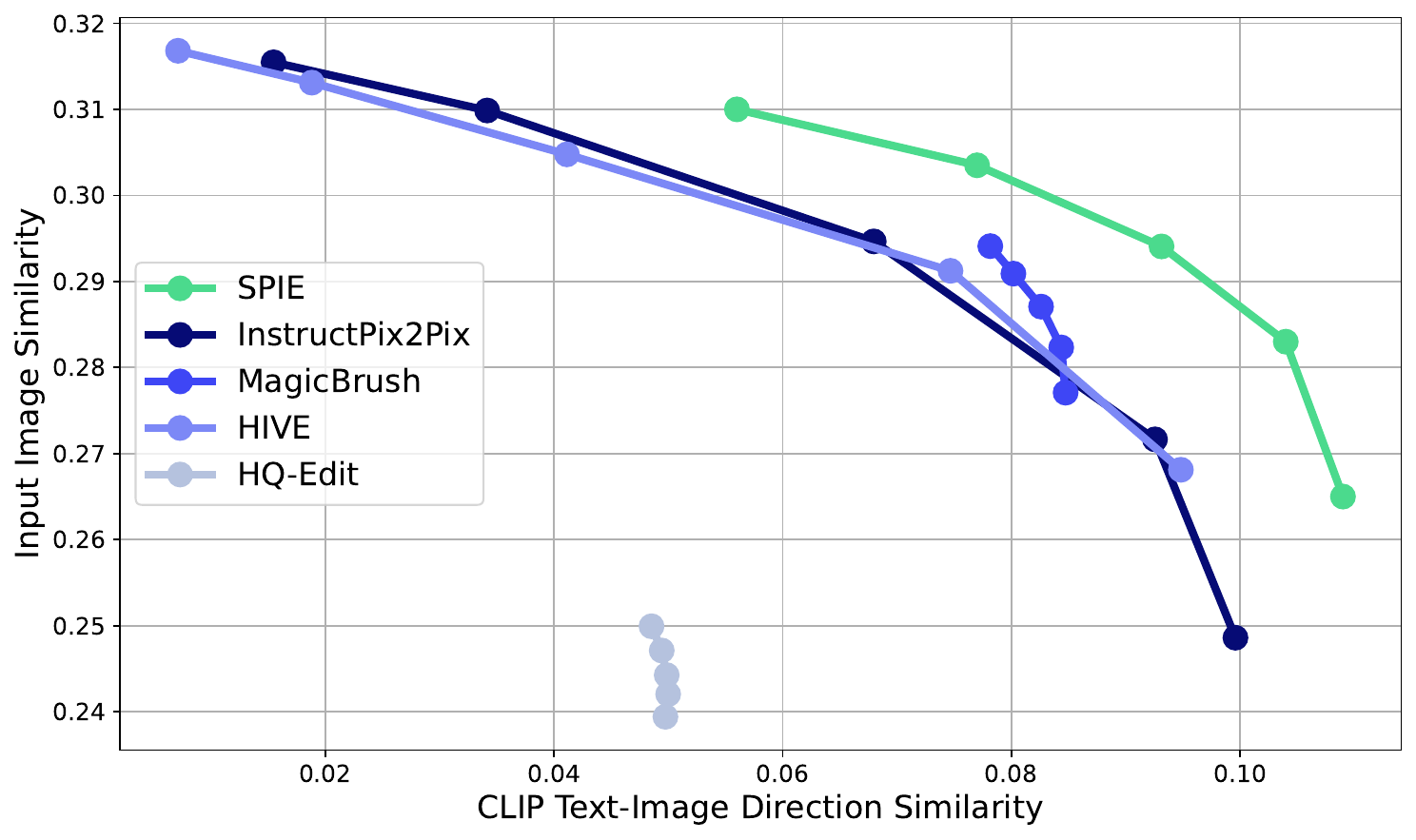}
}
\caption{Quantitative comparison of instructional editing models. We plot the trade-off between input image consistency (Y-axis) and edit consistency (X-axis). Higher values indicate better performance for both metrics. For all methods, we fix the same parameters as in \citet{ip2p} and vary $s_{I_\text{in}} \in [1.0, 2.0]$.}
\label{quant_og_baseline}
\end{figure}

\begin{table}[t]
\centering
\setlength{\tabcolsep}{3pt}
\scalebox{0.62}{
\begin{tabular}{lccccccccc}
\toprule
& \multicolumn{2}{c}{Structural $\downarrow$} & \multicolumn{4}{c}{Semantic $\uparrow$} & \multicolumn{3}{c}{Human Preference $\uparrow$}\\
\cmidrule(lr){2-3} \cmidrule(lr){4-7} \cmidrule(lr){8-10}
Method & Depth & $\text{L}_{2_\text{out}}$ & $\text{CLIP}_{\text{in}}$ & $\text{DINO}_{\text{in}}$ & $\text{DSim}_{\text{in}}$  & $\text{CLIP}_{\text{txt}}$ & ImageReward & PickScore & HPSv2\\
\midrule
IP2P    & 30.06 & 0.034 & 0.430  & 0.058 & \underline{0.152} & 0.195 & -0.648 & 19.18 & 21.60\\
MBrush  & 56.50 & 0.031 & 0.422 & 0.059 & 0.148 & \underline{0.209} & -0.420 & 19.16 & 21.67\\
HQ-Edit & 90.81 & 0.135 & 0.399 & 0.057 & 0.141 & 0.206 & \underline{-0.293} & 19.08 & \textbf{24.79}\\
HIVE    & \underline{18.09} & \textbf{0.022} & \underline{0.438}  & \underline{0.064} & 0.145 & 0.177 & -0.848 & \underline{19.19} & 21.14\\
\textbf{SPIE}    & \textbf{16.55} & \underline{0.025} & \textbf{0.440} & \textbf{0.076} & \textbf{0.219} & \textbf{0.213} & \textbf{-0.284} & \textbf{19.21} & \underline{21.82}\\
\bottomrule
\end{tabular}
}
\caption{Comparison of structural preservation through depth mask alignment and reconstruction metrics (left), semantic alignment with text and visual prompts (center), and human-preference-aligned metrics (right). Some metrics are computed for regions inside and outside the edit mask, denoted by indices ``in" and ``out" respectively. Text alignment is evaluated using descriptive prompts capturing all image information.}
\label{table_structure}
\vspace{-0.75em}
\end{table}

We quantitatively assess the tradeoff between input fidelity and text alignment in Figure \ref{quant_og_baseline}. 
Input fidelity is measured via cosine similarity of image patch embeddings outside the edit region, using grounded SAM2 for high-quality masking. We capture both high and low-level features by averaging scores across DINOv2 \cite{dinov2}, CLIP \cite{clip}, and DreamSim \cite{dreamsim} encoders. Text alignment is evaluated using directional CLIP similarity \cite{clipdir}, which measures how well the change in descriptive captions agrees with the change in input and generated images.
Both metrics are antagonistic, increasing the desired edit strength will reduce the output's faithfulness to the input image. 
Our method achieves better directional similarity for the same level of image consistency compared to baselines. 
These quantitative results hence confirm that counterparts like HIVE emphasize preserving the input image over executing strong edits. 
In contrast, SPIE offers superior balance between edit strength and input fidelity. 
Additionally, Table \ref{table_structure} shows that while HIVE excels at reconstructing unchanged regions, our method closely matches this performance and outperforms others in depth mask alignment, indicating effective preservation of structure, crucial for visual coherence.

\begin{figure}[t]
\centering
\newcommand{\imgwidth}{0.15\textwidth}
\setlength{\tabcolsep}{1pt} 
\renewcommand{\arraystretch}{0.25} 
\newcommand{\stycompression}{0.05}
\newcommand{\styposition}{62}
\begin{tabular}{ccc}
    Input & Sparse Snow & Dense Snow \\

    \includegraphics[width=\imgwidth]{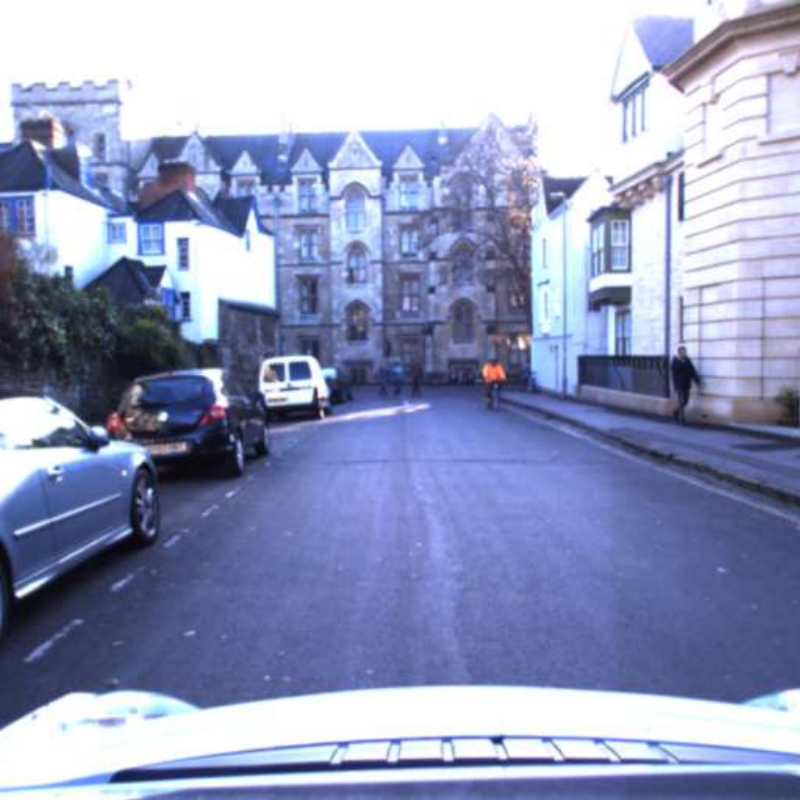} &
    
    \begin{overpic}[width=\imgwidth]{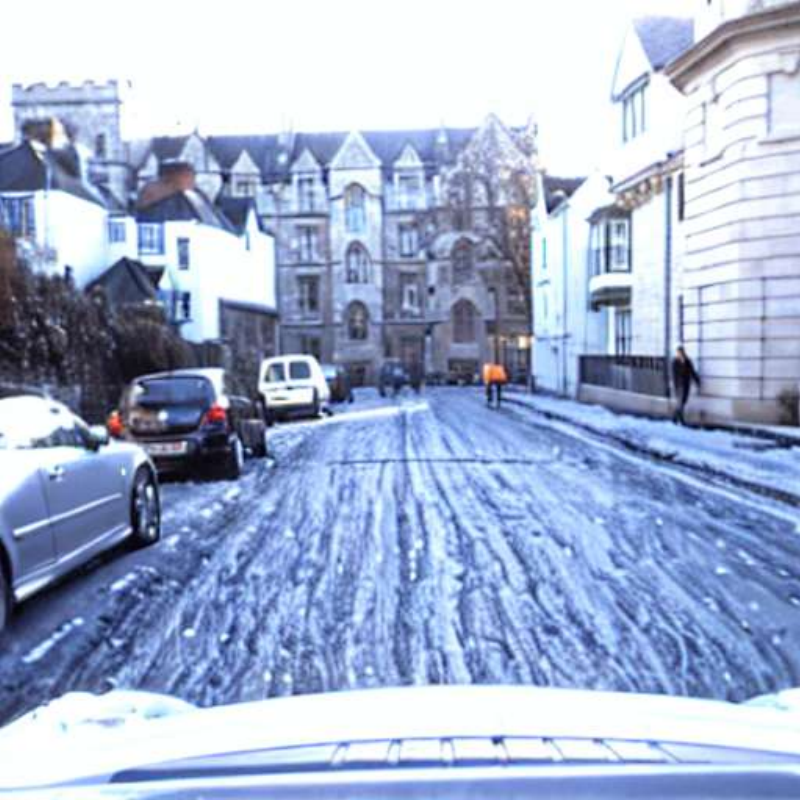}
        \put(-1, 63){
         \begin{tikzpicture}
          \begin{scope}
           \clip[rounded corners=5pt] (0,0) rectangle (\stycompression\textwidth, \stycompression\textwidth);
           \node[anchor=north east, inner sep=0pt] at (\stycompression\textwidth,\stycompression\textwidth) {\includegraphics[width=\stycompression\textwidth]{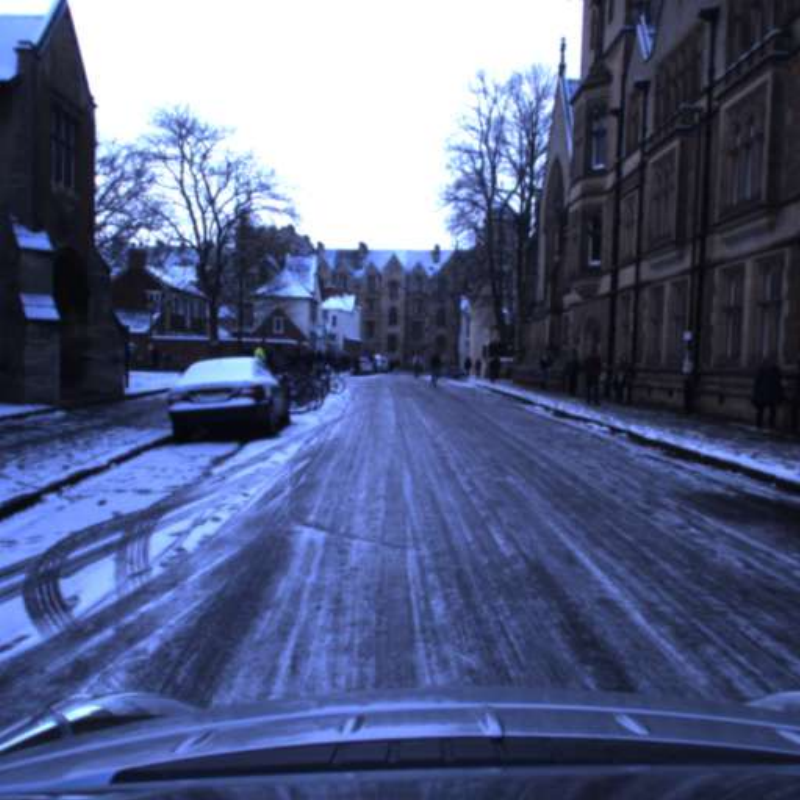}};
        \end{scope}
        \draw[rounded corners=5pt, RubineRed, very thick, dashed] (0,0) rectangle (\stycompression\textwidth,\stycompression\textwidth);
        \end{tikzpicture}}
    \end{overpic}&

    \begin{overpic}[width=\imgwidth]{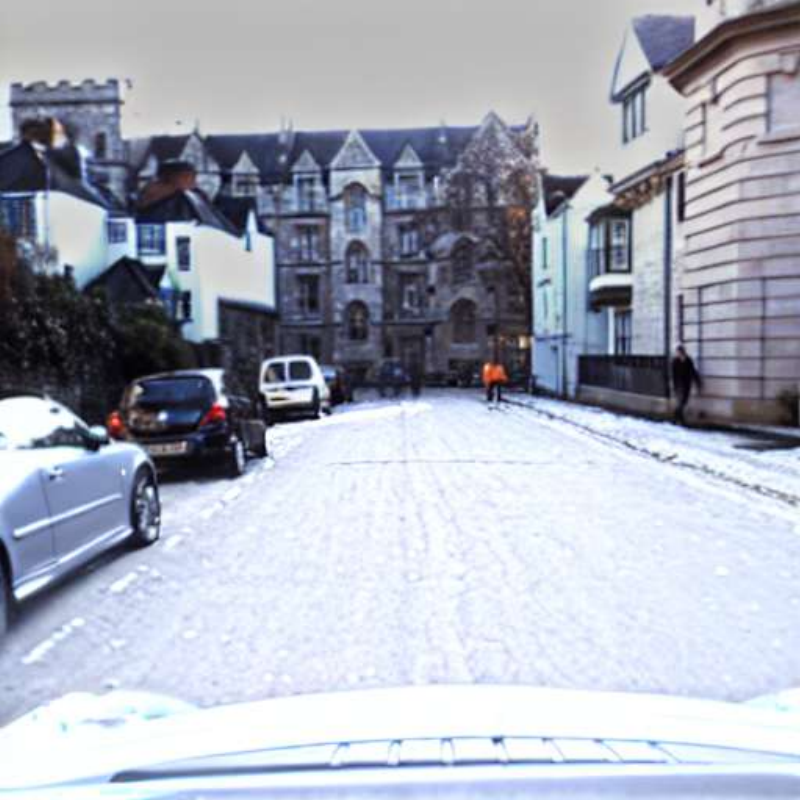}
        \put(-1, 63){
         \begin{tikzpicture}
          \begin{scope}
           \clip[rounded corners=5pt] (0,0) rectangle (\stycompression\textwidth, \stycompression\textwidth);
           \node[anchor=north east, inner sep=0pt] at (\stycompression\textwidth,\stycompression\textwidth) {\includegraphics[width=\stycompression\textwidth]{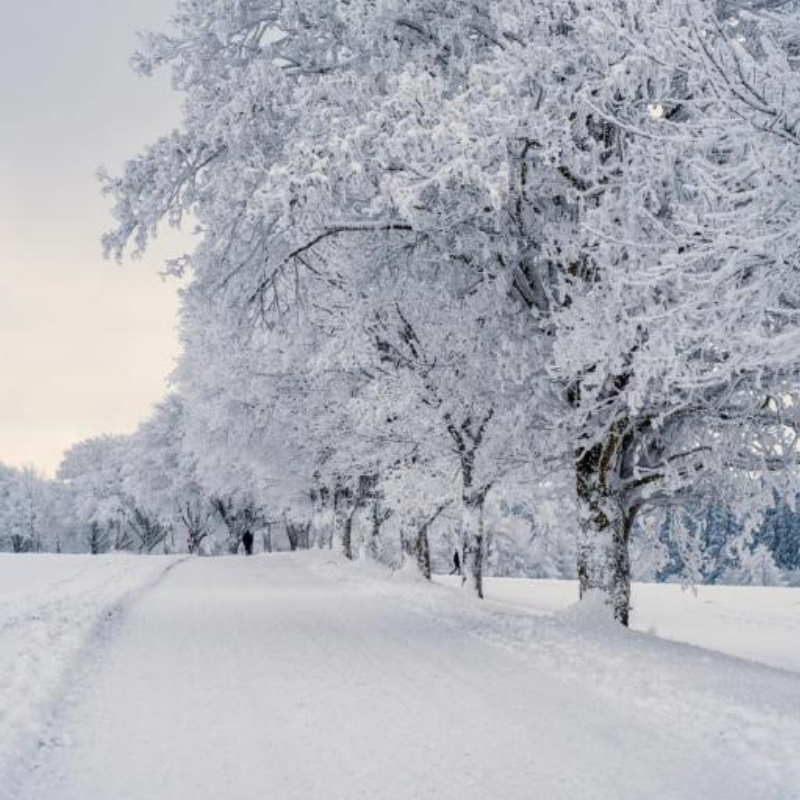}};
        \end{scope}
        \draw[rounded corners=5pt, RubineRed, very thick, dashed] (0,0) rectangle (\stycompression\textwidth,\stycompression\textwidth);
        \end{tikzpicture}}
    \end{overpic} \\[.25em]

    \includegraphics[width=\imgwidth]{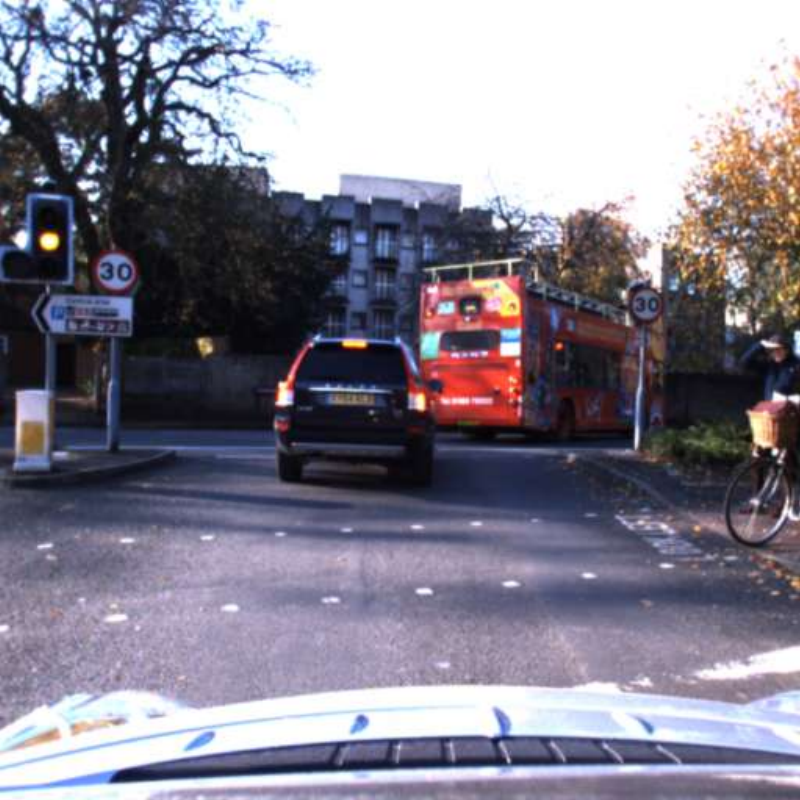} &

    \begin{overpic}[width=\imgwidth]{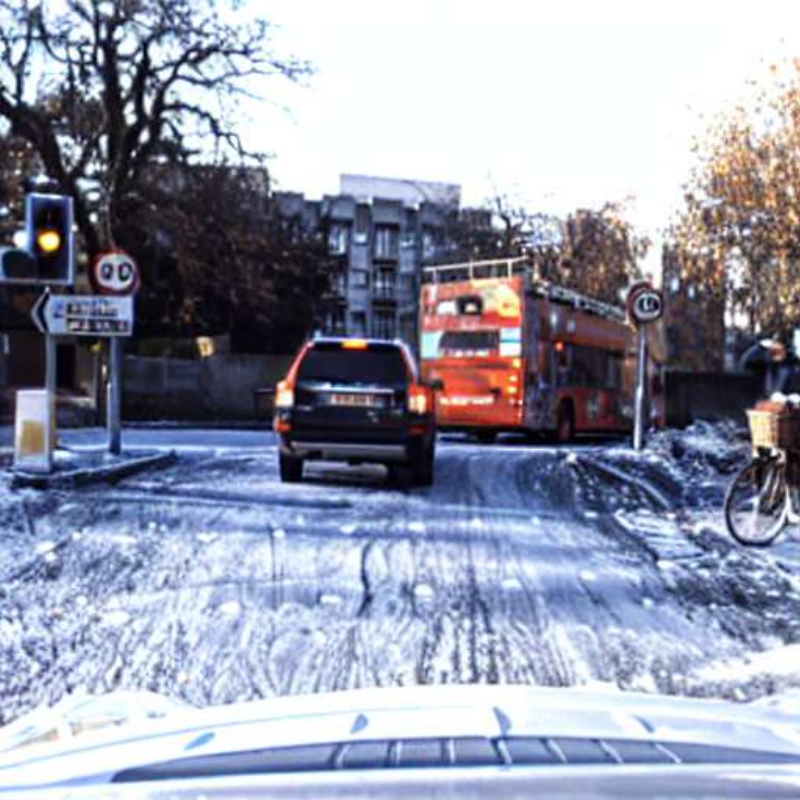}
        \put(60, 63){
         \begin{tikzpicture}
          \begin{scope}
           \clip[rounded corners=5pt] (0,0) rectangle (\stycompression\textwidth, \stycompression\textwidth);
           \node[anchor=north east, inner sep=0pt] at (\stycompression\textwidth,\stycompression\textwidth) {\includegraphics[width=\stycompression\textwidth]{images/qualitative/all_pdf_512/tgt/snow/1422953601799169.pdf}};
        \end{scope}
        \draw[rounded corners=5pt, RubineRed, very thick, dashed] (0,0) rectangle (\stycompression\textwidth,\stycompression\textwidth);
        \end{tikzpicture}}
    \end{overpic} &

    \begin{overpic}[width=\imgwidth]{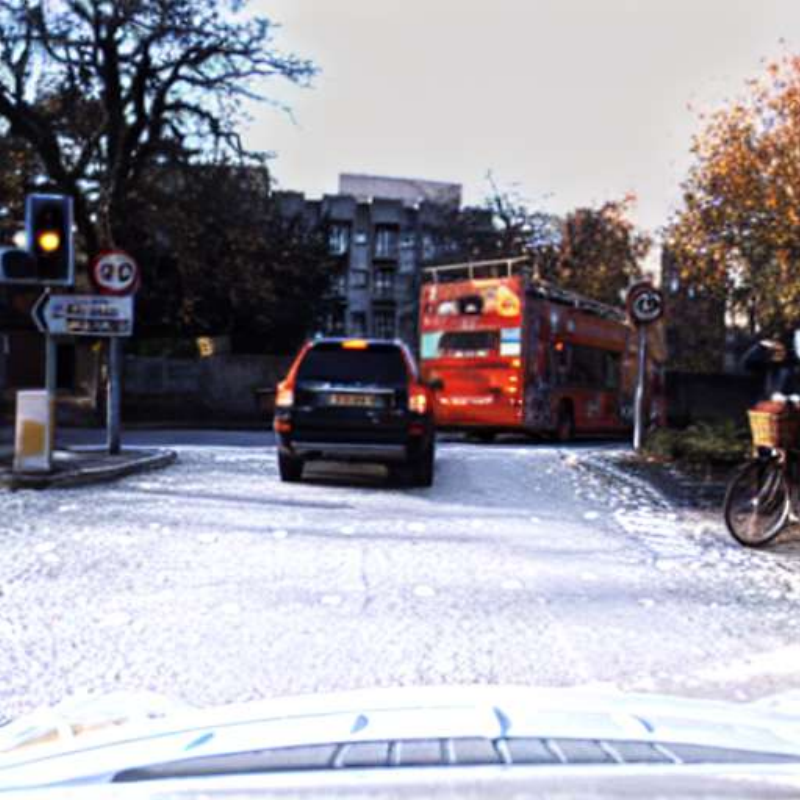}
        \put(60, 63){
         \begin{tikzpicture}
          \begin{scope}
           \clip[rounded corners=5pt] (0,0) rectangle (\stycompression\textwidth, \stycompression\textwidth);
           \node[anchor=north east, inner sep=0pt] at (\stycompression\textwidth,\stycompression\textwidth) {\includegraphics[width=\stycompression\textwidth]{images/qualitative/all_pdf_512/tgt/snow_nolanes/img2.pdf}};
        \end{scope}
        \draw[rounded corners=5pt, RubineRed, very thick, dashed] (0,0) rectangle (\stycompression\textwidth,\stycompression\textwidth);
        \end{tikzpicture}}
    \end{overpic}
\end{tabular}

\caption{Visualization of the impact of the visual prompt (displayed in dashed lines contour) on generated samples when provided the text instruction ``add snow on the road". SPIE effectively captures semantic nuances beyond that described in the text prompt. 
We present further applications of such nuanced control for materials in Fig. \ref{appdx_real_conditioning_qualitative}, demonstrating how our method can effectively handle diverse visual styles and attributes, such as varying colors, while maintaining structural coherence.
}
\label{real_conditioning_qualitative}
\vspace{-0.75em}
\end{figure}

We demonstrate SPIE's ability to interpret subtle details beyond text prompts by training it to add dense and sparse snow layers without explicit text instructions. 
Figure \ref{real_conditioning_qualitative} shows that our model effectively reproduces the visual style hinted at in the text prompt while respecting the input image's spatial composition. 
This underscores the benefit of leveraging a visual prompt to infuse fine-grained nuances without requiring a extensive text descriptions.

We quantitatively validate these observations in Table \ref{table_structure}.
Our evaluation focuses on two key aspects: visual semantic alignment between regions-of-interest, and text-image alignment scores.
For the visual alignment, we compute the cosine similarity between masked regions in the generated and conditioning images, across DINOv2, CLIP, and DreamSim embeddings.
Text-image alignment ($\text{CLIP}_\text{txt}$) is measured through CLIP similarity between the edited image and output caption.
SPIE surpasses all baselines, both in the visual and text-image alignments, confirming our method's efficacy to increase instruction fidelity.

We also compare the scores obtained from T2I synthesis preference prediction models, namely ImageReward \cite{imagereward}, PickScore \cite{pickscore}, and HPSv2 \cite{hpsv2}, which emulate human preferences.
We find in Table \ref{table_structure} that our method outperforms its counterparts across the different edits.
This confirms SPIE's superior ability to preserve the essential structural features of the input image, which are crucial for perceived realism, while also aligning effectively with the specified editing instructions.

\begin{table}[t]
    \centering
    \begin{subtable}{\linewidth}
        \centering
        \scalebox{0.89}{
        \begin{tabular}{lccccc}
            \toprule
            Method & L1 $\downarrow$ & L2 $\downarrow$ & CLIP-I $\uparrow$ & DINO $\uparrow$ & CLIP-T $\uparrow$\\
            \midrule
            IP2P & 0.1122 & 0.0371 & 0.8524 & 0.7428 & 0.2764\\
            MBrush & \underline{0.0740} & \underline{0.0267} & \underline{0.9166} & \underline{0.8649} & \textbf{0.2813} \\
            HIVE & 0.1092 & 0.0341 & 0.8519 & 0.7500 & 0.2752\\
            \textbf{SPIE} & \textbf{0.0673} & \textbf{0.0187} & \textbf{0.9224} & \textbf{0.8743} & \underline{0.2768}\\
            \bottomrule
        \end{tabular}
        }
        \caption{MagicBrush Test Set}
        \label{table_magicbrush}
    \end{subtable}

    \vspace{0.5em}

    \begin{subtable}{\linewidth}
        \centering
        \scalebox{0.85}{
        \begin{tabular}{lccccc}
            \toprule
            Method & CLIP$_\text{dir}$ $\uparrow$ & CLIP$_\text{im}$ $\uparrow$ & CLIP$_\text{out}$ $\uparrow$ & L1 $\downarrow$ & DINO $\uparrow$ \\
            \midrule
            IP2P    & 0.078 & 0.834 & 0.219 & 0.121 & 0.762 \\
            MBrush  & \underline{0.090} & 0.838 & \underline{0.222} & 0.100 & 0.776 \\
            HIVE    & 0.061 & \underline{0.882} & 0.213 & \underline{0.083} & \underline{0.822} \\
            Emu Edit & \textbf{0.109} & 0.859 & \textbf{0.231} & 0.094 & 0.819 \\
            \textbf{SPIE} & 0.063 & \textbf{0.897} & 0.221 & \textbf{0.078} & \textbf{0.858} \\
            \bottomrule
        \end{tabular}
        }
        \caption{Emu Edit Test Set}
        \label{table_emuedit}
    \vspace{-0.5em}
    \end{subtable}
    \caption{Evaluation on benchmarks spanning diverse editing tasks.}
    \label{standard_dataset}
    \vspace{-0.5em}
\end{table}

Finally, we extend our baseline comparisons to standardized benchmarks that encompass a broader range of editing tasks.
We evaluate our method in Tables \ref{table_magicbrush} and \ref{table_emuedit}, without additional training. 
SPIE outperforms all baselines in structure preservation and surpasses IP2P in semantics, demonstrating strong transferability beyond region-based editing.

\subsection{Sim-to-Real Editing}
\label{exp_sim2real_RTX}

SPIE demonstrates utility beyond creative applications by enhancing the visual realism of simulated environments, addressing a critical limitation in their use for robotics research.
In robotics, evaluating generalist manipulation policies poses significant challenges due to the scalability and reproducibility constraints of real-world testing.
SIMPLER \cite{simpler}, a framework for simulation-based evaluation, aims to provide a reliable proxy for real-world assessments.
A major challenge highlighted by \citet{simpler} is the visual disparity between simulated environments and their real-world counterparts, which can undermine the accuracy of policy evaluation. 
They mitigate this gap with a 2-step approach called visual matching (VisMatch), which overlays simulated elements onto real-world backgrounds, and bakes their textures and colors from real-world images.
However, this method has notable limitations: it relies heavily on human effort, as texture matching is not automated and requires extensive curation of visual assets alongside access to 3D modeling tools. 
Hence, it does not scale efficiently, as assembling new scenes demands additional human input for each instance.

To further reduce the sim-to-real gap, we propose applying SPIE to simulated scenes.
Specifically, we finetune an editing diffusion model using only 5 reference images to produce realistic edits in a given style within the environment.
Such model can then be used to modify the robots' observations of their simulated environment during evaluation, introducing realistic textures and materials that better align with real-world settings.
We conduct experiments in the opening/closing drawer task of the Google Robot environment, as the table texture domain shift is known for being one of the most difficult for policies to generalize effectively \cite{finn_visualrobot}.
To evaluate our method, we specialize expert models to generate variants of the cabinet in diverse materials: wood, gold, leather, stone, steel, and marble.

\begin{figure}[t]
\centering
\newcommand{\imgwidth}{0.094}
\setlength{\tabcolsep}{0.5pt} 
\renewcommand{\arraystretch}{0.25} 
\begin{tabular}{ccccc} 
    Input & Wood & Steel & Marble & Leather \\

    \includegraphics[width=\imgwidth\textwidth]{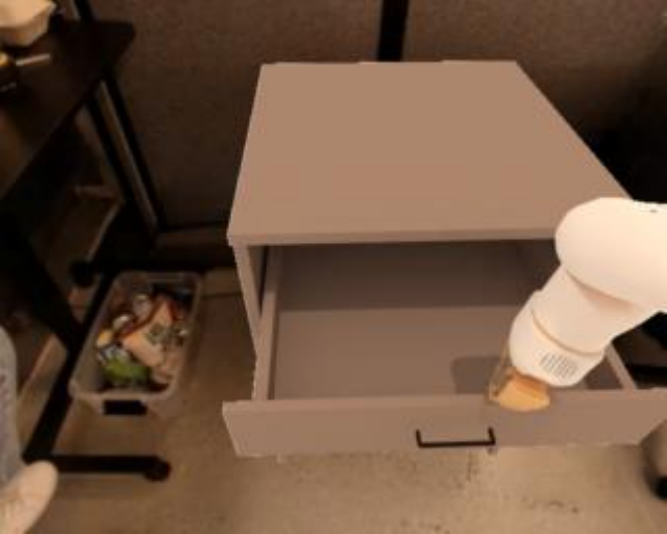} &
    \includegraphics[width=\imgwidth\textwidth]{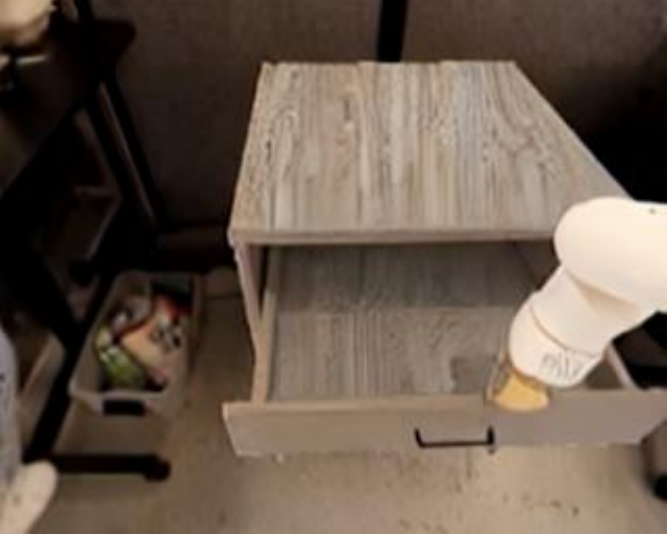} &
    \includegraphics[width=\imgwidth\textwidth]{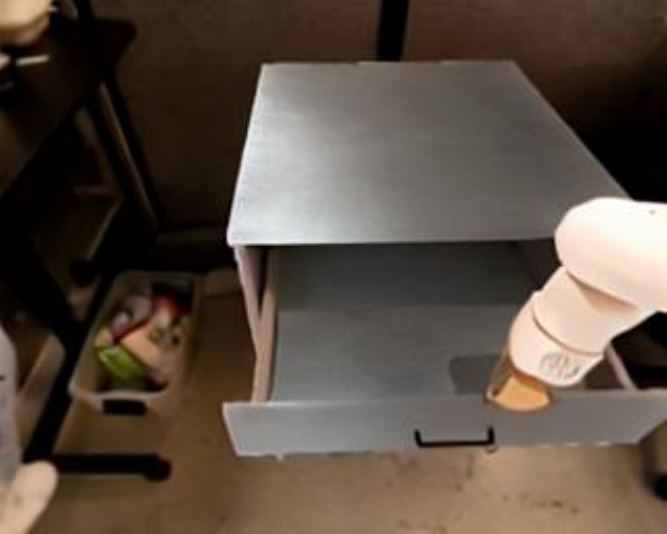} &
    \includegraphics[width=\imgwidth\textwidth]{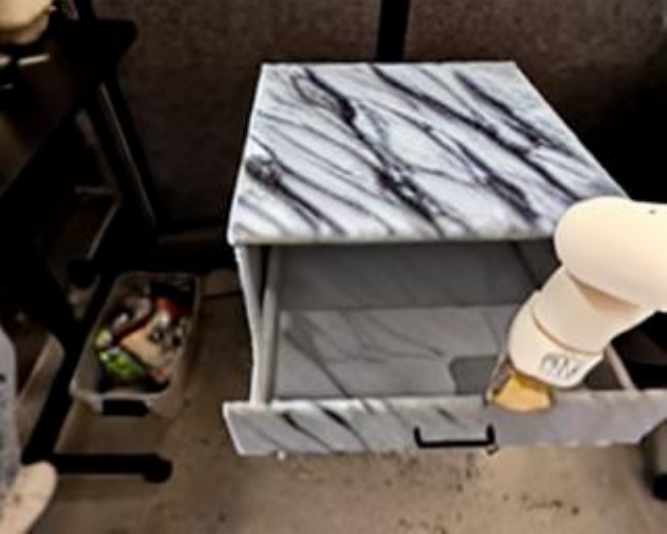} &
    \includegraphics[width=\imgwidth\textwidth]{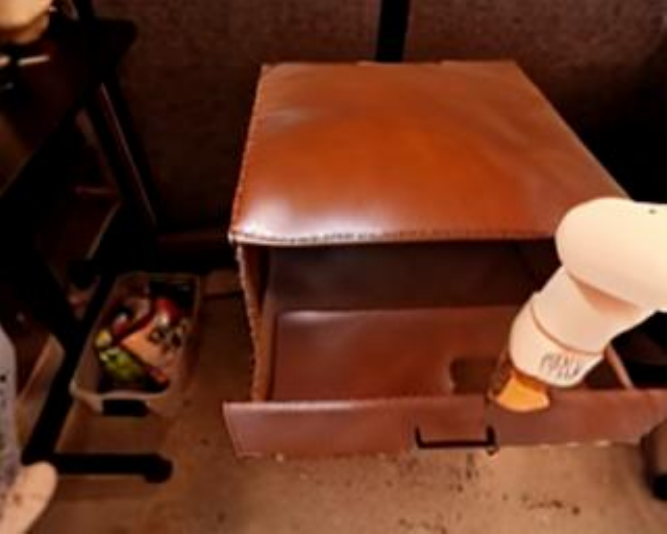} \\

\end{tabular}
\caption{Examples of a simulated scene edited by our method, showcasing enhanced realism compared to the original image across various styles. See Appendix \ref{appendix_experiment_RTX_details} for more variants.}

\label{rtx_main_comp}
\end{figure}

\begin{table}[t]
    \centering
    \scalebox{0.87}{
    \begin{tabular}{llccc}
        \toprule
          & Visual Domain & Open & Close & Average \\
        \midrule
        & SIMPLER-VarAgg & 0.000 & 0.130 & 0.083 \\
        \textbf{MMRV$\downarrow$} & SIMPLER-VisMatch & 0.000 & 0.130 & 0.083 \\
        & \textbf{SPIE} & 0.000 & 0.000 & 0.000 \\
        \cmidrule(r){1-5}
        & SIMPLER-VarAgg & 0.915 & 0.756 & 0.964 \\
        \textbf{Pearson $r$$\uparrow$} & SIMPLER-VisMatch & 0.987 & 0.891 & 0.972 \\
        & \textbf{SPIE} & 0.917 & 0.978 & 0.966 \\
        \bottomrule
    \end{tabular}
    }
    \caption{Comparison of visual domains for RT-1 policy evaluation on Google Robot tasks. 
    Using our method results in much stronger correlation with real evaluation than using the SIMPLER methods. 
    See Table \ref{appdx_rtx_full} for a detailed breakdown of results per policy.}
    \label{rtx_short}
    \vspace{-0.5em}
\end{table}

Visual exemplars in Fig. \ref{rtx_main_comp} demonstrate that our method can convincingly alter simulated scenes to resemble real-world counterparts across various styles.
Quantitatively, we assess the realism of our generated images by evaluating robot policies trained in the real-world on our edited images.
Following \citet{simpler}, we conduct evaluations using multiple RT-1 \cite{rt1model, openx} checkpoints (see Appendix Sec \ref{appendix_experiment_RTX_details} for further experiment details).
Table \ref{rtx_short} summarizes the results on two metrics: Mean Maximum Rank Violation (MMRV) and Pearson correlation coefficient (Pearson \textit{r}), which respectively measure the ranking and linear consistency between simulated and real-world performance. We evaluate SPIE, VisMatch, and the variant aggregation method (VarAgg) \cite{simpler}, which combines many visually randomized versions of a simulated scene, including variations in drawer texture. 
Some policies are highly sensitive to visual discrepancies between simulation and reality, which exacerbates the challenges introduced by domain shifts and causes VarAgg to perform worse than VisMatch. 
Notably, SPIE outperforms both VarAgg and VisMatch, achieving higher MMRV scores—identified by \citet{simpler} as the more robust metric—and higher Pearson \textit{r} scores for the closing task.
These stronger results indicate that our edits provide a more realistic proxy for real-world evaluations.
Improving finer details, such as drawer handles, presents a promising way to further enhance alignment between simulated and real-world environments, as suggested by the lower performance on the opening task against VisMatch.
Overall, SPIE offers a scalable solution for bridging the visual realism gap in robotics policy testing by reducing reliance on manual efforts and enabling automated generation of realistic edits across diverse styles.

\subsection{Ablation Study}
\label{exp_ablation}

\begin{figure}[t]
\centering
\adjustbox{trim=1.5em 0em 2em 1.1em, clip}{
\includegraphics[width=0.55\textwidth]{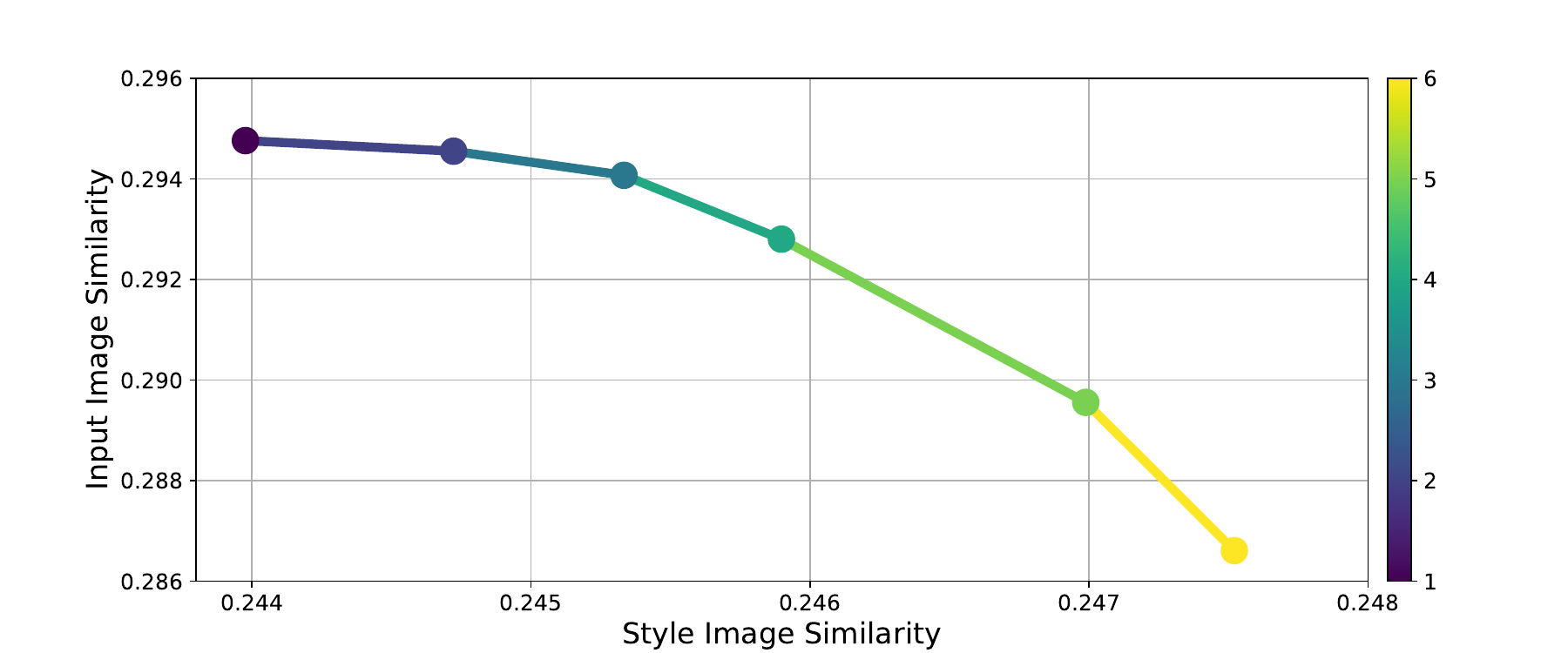}
}
\caption{We plot the trade-off between consistency with the input image (Y-axis) and consistency with the visual prompt (X-axis). For both metrics, higher is better. We fix the same parameters as in \citet{ip2p} and vary $s_{I_\text{sty}} \in [1.0, 6.0]$}
\label{quant_baseline}
\vspace{-0.25em}
\end{figure}

\paragraph{Classifier-free-guidance scale.}
In Figure \ref{quant_baseline}, we examine how varying the classifier-free-guidance scale $s_{I_\text{sty}}$ from Eq. \ref{final_cfg_eq} affects the edit's alignment with the visual prompt. 
Increasing its value enhances alignment with the visual prompt but reduces similarity with the input image. We find that values between 2 to 5 yield the best results, and use $s_{I_\text{sty}}=3$ for quantitative evaluations in Sec. \ref{exp_baseline}.
In practice, and for qualitative results shown in the paper, we find it beneficial to adjust this guidance weight for each edit type to obtain an optimal balance between faithfulness to the input and alignment with the visual prompt.

\begin{figure}[t]
\centering
\newcommand{\imgwidth}{0.094}
\setlength{\tabcolsep}{.5pt} 
\renewcommand{\arraystretch}{0.25} 
\begin{tabular}{ccccc} 
    Input & Style & DINOv2 & CLIP & DreamSim \\

    \includegraphics[width=\imgwidth\textwidth]{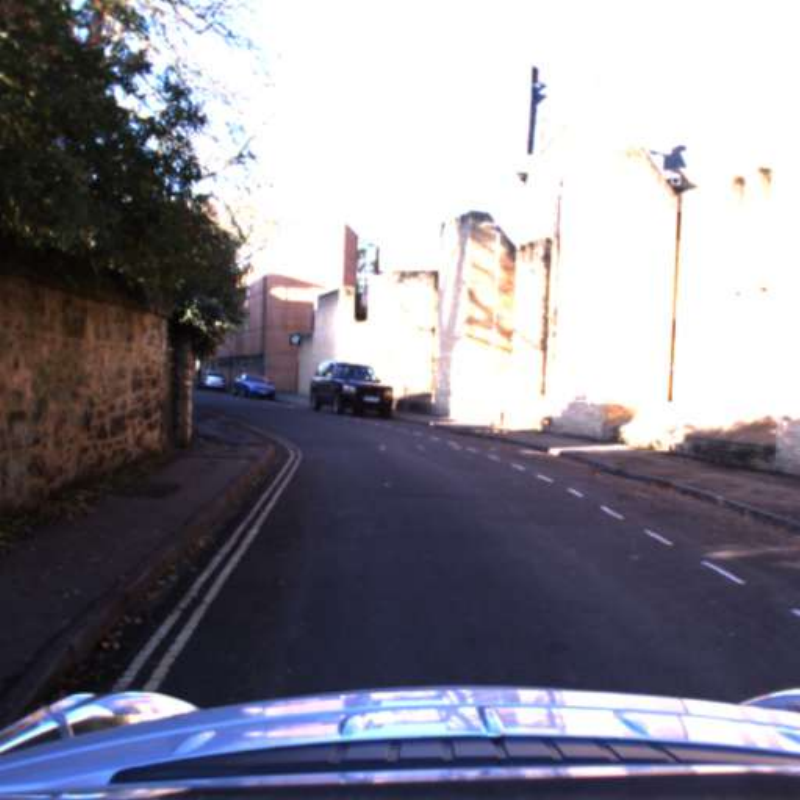} &
    \includegraphics[width=\imgwidth\textwidth]{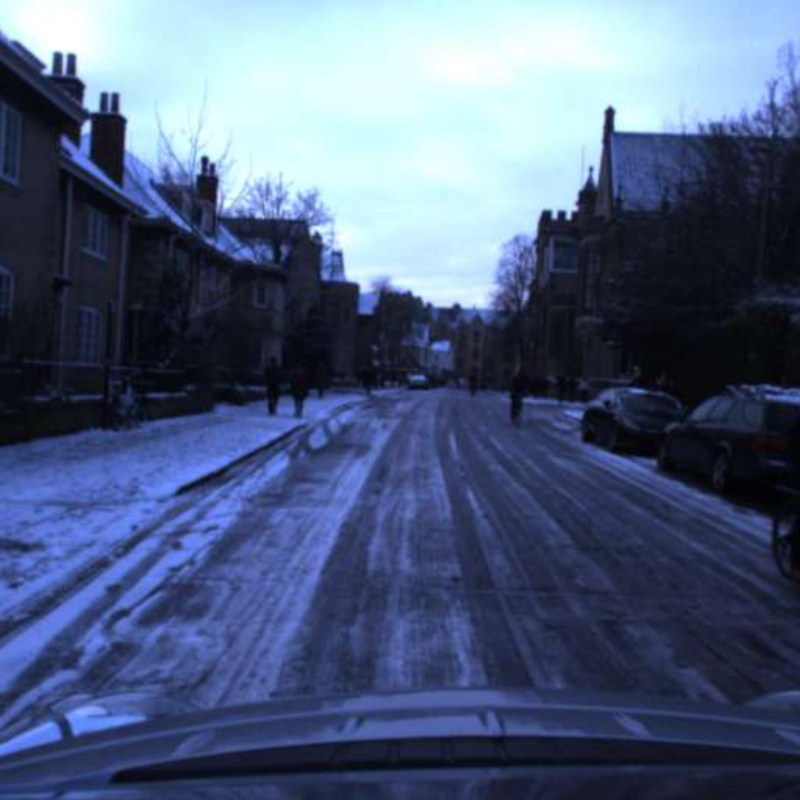} &
    \includegraphics[width=\imgwidth\textwidth]{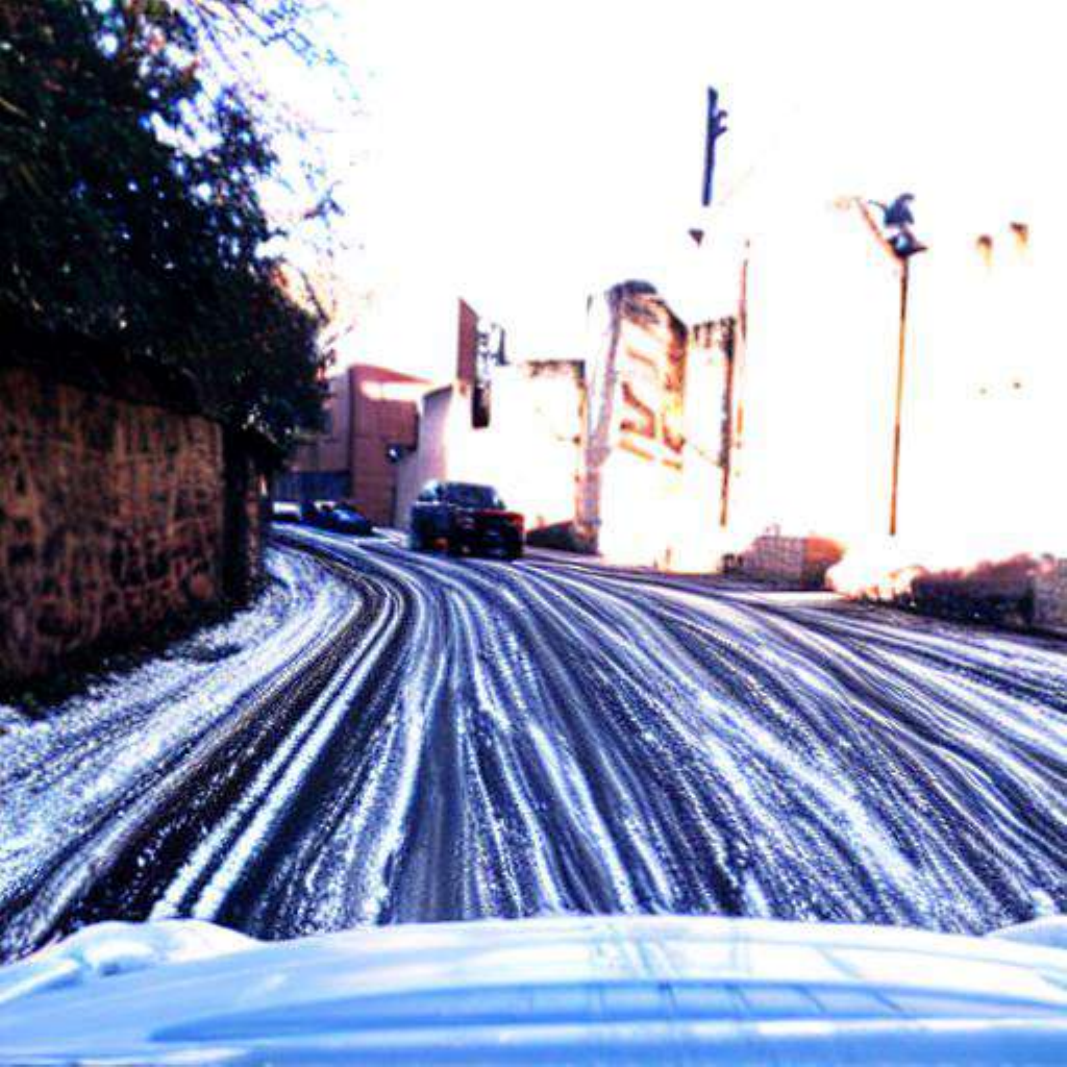} &
    \includegraphics[width=\imgwidth\textwidth]{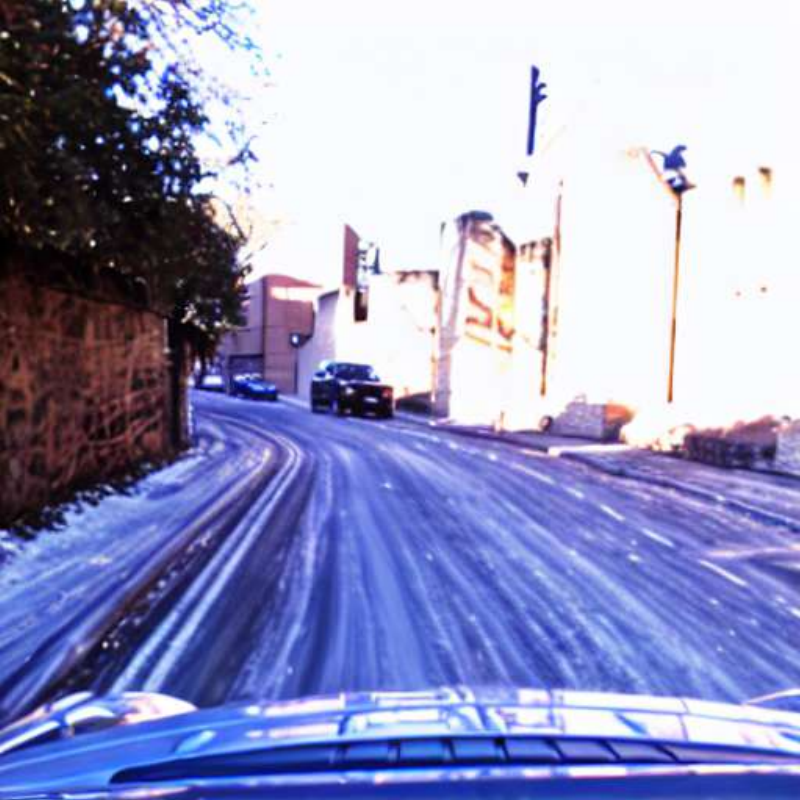} &
    \includegraphics[width=\imgwidth\textwidth]{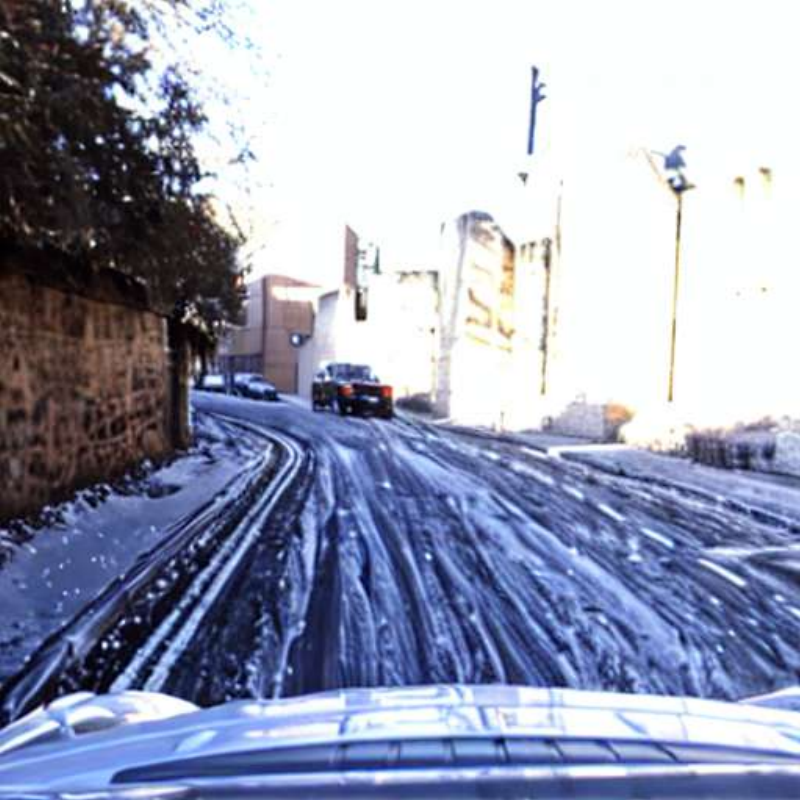} \\

\end{tabular}
\caption{Qualitative comparison of the visual prompt's reproduction induced by different encoders. Best viewed zoomed-in.
}
\label{abl_encoder_qualitative}
\vspace{-0.75em}
\end{figure}

\vspace{-1em}

\paragraph{Encoder Choice.}
Different encoders capture distinct information, influencing the learning process and output quality.
Other works commonly use DINOv2 and CLIP as evaluation metrics for generated sample quality \cite{emuedit, magicbrush, ip2p}.
However, DreamSim recently showed to outperform those encoders in alignment with human preferences.
To assess the effect of encoder choice, we train three versions of SPIE on the same task—editing sparse snow—systematically replacing the encoder in Eq. \ref{eq:loss_semantic}.
We focus on qualitative results, as semantic visual alignment metrics based on these same encoders lack impartiality.
We find in Fig. \ref{abl_encoder_qualitative} that the images generated by the DINOv2-guided model possess a grainy texture that is not present in the visual prompt.
Also, the CLIP-guided model reproduces excessively smooth and vaguely defined snow lanes compared to the visual prompt, with an unwanted purple tint across the frame.
Contrastingly, DreamSim better enforces the color and structure of the visual prompt.
It does not lead to learning spurious cues like an unrelated tint or saturated colors, and best reproduces the structure of the snow stripes.
This results in more realistic samples with stronger alignment to the prompt.

\section{Conclusion}
\label{sec:conclusion}

In this paper, we introduce SPIE, a novel approach to instruction-based image editing that enhances semantic alignment and structural preservation through few-steps post-training, effectively mimicking human preferences without direct feedback.
Our method demonstrates that these improvements can be achieved by leveraging AI-generated supervision, circumventing the need for extensive human annotations or large-scale datasets. 
SPIE learns to capture and reproduce intricate details in visual prompts, with only 5 examples per concept, further reducing the reliance on elaborate textual prompts.
Our approach significantly improves upon previous state-of-the-art methods in balancing faithfulness to the input image and alignment with instruction prompts, resulting in samples with higher perceived realism. 
The efficient finetuning approach with visual prompts enables complex sim-to-real edits using minimal reference images, demonstrating potential for high-quality simulated evaluation environments in robotics.

While our approach shows significant improvements, we acknowledge certain limitations.
The current method primarily excels at modifying textures and surfaces rather than altering global shapes. 
Future work could explore flexible constraints in masking and depth alignment operations, allowing for more substantial structural modifications like adding and removing elements.
The model may inherit biases from the pre-trained InstructPix2Pix model and the AI models providing the alignment supervision signal.
However, this limitation can be mitigated by substituting these components with suitable alternatives in a modular fashion.

We hope to inspire further research in online reinforcement learning for T2I models and additional studies on crafting AI-generated rewards for objectives that better align with human intent.

\clearpage
\section*{Acknowledgments}
\label{acknowledgments}

We would like to thank Sotiris Anagnostidis, Victor Miara and Kevin Kasa for insightful conversations, and Litu Rout for his generous feedback on this work.

{
    \small
    \bibliographystyle{ieeenat_fullname}
    \bibliography{main}
}

\clearpage
\onecolumn 
\section{Appendix}
\label{appendix}

\subsection{Derivation for Classifier-free Guidance with Three Conditionings}
\label{appendix_cfg}

We introduce separate guidance scales like InstructPix2Pix to enable separately trading off the strength of each conditioning. 
The modified score estimate for our model is derived as follows.
Our generative model learns $P(z \vert c_T,\; c_{I_\text{sty}},\; c_{I_\text{in}})$, which corresponds to the probability distribution of the image latents $z=\mathcal{E}(x)$ conditioned on an input image $c_{I_\text{in}}$, a reference style image $c_{I_\text{sty}}$, and a text instruction $c_T$. We arrive at our particular classifier-free guidance formulation by expressing the conditional probability as follows:

\begin{equation}
\scalebox{1}{$
    \begin{split}
    P(z \vert c_T,\; c_{I_\text{sty}},\; c_{I_\text{in}}) & = \frac{P(z, c_T,\; c_{I_\text{sty}},\; c_{I_\text{in}})}{P(c_T,\; c_{I_\text{sty}},\; c_{I_\text{in}})} \\
    & = \frac{P(c_T \vert c_{I_\text{sty}}, c_{I_\text{in}}, z) P(c_{I_\text{sty}} \vert c_{I_\text{in}}, z) P(c_{I_\text{in}} \vert z) P(z)}{P(c_T,\; c_{I_\text{sty}},\; c_{I_\text{in}})}\\
    \end{split}$}
\end{equation}

\noindent Diffusion models estimate the score \cite{hyvarinen05a} of the data distribution, i.e. the derivative of the log probability. Taking the logarithm of the expression above yields the following:

\begin{equation}
\scalebox{1}{$
    \begin{split}
    \log(P(z \vert c_T,\; c_{I_\text{sty}},\; c_{I_\text{in}})) = & \;\log(P(c_T \vert c_{I_\text{sty}}, c_{I_\text{in}}, z)) \\
                                                    & + \log(P(c_{I_\text{sty}} \vert c_{I_\text{in}}, z)) \\
                                                    & + \log(P(c_{I_\text{in}} \vert z)) \\
                                                    & + \log(P(z)) \\
                                                    & - \log(P(c_T,\; c_{I_\text{sty}},\; c_{I_\text{in}}))
    \end{split}$}
\end{equation}

\noindent Taking the derivative and rearranging, we obtain:
\begin{equation}
\scalebox{1}{$
    \begin{split}
    \nabla_{z}\log(P(z \vert c_T,\; c_{I_\text{sty}},\; c_{I_\text{in}})) = & \;\nabla_{z}\log(P(z)) \\
                                                                & + \nabla{z}\log(P(c_{I_\text{in}} \vert z)) \\
                                                                & + \nabla{z}\log(P(c_{I_\text{sty}} \vert c_{I_\text{in}}, z)) \\
                                                                & + \nabla_{z}\log(P(c_T \vert c_{I_\text{sty}}, c_{I_\text{in}}, z)) \\
    \end{split}$}
\end{equation}

\noindent This corresponds to the following formulation of classifier-free guidance, with three classifier-free-guidance scales:

\begin{equation}
\scalebox{1}{$
    \begin{aligned}
    \tilde{e}_{\theta}(z_t, c_{I_\text{in}}, c_{I_\text{sty}}, c_T) &= e_{\theta}(z_t, \varnothing, \varnothing, \varnothing) \\
                                                                   &\quad + s_{I_\text{in}} \cdot \left(e_{\theta}(z_t, c_{I_\text{in}}, \varnothing, \varnothing) - e_{\theta}(z_t, \varnothing, \varnothing, \varnothing)\right) \\
                                                                   &\quad + s_{I_\text{sty}} \cdot \left(e_{\theta}(z_t, c_{I_\text{in}}, c_{I_\text{sty}}, \varnothing) - e_{\theta}(z_t, c_{I_\text{in}}, \varnothing, \varnothing)\right) \\
                                                                   &\quad + s_T \cdot \left(e_{\theta}(z_t, c_{I_\text{in}}, c_{I_\text{sty}}, c_T) - e_{\theta}(z_t, c_{I_\text{in}}, c_{I_\text{sty}}, \varnothing)\right)
    \end{aligned}$}
\label{final_cfg_eq}
\end{equation}

\subsection{Training Details}
\label{appendix_trainingdetails}
In training, we initialize our model from the InstructPix2Pix checkpoint. Depending on the specialization, we train as few as six steps at a resolution of $256 \times 256$, with a total batch size of 256 samples. For a fair comparison, we maintain the same RL training hyperparameters as in D3PO \cite{d3po} without conducting hyperparameter optimization. Additionally, we do not optimize for the best text prompt for each edit type, as we design SPIE to leverage the visual prompt as the primary carrier of semantic information, minimizing reliance on extensive textual descriptions.

The selection of hyperparameters $\lambda$ and $\alpha$ requires balancing multiple objectives for optimal performance. While our recommended values work well across most scenarios, these parameters can be slightly fine-tuned based on the model's zero-shot capabilities to further optimize learning convergence. This flexibility allows practitioners to adapt the framework to specific task requirements while maintaining robust performance.

\subsection{Real Image Experiment Details}
\label{real_eval_details}

Our evaluations focus on the ability to perform localized edits in complex scenes containing multi-level information, including local objects, global layout, and background environments that must remain unchanged unless explicitly instructed. We use two datasets for our experiments.

With the Oxford RobotCar Dataset \cite{RobotCarDatasetIJRR}, we train SPIE to perform seven types of edits: adding dense snow on the road, adding sparse snow on the road, adding rain on the road, adding sand on the road, changing the road to gold, changing the road to wood, and changing the entire scene to nighttime (using a full-frame mask).

With the Places Dataset \cite{places}, we train our model on six additional edit types: changing water to gold, changing a bed to leather, changing a building facade to steel, changing a telephone booth to stone, changing a lighthouse to terracotta, changing a train to wood.

When selecting edit types for our experiments, we aimed to cover a diverse range of materials, textures, and environmental modifications to demonstrate the versatility and robustness of our approach.
This diverse selection allows us to assess our method's performance across both realistic modifications (weather changes) and more stylistic transformations (material changes), while testing its ability to maintain structural coherence across different scene types, object scales, and edit complexities.

To prevent overfitting on spurious cues, we alternate between five conditioning style images related to the same text prompt during training. Our evaluations cover 29,500 images at a resolution of $512 \times 512$ across both datasets and edit types (2,500 images for each of the seven Oxford RobotCar edits and 2,000 images for each of the six Places edits).

For baseline comparisons, we evaluate the InstructPix2Pix checkpoint from which we initialize our model, and the best publicly available versions of HIVE, HQ-Edit, and MagicBrush based on StableDiffusion v1.5.
Some evaluation results on standardized benchmarks were taken directly from the ones reported in the original MagicBrush \cite{magicbrush} and Emu Edit \cite{emuedit} papers.

While our model is trained at $256 \times 256$ resolution, we find it generalizes well to $512 \times 512$ resolution at inference time. We generate qualitative results at $512 \times 512$ resolution with 100 denoising steps using an Euler ancestral sampler with denoising variance schedule proposed by \citet{eulerancestral}. Editing an image with our model takes approximately 9 seconds on an A100 GPU.

\subsection{Sim-to-real Experiment Details}
\label{appendix_experiment_RTX_details}

\begin{table}[h!]
    \centering
    \scalebox{0.78}{
    \begin{tabular}{llccc}
        \toprule
        \multirow{2.5}{*}{\begin{tabular}[c]{@{}l@{}}Google Robot \\ Evaluation Setup\end{tabular}} & \multirow{2.5}{*}{Policy} & \multicolumn{3}{c}{Open / Close Drawer} \\
        \cmidrule(lr){3-5}
        & & Open & Close & Average \\
        \midrule
        \multirow{4}{*}{Real Eval}
        & RT-1 (Converged) & 0.815 & 0.926 & 0.870 \\
        & RT-1 (15\%)      & 0.704 & 0.889 & 0.796 \\
        & RT-1-X           & 0.519 & 0.741 & 0.630 \\
        & RT-1 (Begin)     & 0.000 & 0.000 & 0.000 \\
        \midrule
        \multirow{6}{*}{\begin{tabular}[c]{@{}l@{}}SIMPLER Eval \\ (Variant Aggregation)\end{tabular}}        
        & RT-1 (Converged) & 0.270 & 0.376 & 0.323 \\
        & RT-1 (15\%)      & 0.212 & 0.323 & 0.267 \\
        & RT-1-X           & 0.069 & 0.519 & 0.294 \\
        & RT-1 (Begin)     & 0.005 & 0.132 & 0.069 \\
        \cmidrule{2-5}
        & \textbf{MMRV$\downarrow$}      & 0.000 & 0.130 & 0.083 \\
        & \textbf{Pearson $r$$\uparrow$} & 0.915 & 0.756 & 0.964 \\
        \midrule
        \multirow{6}{*}{\begin{tabular}[c]{@{}l@{}}SIMPLER Eval \\ (Visual Matching)\end{tabular}}
        & RT-1 (Converged) & 0.601 & 0.861 & 0.730 \\
        & RT-1 (15\%)      & 0.463 & 0.667 & 0.565 \\
        & RT-1-X           & 0.296 & 0.891 & 0.597 \\
        & RT-1 (Begin)     & 0.000 & 0.278 & 0.139 \\
        \cmidrule{2-5}
        & \textbf{MMRV$\downarrow$}      & 0.000 & 0.130 & 0.083 \\
        & \textbf{Pearson $r$$\uparrow$} & 0.987 & 0.891 & 0.972 \\
        \midrule
        \multirow{6}{*}{\textbf{SPIE}}
        & RT-1 (Converged) & 0.471 & 0.810 & 0.640 \\
        & RT-1 (15\%)      & 0.259 & 0.619 & 0.439 \\
        & RT-1-X           & 0.180 & 0.608 & 0.394 \\
        & RT-1 (Begin)     & 0.021 & 0.058 & 0.040 \\
        \cmidrule{2-5}
        & \textbf{MMRV$\downarrow$} & 0.000 & 0.000 & 0.000 \\
        & \textbf{Pearson $r$$\uparrow$} & 0.917 & 0.978 & 0.966 \\
        \bottomrule
    \end{tabular}
    }
    \caption{
    Real-world, standard SIMPLER environment, and our visually-edited environment evaluation results across different policies on the Google Robot ``(open/close) (top/middle/bottom) drawer" task. We present success rates for the `Variant Aggregation' and `Visual Matching' approaches, as well as our novel visually-edited environments with seven material styles. We calculate the Mean Maximum Rank Violation (`MMRV', lower is better) and the Pearson correlation coefficient (`Pearson \textit{r}', higher is better) to assess the alignment between simulation and real-world relative performances across different policies.
    }
    \label{appdx_rtx_full}
\end{table}

\noindent In this section, we provide detailed descriptions of our robotics experiments using the SIMPLER environments.
We follow the same evaluation protocol as established in SIMPLER \cite{simpler}, focusing on a language-conditioned drawer manipulation task where the robot must ``(open/close) (top/middle/bottom) drawer." The robot is positioned in front of a cabinet with three drawers and must manipulate the specified drawer according to the instruction. For our simulation experiments, we also place the robot at 9 different grid positions within a rectangular area on the floor, resulting in 9 × 3 × 2 = 54 total trials.

Following SIMPLER, we conduct experiments on RT-1 checkpoints at various training stages: RT-1-X, RT-1 trained to convergence (RT-1 Converged), RT-1 at 15\% of training steps (RT-1 15\%), and RT-1 at the beginning of training (RT-1 Begin).

We train our model to modify the cabinet's material using seven different styles:
gold, leather, white marble, black marble, steel, stone, and wood.
Importantly, we only modify the visual appearance of the cabinet without changing any physical properties such as friction coefficients, material density, center of mass, or static and dynamic friction.
Since our method involves a non-deterministic diffusion process, we extend the SIMPLER protocol by averaging success rates across three different random seeds and across the seven different edit styles to produce lower-variance performance estimates.

For the VisMatch, VarAgg and Real evaluation results presented in Table \ref{appdx_rtx_full}, we directly reference the values reported in SIMPLER.

\subsection{Additional Qualitative Results}
\label{appendix_more_quali}

\begin{figure*}[h]
\centering
\newcommand{\imgwidth}{0.16}
\setlength{\tabcolsep}{2pt} 
\renewcommand{\arraystretch}{1} 
\newcommand{\textspace}{0.8em}

\begin{tabular}{cccccc} 
    Input & InstructPix2Pix & MagicBrush & HQ-Edit & HIVE & \textbf{SPIE} \\
    \includegraphics[width=\imgwidth\textwidth]{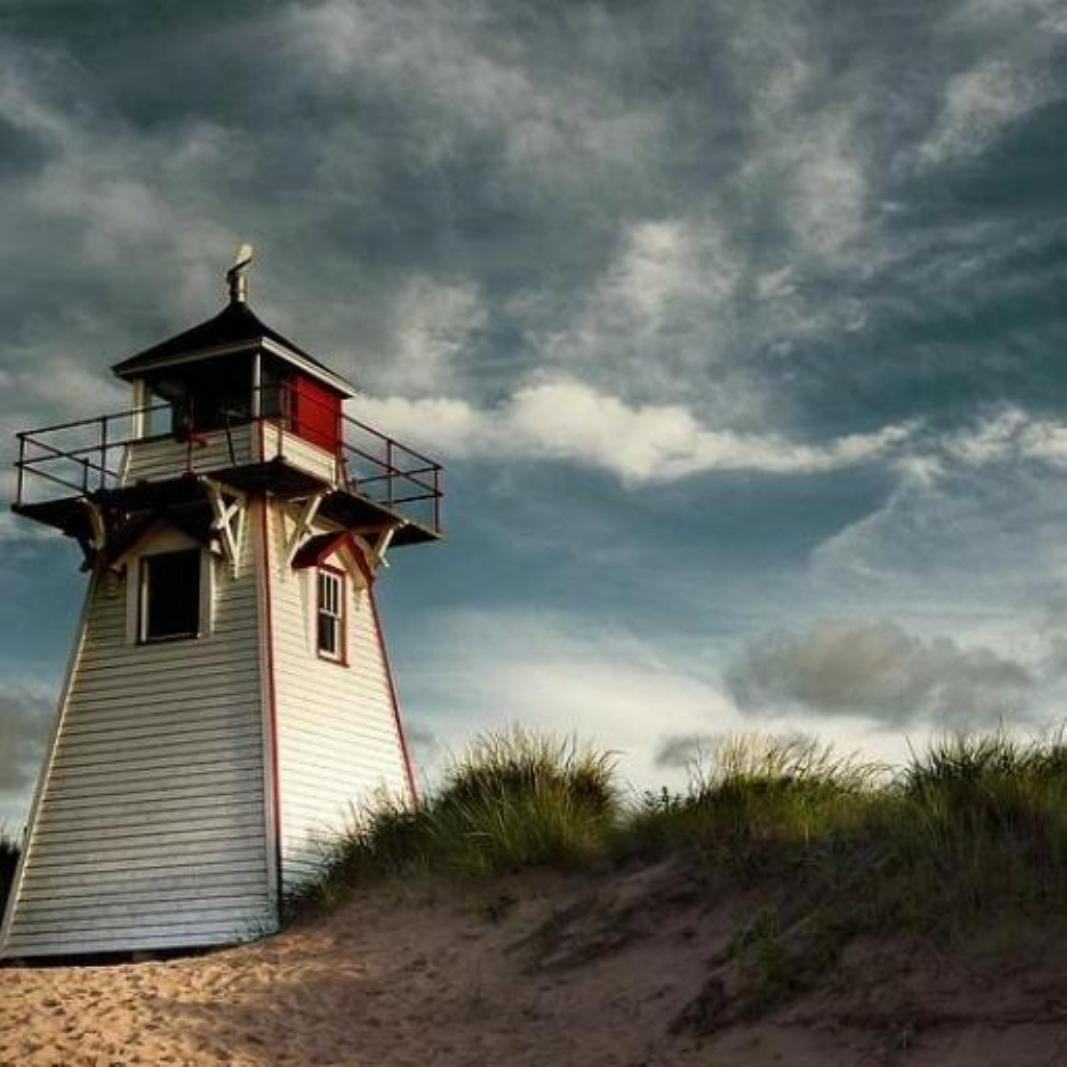} &
    \includegraphics[width=\imgwidth\textwidth]{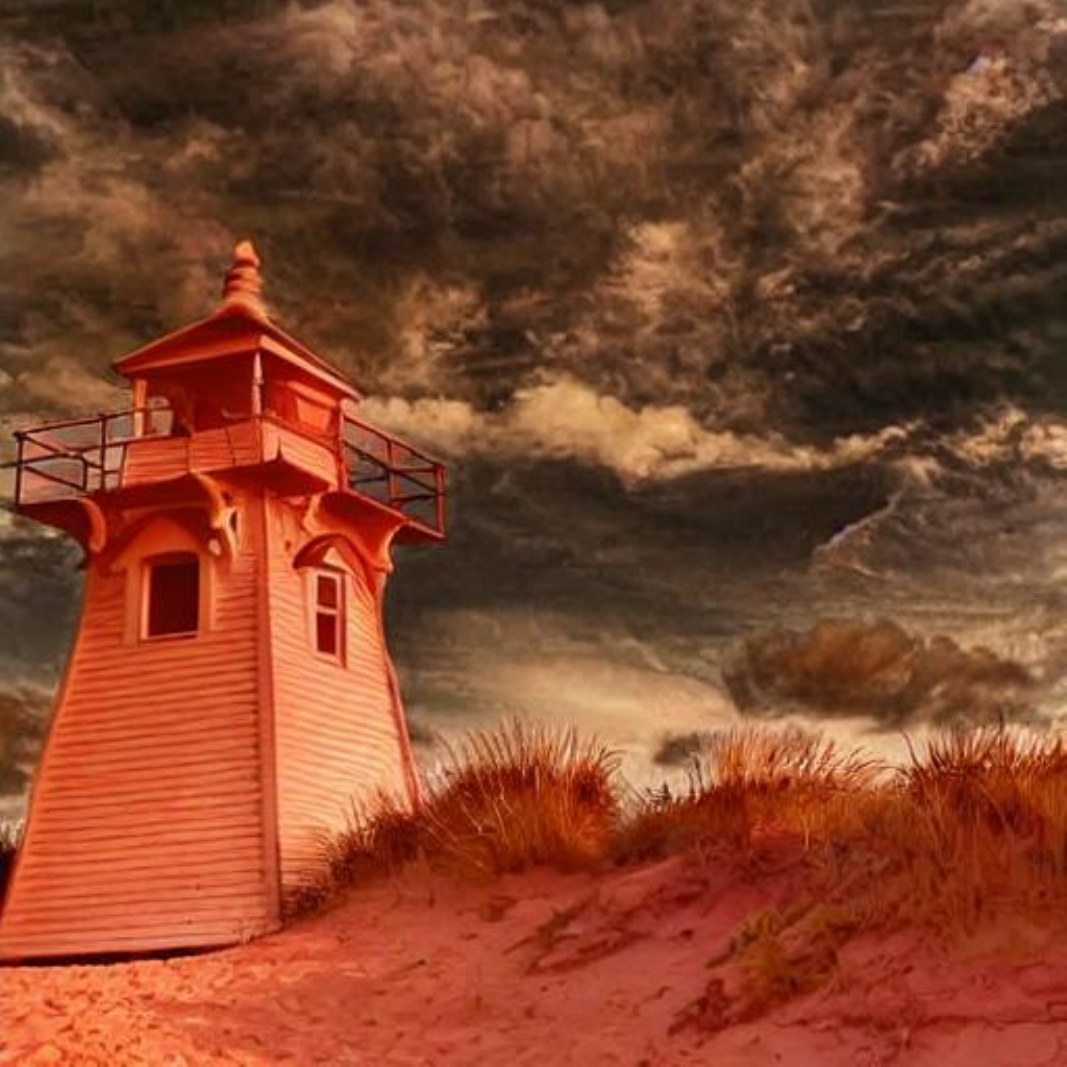} &
    \includegraphics[width=\imgwidth\textwidth]{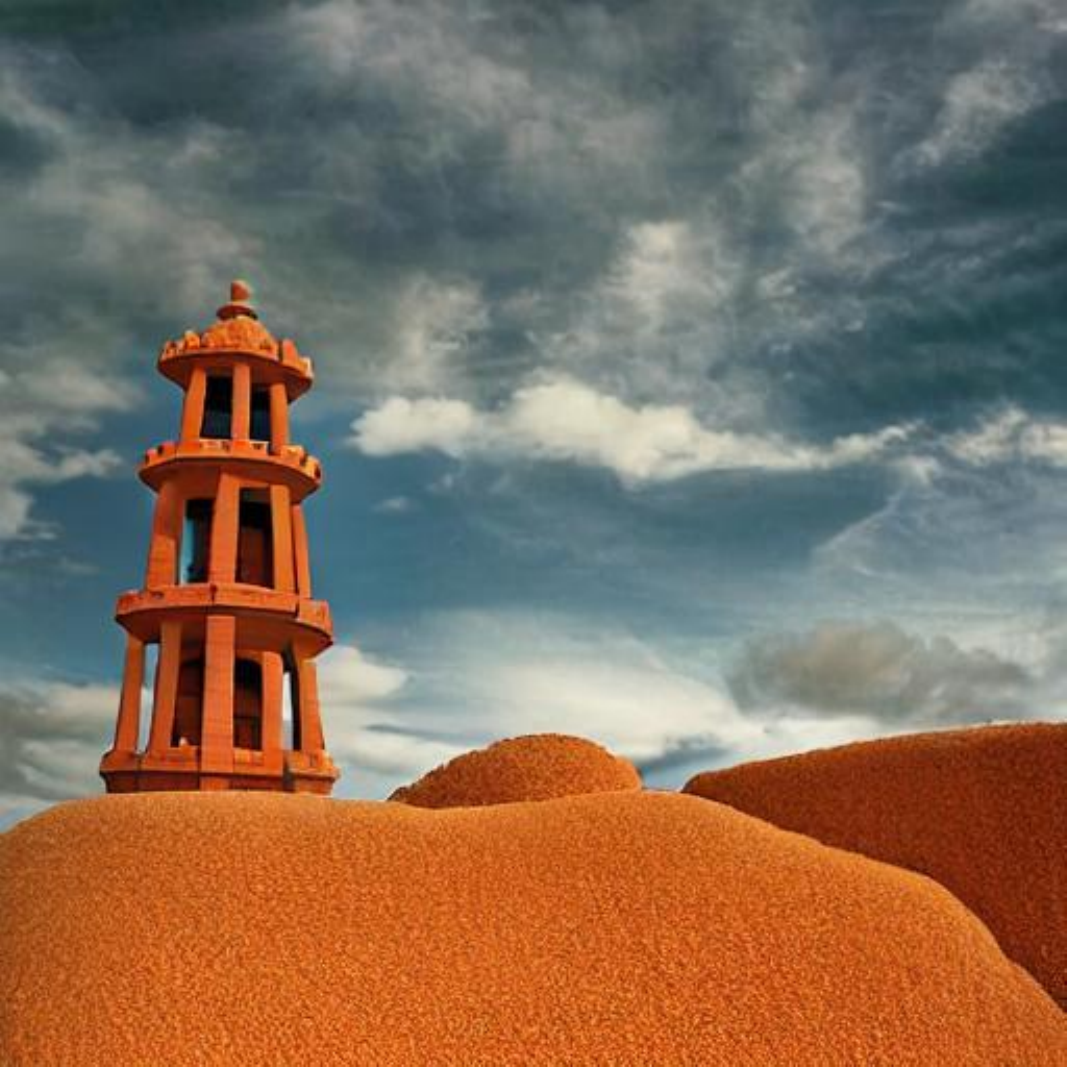}&
    \includegraphics[width=\imgwidth\textwidth]{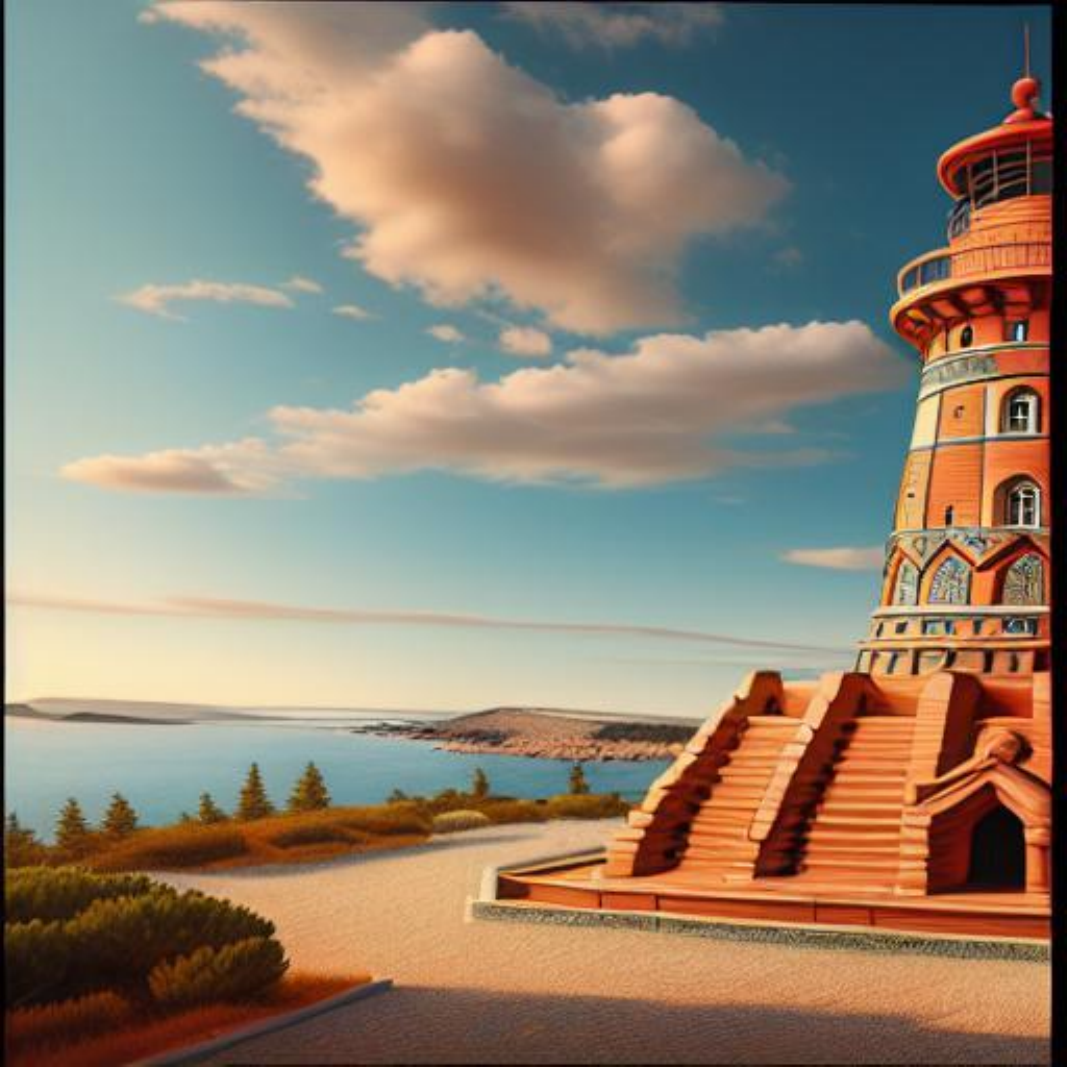}&
    \includegraphics[width=\imgwidth\textwidth]{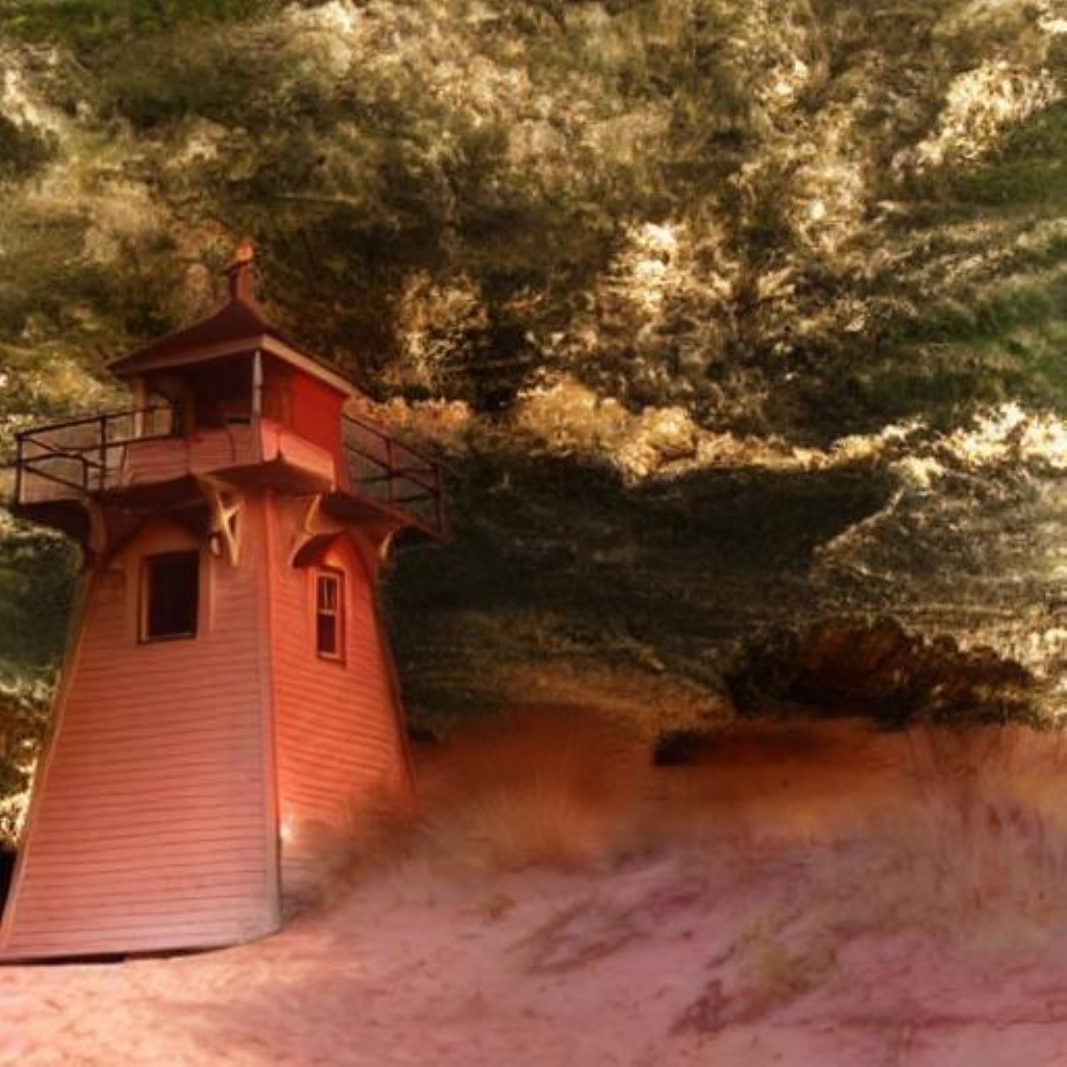} &
    \includegraphics[width=\imgwidth\textwidth]{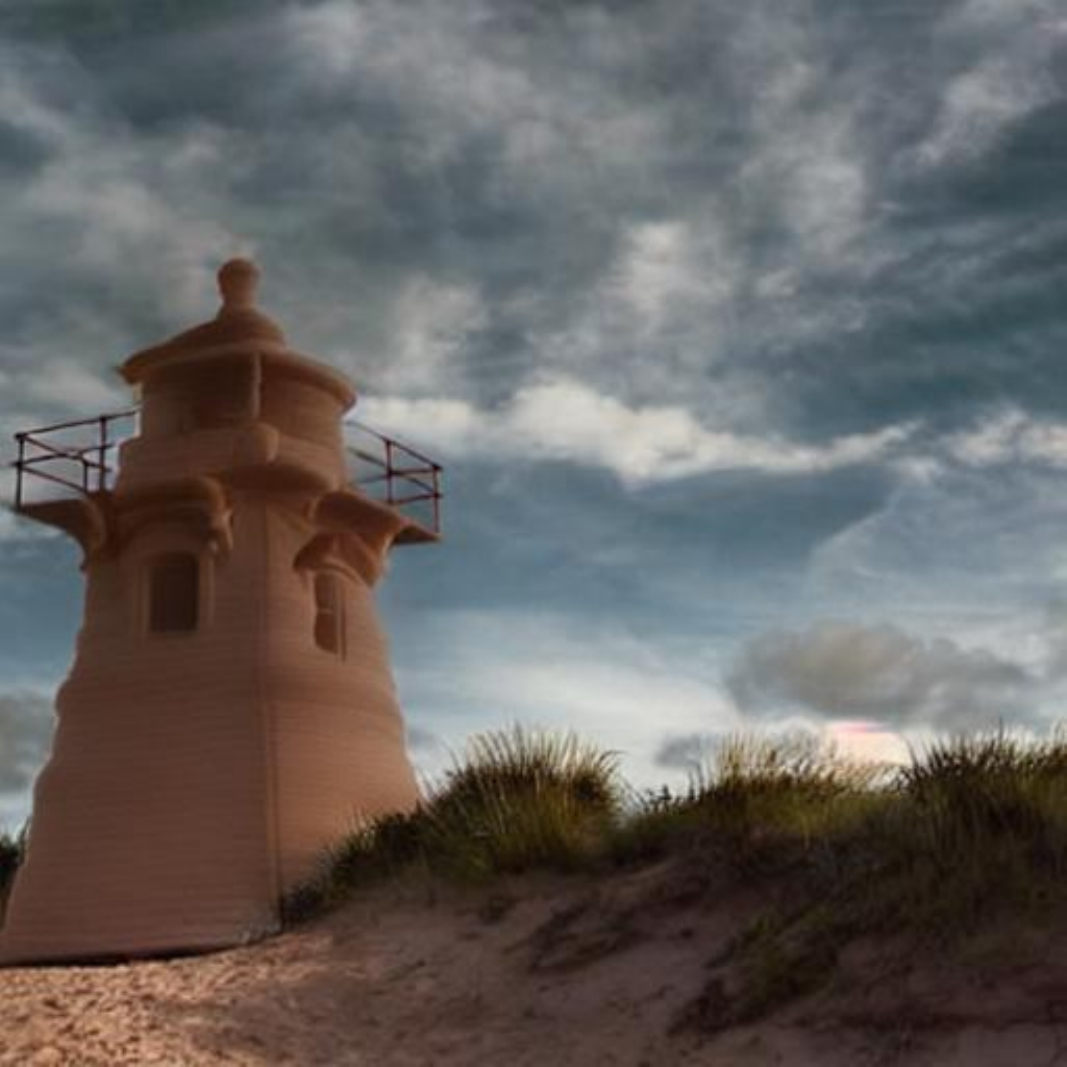} \\
    \multicolumn{6}{c}{``Change the lighthouse into terracotta"} \\[\textspace]

    \includegraphics[width=\imgwidth\textwidth]{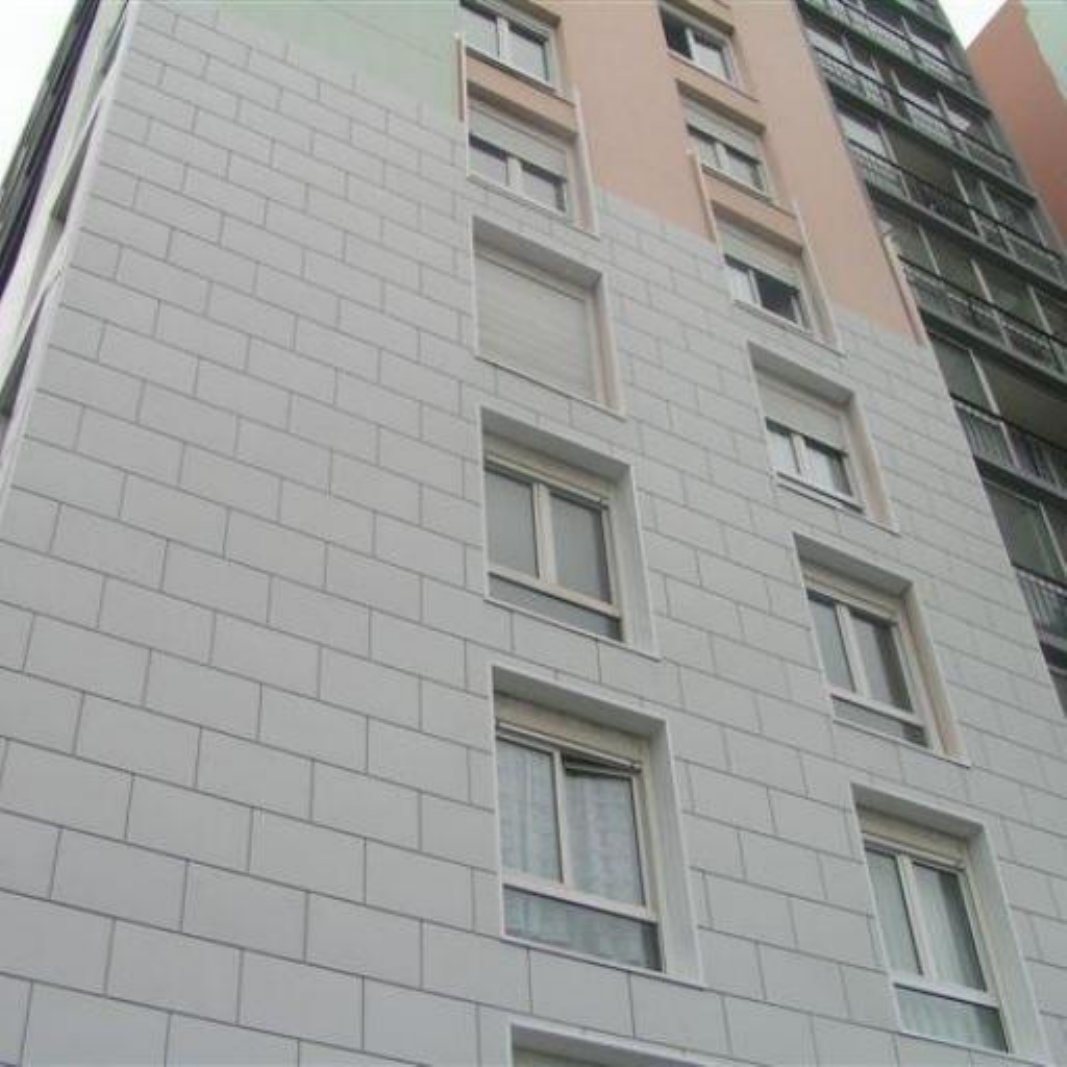} &
    \includegraphics[width=\imgwidth\textwidth]{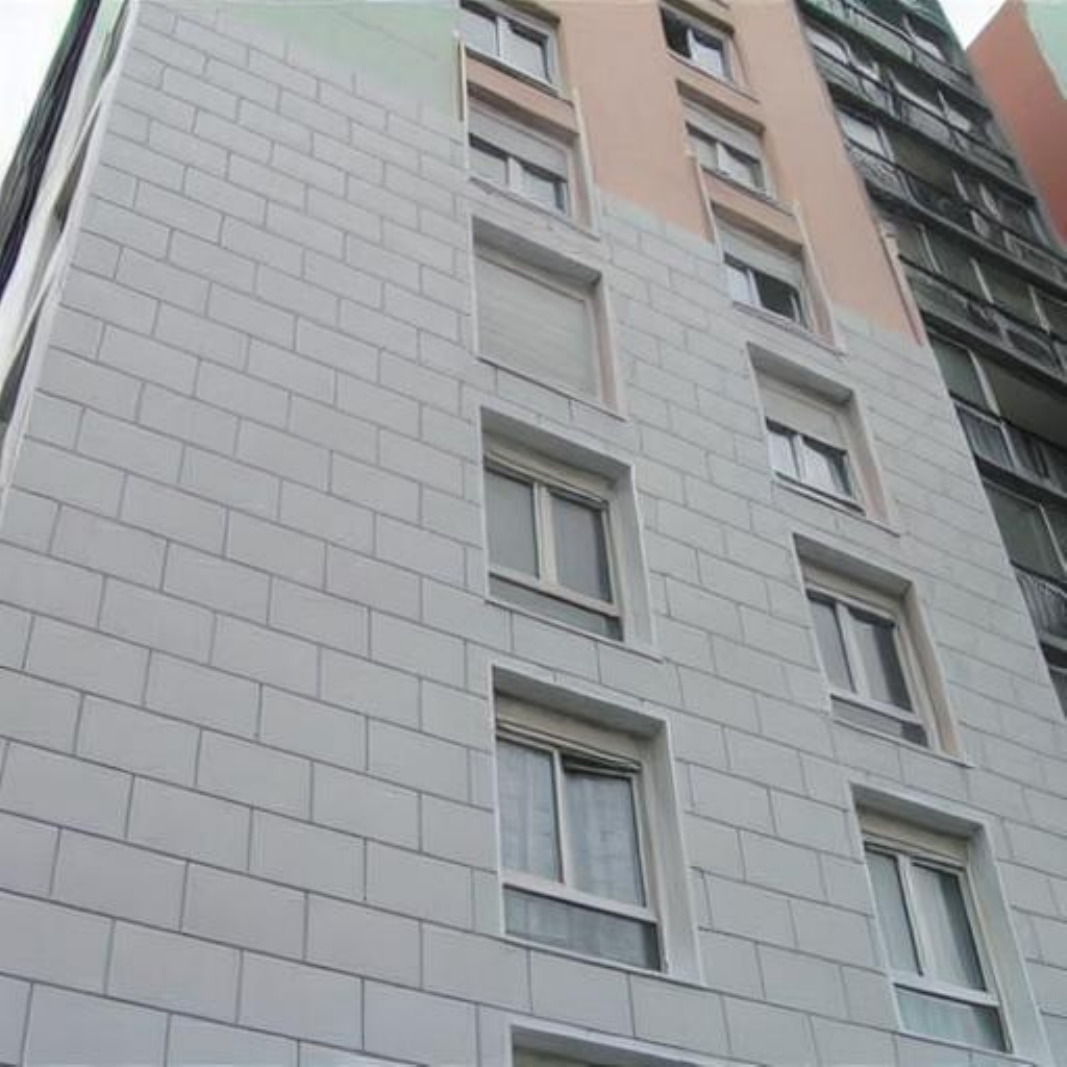} &
    \includegraphics[width=\imgwidth\textwidth]{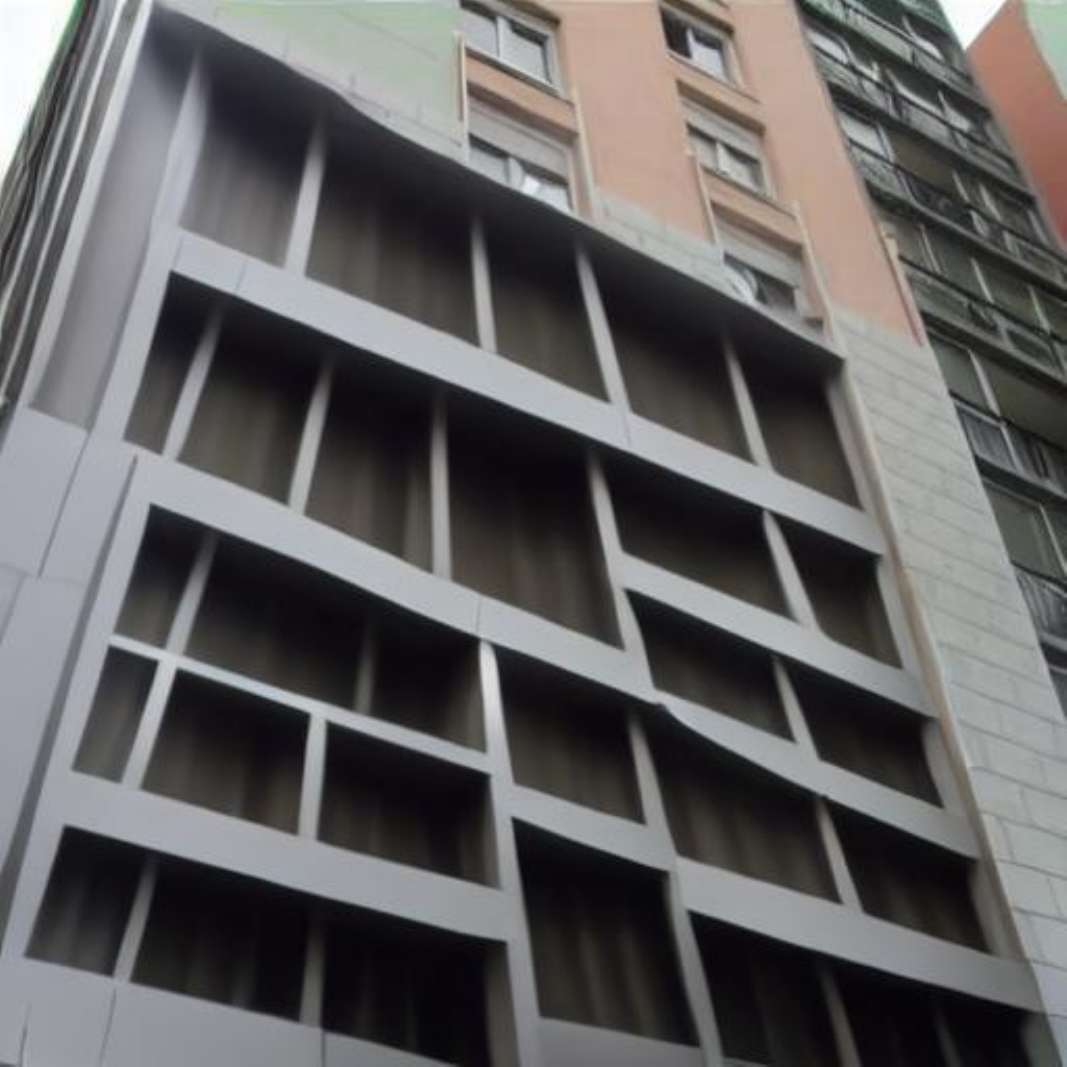}&
    \includegraphics[width=\imgwidth\textwidth]{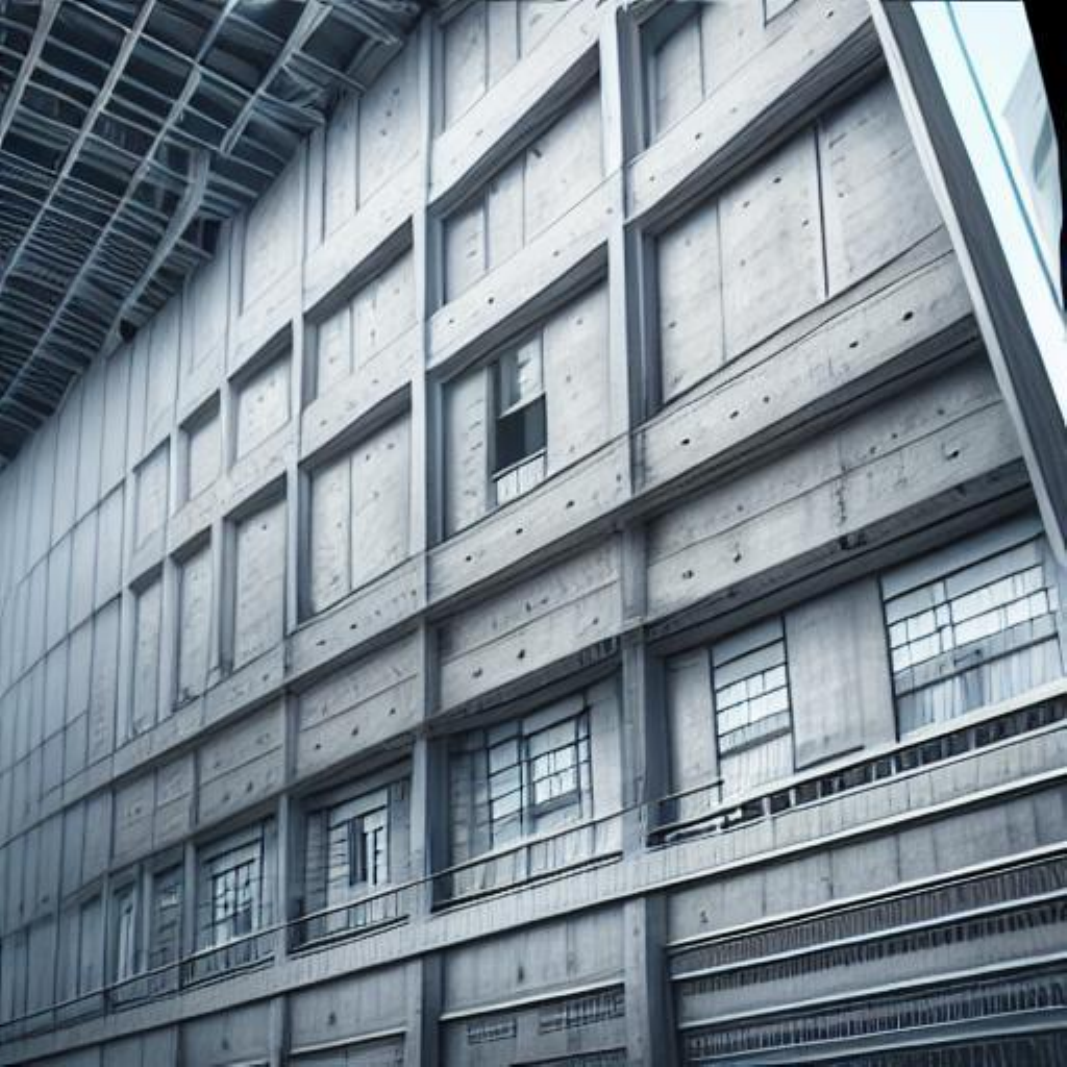}&
    \includegraphics[width=\imgwidth\textwidth]{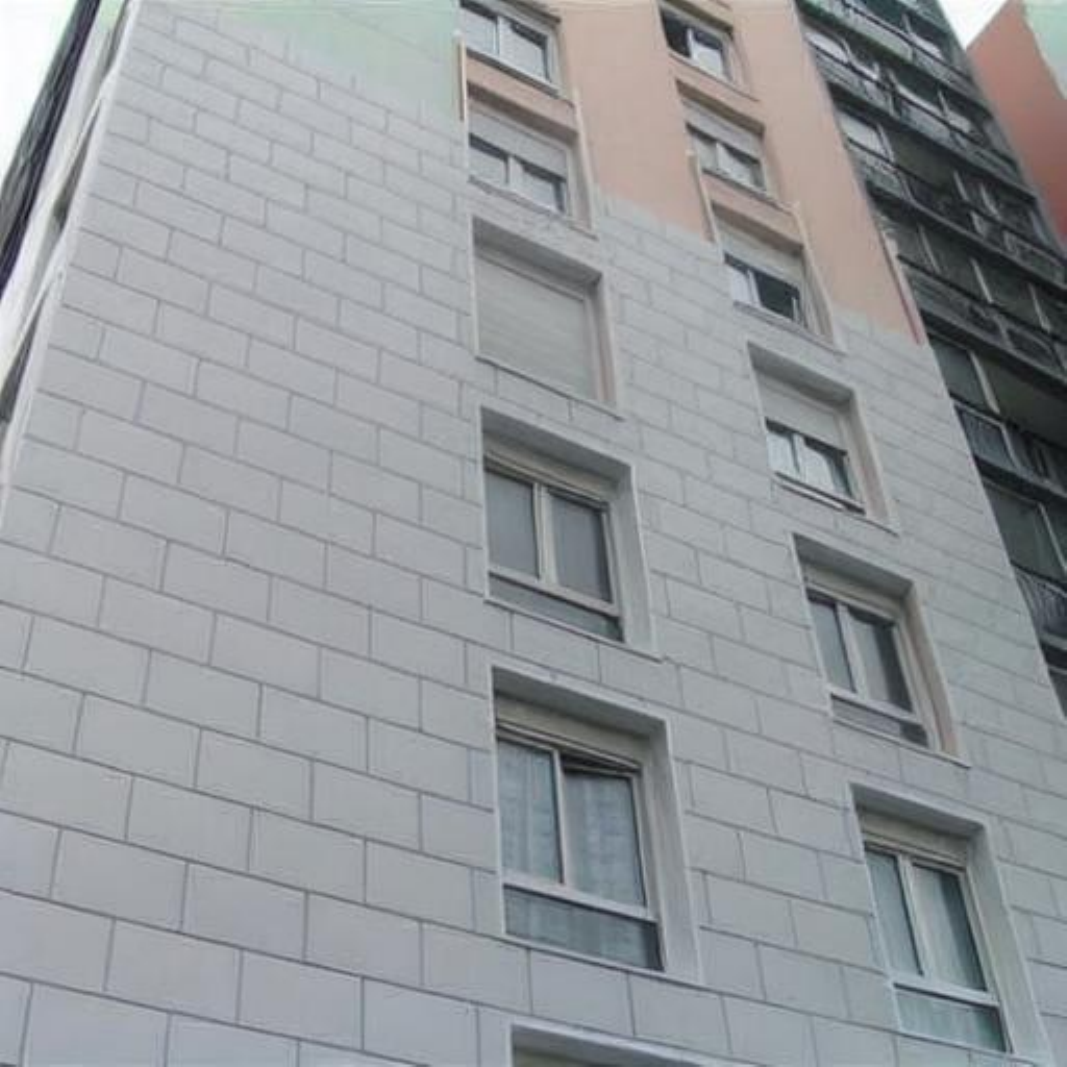}&
    \includegraphics[width=\imgwidth\textwidth]{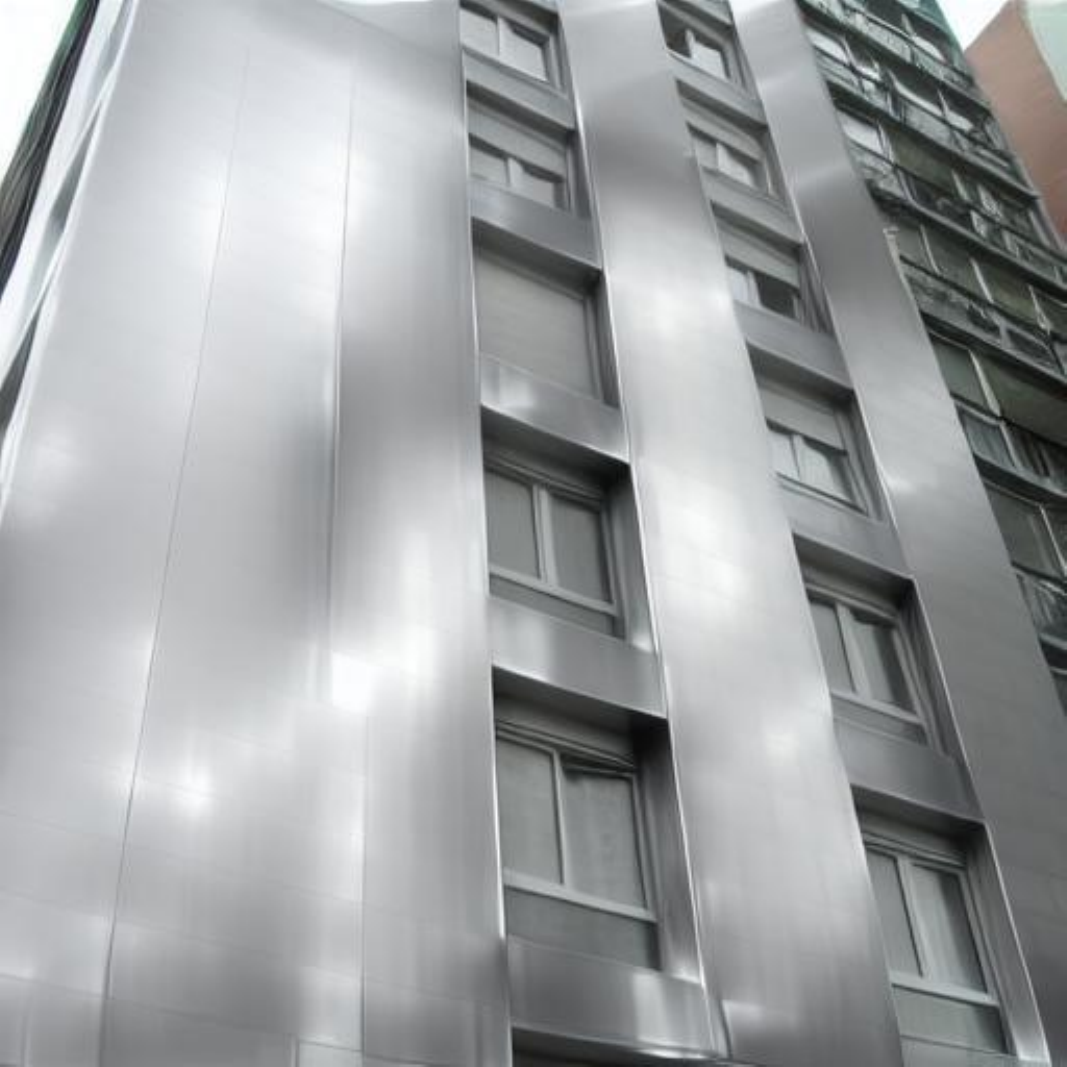} \\
    \multicolumn{6}{c}{``Turn the building into steel"} \\[\textspace]

    \includegraphics[width=\imgwidth\textwidth]{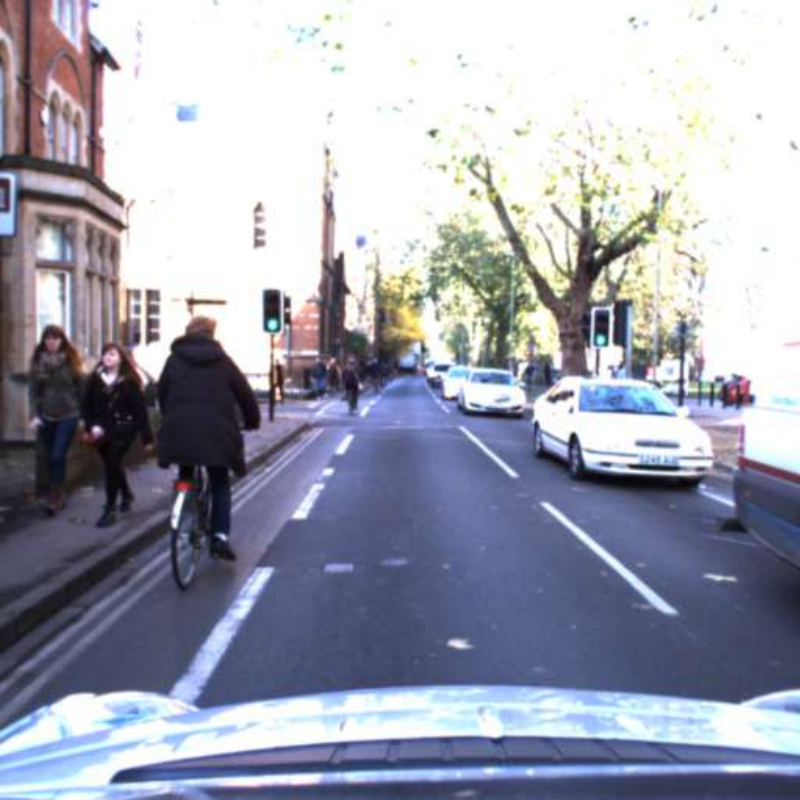} &
    \includegraphics[width=\imgwidth\textwidth]{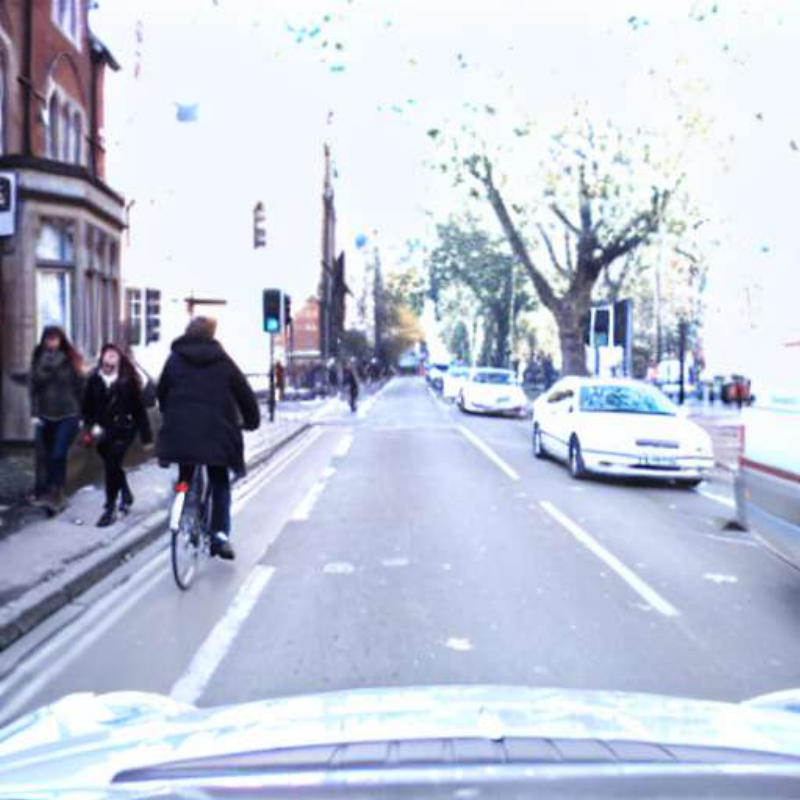} &
    \includegraphics[width=\imgwidth\textwidth]{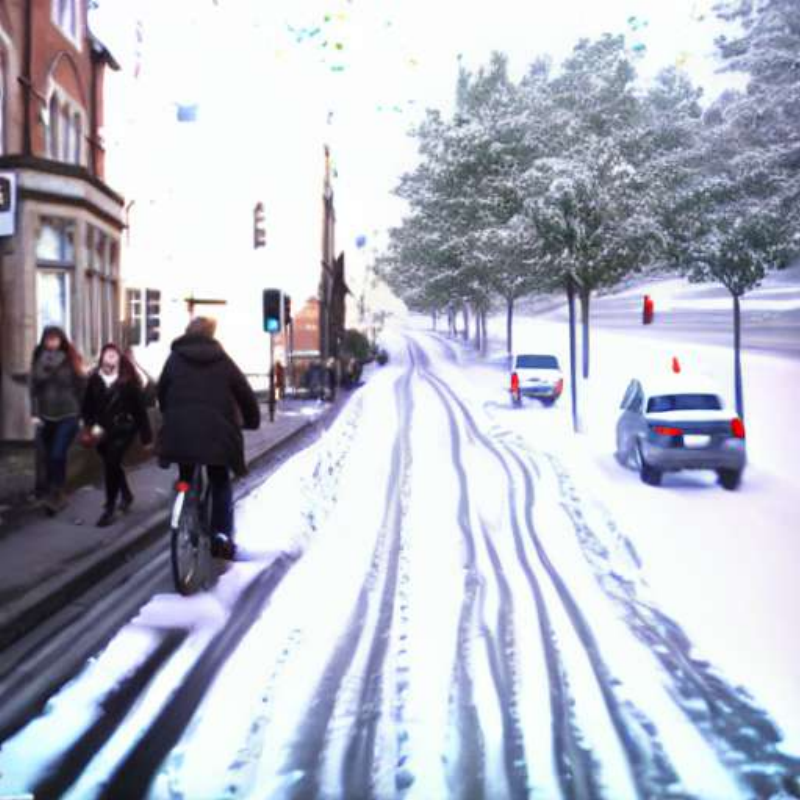} &
    \includegraphics[width=\imgwidth\textwidth]{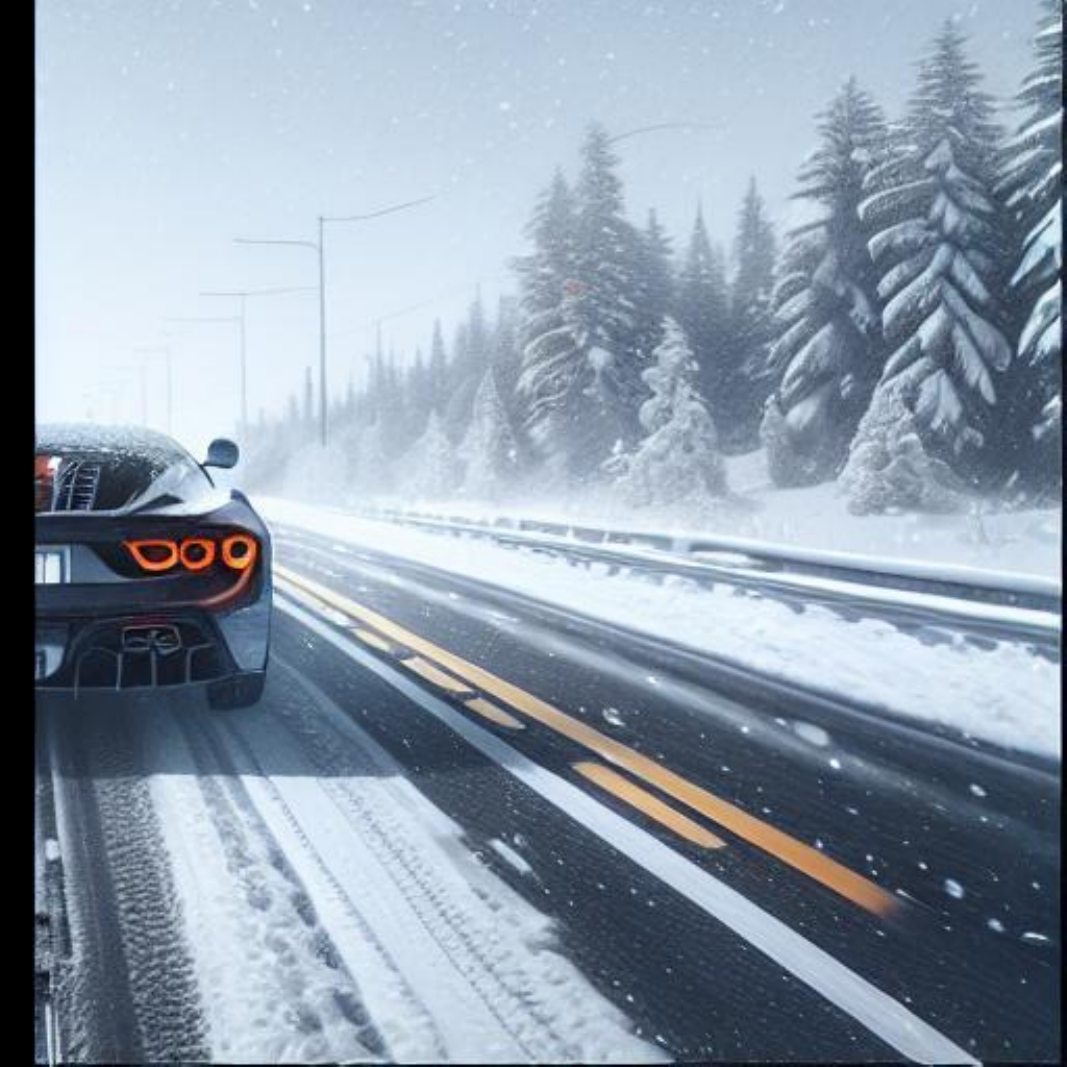} &
    \includegraphics[width=\imgwidth\textwidth]{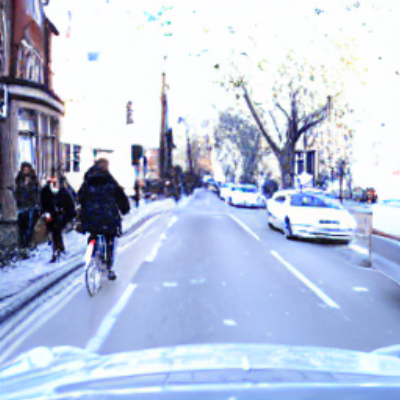} &
    \includegraphics[width=\imgwidth\textwidth]{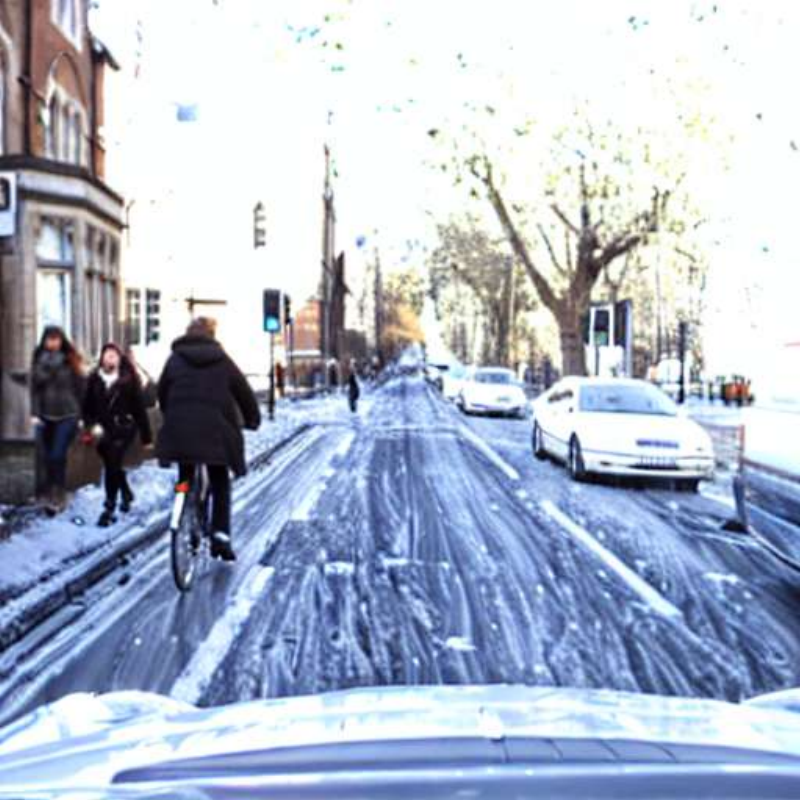} \\
    \multicolumn{6}{c}{``Add snow on the road"} \\[\textspace]

    \includegraphics[width=\imgwidth\textwidth]{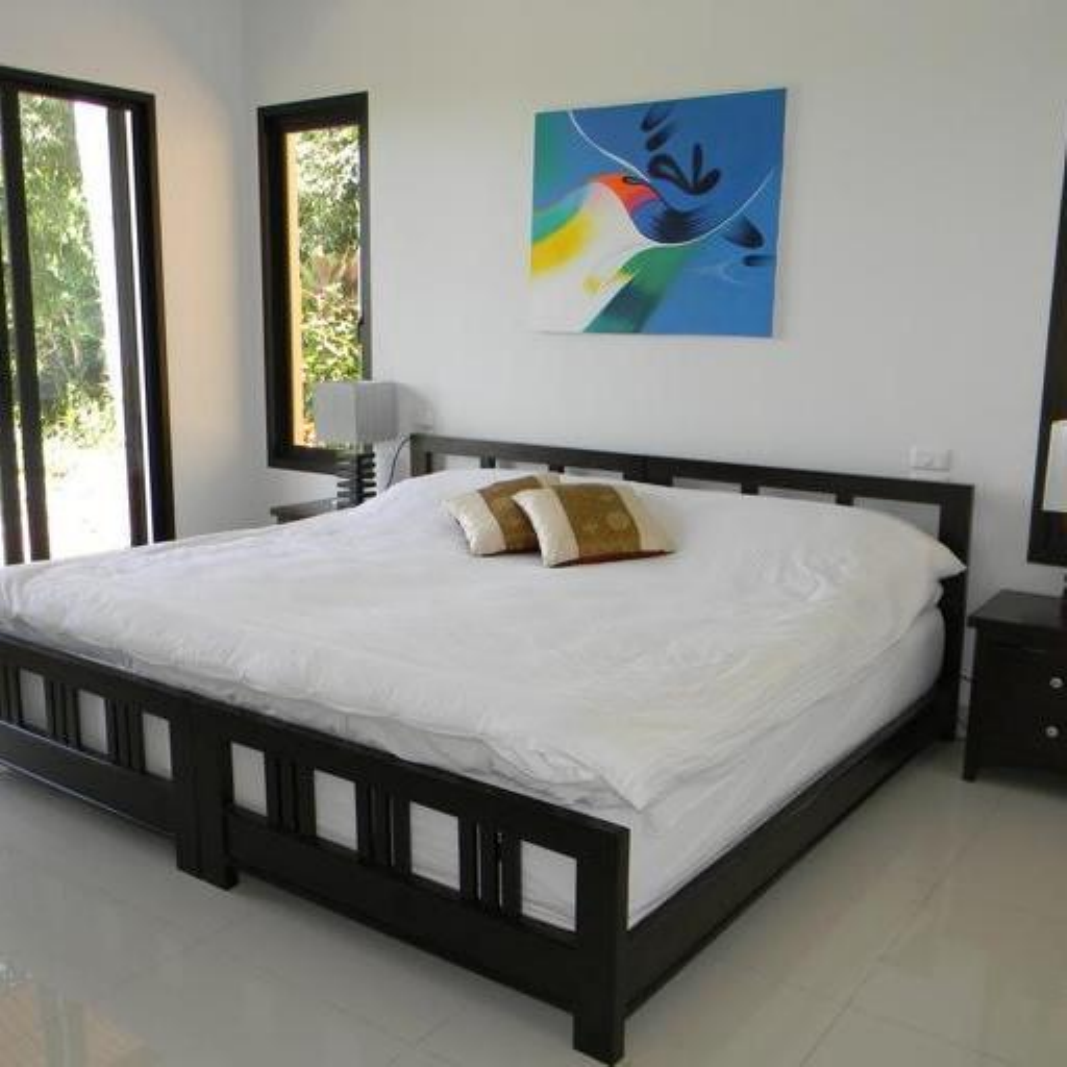} &
    \includegraphics[width=\imgwidth\textwidth]{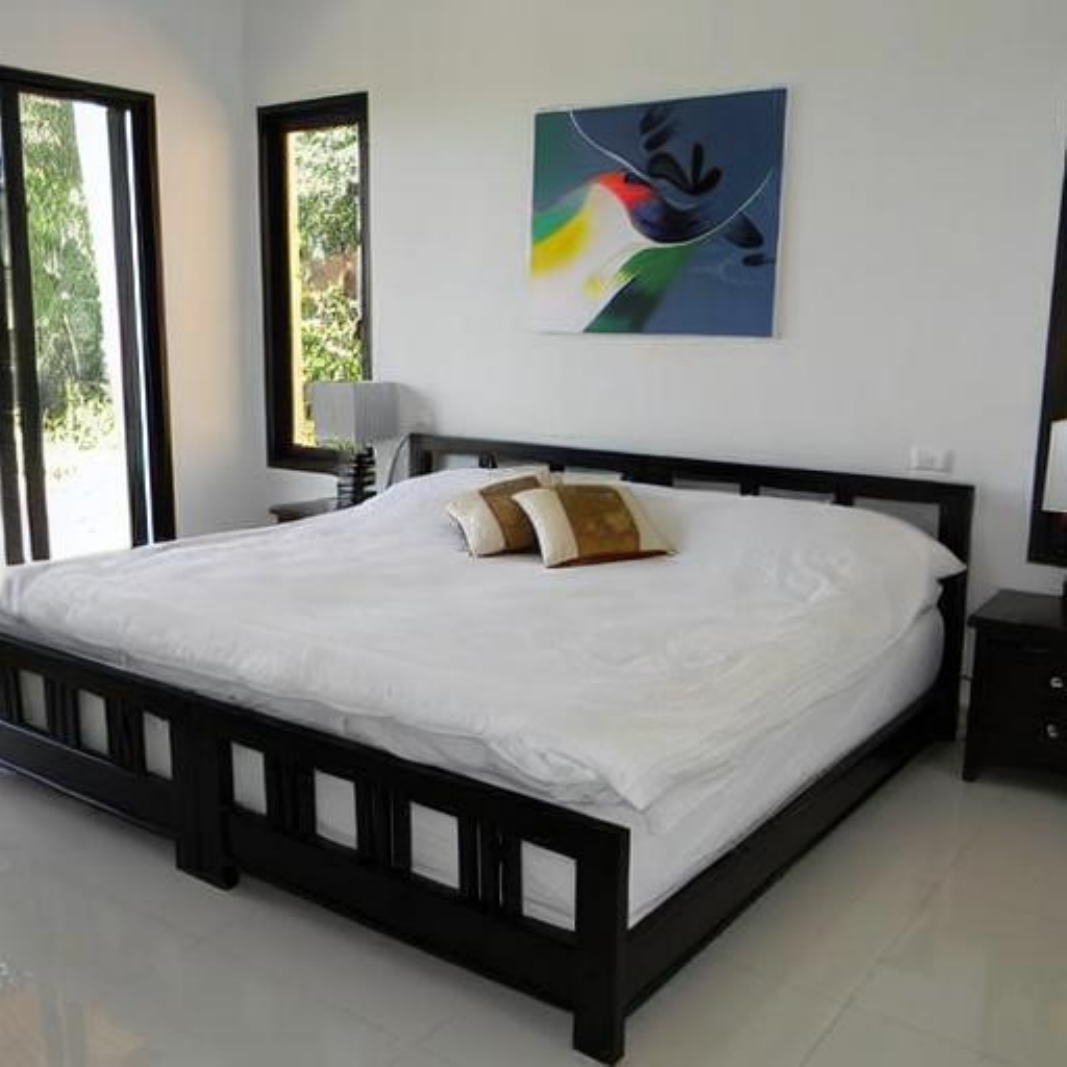} &
    \includegraphics[width=\imgwidth\textwidth]{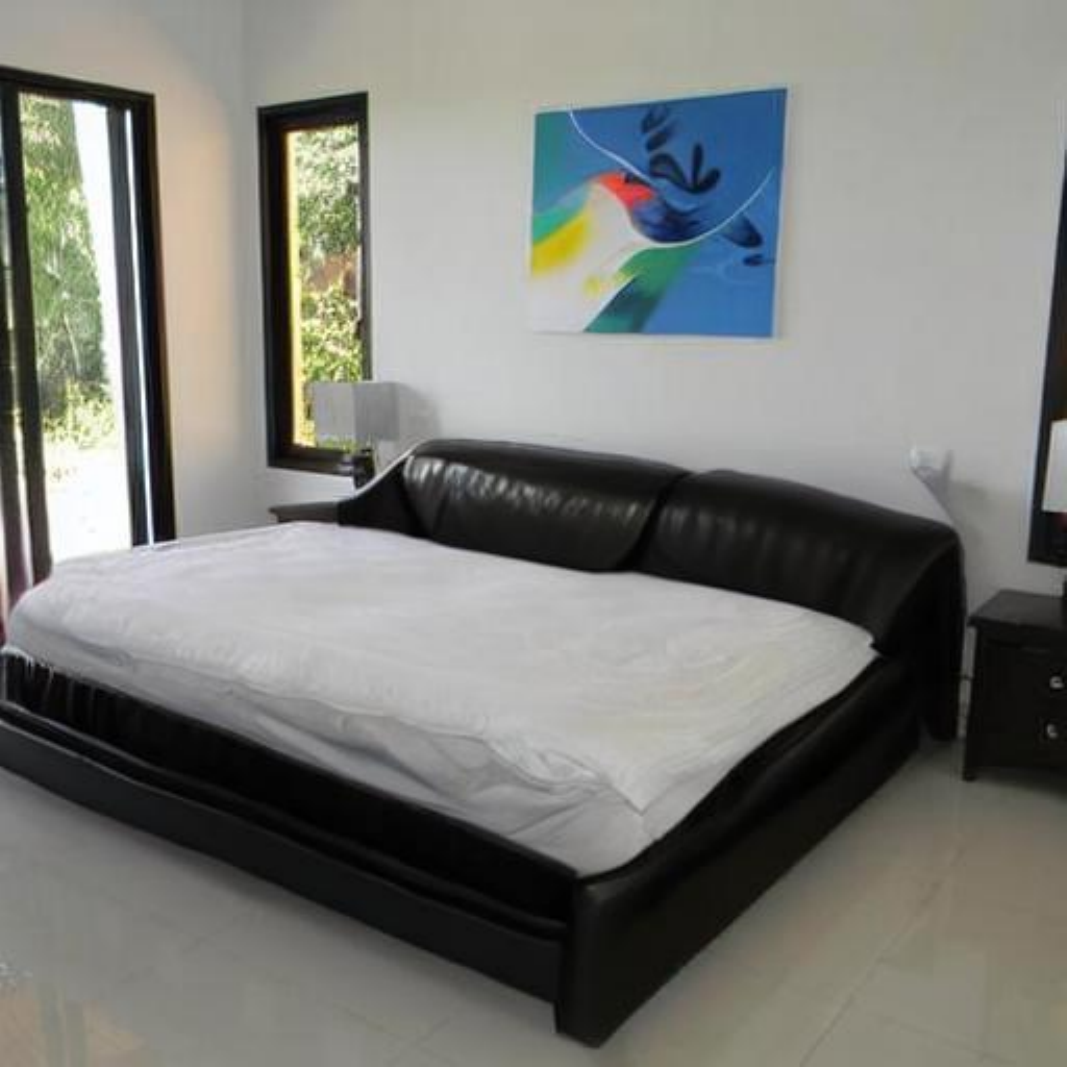}&
    \includegraphics[width=\imgwidth\textwidth]{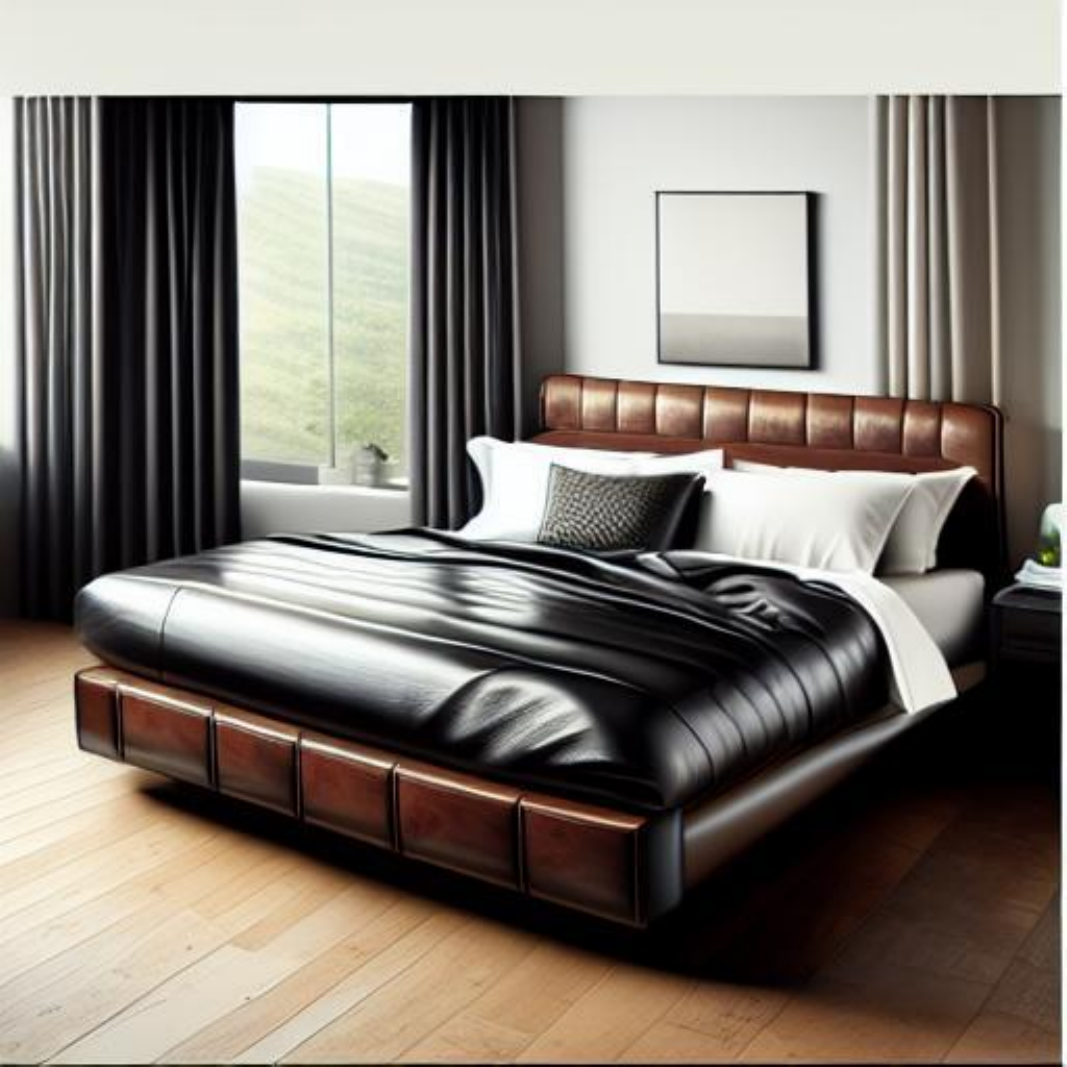}&
    \includegraphics[width=\imgwidth\textwidth]{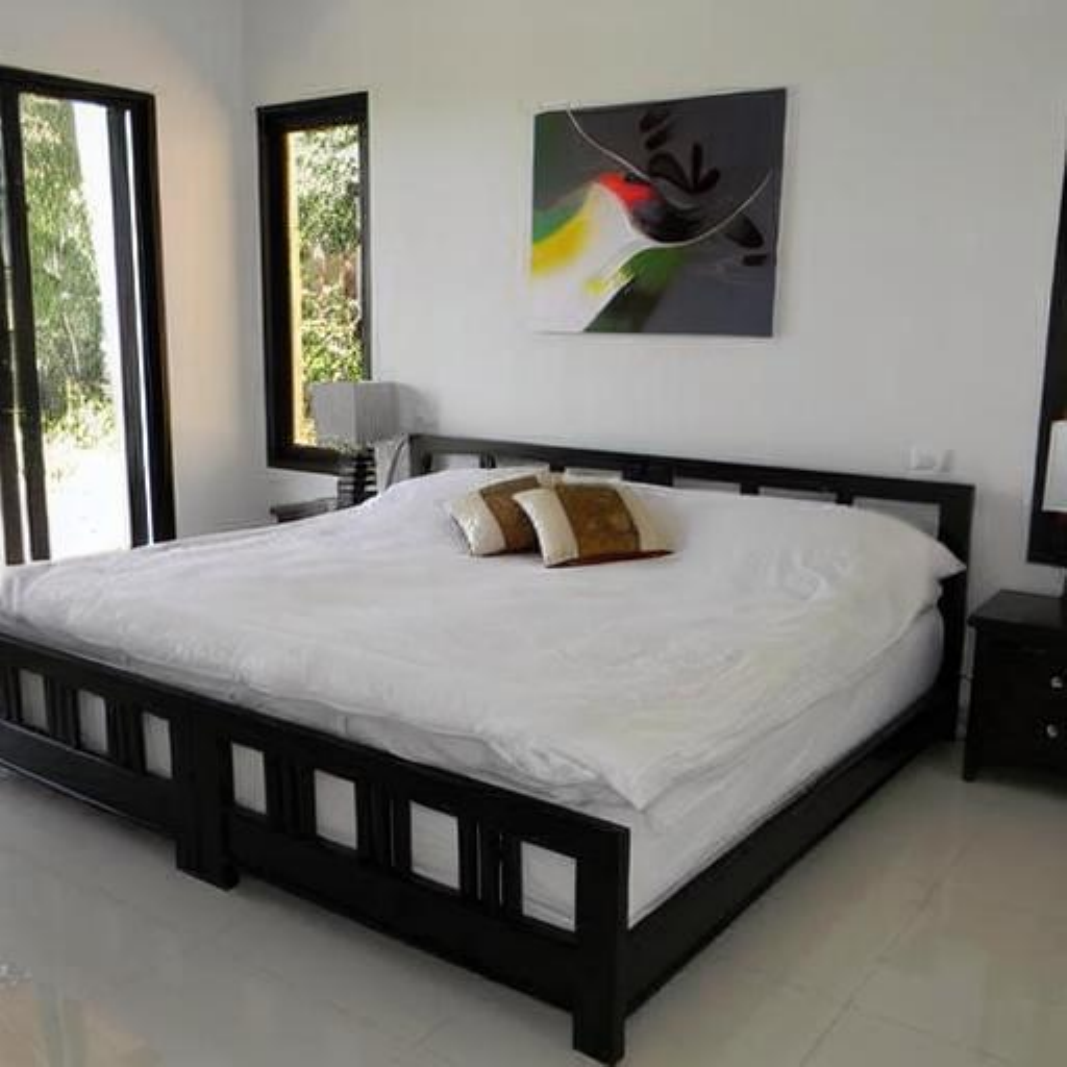}&
    \includegraphics[width=\imgwidth\textwidth]{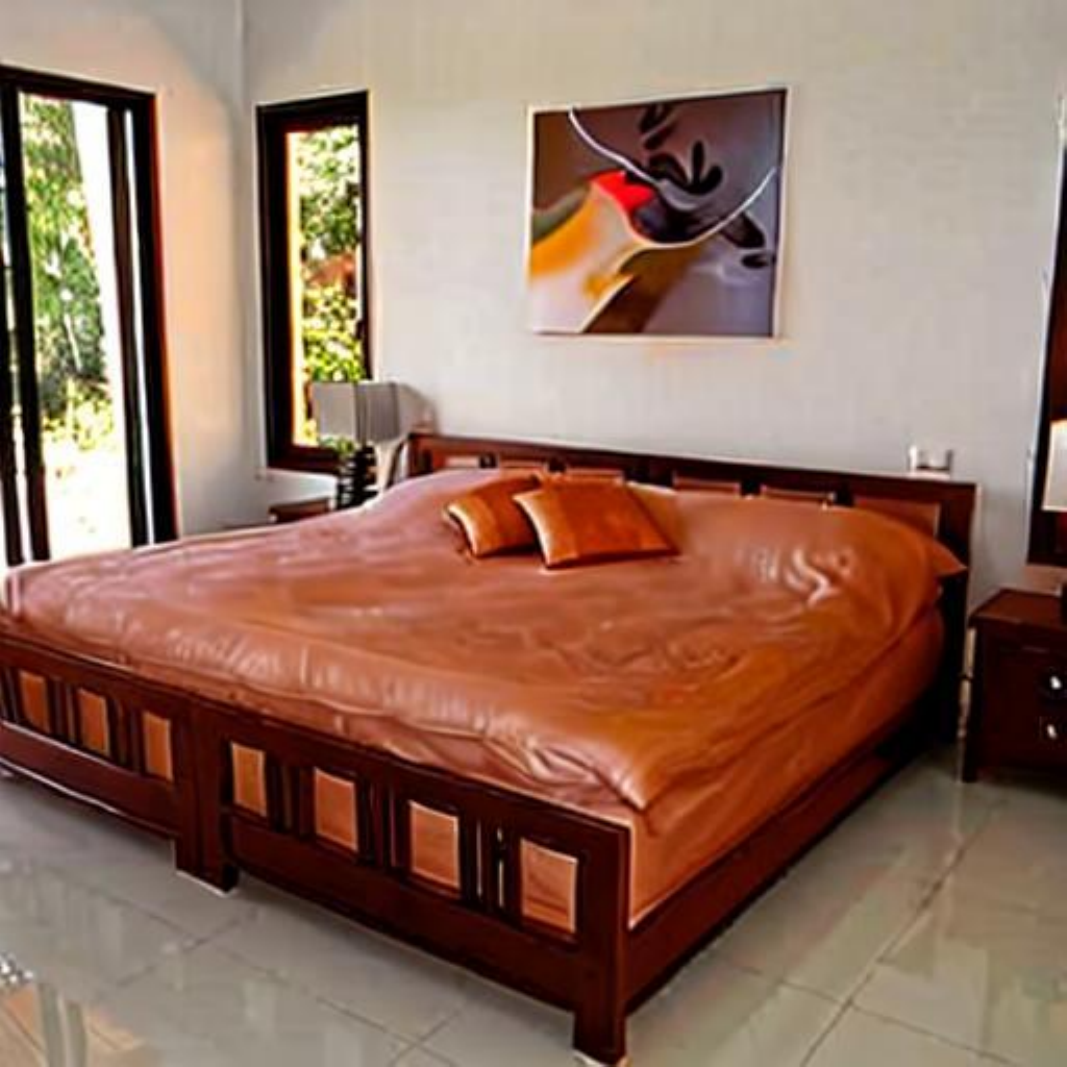} \\
    \multicolumn{6}{c}{``Make the bed out of leather"} \\[\textspace]

    \includegraphics[width=\imgwidth\textwidth]{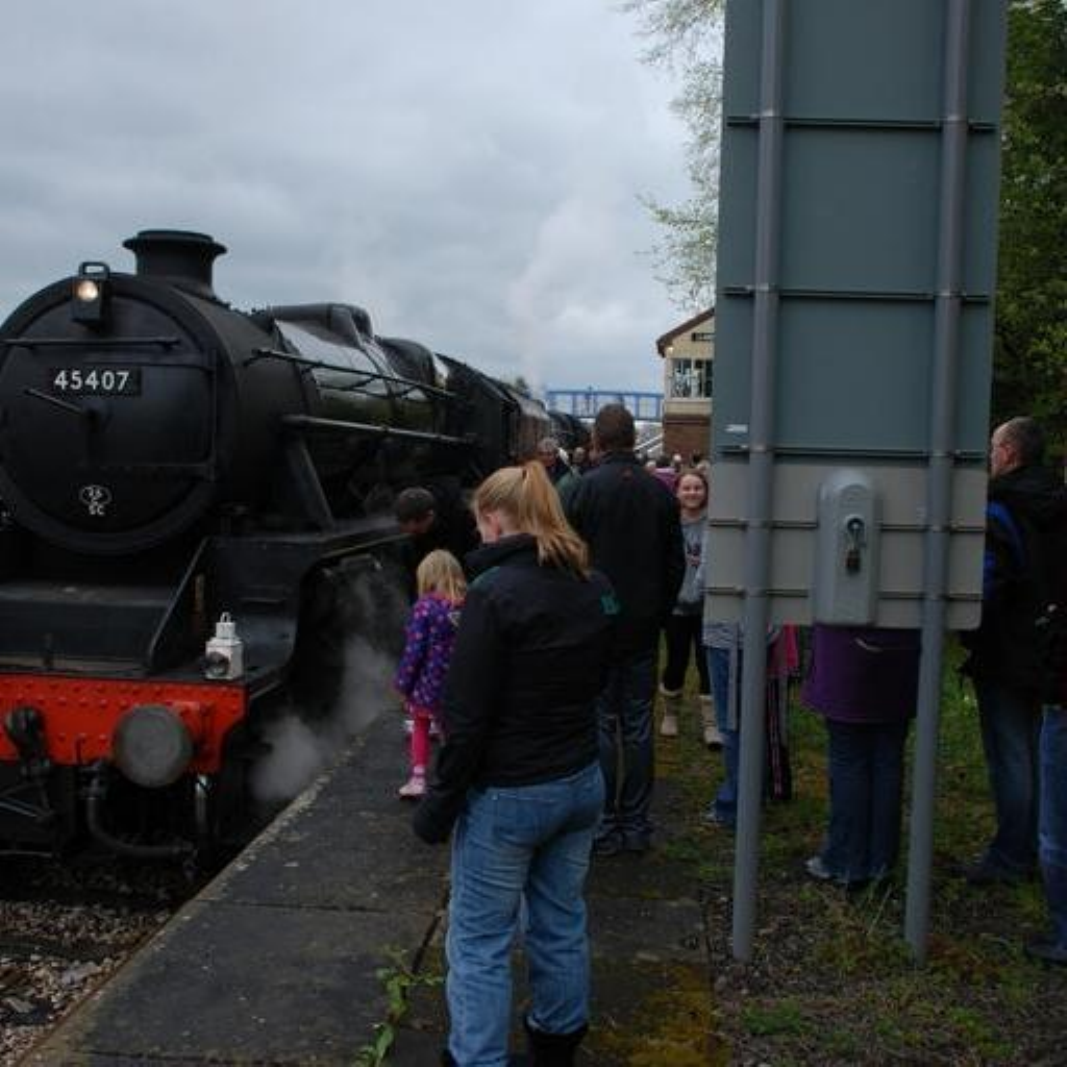} &
    \includegraphics[width=\imgwidth\textwidth]{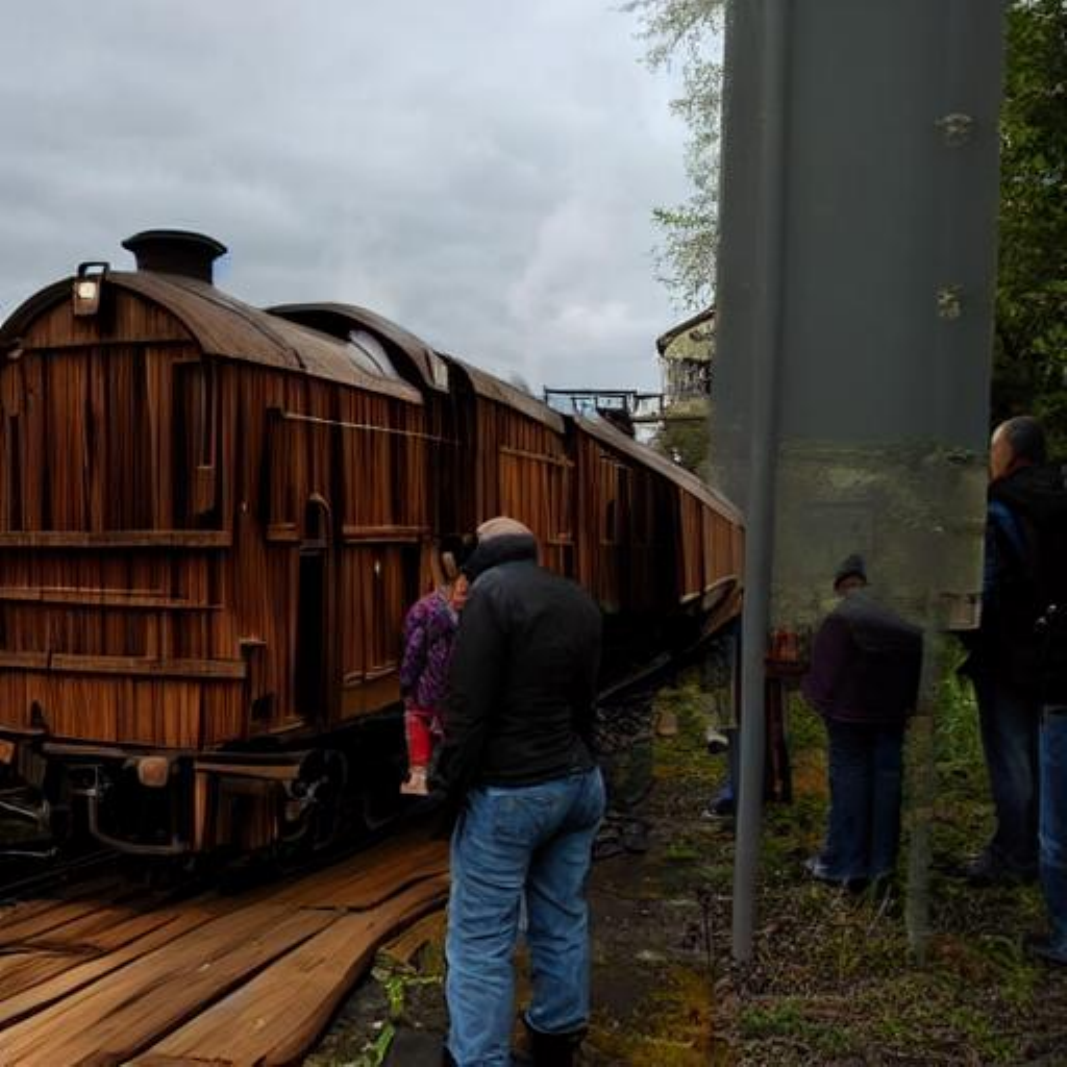} &
    \includegraphics[width=\imgwidth\textwidth]{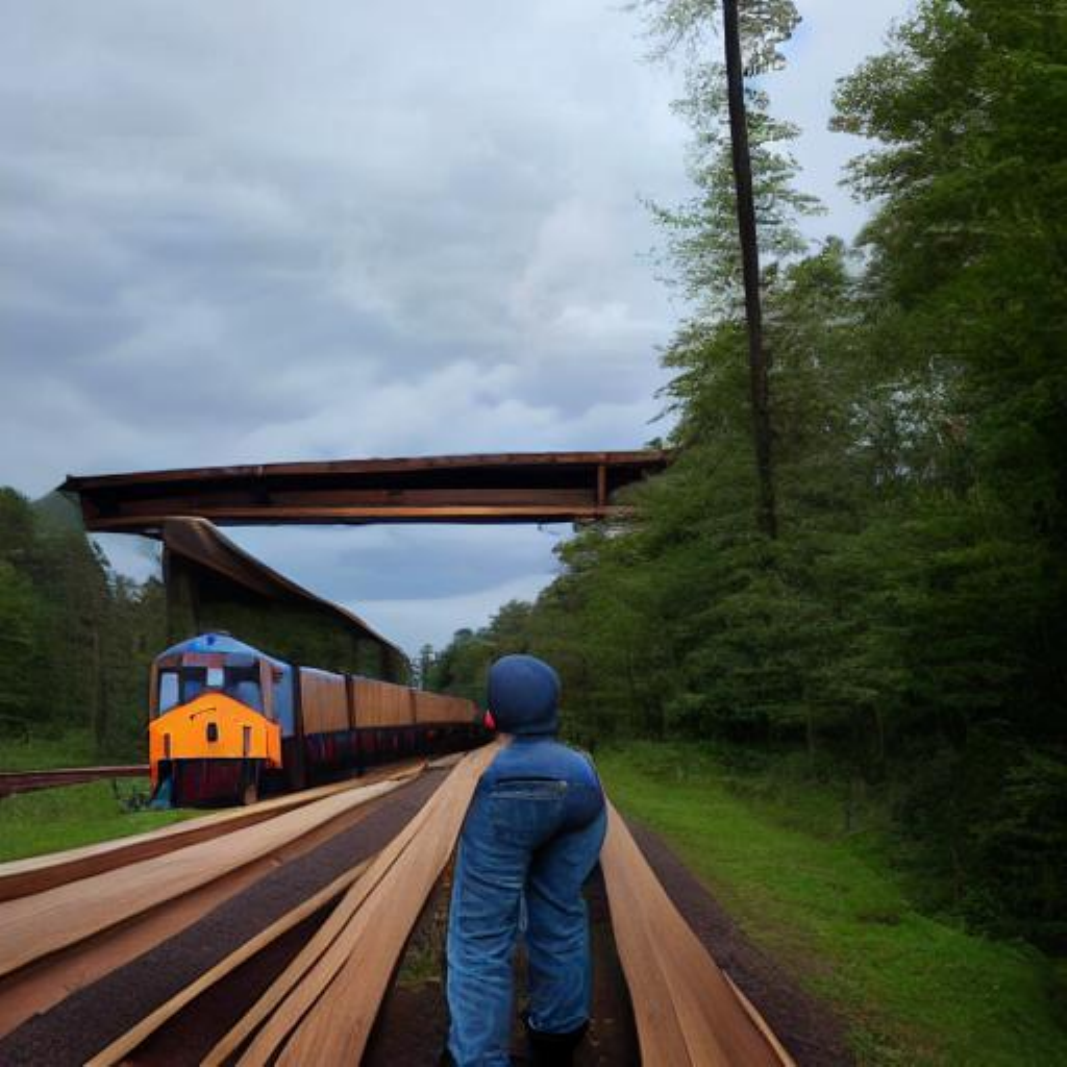}&
    \includegraphics[width=\imgwidth\textwidth]{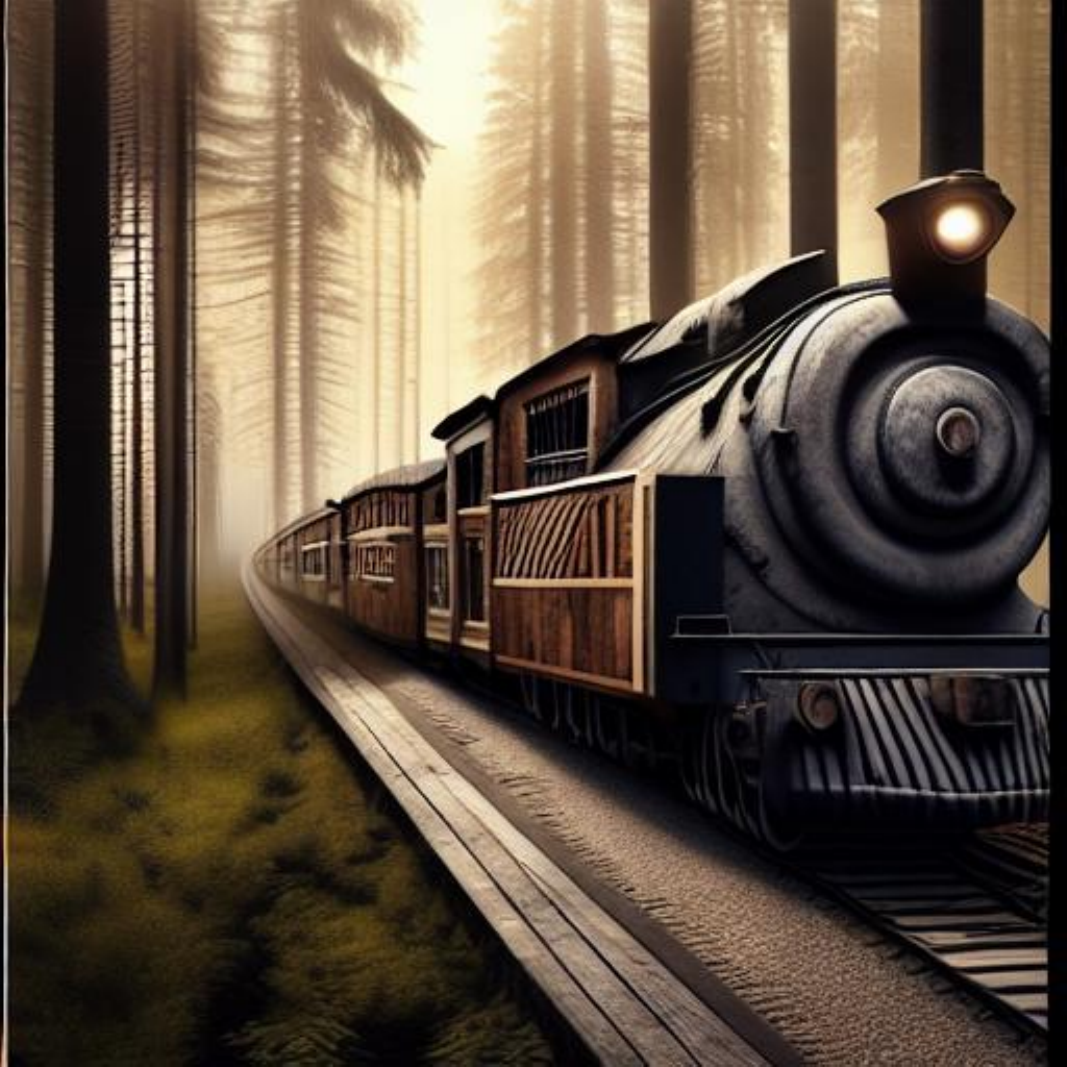}&
    \includegraphics[width=\imgwidth\textwidth]{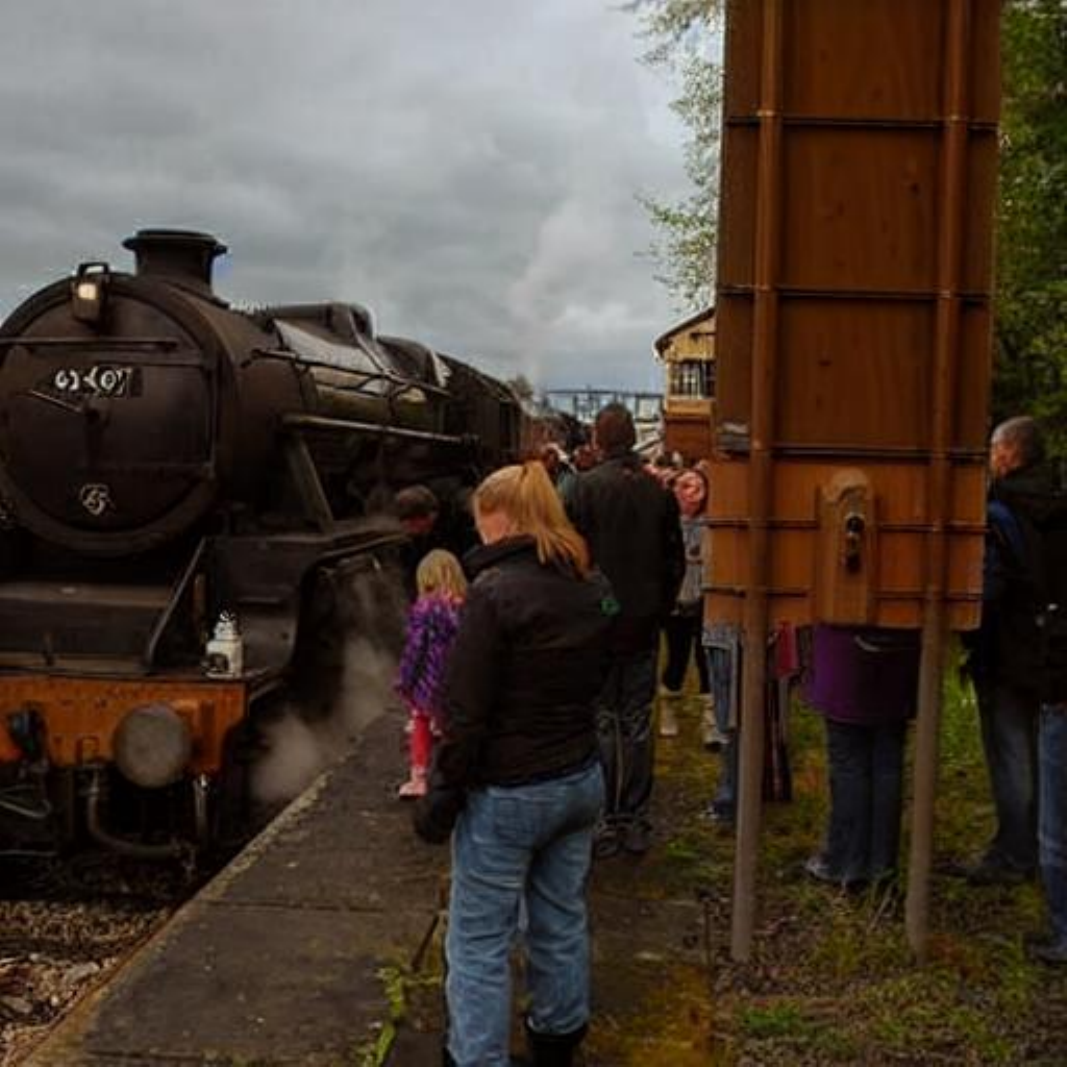}&
    \includegraphics[width=\imgwidth\textwidth]{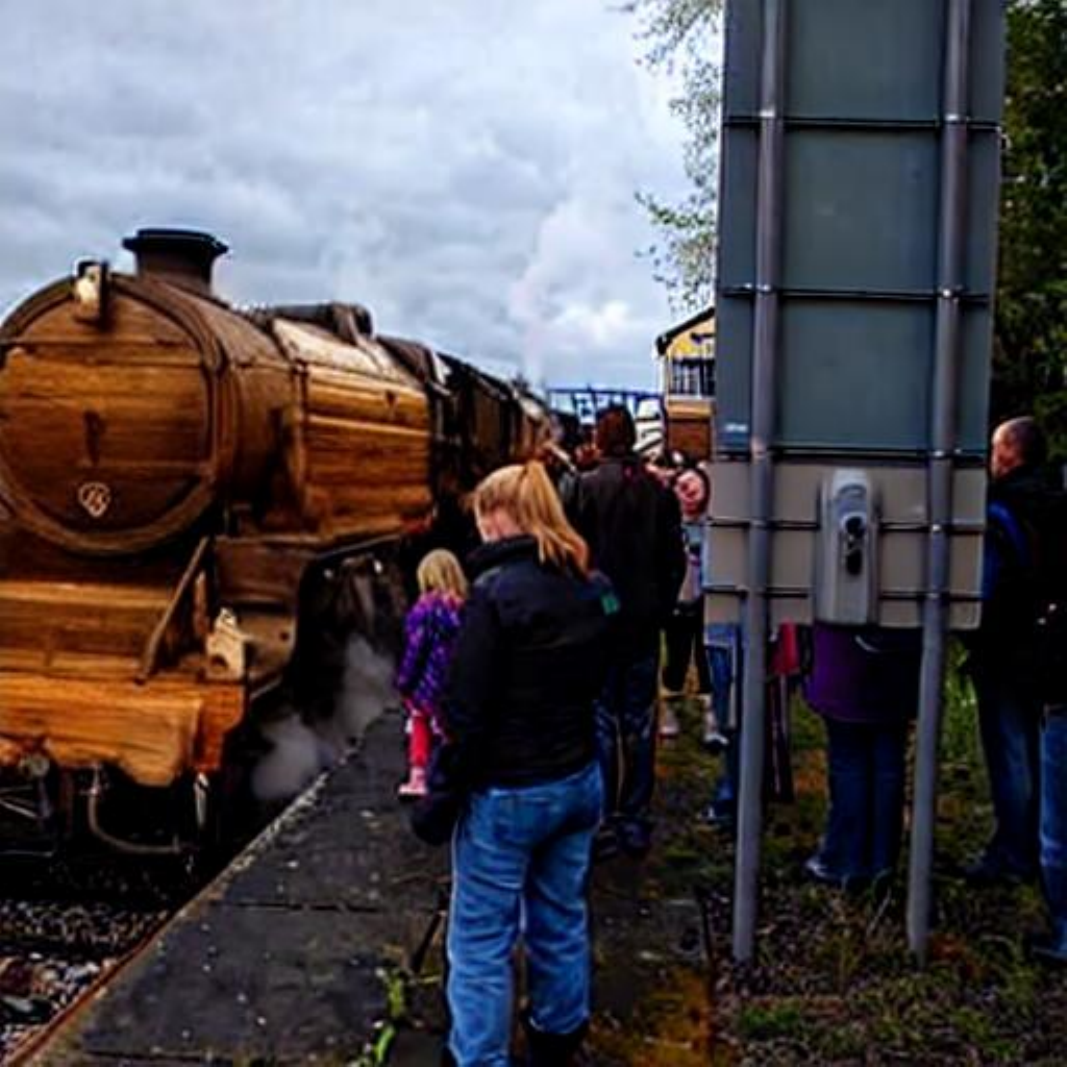} \\
    \multicolumn{6}{c}{``Make the train out of wood"} \\[\textspace]

    \includegraphics[width=\imgwidth\textwidth]{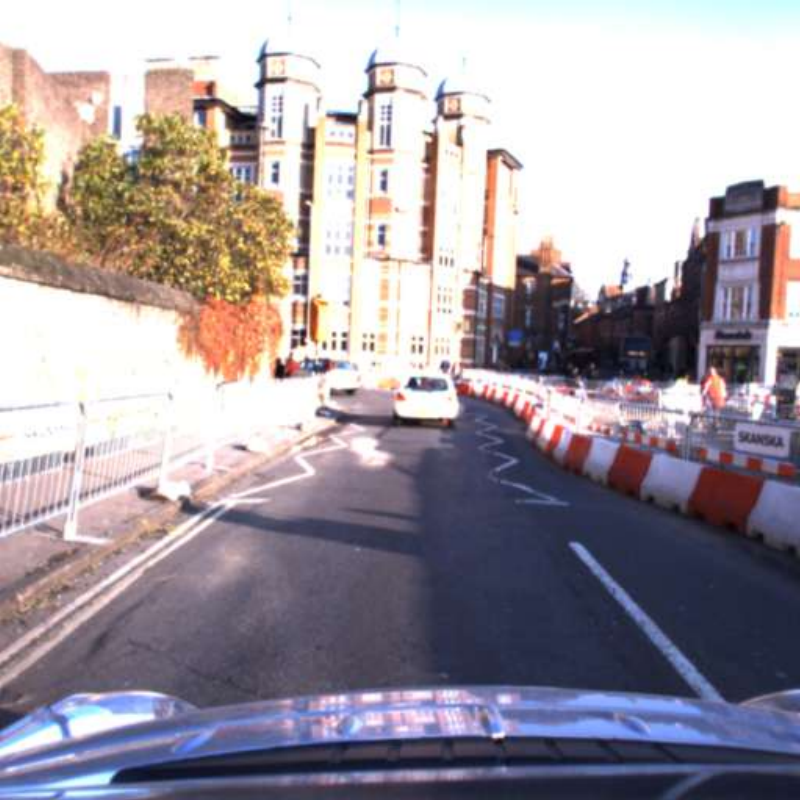} &
    \includegraphics[width=\imgwidth\textwidth]{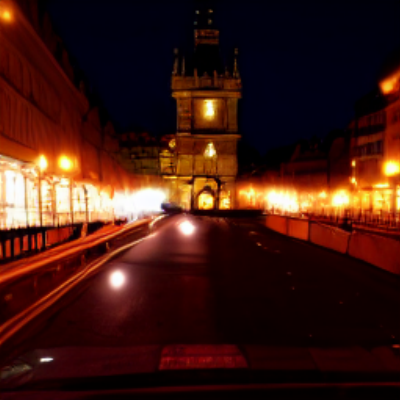} &
    \includegraphics[width=\imgwidth\textwidth]{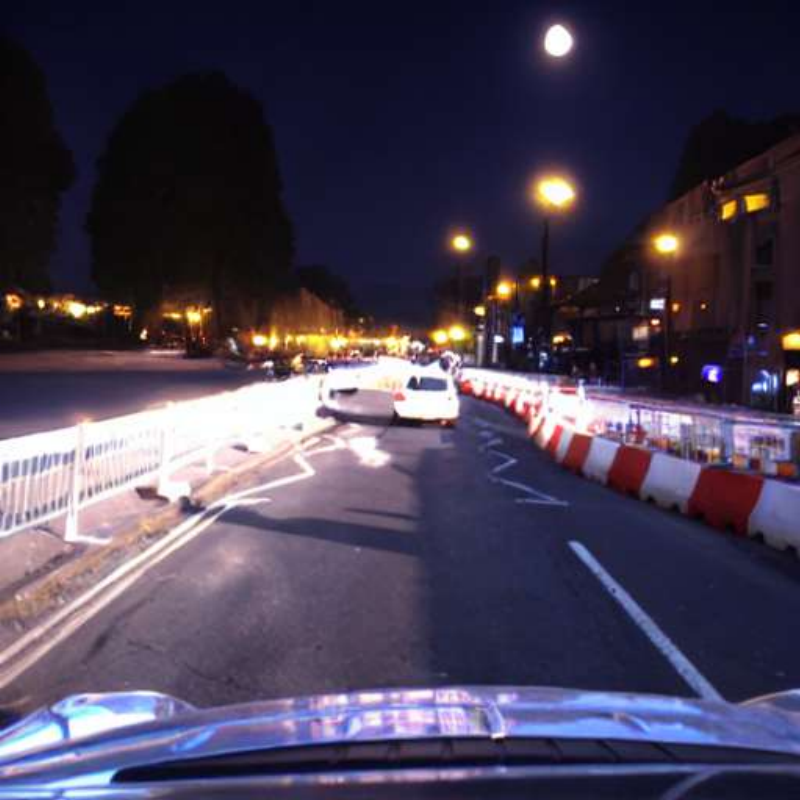} &
    \includegraphics[width=\imgwidth\textwidth]{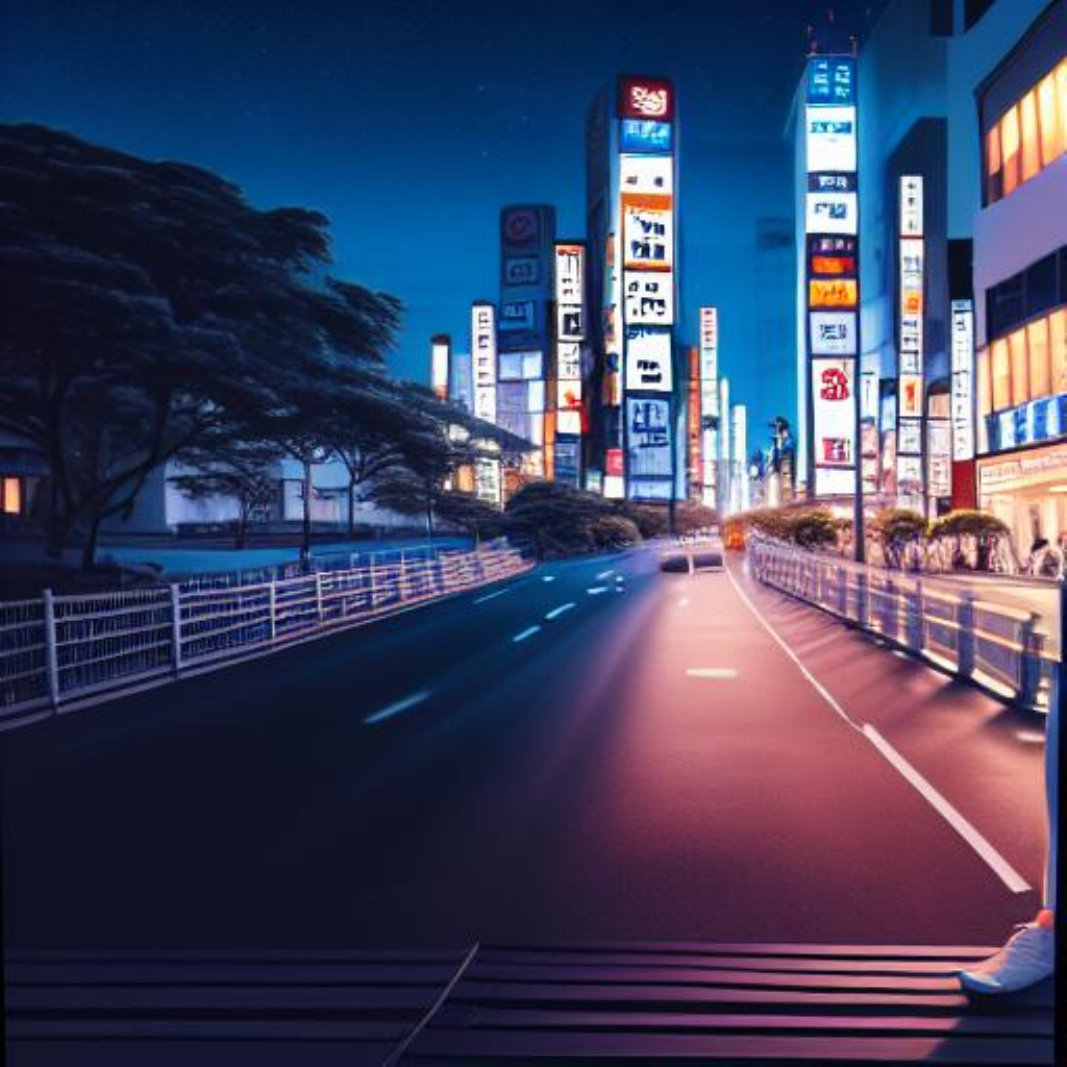} &
    \includegraphics[width=\imgwidth\textwidth]{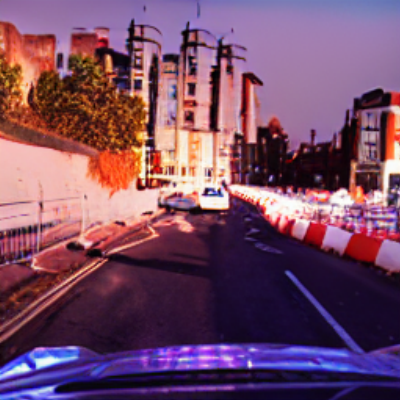} &
    \includegraphics[width=\imgwidth\textwidth]{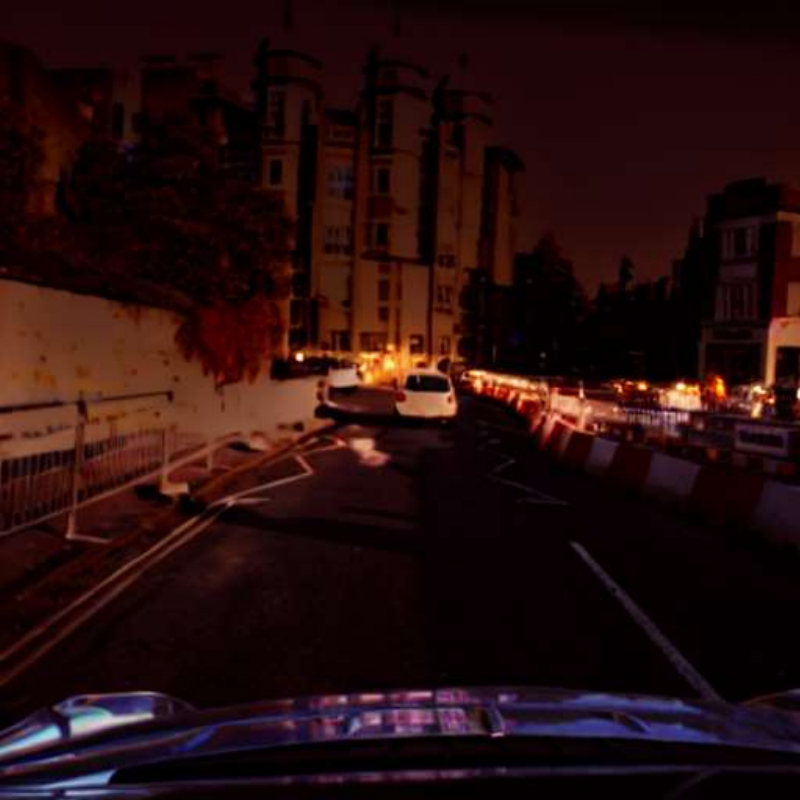} \\
    \multicolumn{6}{c}{``Change the time to nighttime"} \\[\textspace]

\end{tabular}
\vspace{-1em}
\caption{Comparative evaluation of our method against baselines on a diverse set of prompts and images, highlighting superior performance in structural preservation, semantic alignment, and realism.}
\label{appdx_baseline1}
\end{figure*}

\begin{figure*}[h]
\centering
\newcommand{\imgwidth}{0.16}
\setlength{\tabcolsep}{2pt} 
\renewcommand{\arraystretch}{1} 
\newcommand{\textspace}{0.7em}
\begin{tabular}{cccccc} 
    Input & InstructPix2Pix & MagicBrush & HQ-Edit & HIVE & \textbf{SPIE} \\

    \includegraphics[width=\imgwidth\textwidth]{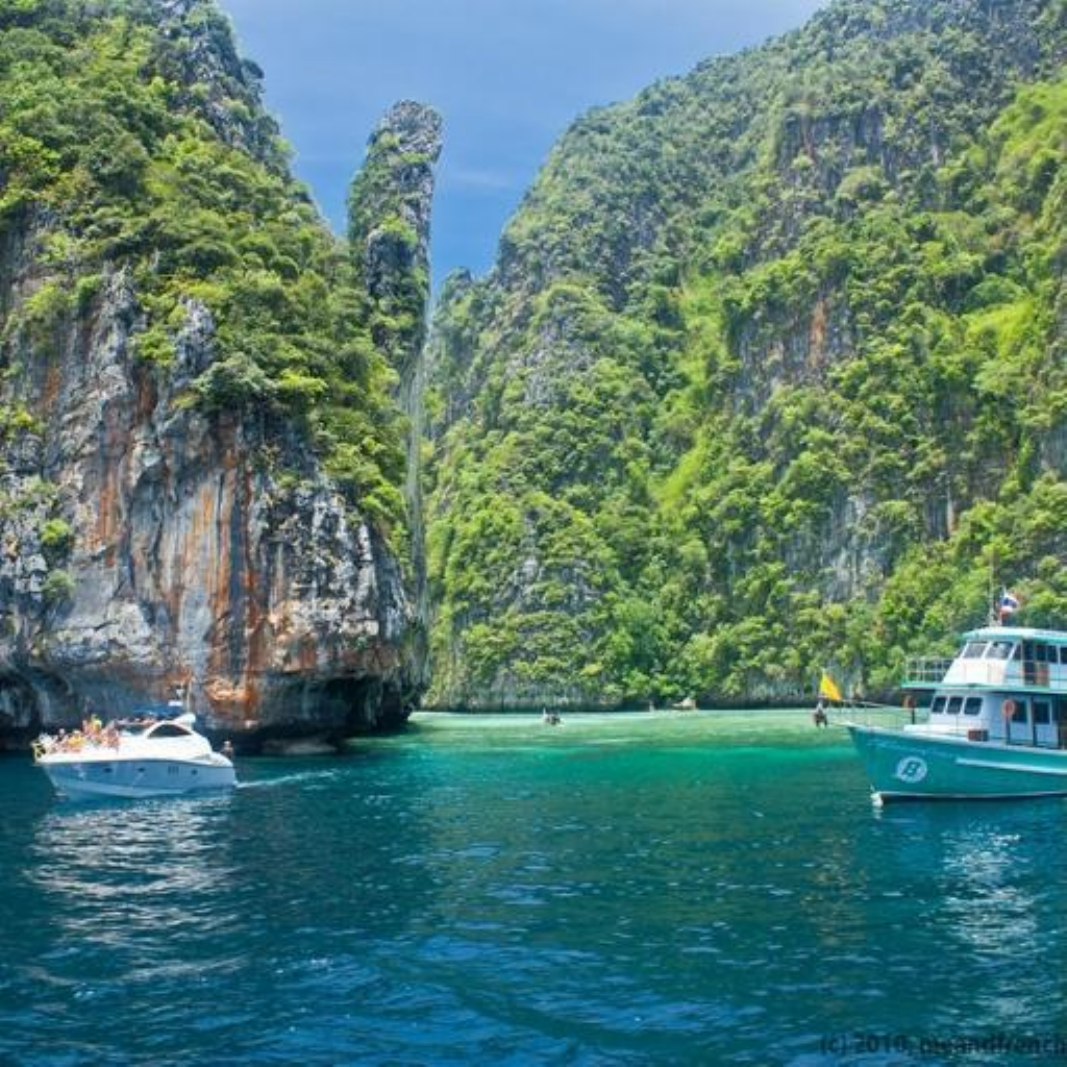} &
    \includegraphics[width=\imgwidth\textwidth]{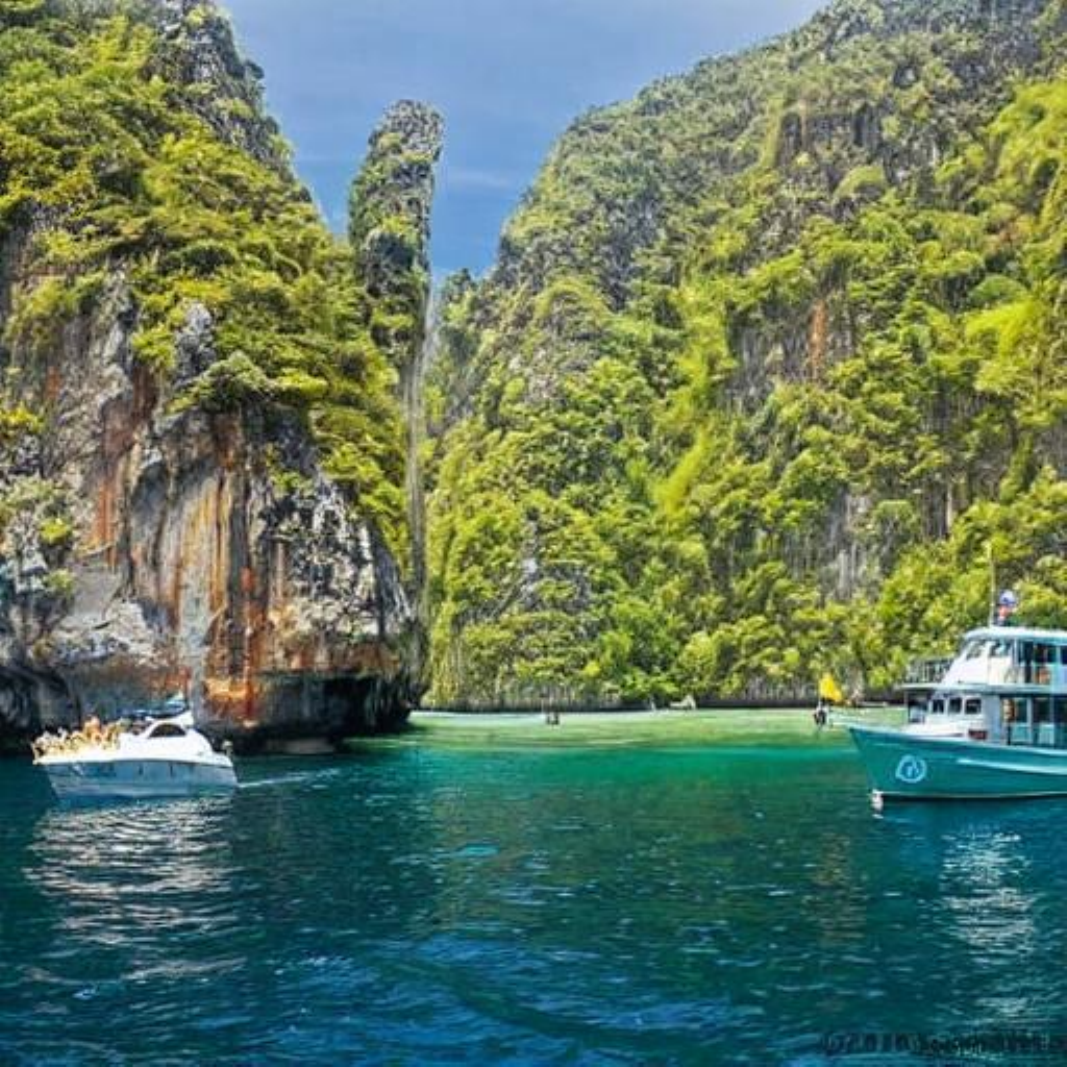} &
    \includegraphics[width=\imgwidth\textwidth]{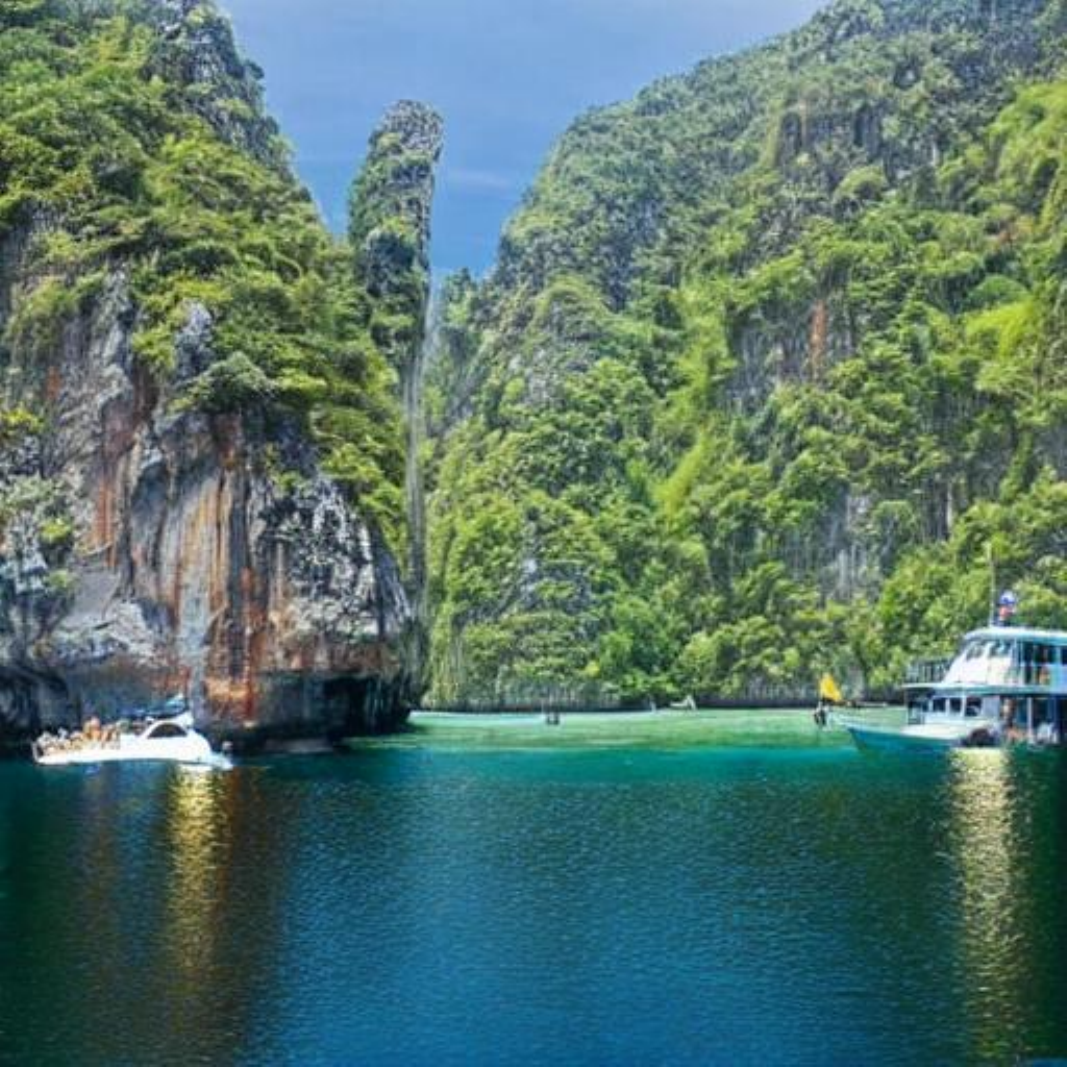}&
    \includegraphics[width=\imgwidth\textwidth]{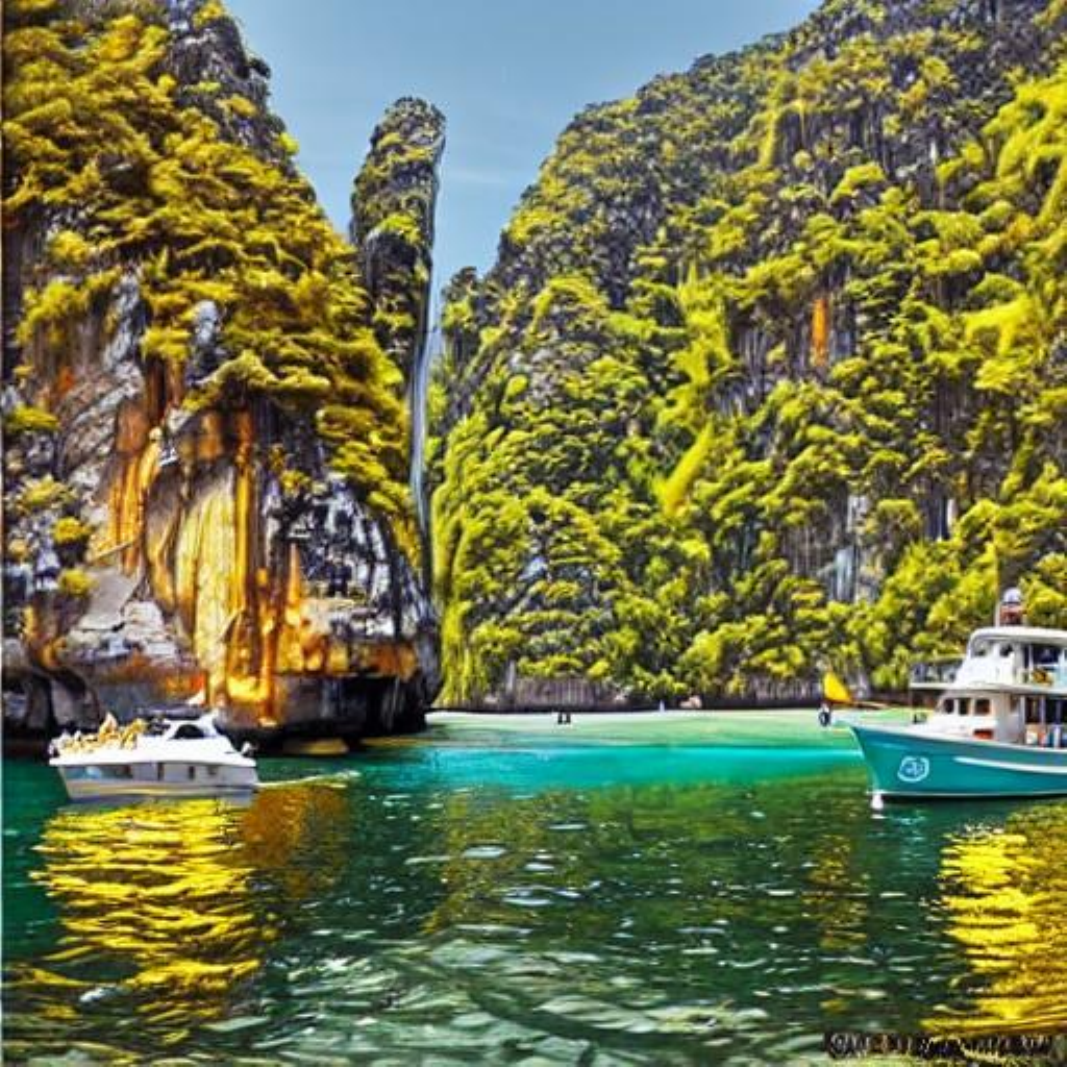}&
    \includegraphics[width=\imgwidth\textwidth]{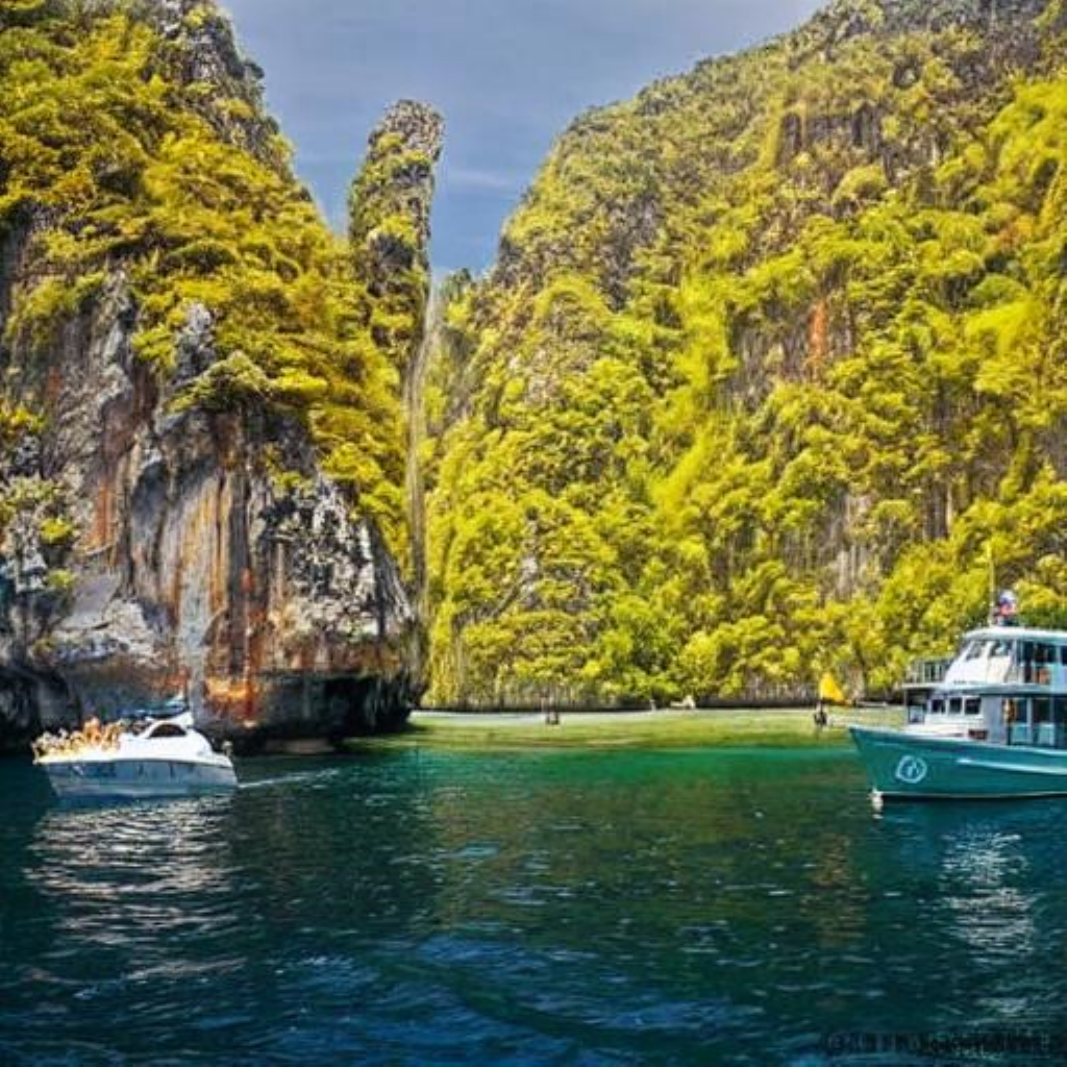}&
    \includegraphics[width=\imgwidth\textwidth]{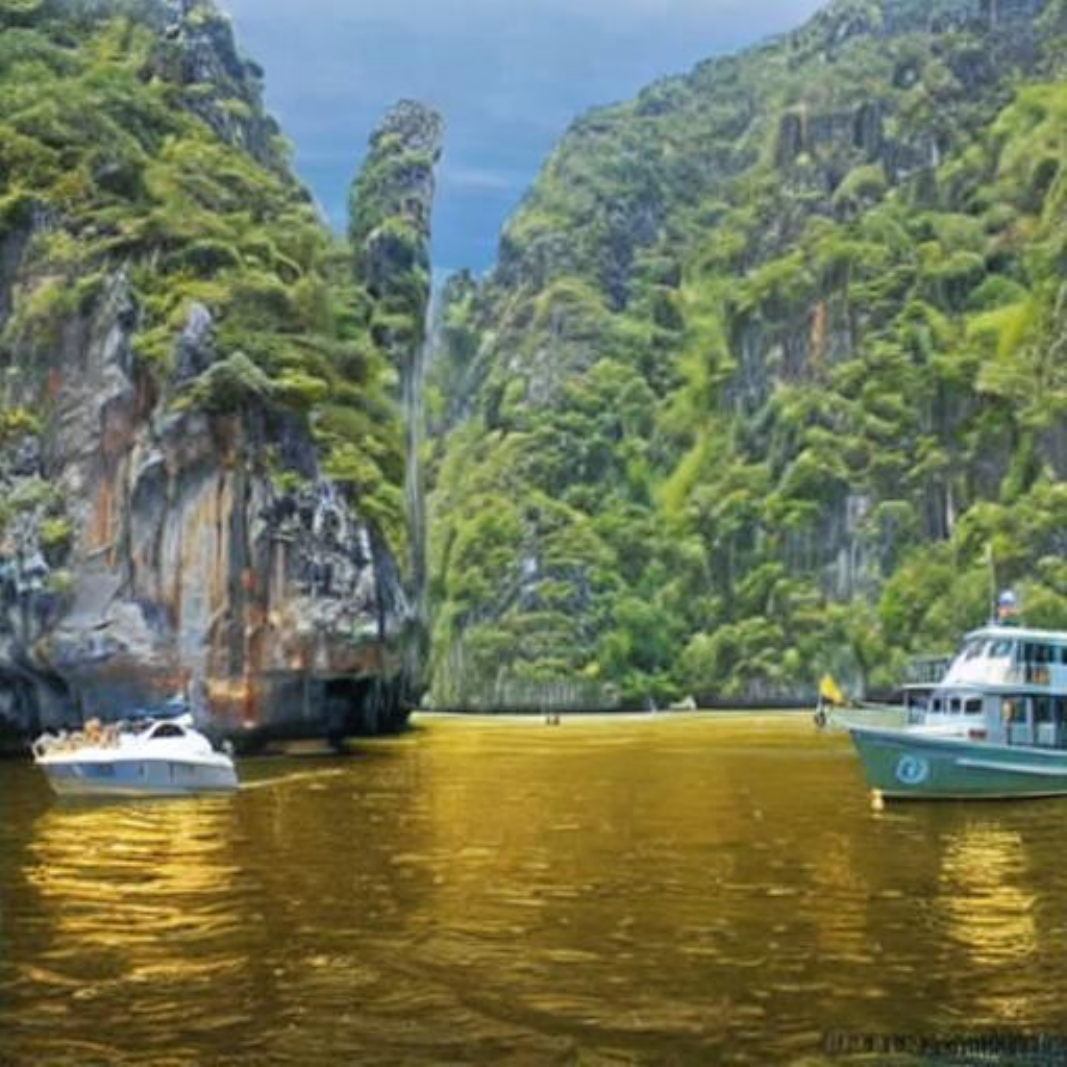} \\
    \multicolumn{6}{c}{``Change the water for gold"} \\[\textspace]

    \includegraphics[width=\imgwidth\textwidth]{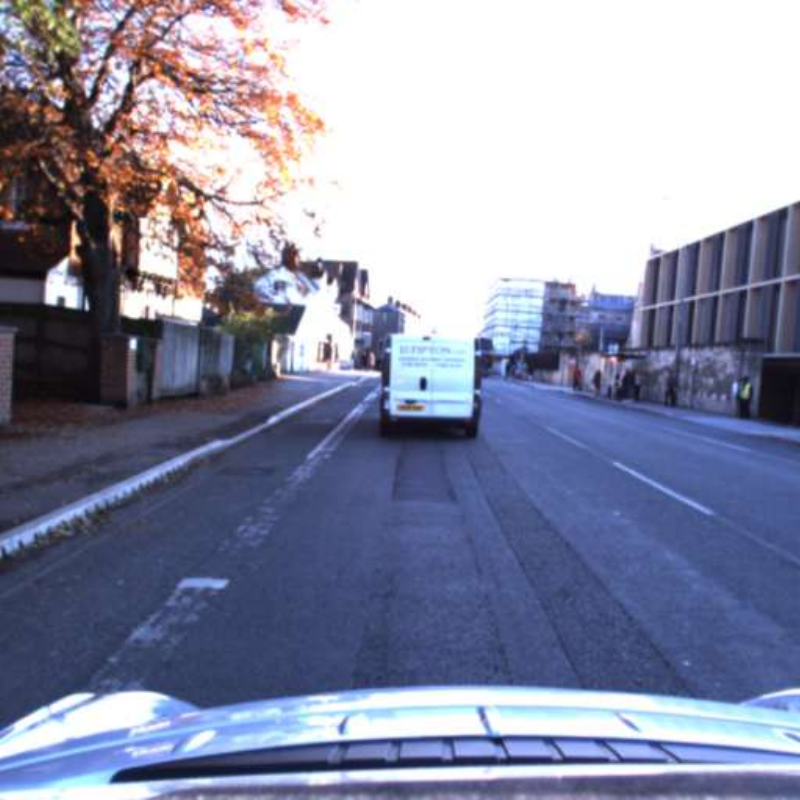}&
    \includegraphics[width=\imgwidth\textwidth]{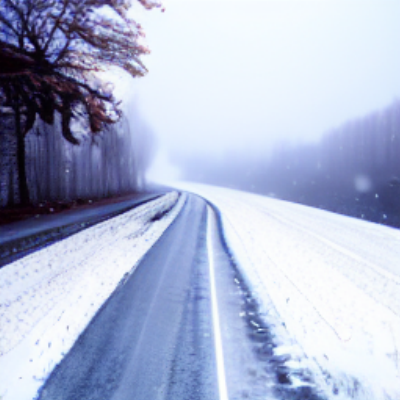} &
    \includegraphics[width=\imgwidth\textwidth]{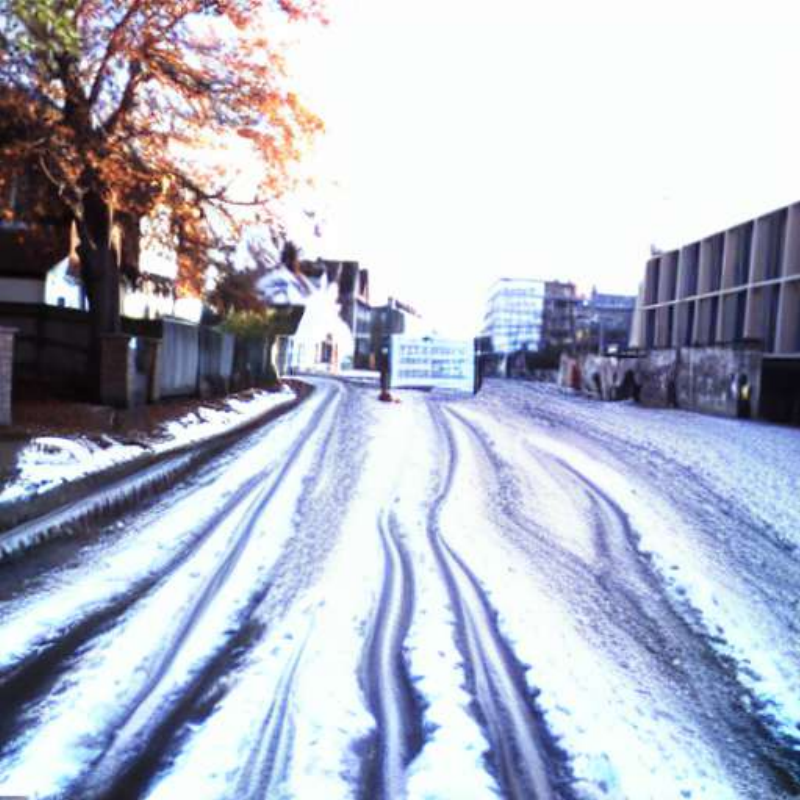} &
    \includegraphics[width=\imgwidth\textwidth]{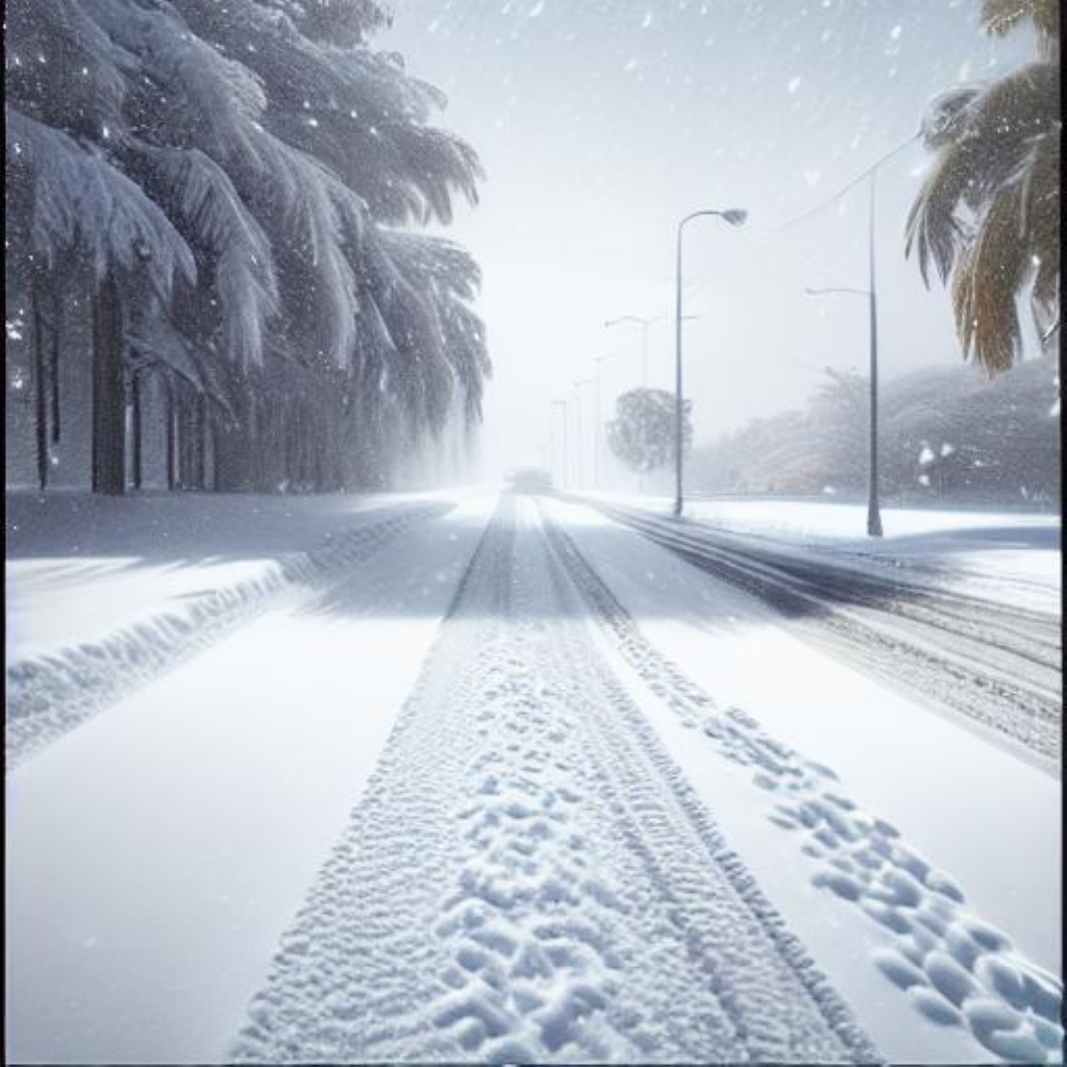} &
    \includegraphics[width=\imgwidth\textwidth]{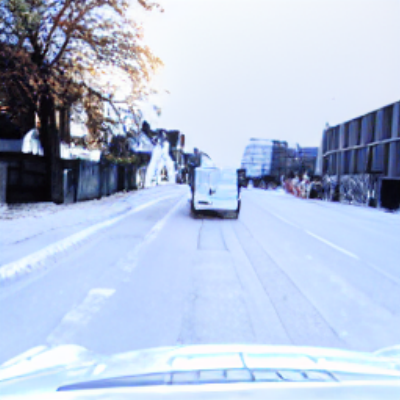} &
    \includegraphics[width=\imgwidth\textwidth]{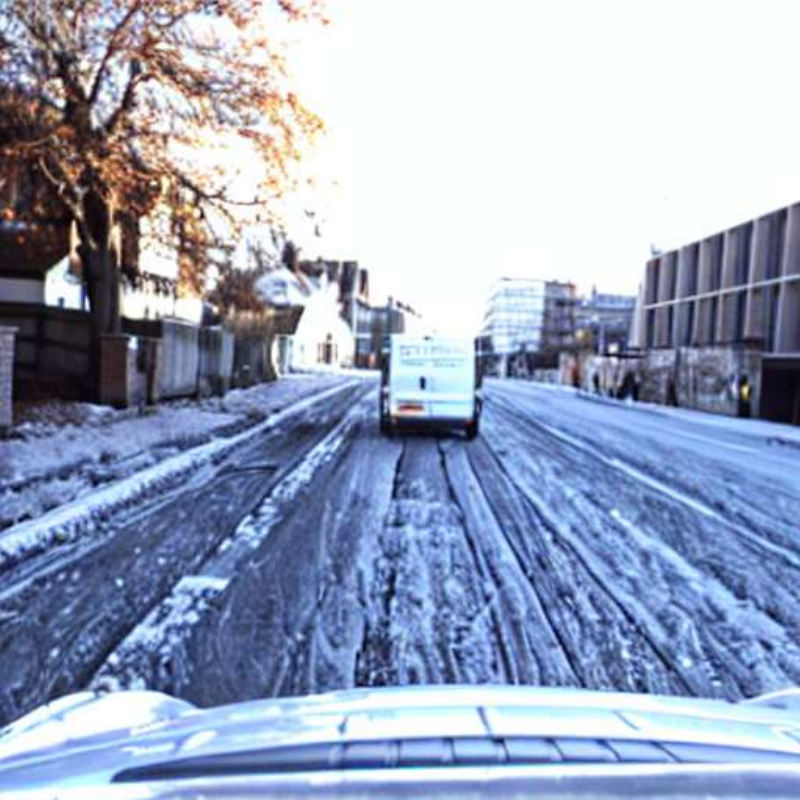}\\
    \multicolumn{6}{c}{``Add snow on the road"} \\[\textspace]

    \includegraphics[width=\imgwidth\textwidth]{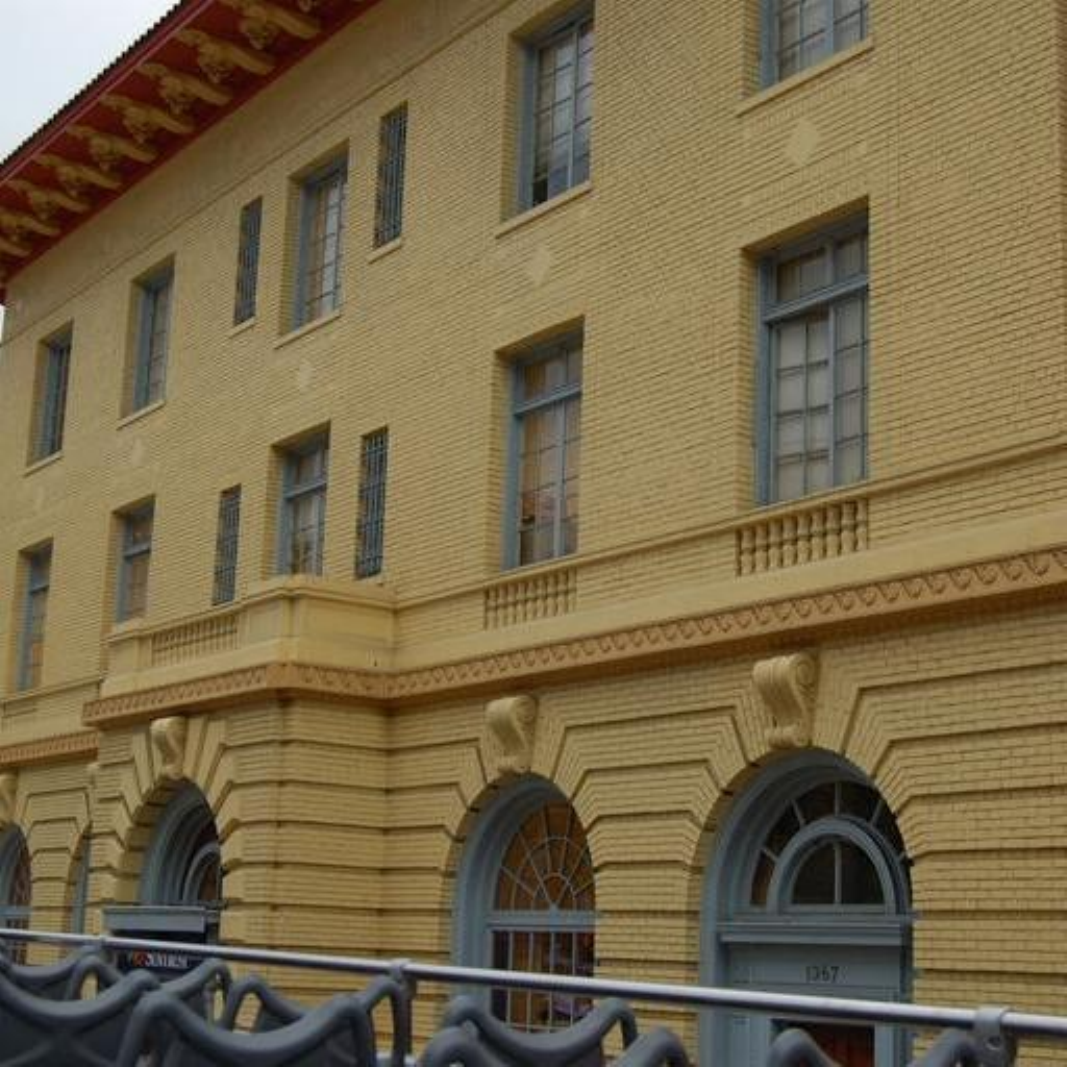} &
    \includegraphics[width=\imgwidth\textwidth]{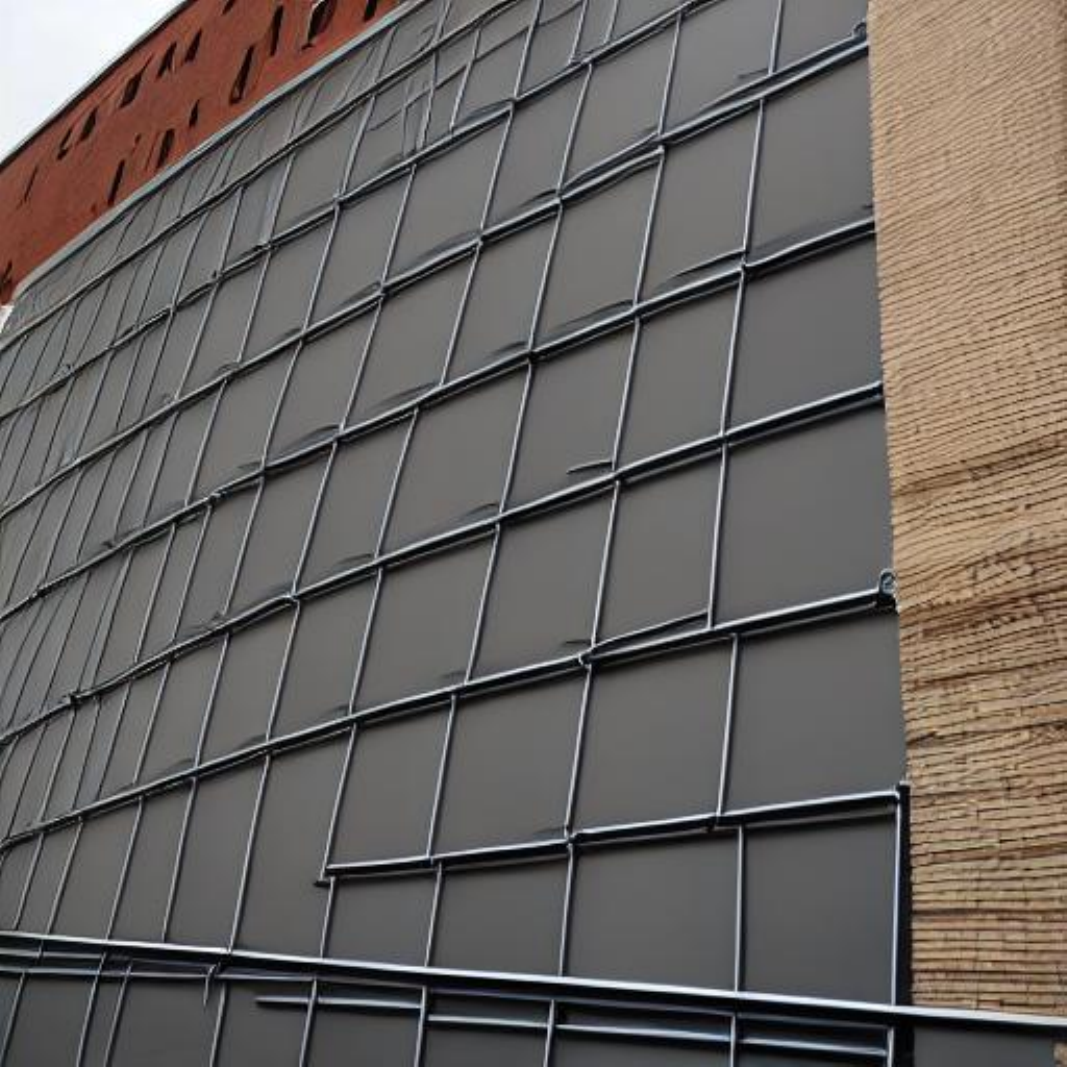} &
    \includegraphics[width=\imgwidth\textwidth]{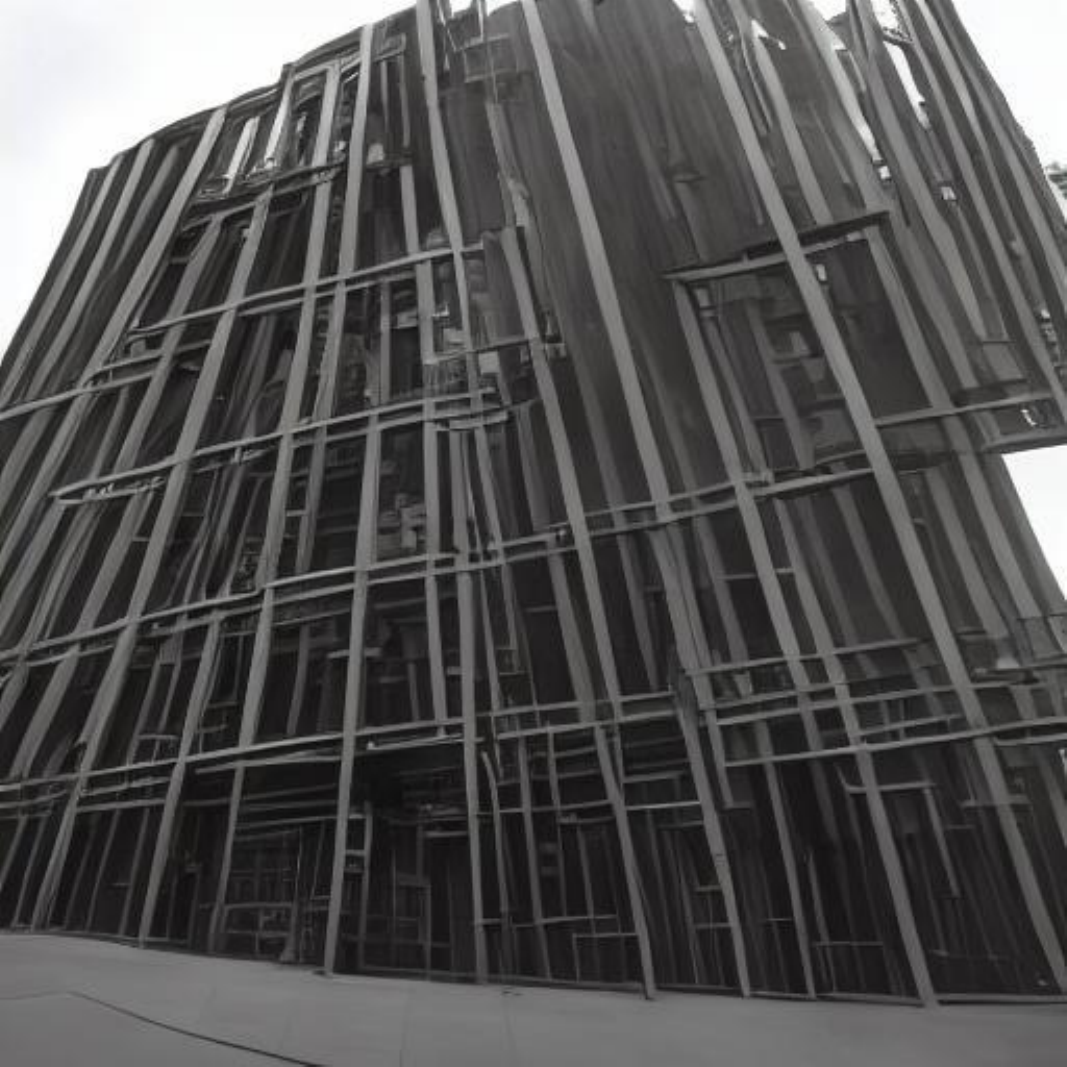}&
    \includegraphics[width=\imgwidth\textwidth]{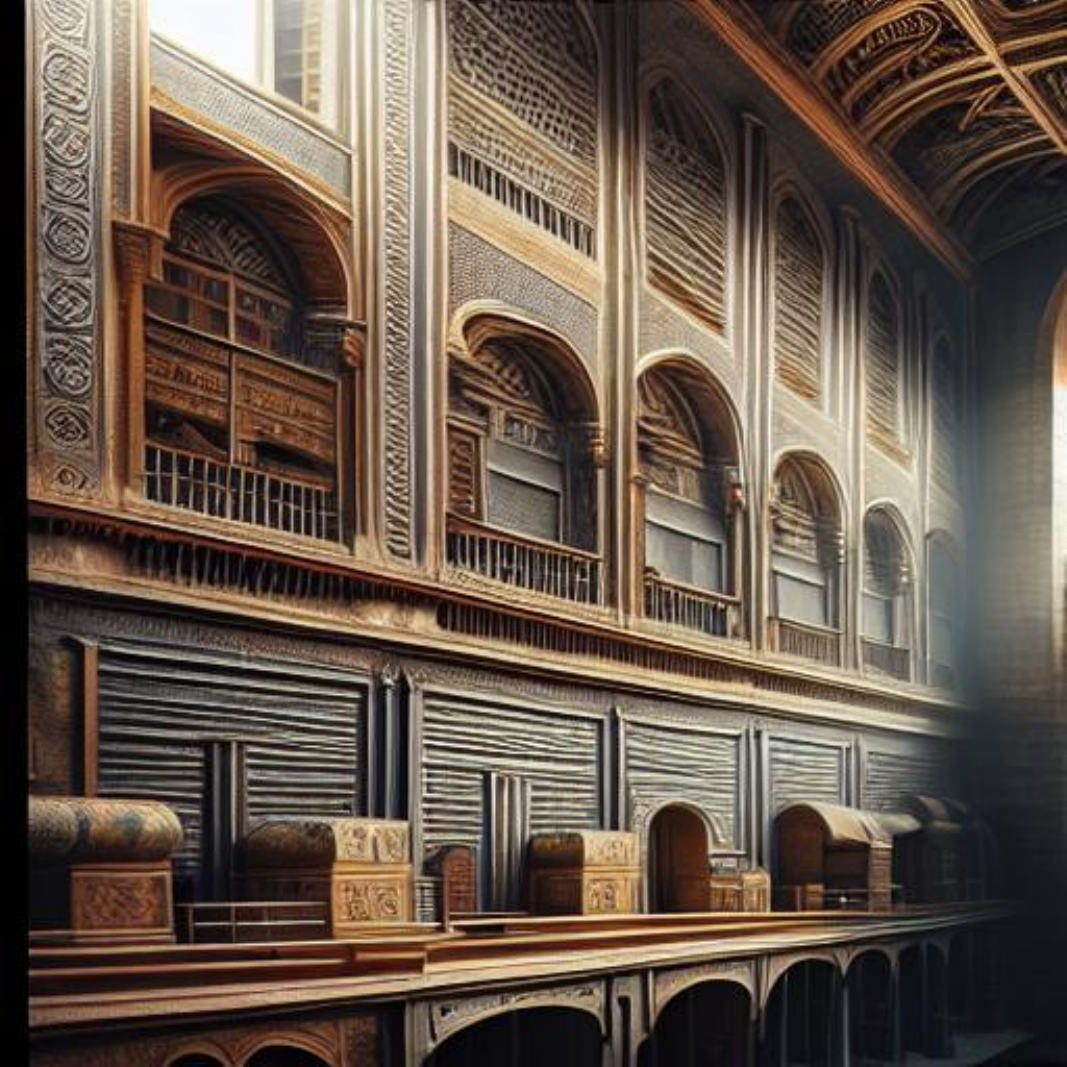}&
    \includegraphics[width=\imgwidth\textwidth]{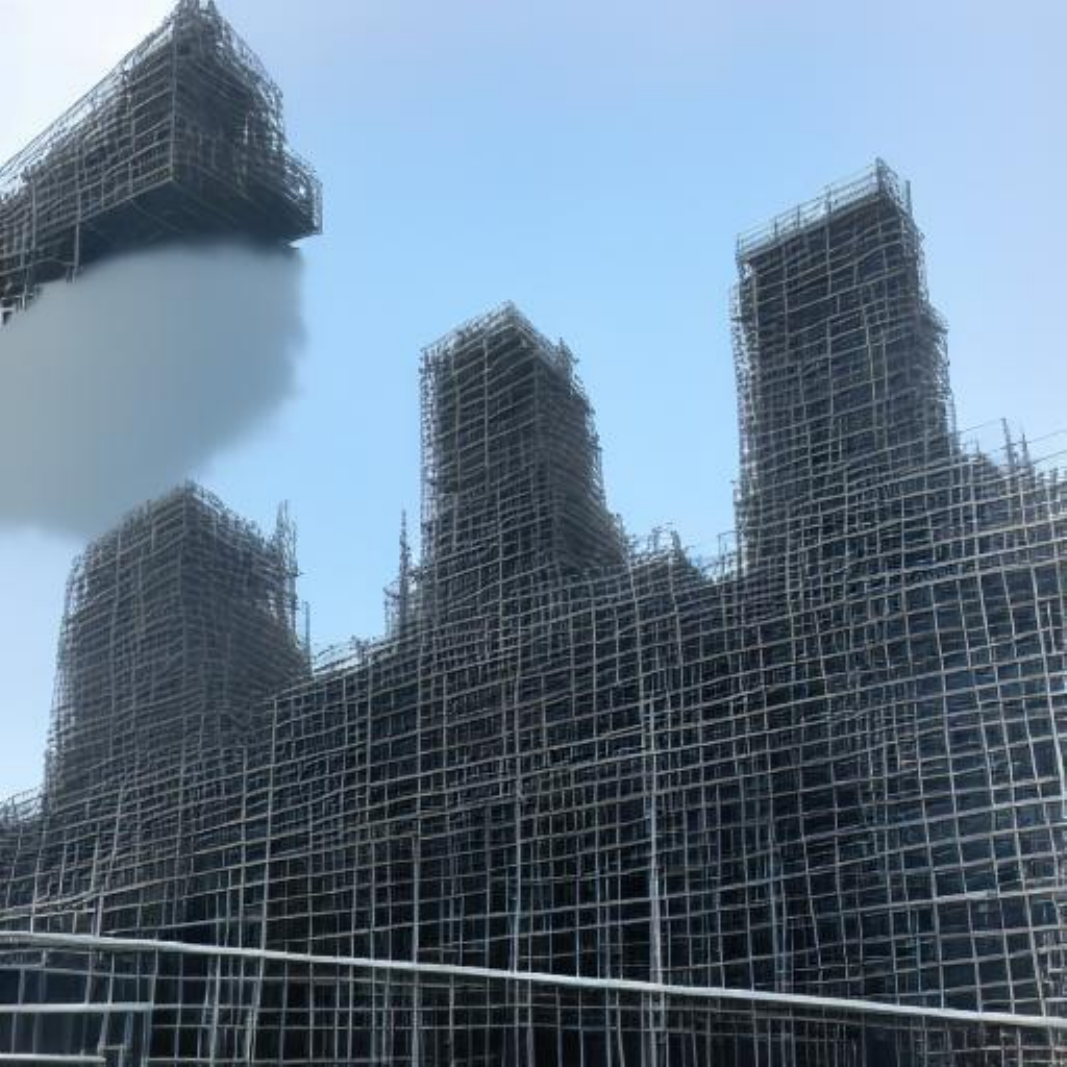}&
    \includegraphics[width=\imgwidth\textwidth]{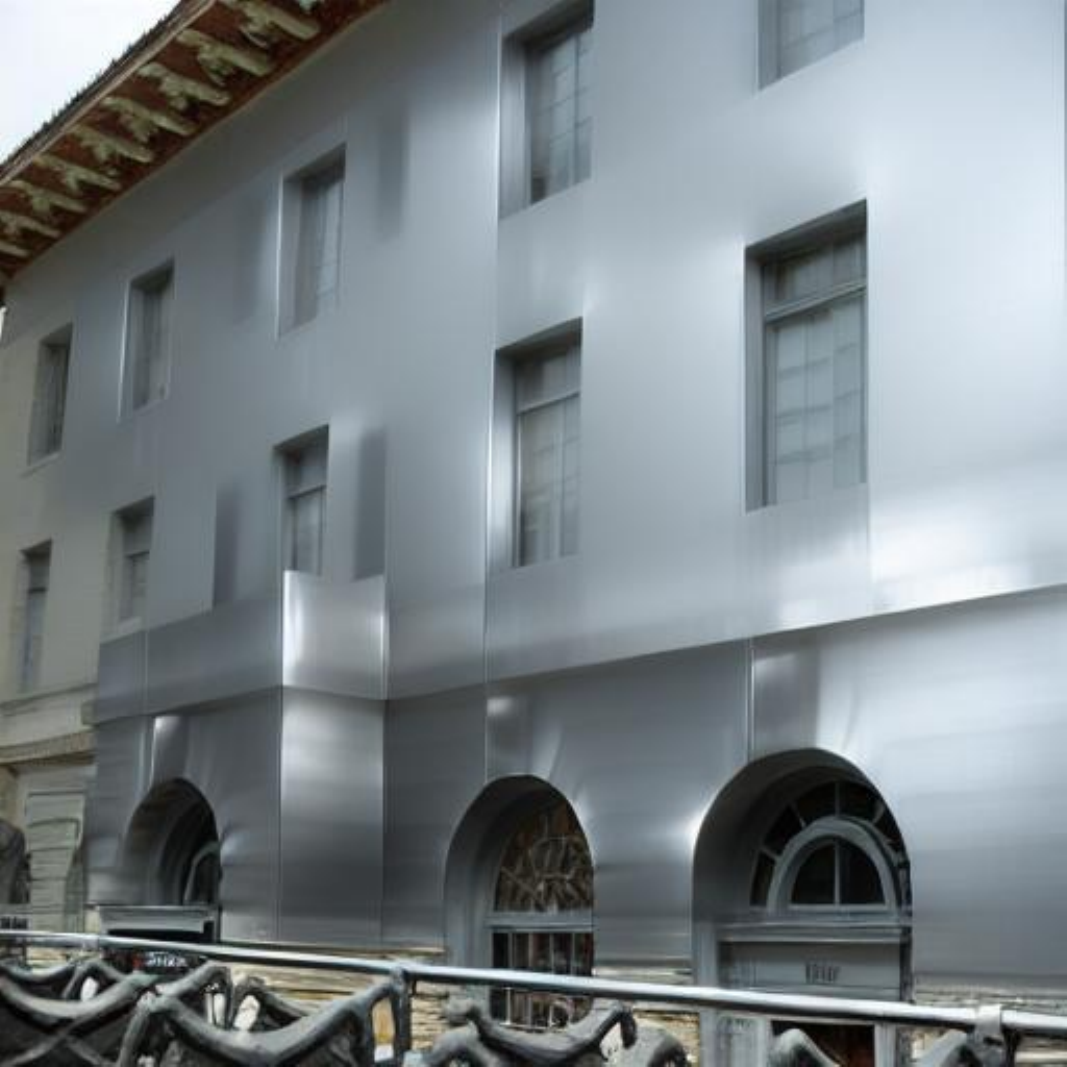} \\
    \multicolumn{6}{c}{``Turn the building into steel"} \\[\textspace]

    \includegraphics[width=\imgwidth\textwidth]{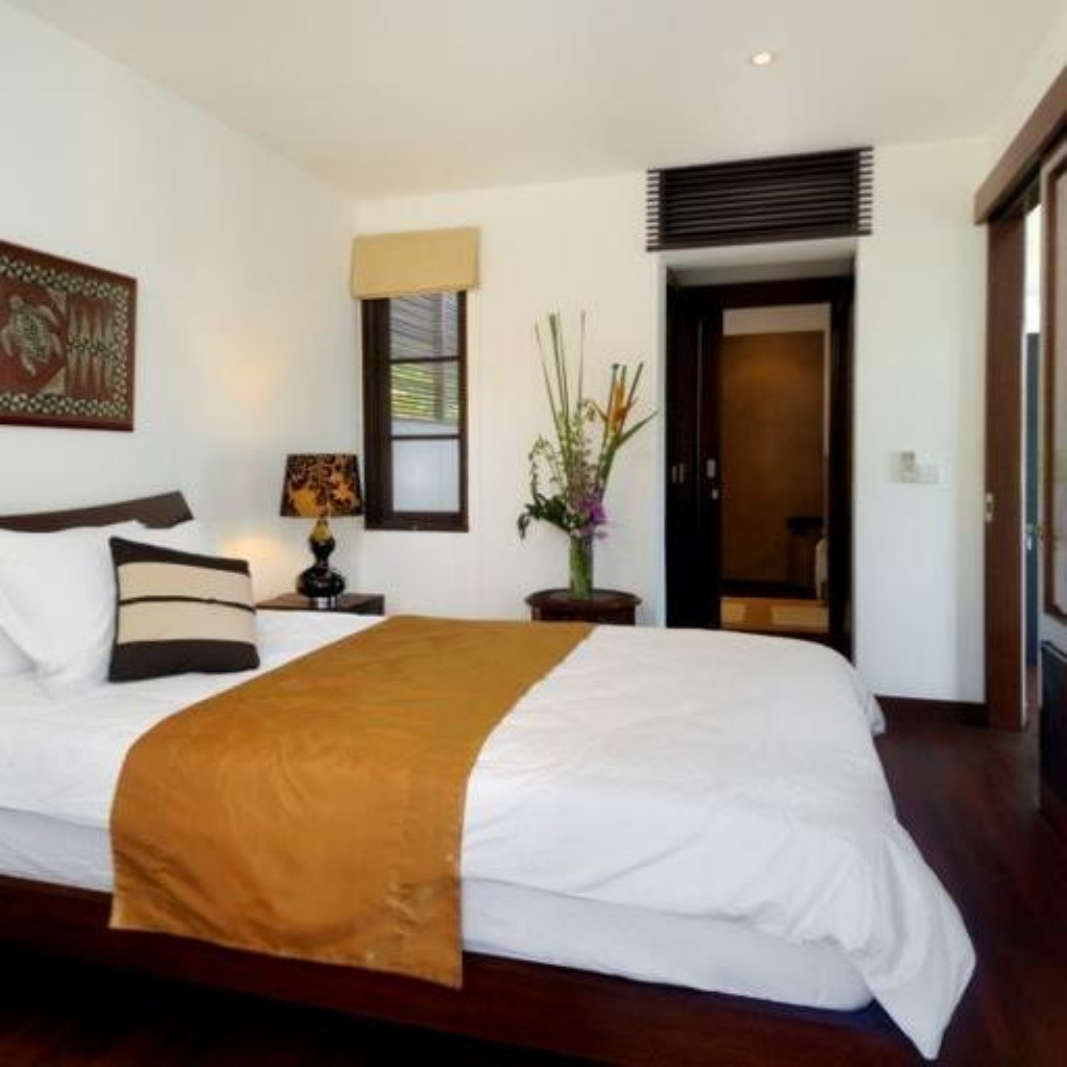} &
    \includegraphics[width=\imgwidth\textwidth]{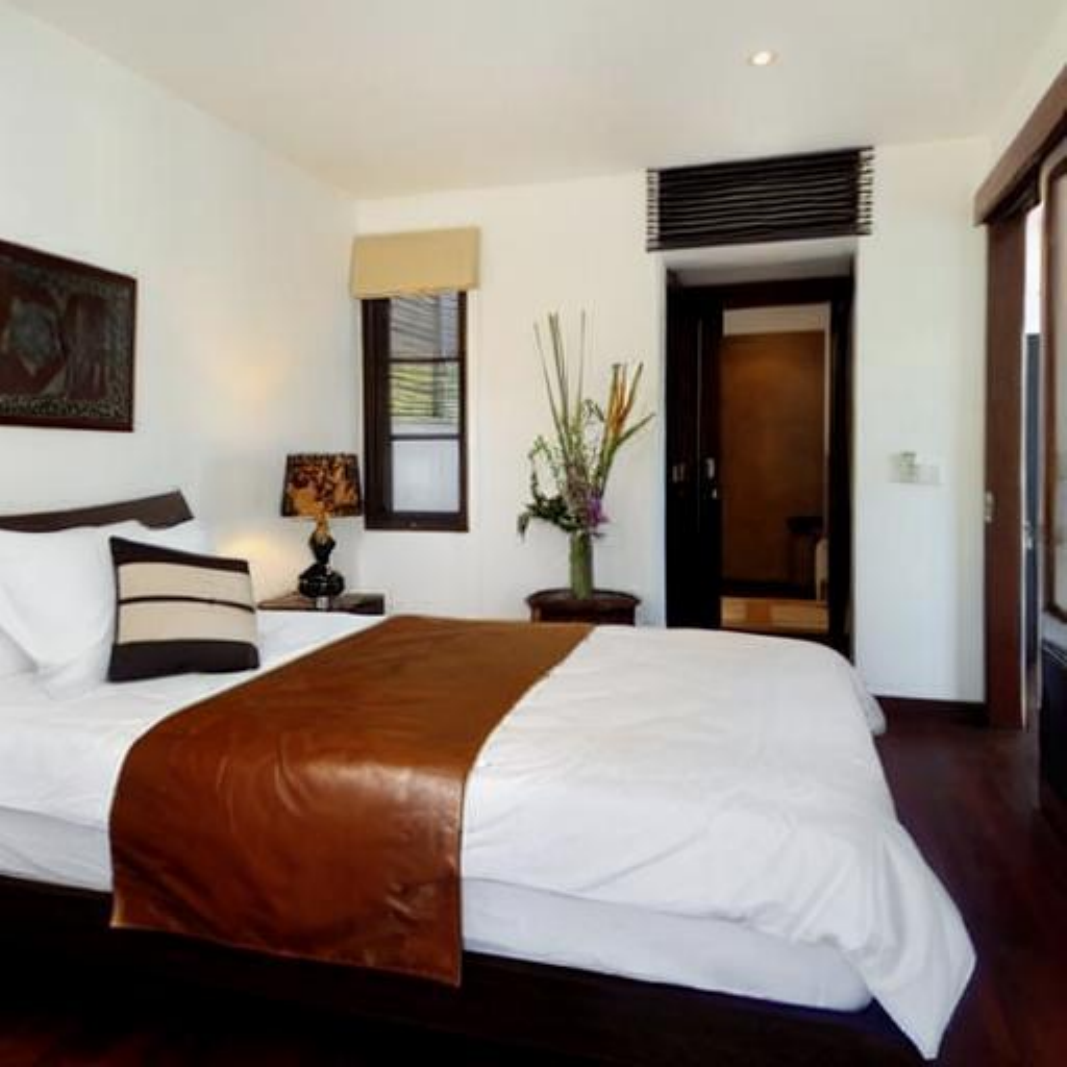} &
    \includegraphics[width=\imgwidth\textwidth]{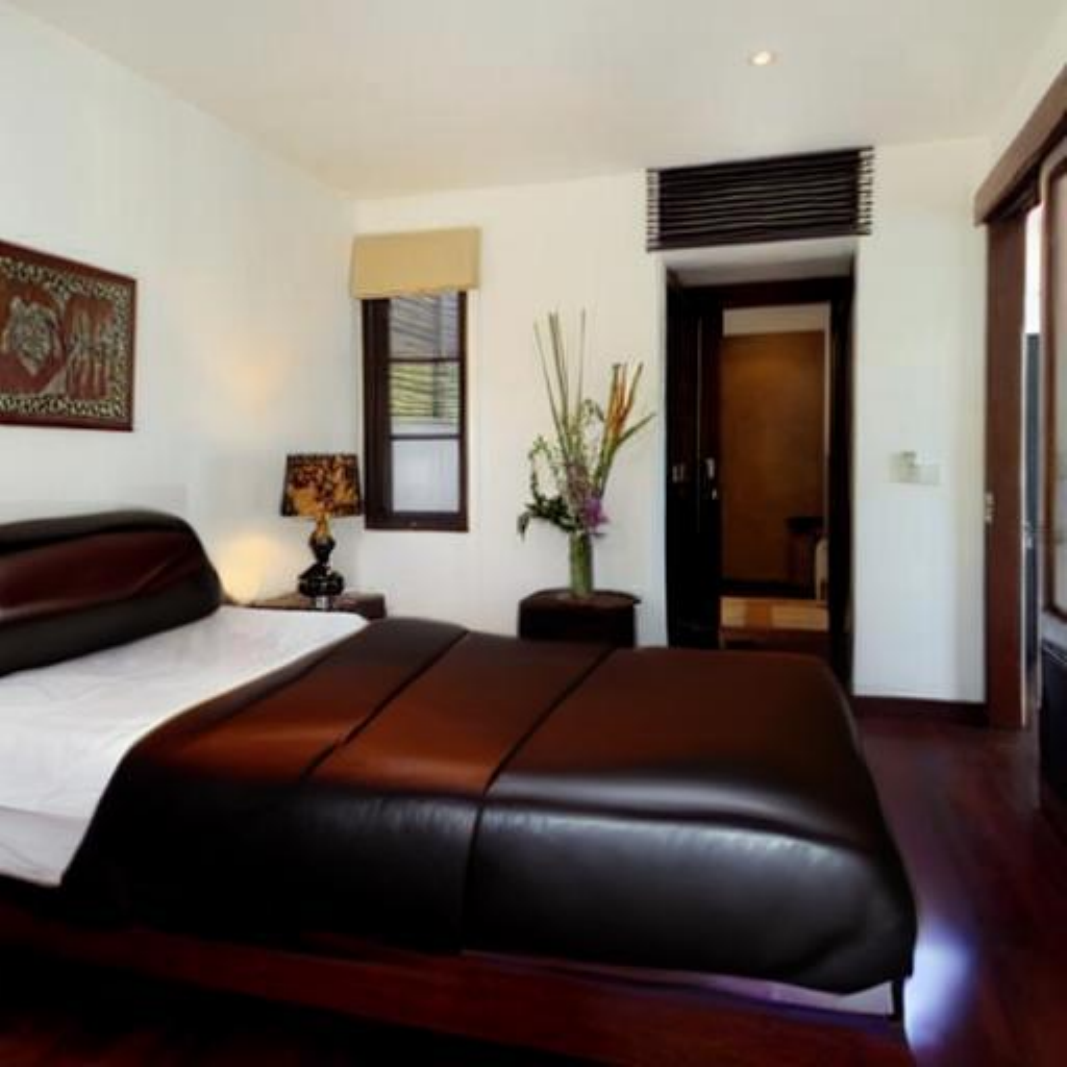}&
    \includegraphics[width=\imgwidth\textwidth]{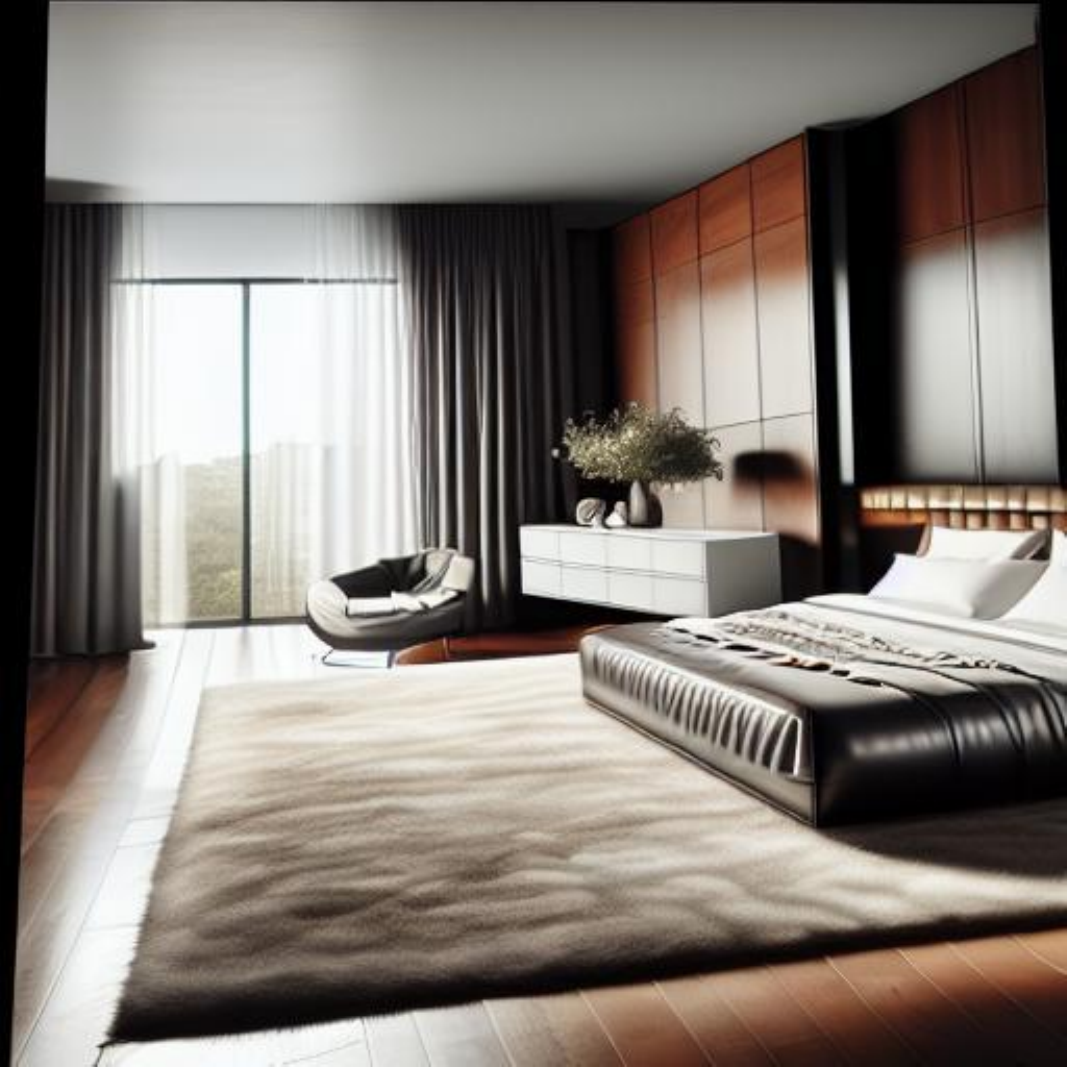}&
    \includegraphics[width=\imgwidth\textwidth]{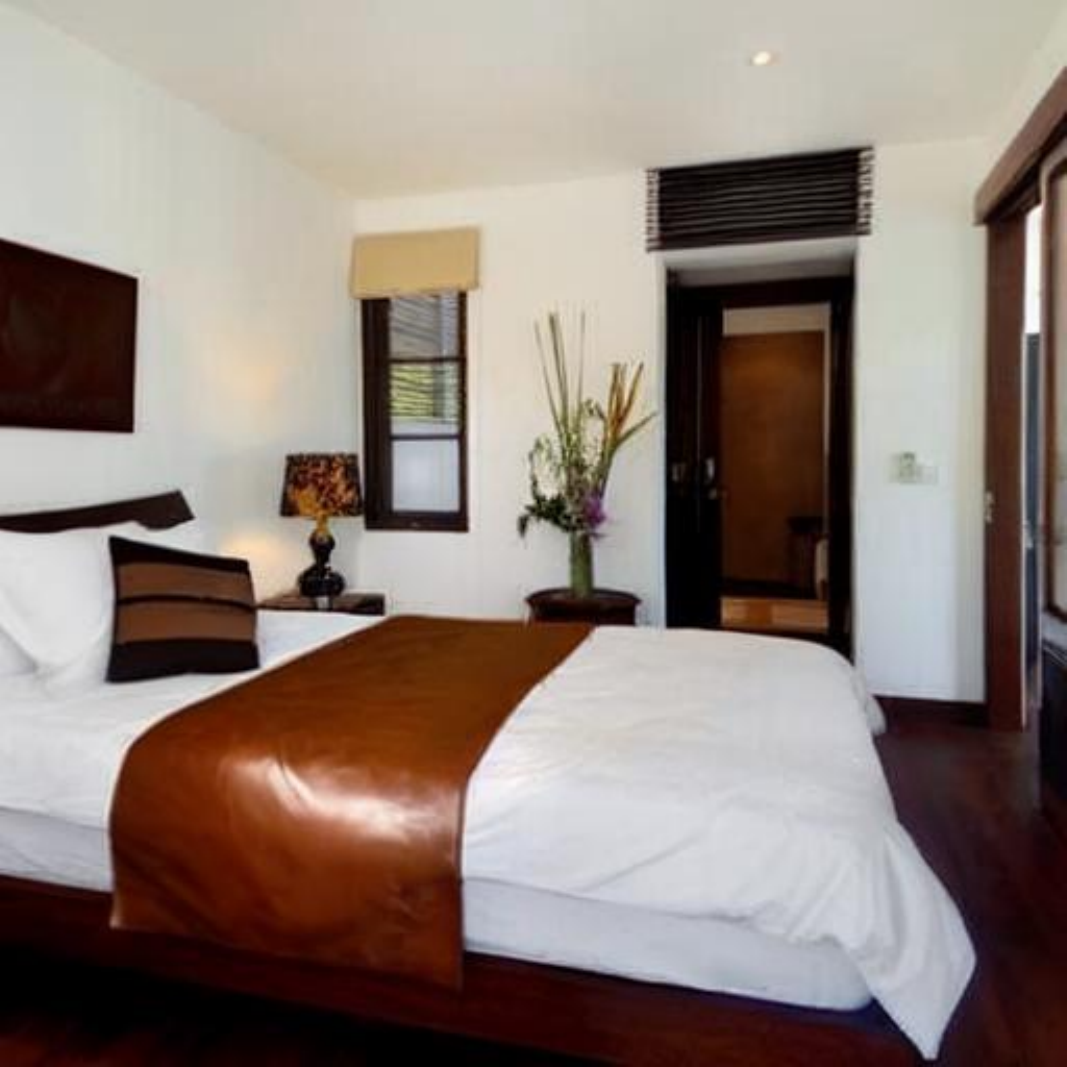}&
    \includegraphics[width=\imgwidth\textwidth]{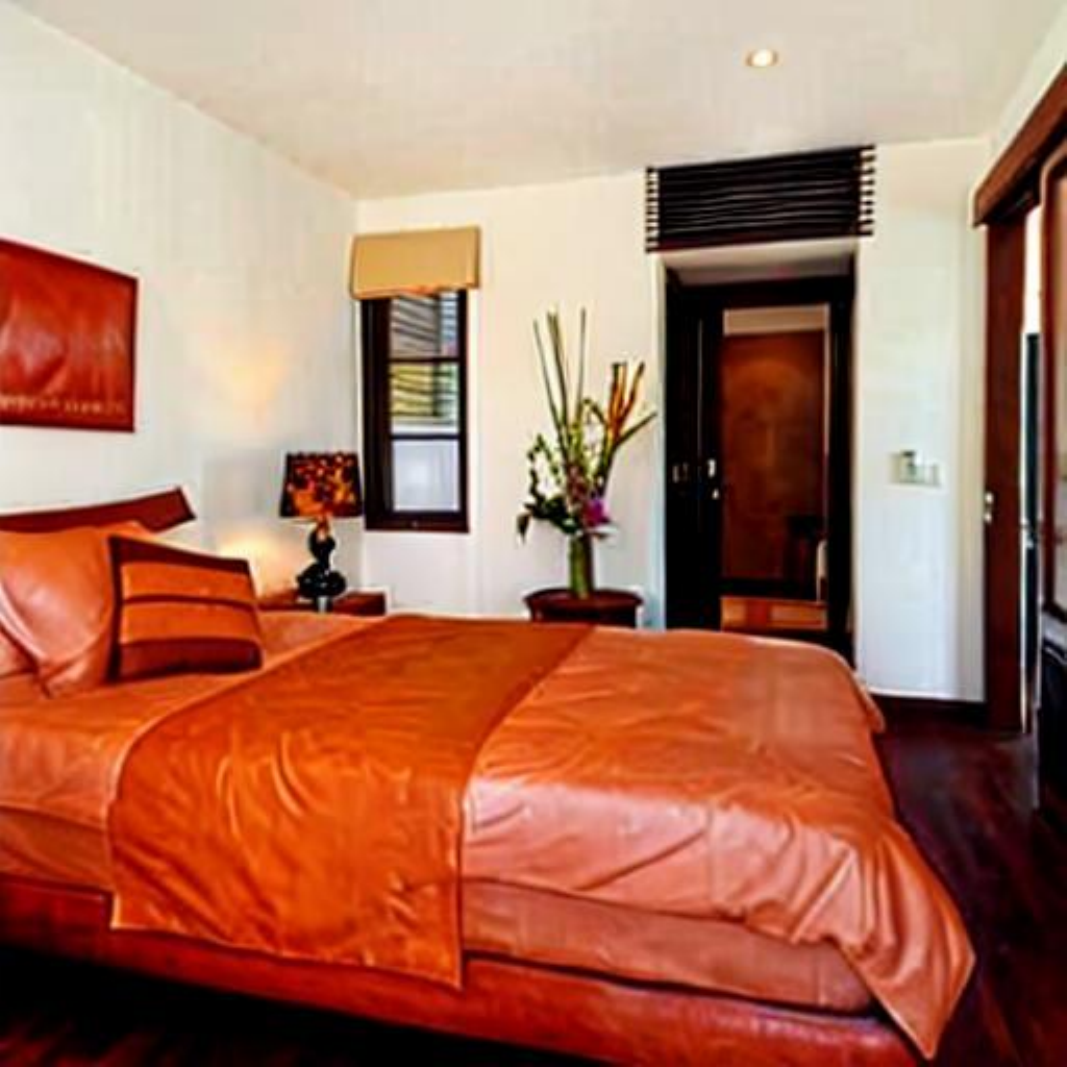} \\
    \multicolumn{6}{c}{``Make the bed out of leather"} \\[\textspace]

    \includegraphics[width=\imgwidth\textwidth]{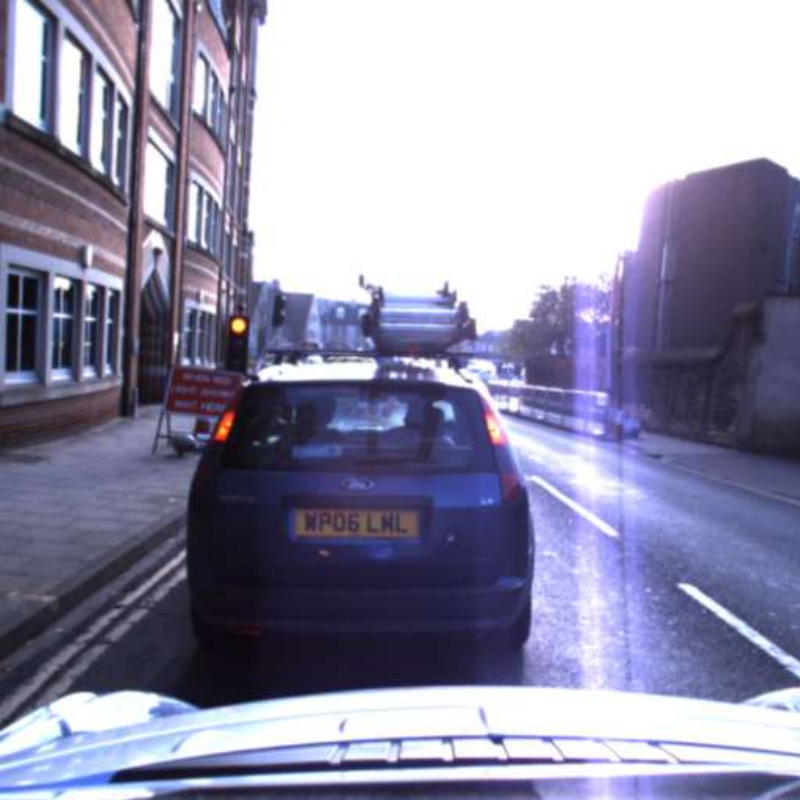} &
    \includegraphics[width=\imgwidth\textwidth]{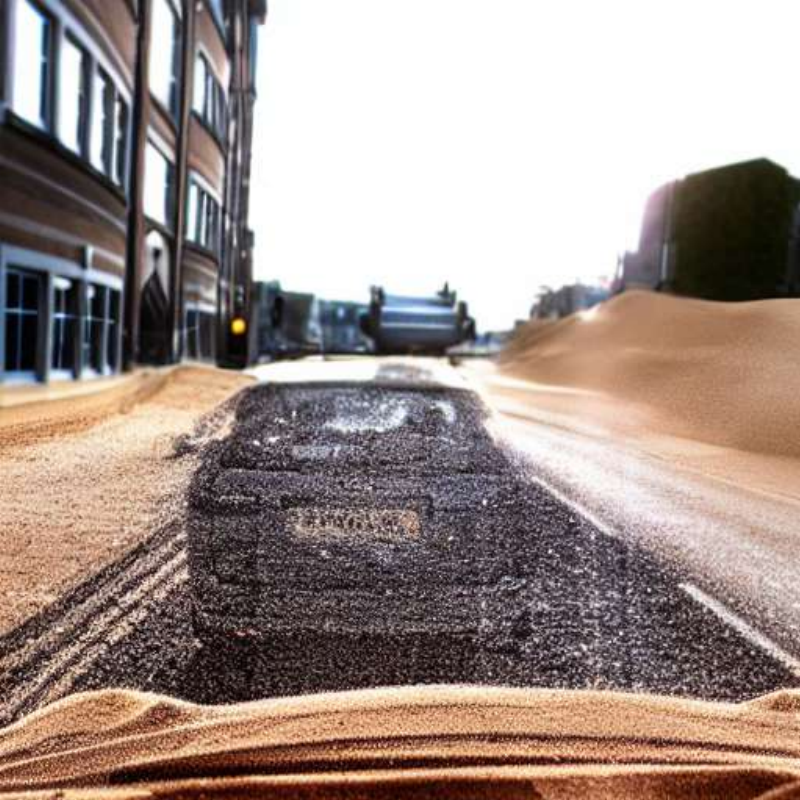} &
    \includegraphics[width=\imgwidth\textwidth]{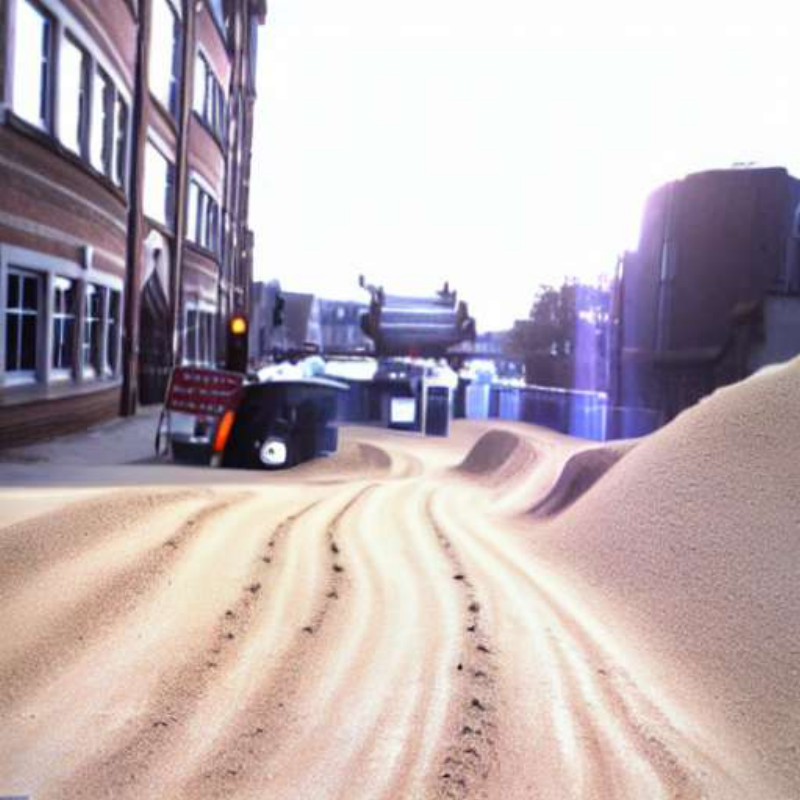} &
    \includegraphics[width=\imgwidth\textwidth]{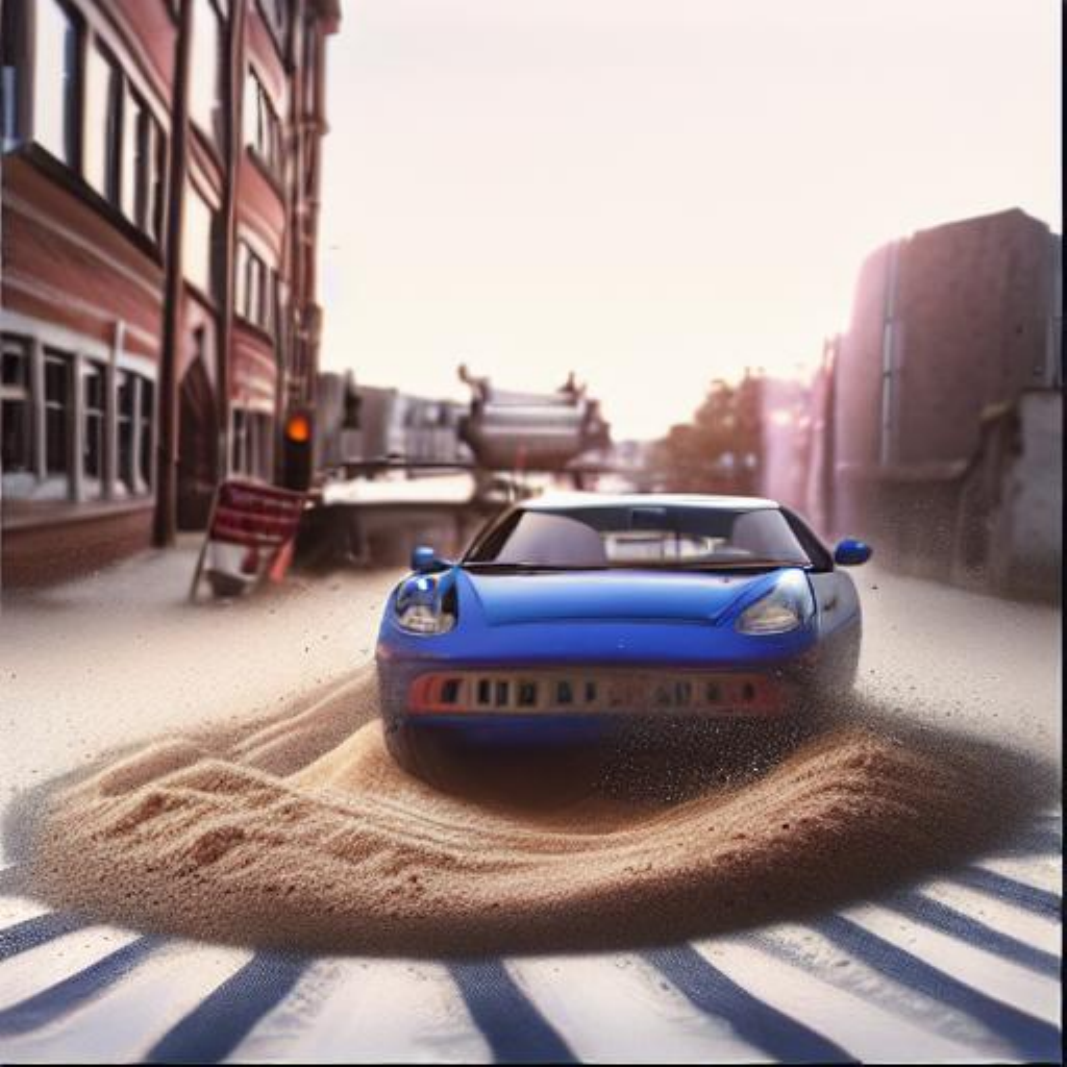} &
    \includegraphics[width=\imgwidth\textwidth]{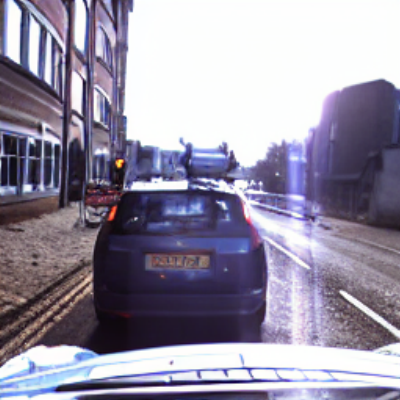} &
    \includegraphics[width=\imgwidth\textwidth]{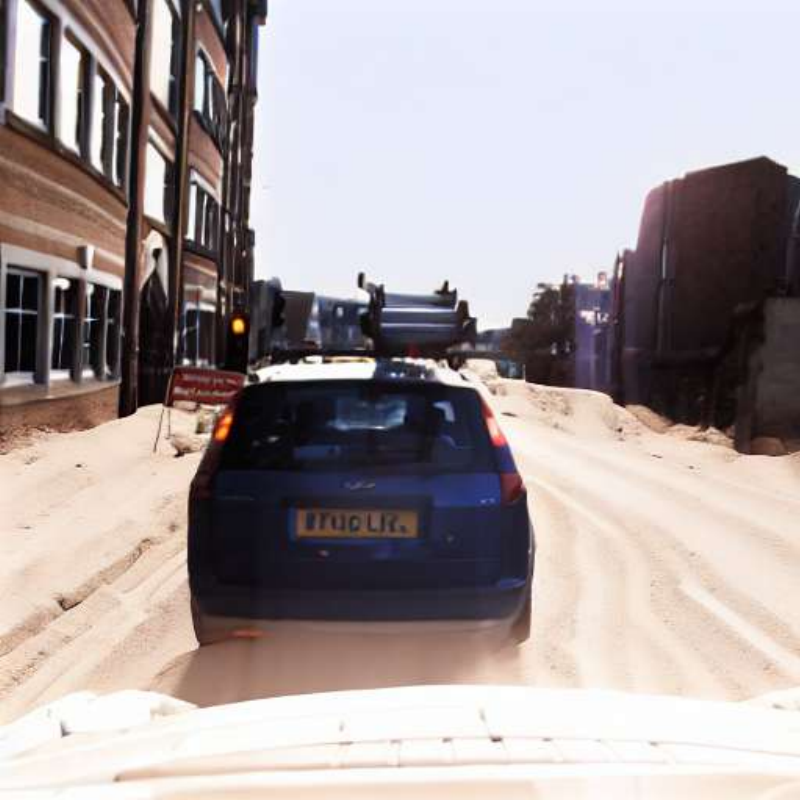} \\
    \multicolumn{6}{c}{``Add sand on the road"} \\[\textspace]

    \includegraphics[width=\imgwidth\textwidth]{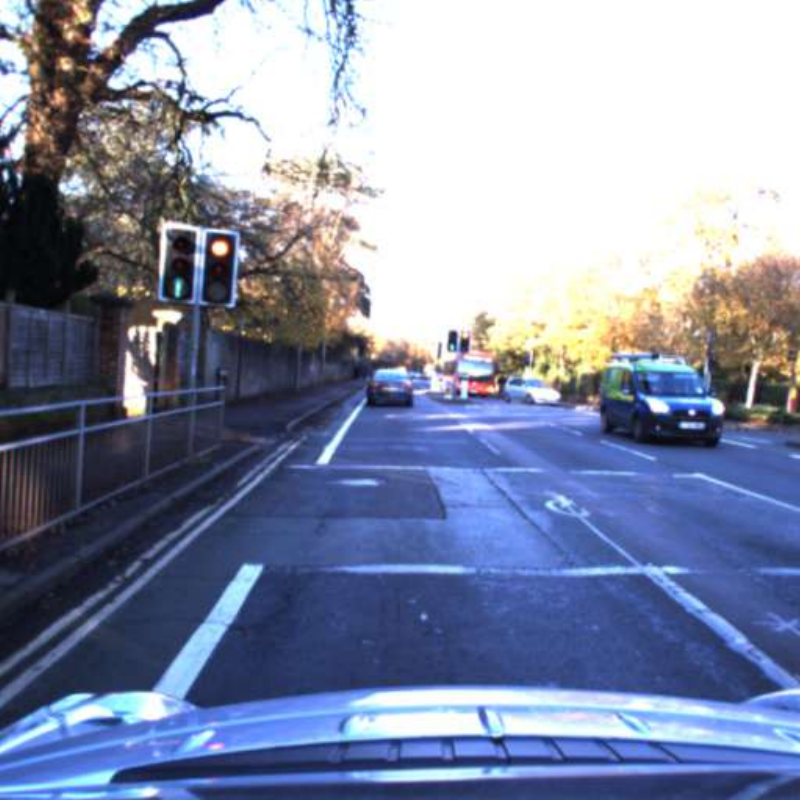} &
    \includegraphics[width=\imgwidth\textwidth]{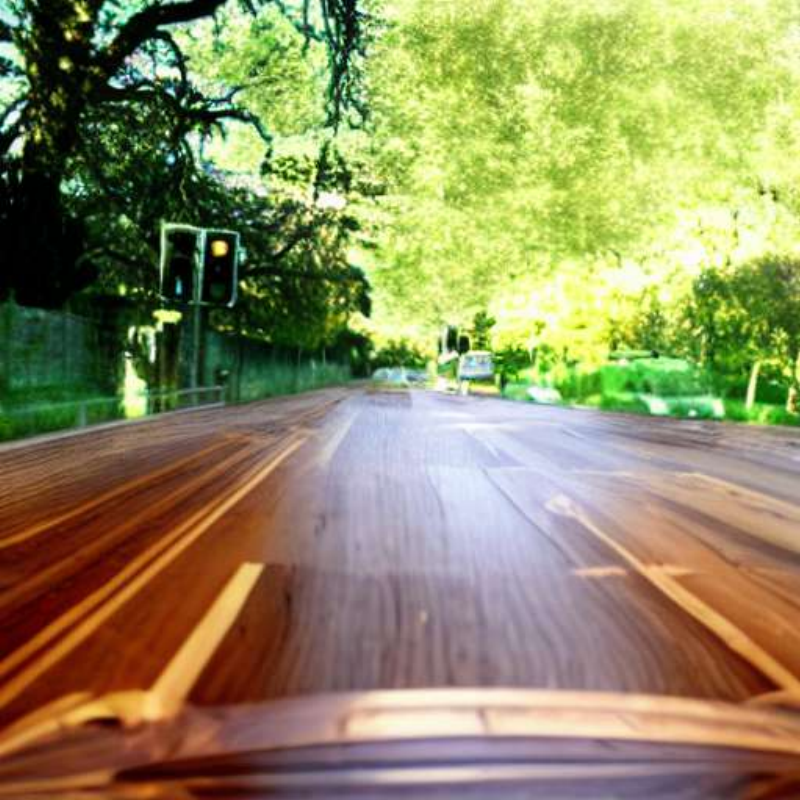} &
    \includegraphics[width=\imgwidth\textwidth]{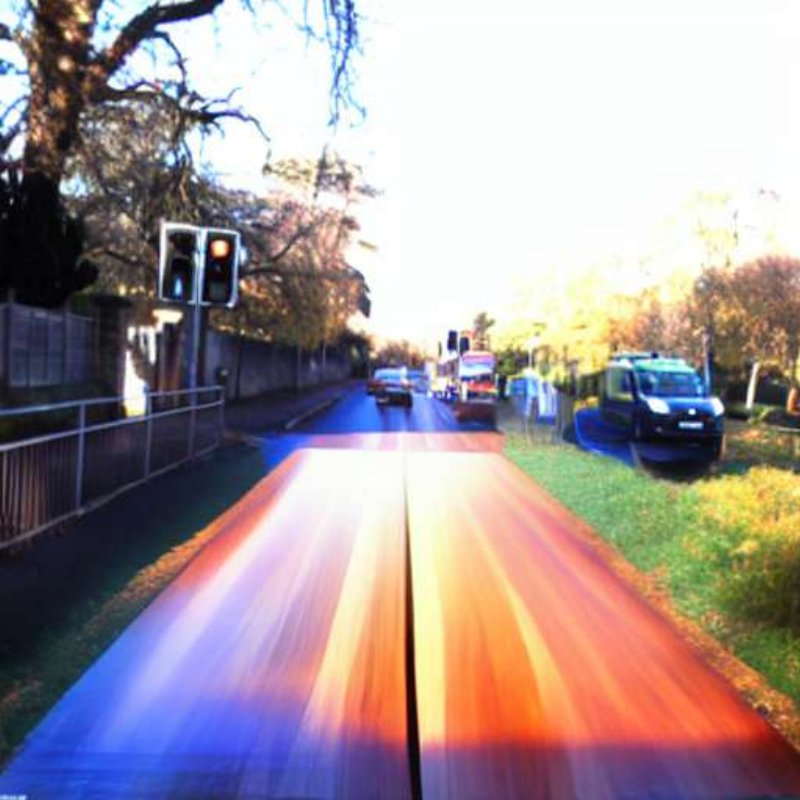} &
    \includegraphics[width=\imgwidth\textwidth]{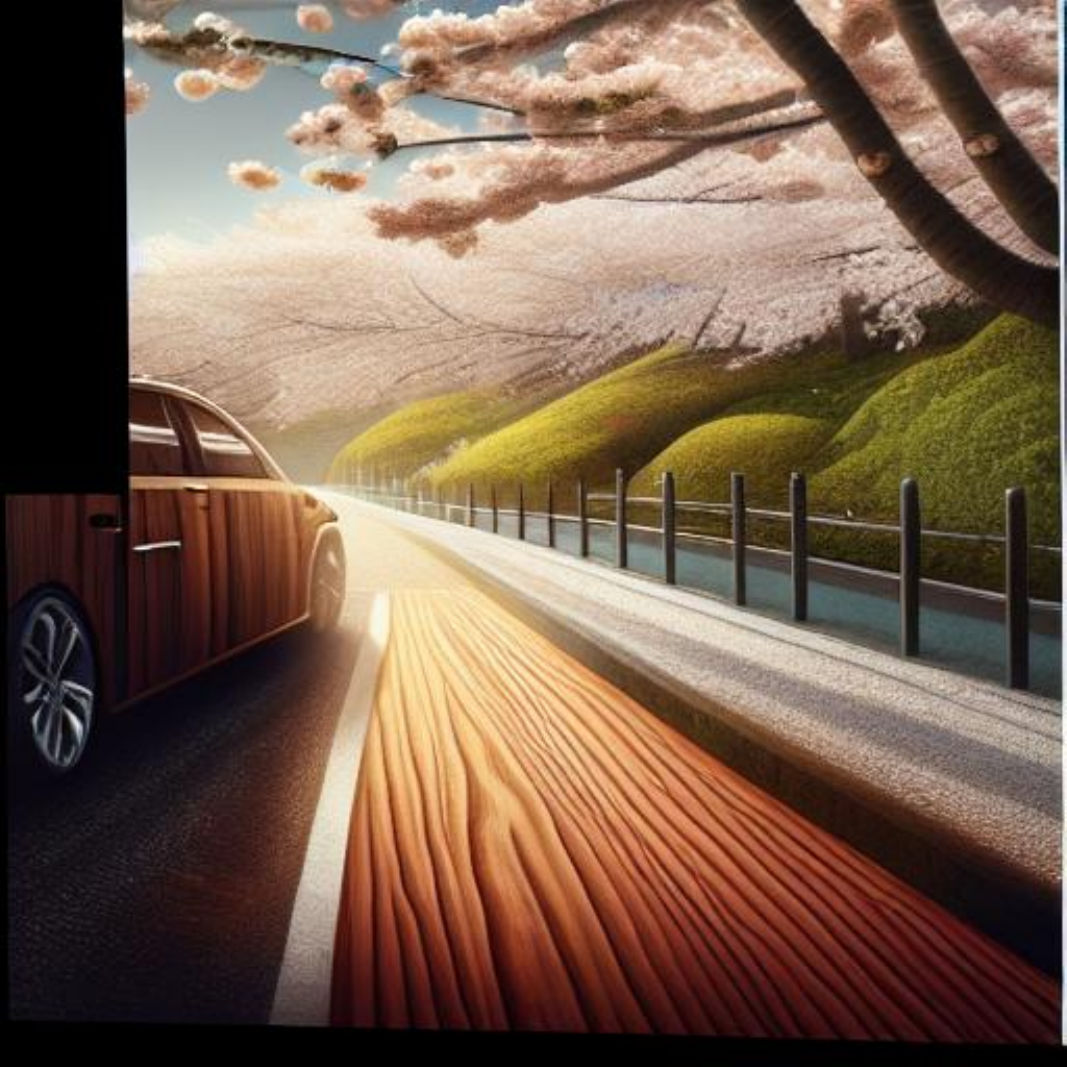} &
    \includegraphics[width=\imgwidth\textwidth]{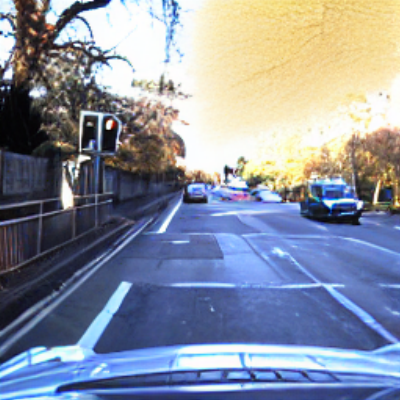} &
    \includegraphics[width=\imgwidth\textwidth]{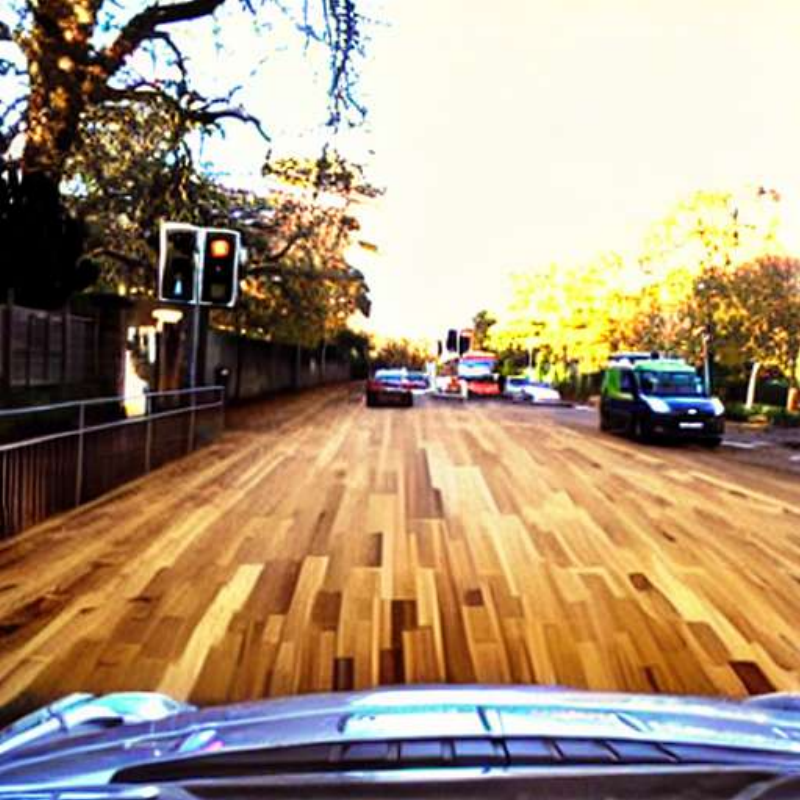} \\
    \multicolumn{6}{c}{``Change the road into wood"} \\[\textspace]

\end{tabular}
\vspace{-1em}
\caption{Comparative evaluation of our method against baselines on a diverse set of prompts and images, highlighting superior performance in structural preservation, semantic alignment, and realism.}
\label{appdx_baseline2}
\end{figure*}

\begin{figure}[t]
\centering
\newcommand{\imgwidth}{0.25\textwidth}
\setlength{\tabcolsep}{1pt} 
\renewcommand{\arraystretch}{0.25} 
\newcommand{\stycompression}{0.05}
\newcommand{\styposition}{62}
\begin{tabular}{ccc}
    Input & White Marble & Black Marble \\ [0.5em]
    \includegraphics[width=\imgwidth]{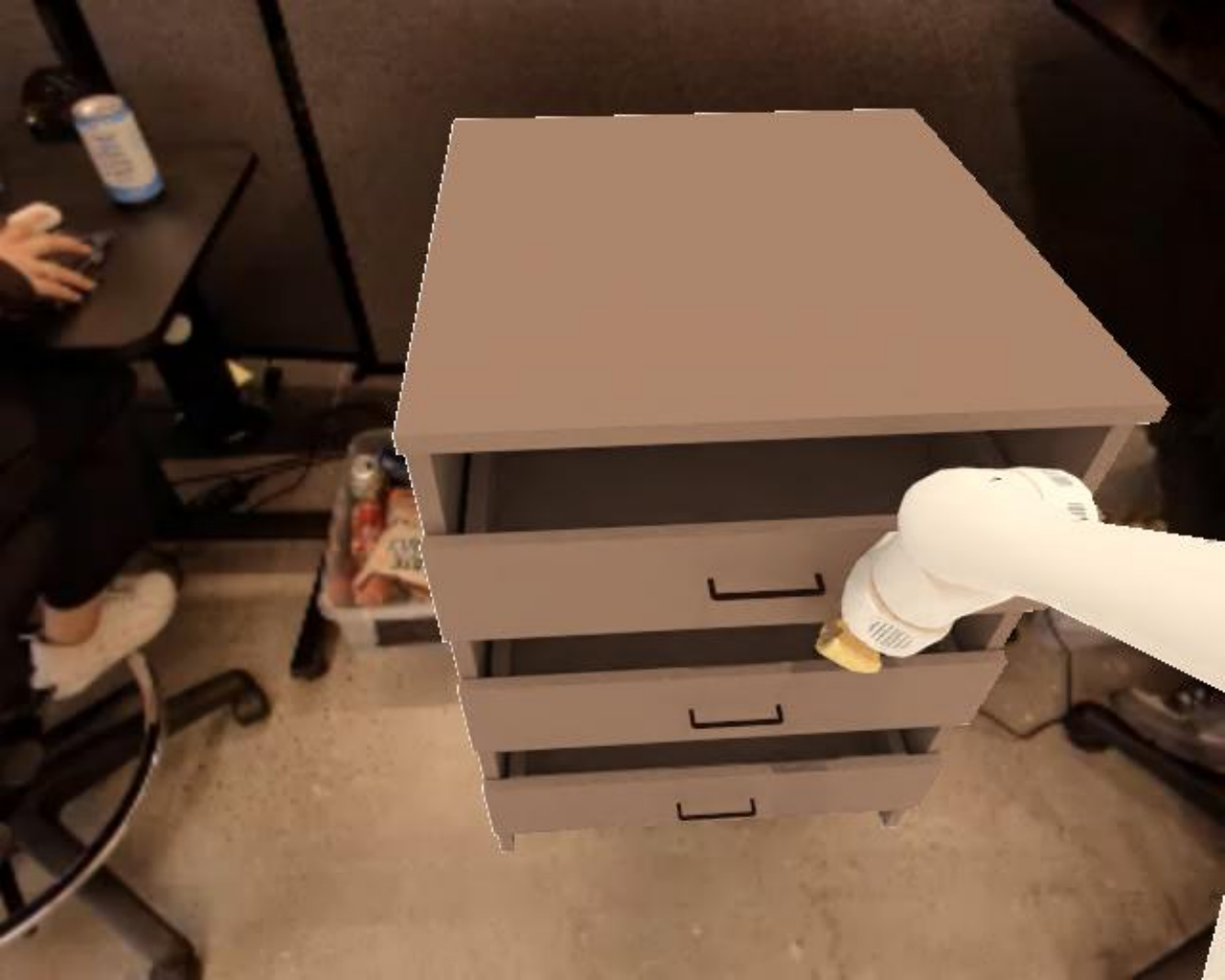} &
    \begin{overpic}[width=\imgwidth]{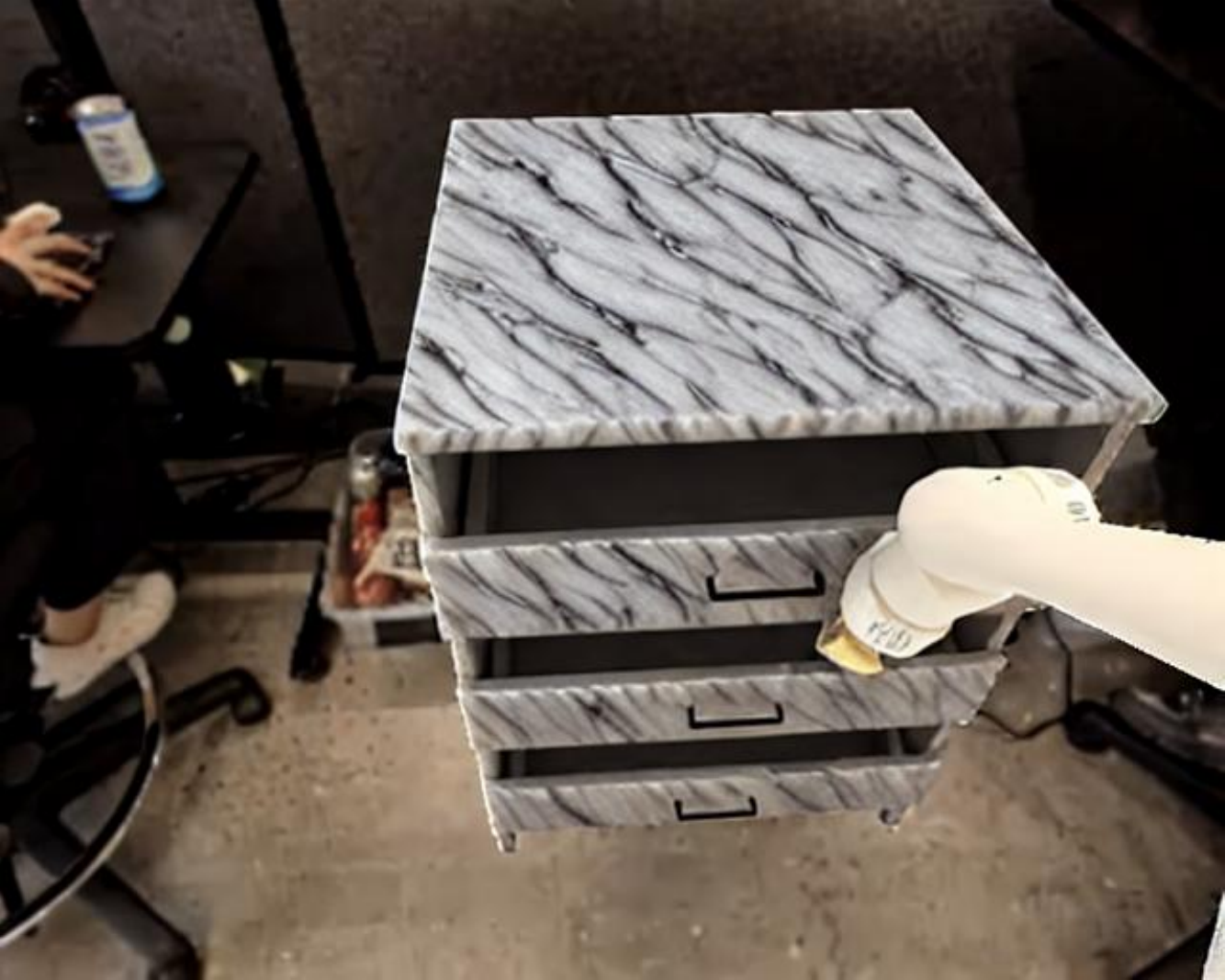}
        \put(1, 2){
         \begin{tikzpicture}
          \begin{scope}
           \clip[rounded corners=5pt] (0,0) rectangle (\stycompression\textwidth, \stycompression\textwidth);
           \node[anchor=north east, inner sep=0pt] at (\stycompression\textwidth,\stycompression\textwidth) {\includegraphics[width=\stycompression\textwidth]{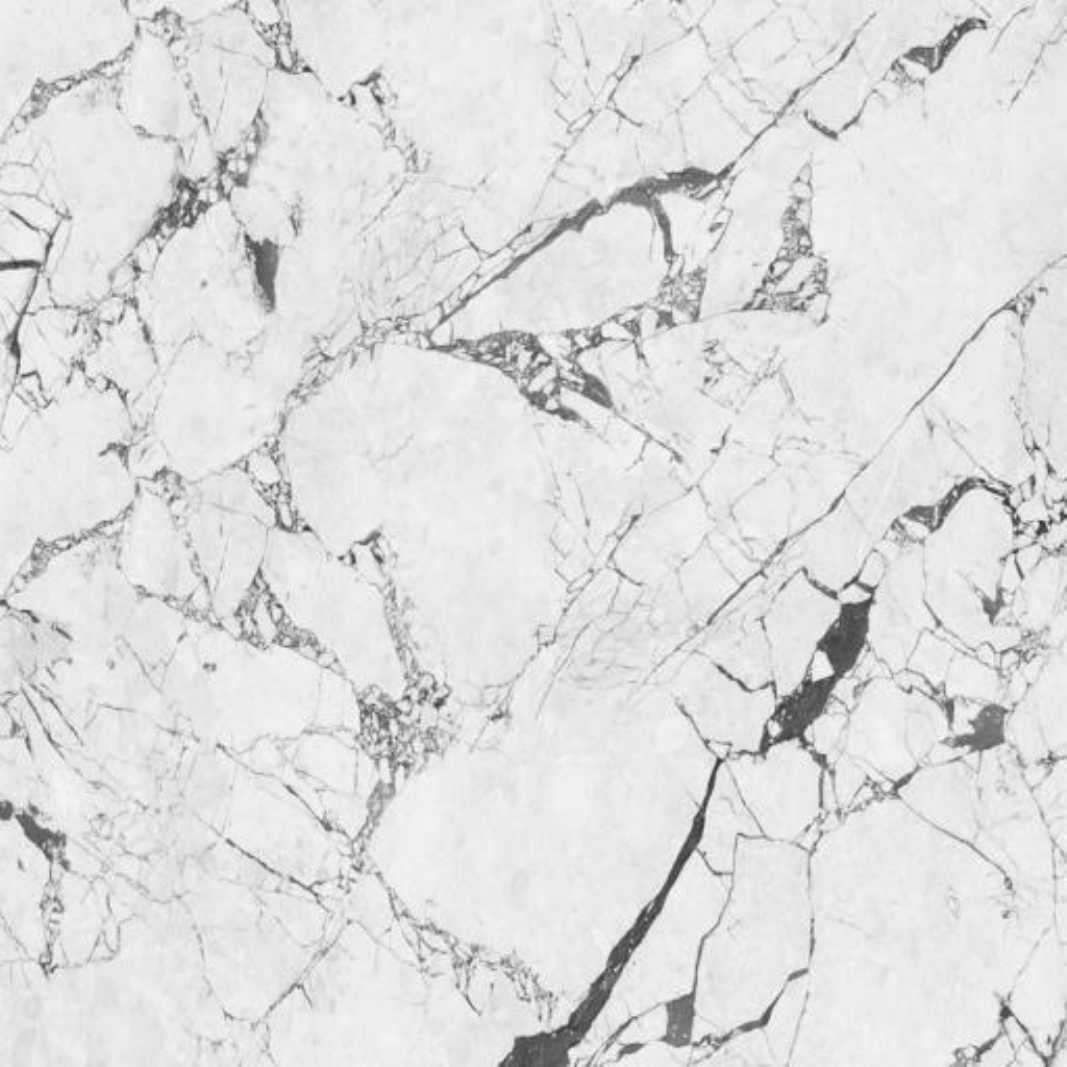}};
        \end{scope}
        \draw[rounded corners=5pt, RubineRed, very thick, dashed] (0,0) rectangle (\stycompression\textwidth,\stycompression\textwidth);
        \end{tikzpicture}}
    \end{overpic}&
    \begin{overpic}[width=\imgwidth]{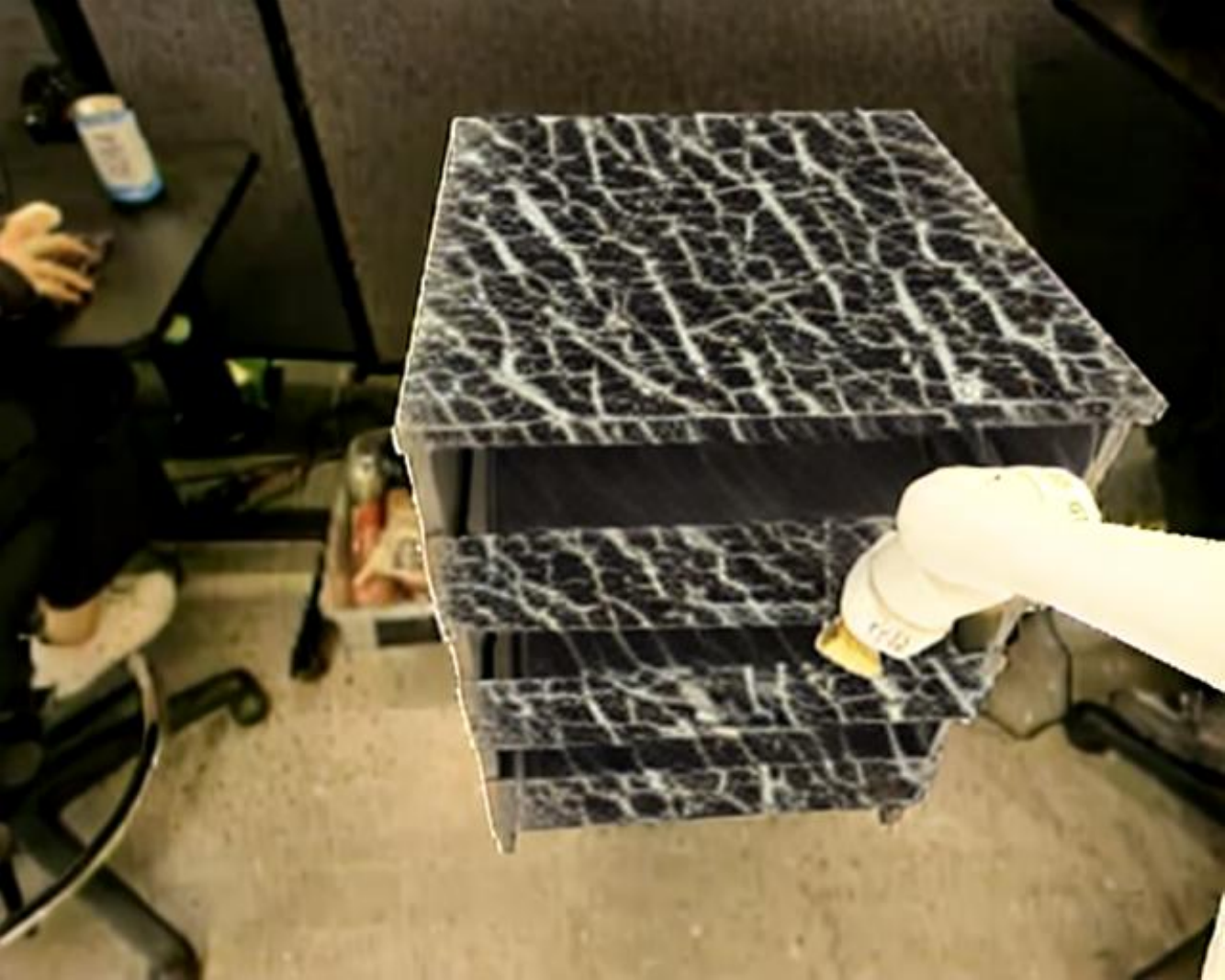}
        \put(1, 2){
         \begin{tikzpicture}
          \begin{scope}
           \clip[rounded corners=5pt] (0,0) rectangle (\stycompression\textwidth, \stycompression\textwidth);
           \node[anchor=north east, inner sep=0pt] at (\stycompression\textwidth,\stycompression\textwidth) {\includegraphics[width=\stycompression\textwidth]{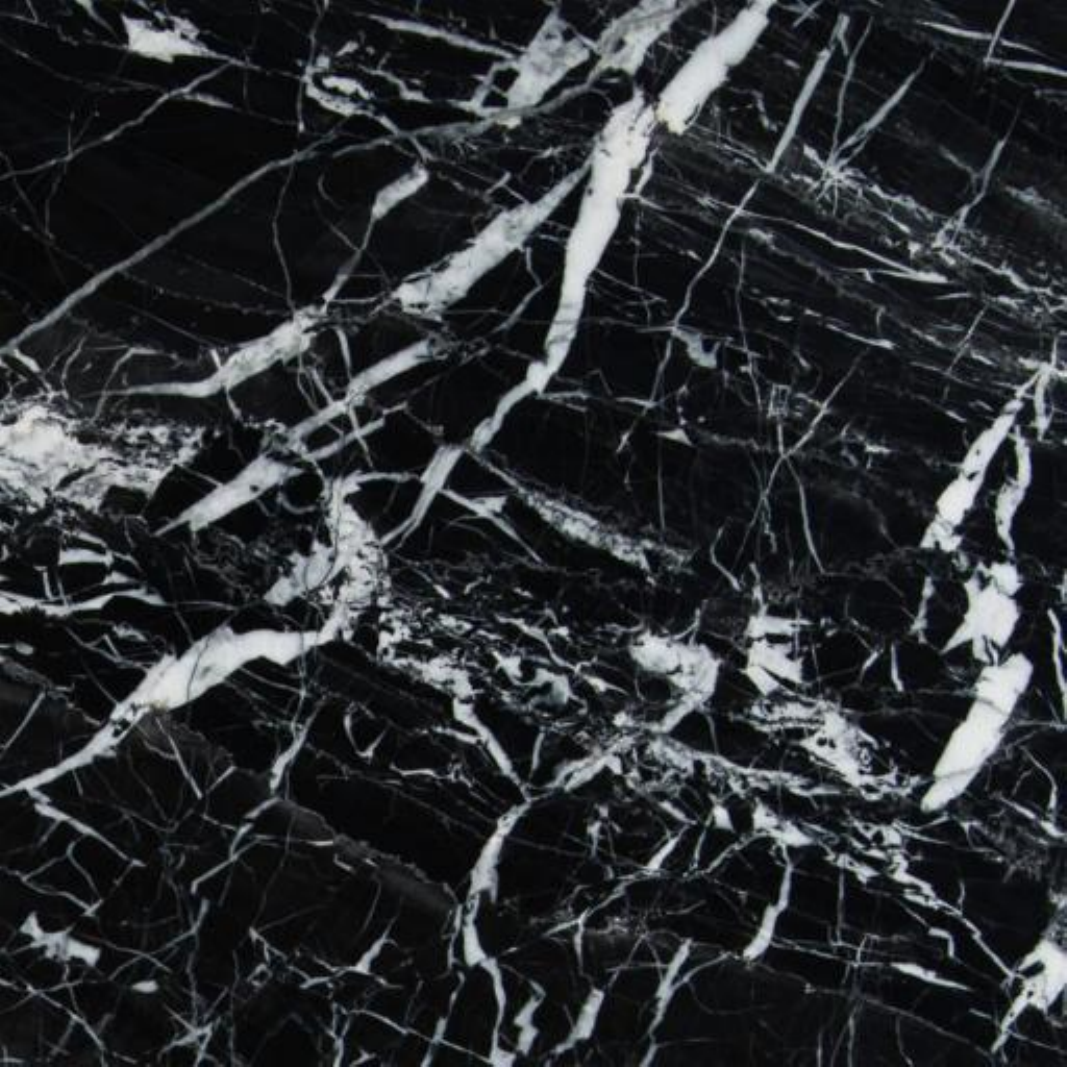}};
        \end{scope}
        \draw[rounded corners=5pt, RubineRed, very thick, dashed] (0,0) rectangle (\stycompression\textwidth,\stycompression\textwidth);
        \end{tikzpicture}}
    \end{overpic} \\[.25em]
    \includegraphics[width=\imgwidth]{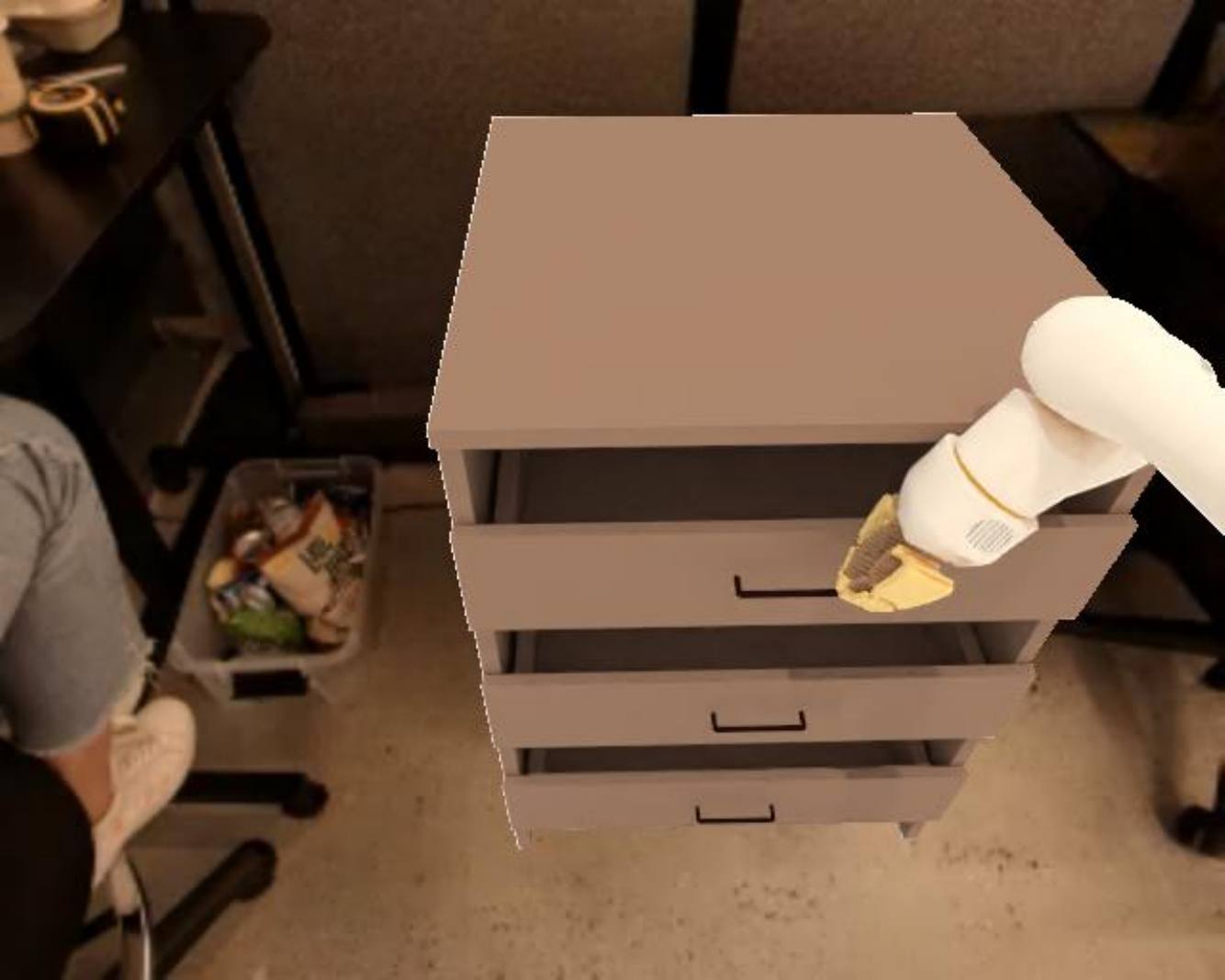} &
    \begin{overpic}[width=\imgwidth]{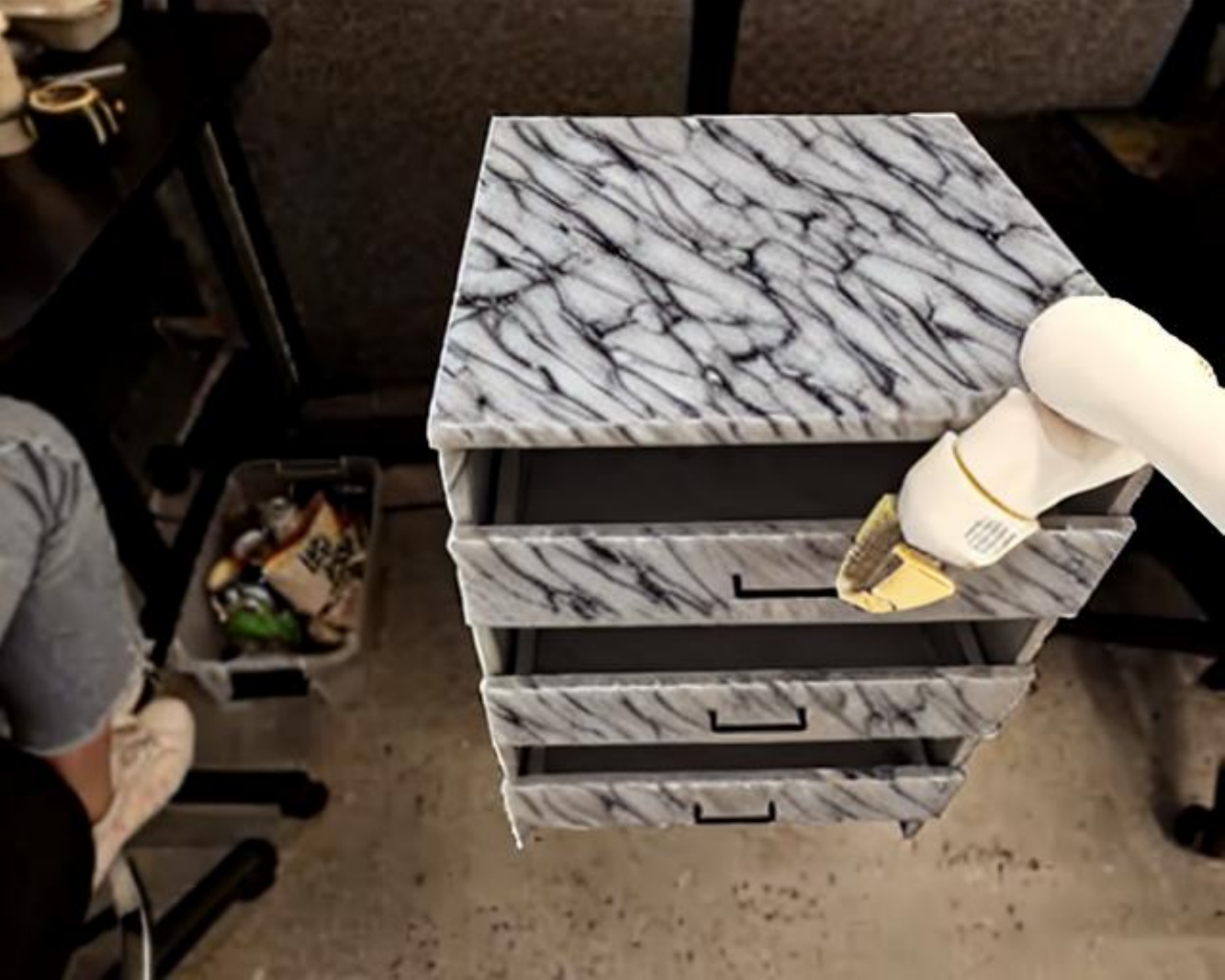}
        \put(1, 50){
         \begin{tikzpicture}
          \begin{scope}
           \clip[rounded corners=5pt] (0,0) rectangle (\stycompression\textwidth, \stycompression\textwidth);
           \node[anchor=north east, inner sep=0pt] at (\stycompression\textwidth,\stycompression\textwidth) {\includegraphics[width=\stycompression\textwidth]{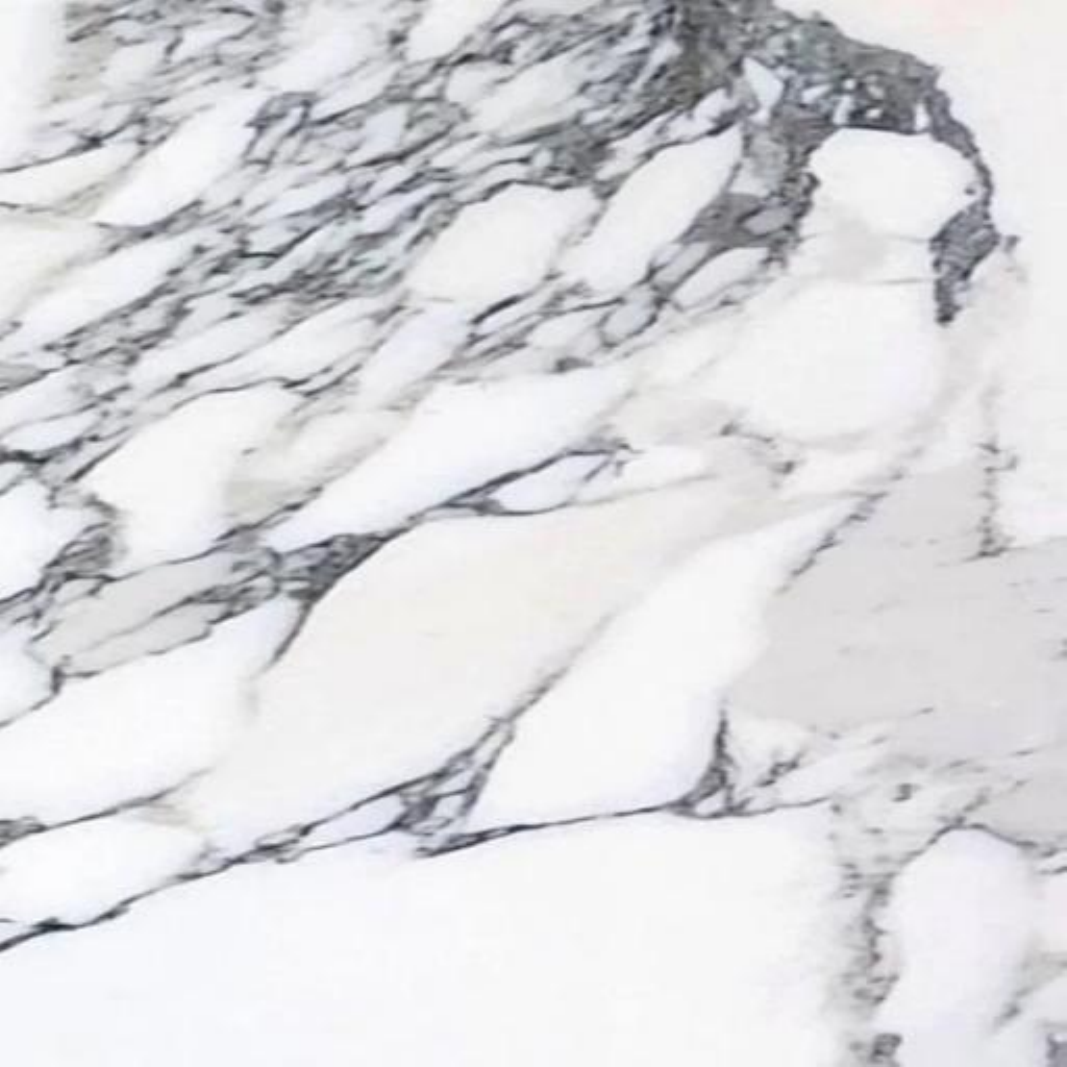}};
        \end{scope}
        \draw[rounded corners=5pt, RubineRed, very thick, dashed] (0,0) rectangle (\stycompression\textwidth,\stycompression\textwidth);
        \end{tikzpicture}}
    \end{overpic}&
    \begin{overpic}[width=\imgwidth]{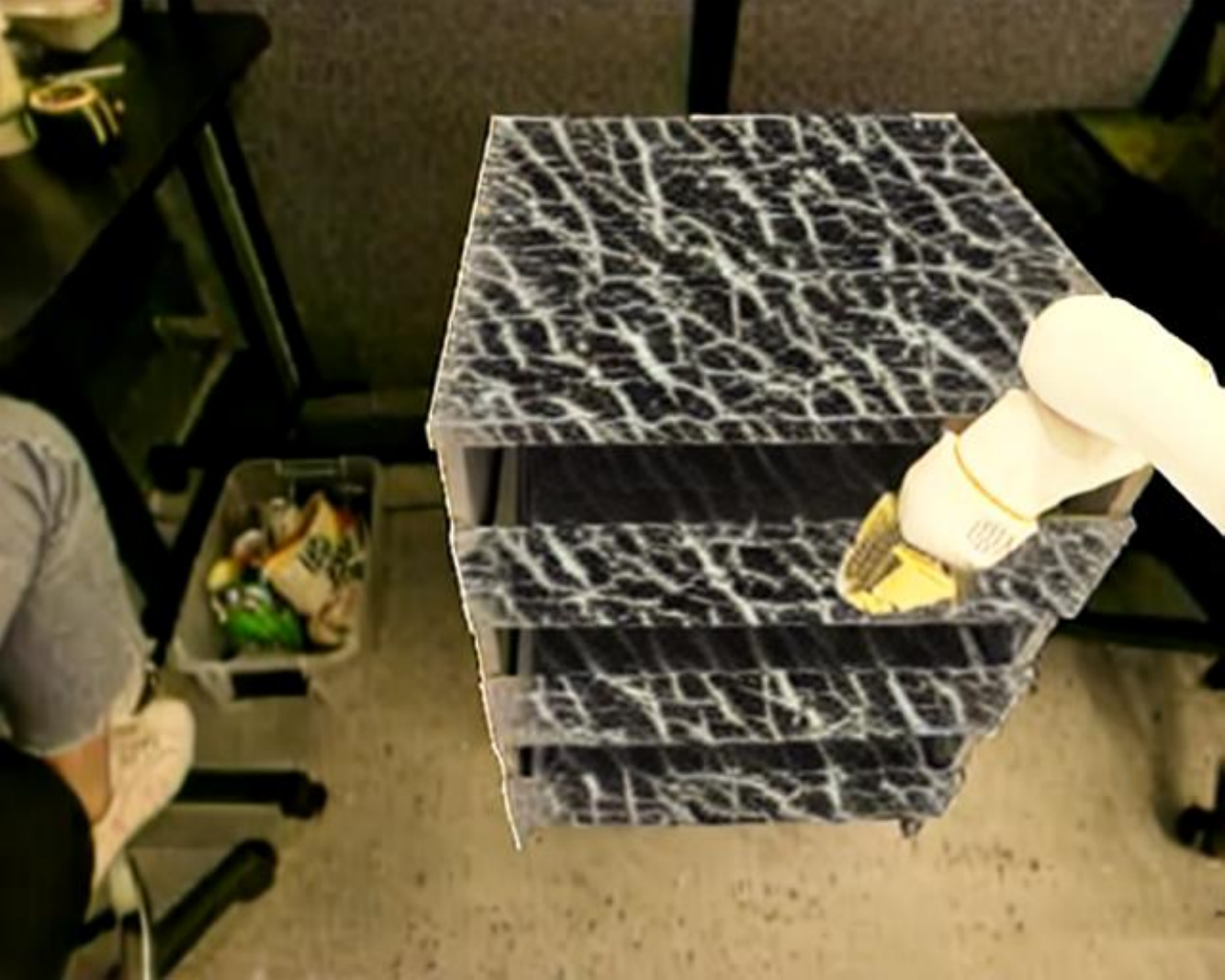}
        \put(1, 50){
         \begin{tikzpicture}
          \begin{scope}
           \clip[rounded corners=5pt] (0,0) rectangle (\stycompression\textwidth, \stycompression\textwidth);
           \node[anchor=north east, inner sep=0pt] at (\stycompression\textwidth,\stycompression\textwidth) {\includegraphics[width=\stycompression\textwidth]{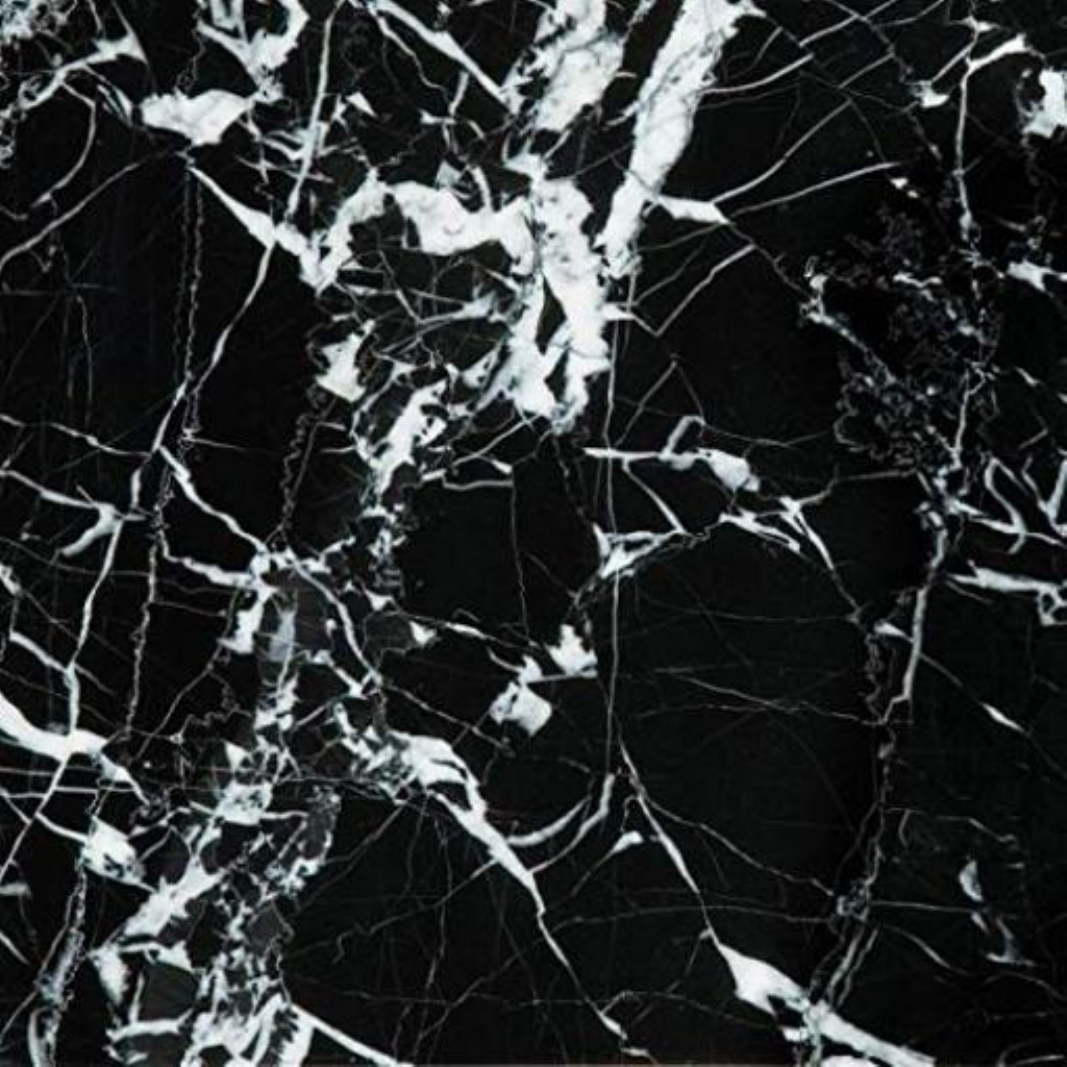}};
        \end{scope}
        \draw[rounded corners=5pt, RubineRed, very thick, dashed] (0,0) rectangle (\stycompression\textwidth,\stycompression\textwidth);
        \end{tikzpicture}}
    \end{overpic} \\[2em]

    Input & Sparse Snow & Dense Snow \\[0.5em]
    \includegraphics[width=\imgwidth]{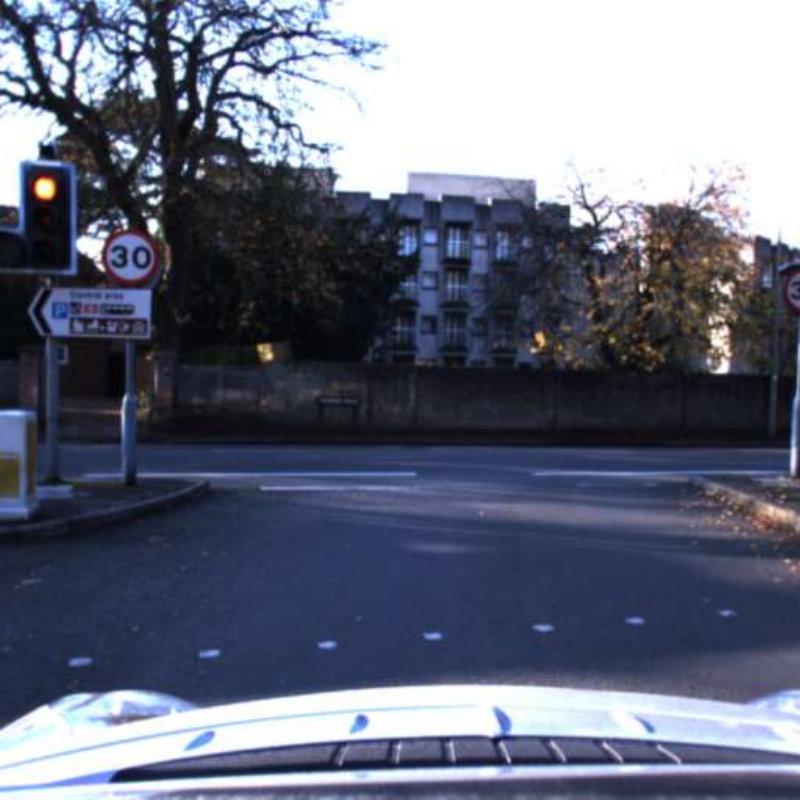} &
    \begin{overpic}[width=\imgwidth]{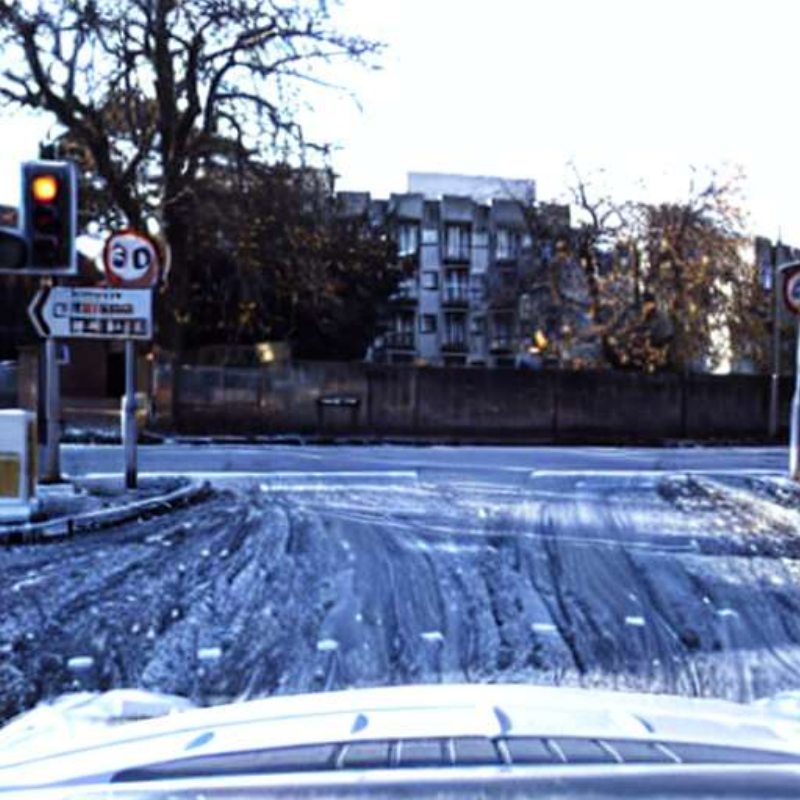}
        \put(73, 75){
         \begin{tikzpicture}
          \begin{scope}
           \clip[rounded corners=5pt] (0,0) rectangle (\stycompression\textwidth, \stycompression\textwidth);
           \node[anchor=north east, inner sep=0pt] at (\stycompression\textwidth,\stycompression\textwidth) {\includegraphics[width=\stycompression\textwidth]{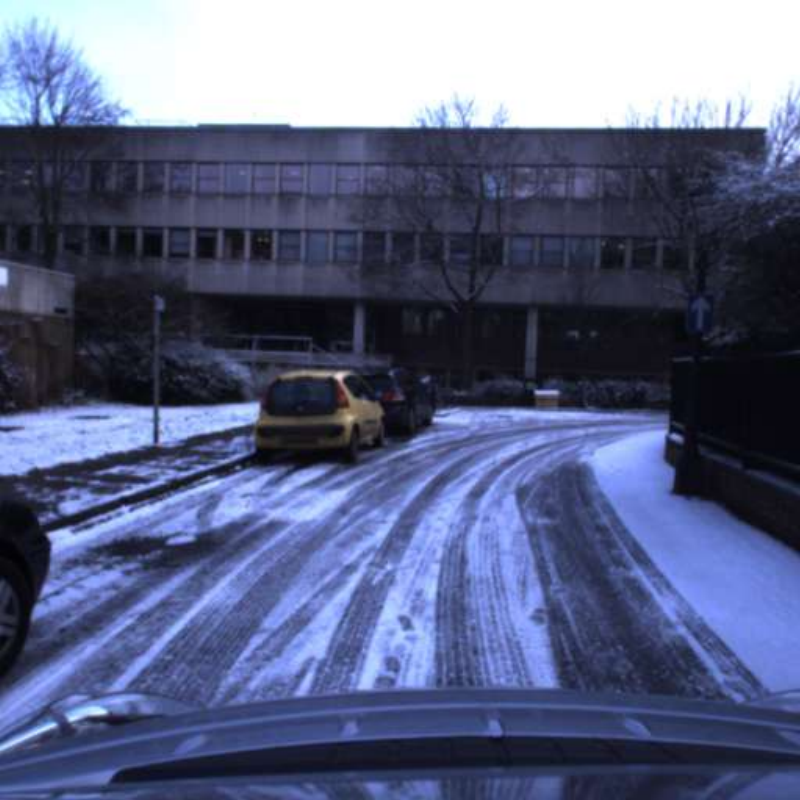}};
        \end{scope}
        \draw[rounded corners=5pt, RubineRed, very thick, dashed] (0,0) rectangle (\stycompression\textwidth,\stycompression\textwidth);
        \end{tikzpicture}}
    \end{overpic}&
    \begin{overpic}[width=\imgwidth]{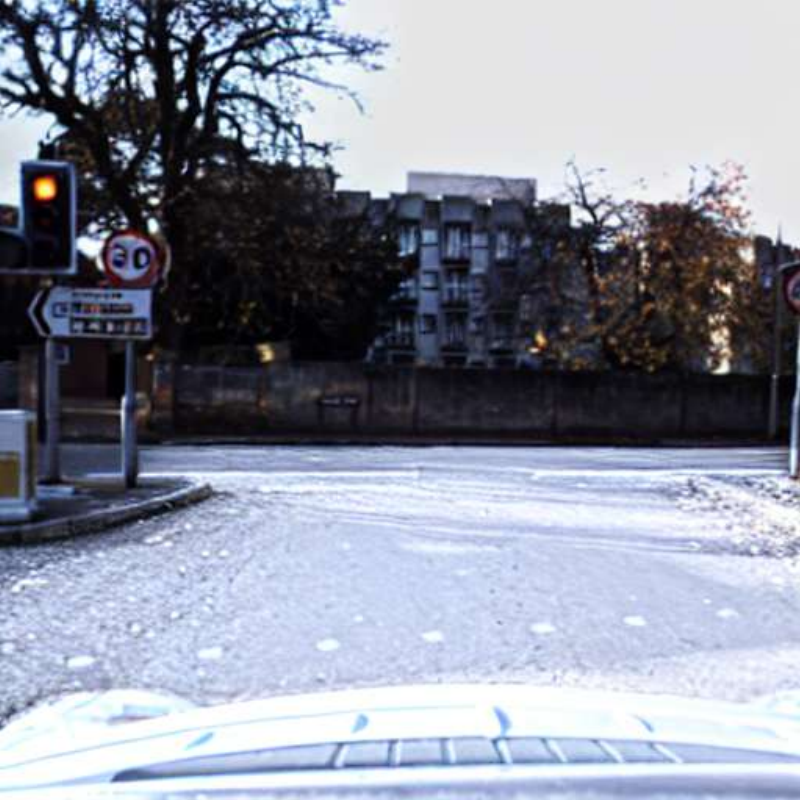}
        \put(73, 75){
         \begin{tikzpicture}
          \begin{scope}
           \clip[rounded corners=5pt] (0,0) rectangle (\stycompression\textwidth, \stycompression\textwidth);
           \node[anchor=north east, inner sep=0pt] at (\stycompression\textwidth,\stycompression\textwidth) {\includegraphics[width=\stycompression\textwidth]{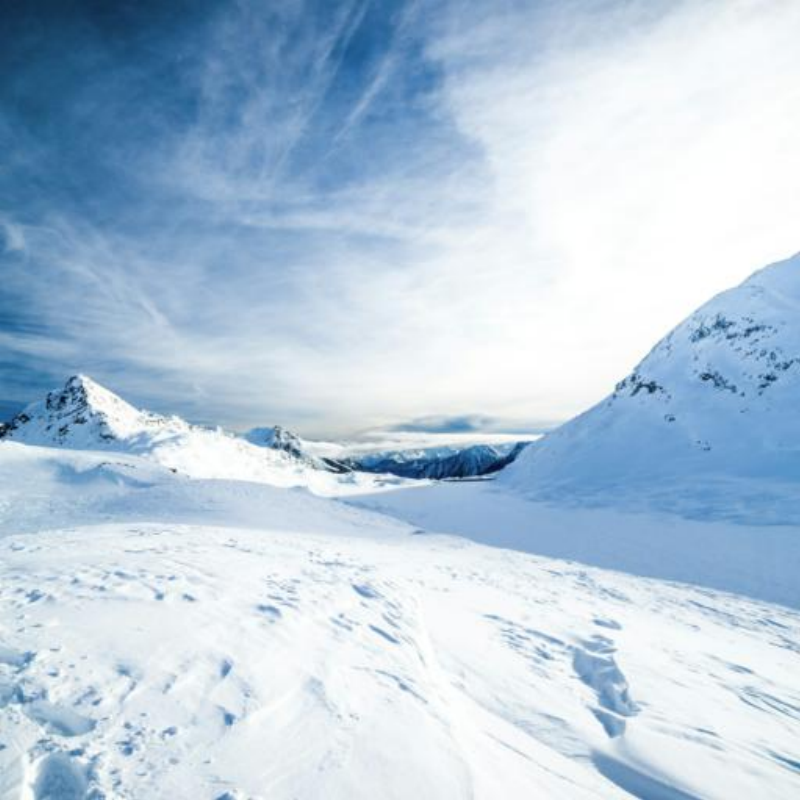}};
        \end{scope}
        \draw[rounded corners=5pt, RubineRed, very thick, dashed] (0,0) rectangle (\stycompression\textwidth,\stycompression\textwidth);
        \end{tikzpicture}}
    \end{overpic} \\[.25em]
    \includegraphics[width=\imgwidth]{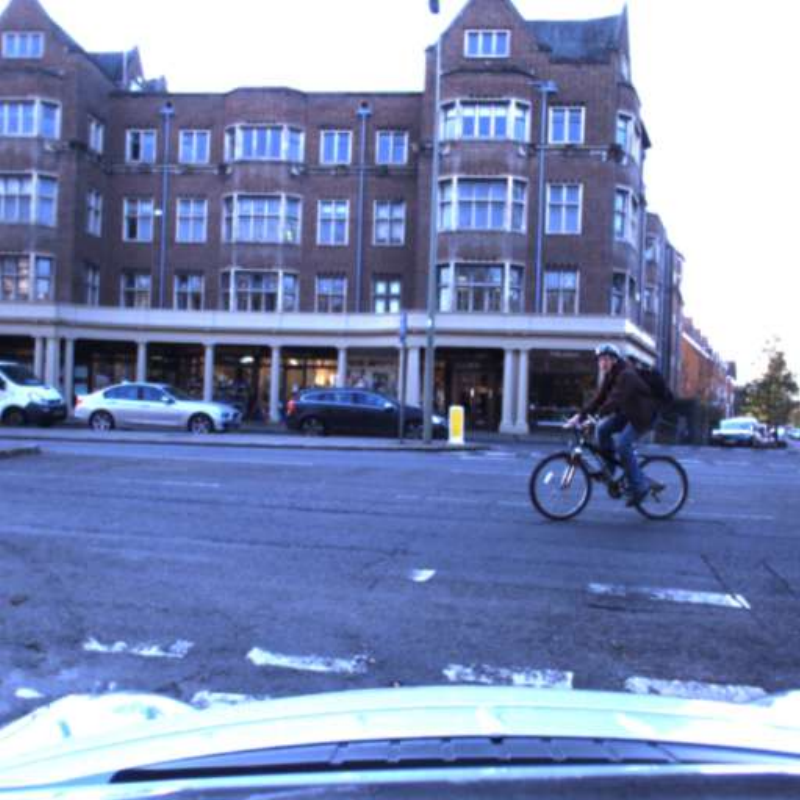} &
    \begin{overpic}[width=\imgwidth]{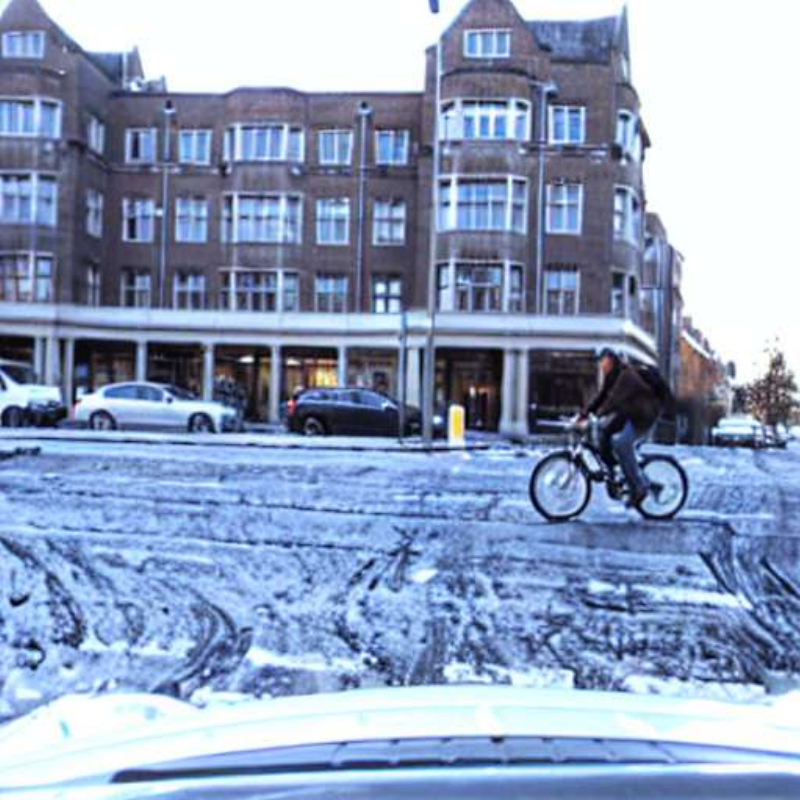}
        \put(73, 75){
         \begin{tikzpicture}
          \begin{scope}
           \clip[rounded corners=5pt] (0,0) rectangle (\stycompression\textwidth, \stycompression\textwidth);
           \node[anchor=north east, inner sep=0pt] at (\stycompression\textwidth,\stycompression\textwidth) {\includegraphics[width=\stycompression\textwidth]{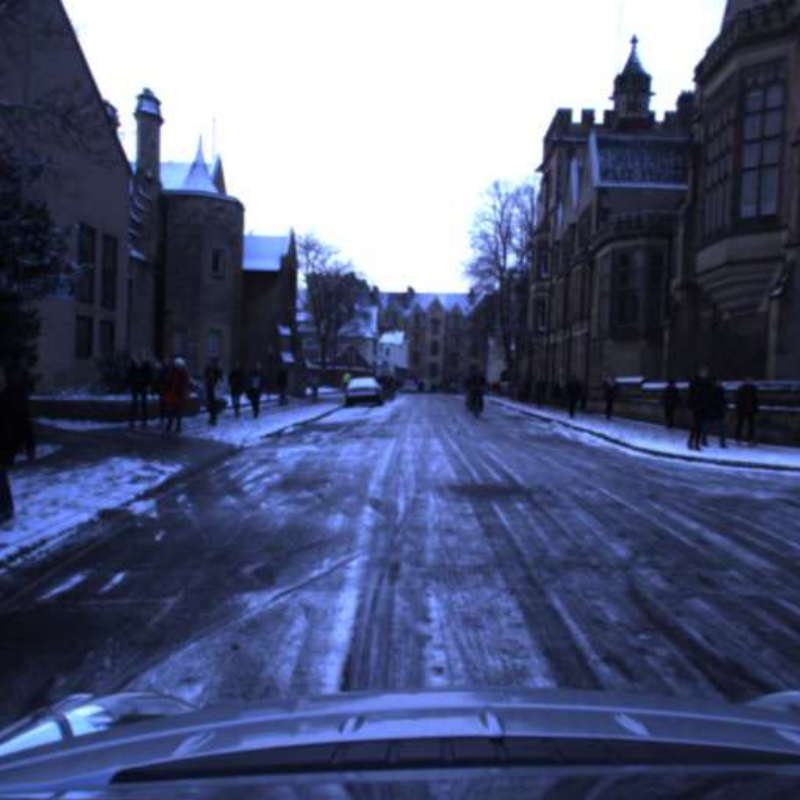}};
        \end{scope}
        \draw[rounded corners=5pt, RubineRed, very thick, dashed] (0,0) rectangle (\stycompression\textwidth,\stycompression\textwidth);
        \end{tikzpicture}}
    \end{overpic} &
    \begin{overpic}[width=\imgwidth]{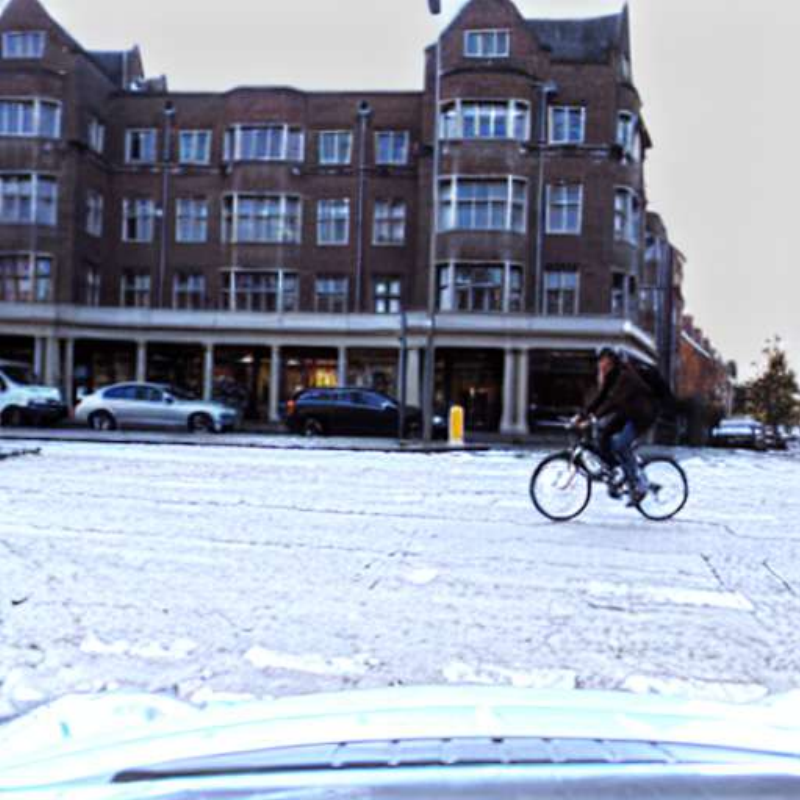}
        \put(73, 75){
         \begin{tikzpicture}
          \begin{scope}
           \clip[rounded corners=5pt] (0,0) rectangle (\stycompression\textwidth, \stycompression\textwidth);
           \node[anchor=north east, inner sep=0pt] at (\stycompression\textwidth,\stycompression\textwidth) {\includegraphics[width=\stycompression\textwidth]{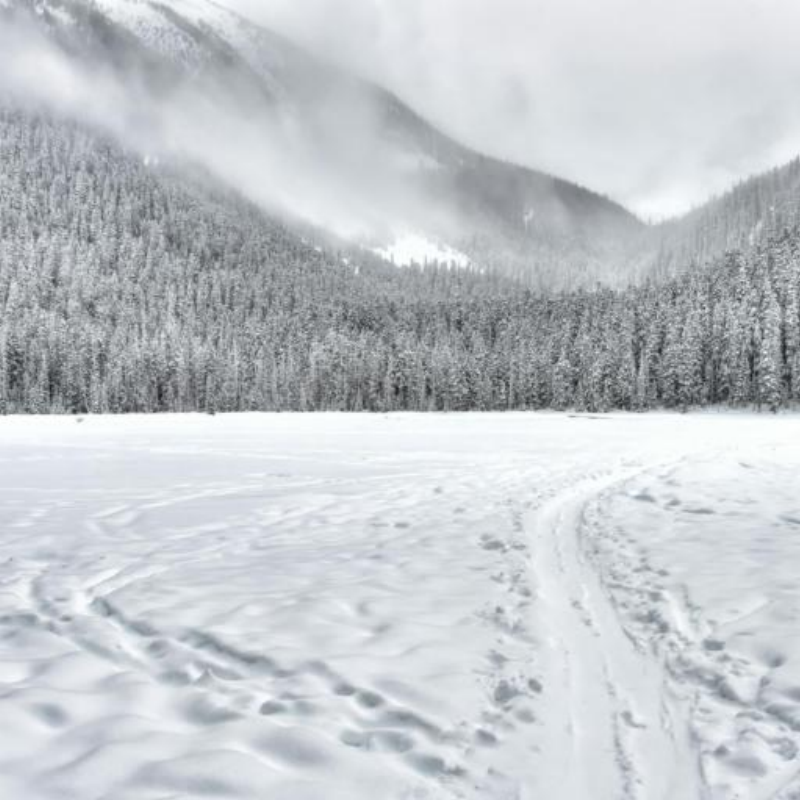}};
        \end{scope}
        \draw[rounded corners=5pt, RubineRed, very thick, dashed] (0,0) rectangle (\stycompression\textwidth,\stycompression\textwidth);
        \end{tikzpicture}}
    \end{overpic}
\end{tabular}

\caption{Visualization of the impact of different visual prompts on generated samples when provided a same text instruction. 
We show edits of both simulation and real-world images. 
Displayed within the dashed frame is one of the 5 style conditioning images relevant to this edit. 
Our method effectively captures semantic nuances beyond that described in the text prompt, understanding that the generated marble should be white or black, and that the generated snow should be sparse or dense.}
\label{appdx_real_conditioning_qualitative}
\end{figure}

\begin{figure}[t]
\centering
\newcommand{\imgwidth}{0.19}
\setlength{\tabcolsep}{1.5pt} 
\renewcommand{\arraystretch}{0.25} 
\newcommand{\stycompression}{0.05}

\begin{tabular}{ccccc} 
    Input & Steel & Marble & Stone & Leather \\

    \includegraphics[width=\imgwidth\textwidth]{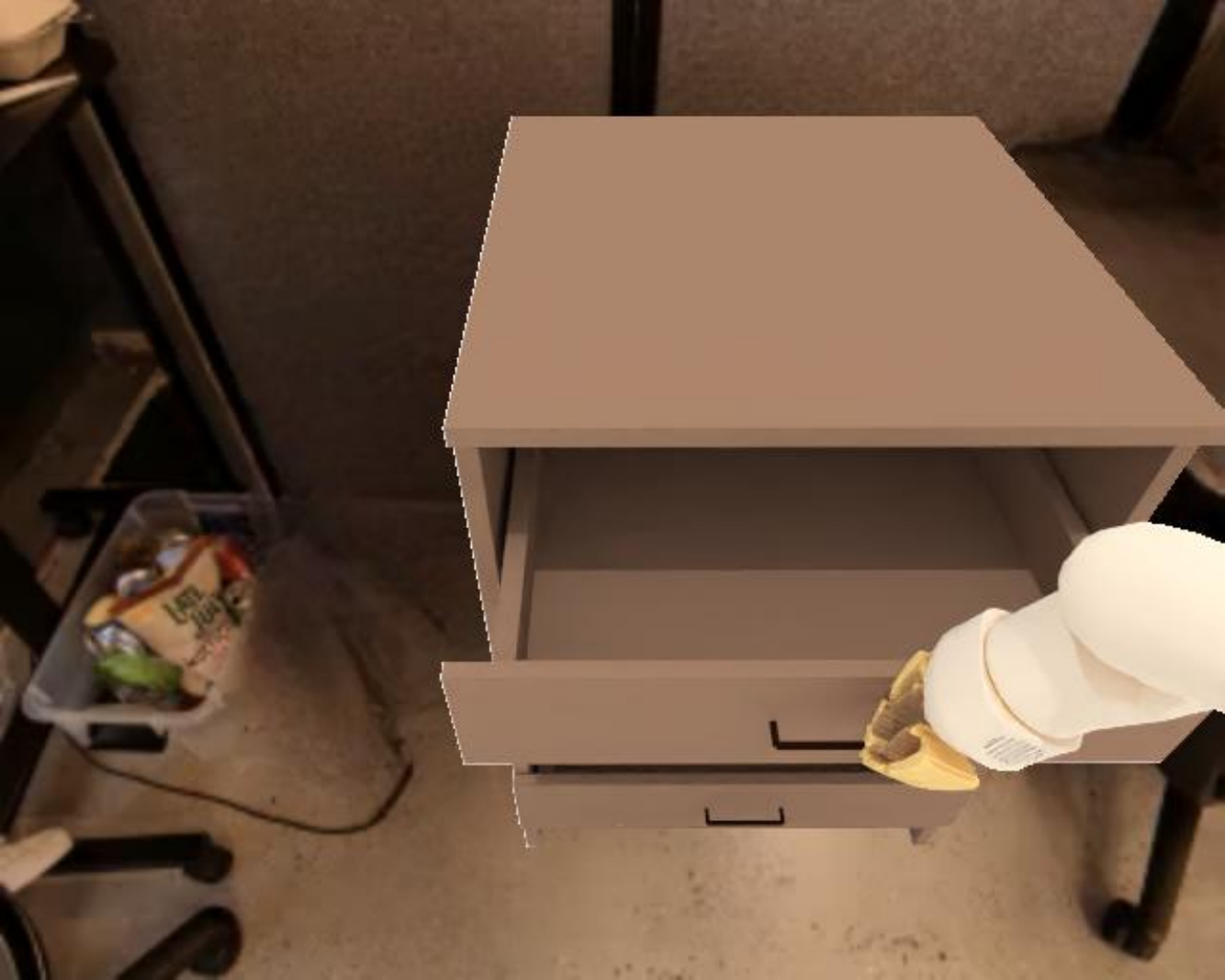} &    
    \begin{overpic}[width=\imgwidth\textwidth]{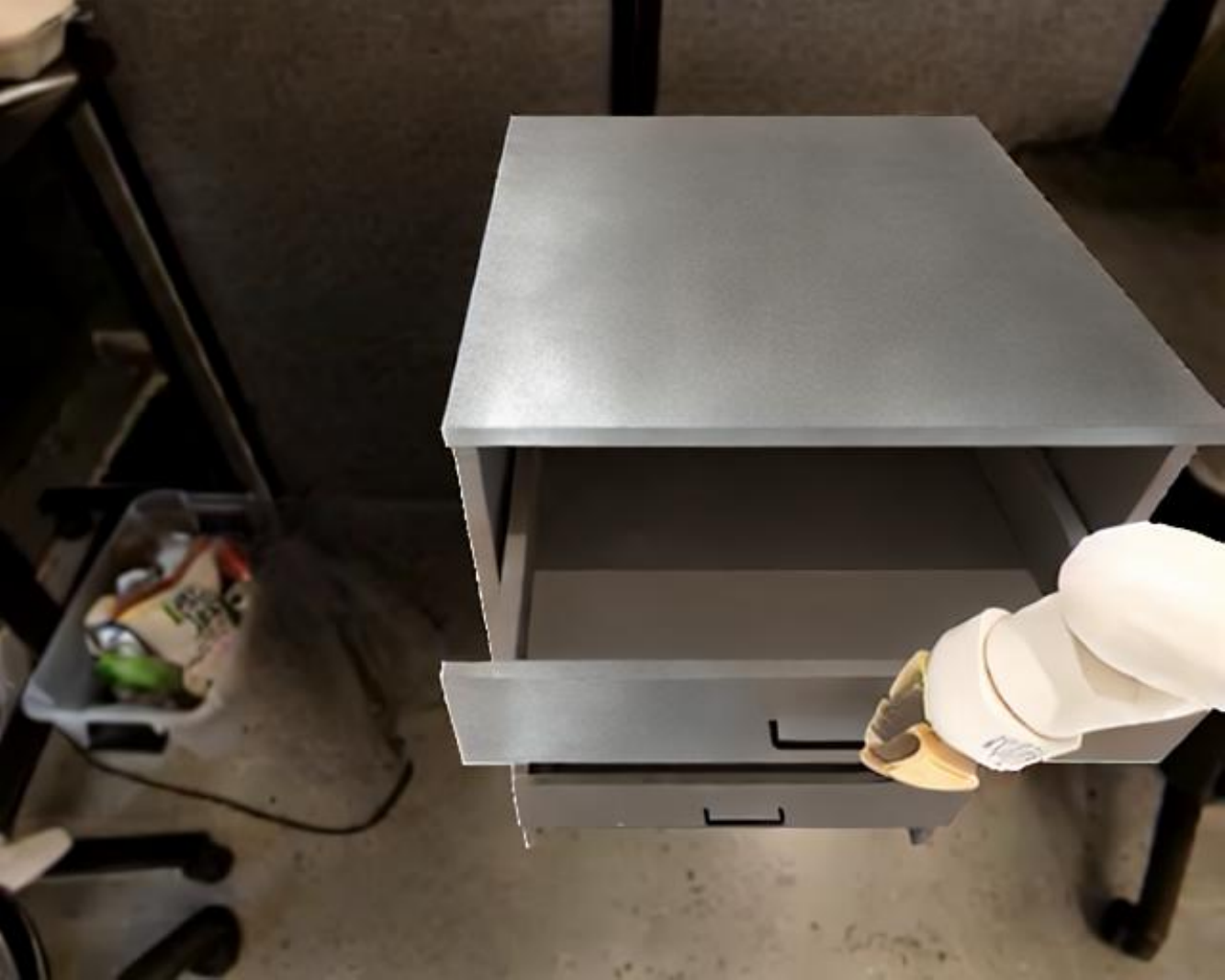}
        \put(1, 50){
         \begin{tikzpicture}
          \begin{scope}
           \clip[rounded corners=5pt] (0,0) rectangle (\stycompression\textwidth, \stycompression\textwidth);
           \node[anchor=north east, inner sep=0pt] at (\stycompression\textwidth,\stycompression\textwidth) {\includegraphics[width=\stycompression\textwidth]{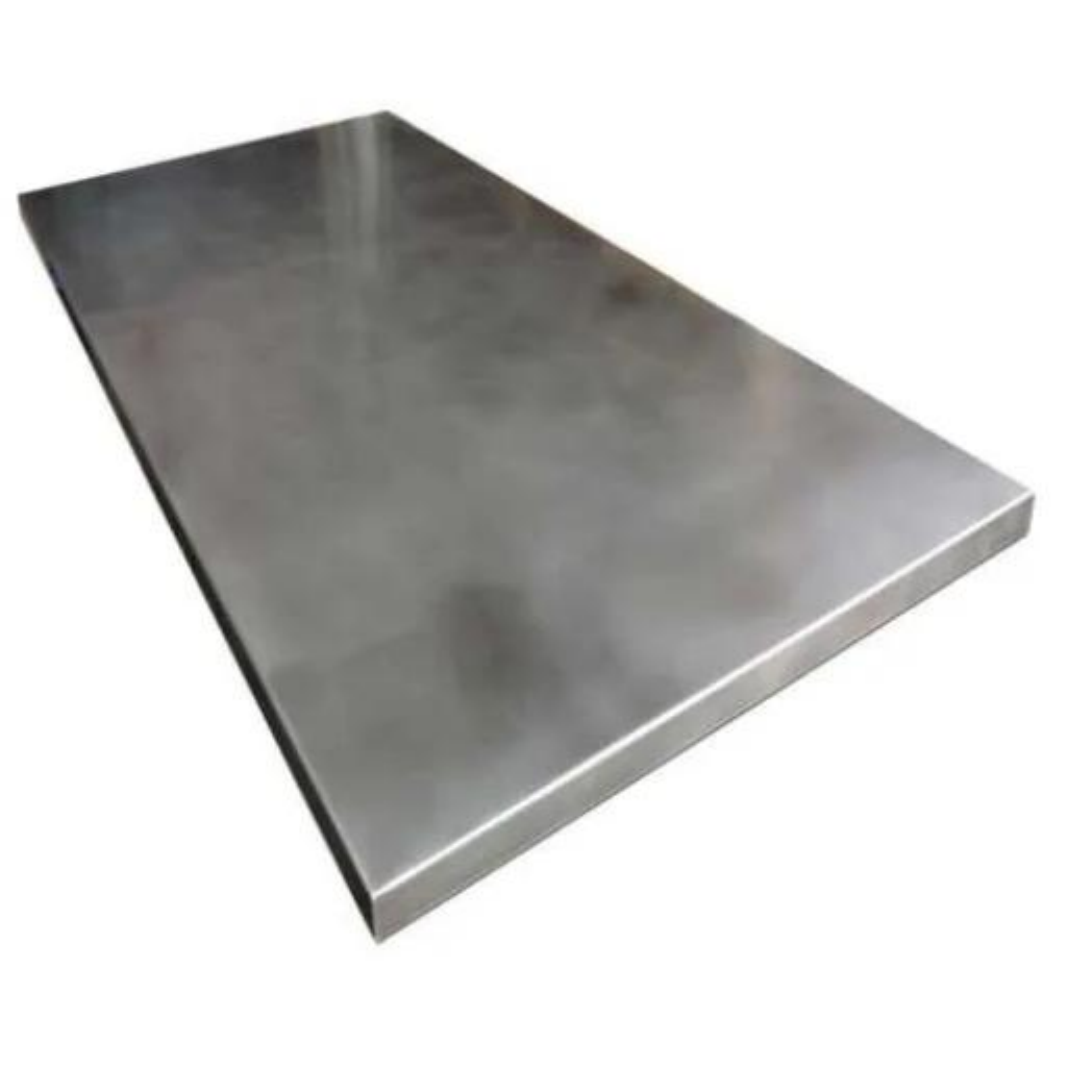}};
        \end{scope}
        \draw[rounded corners=5pt, RubineRed, very thick, dashed] (0,0) rectangle (\stycompression\textwidth,\stycompression\textwidth);
        \end{tikzpicture}}
    \end{overpic}&
    \begin{overpic}[width=\imgwidth\textwidth]{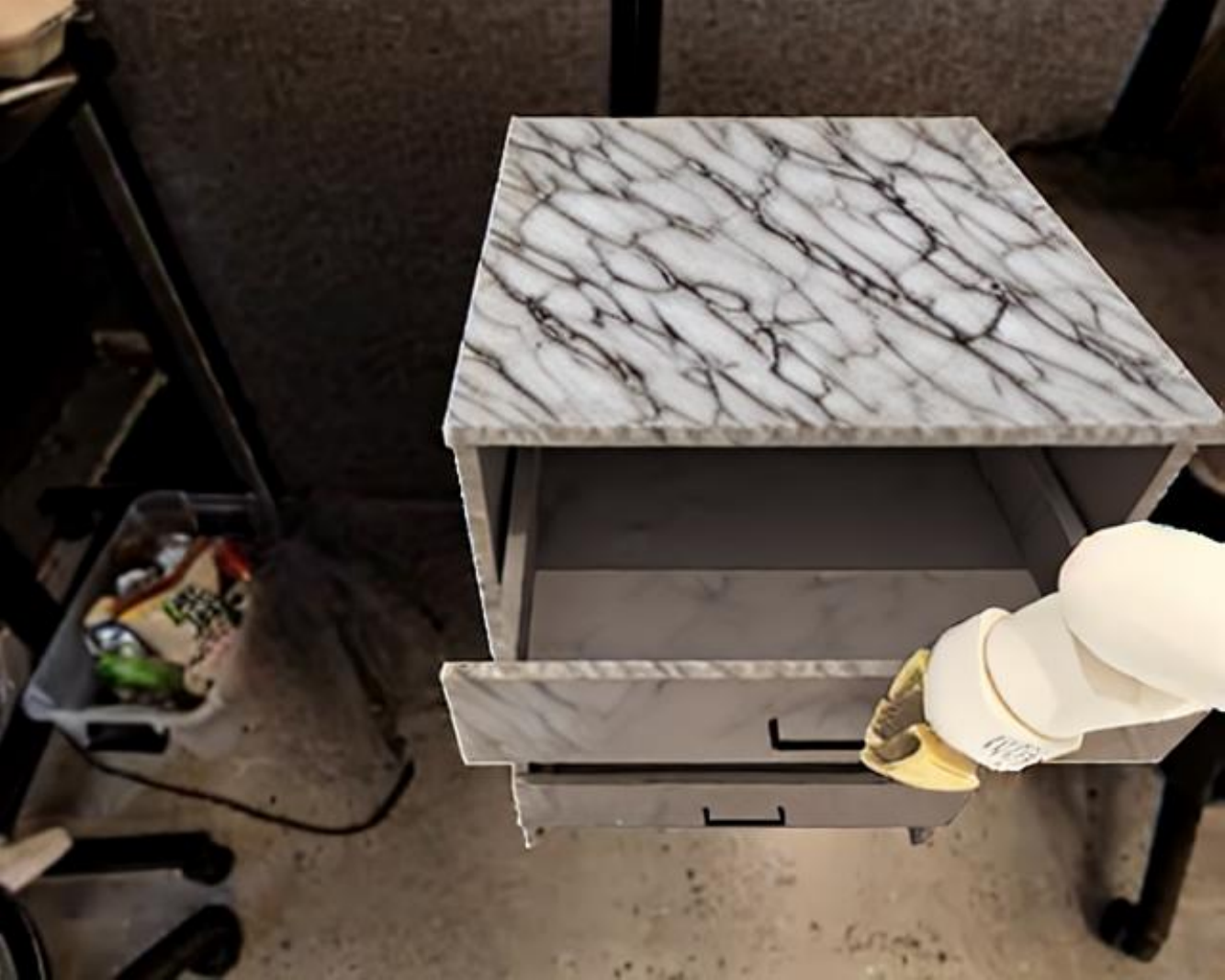}
        \put(1, 50){
         \begin{tikzpicture}
          \begin{scope}
           \clip[rounded corners=5pt] (0,0) rectangle (\stycompression\textwidth, \stycompression\textwidth);
           \node[anchor=north east, inner sep=0pt] at (\stycompression\textwidth,\stycompression\textwidth) {\includegraphics[width=\stycompression\textwidth]{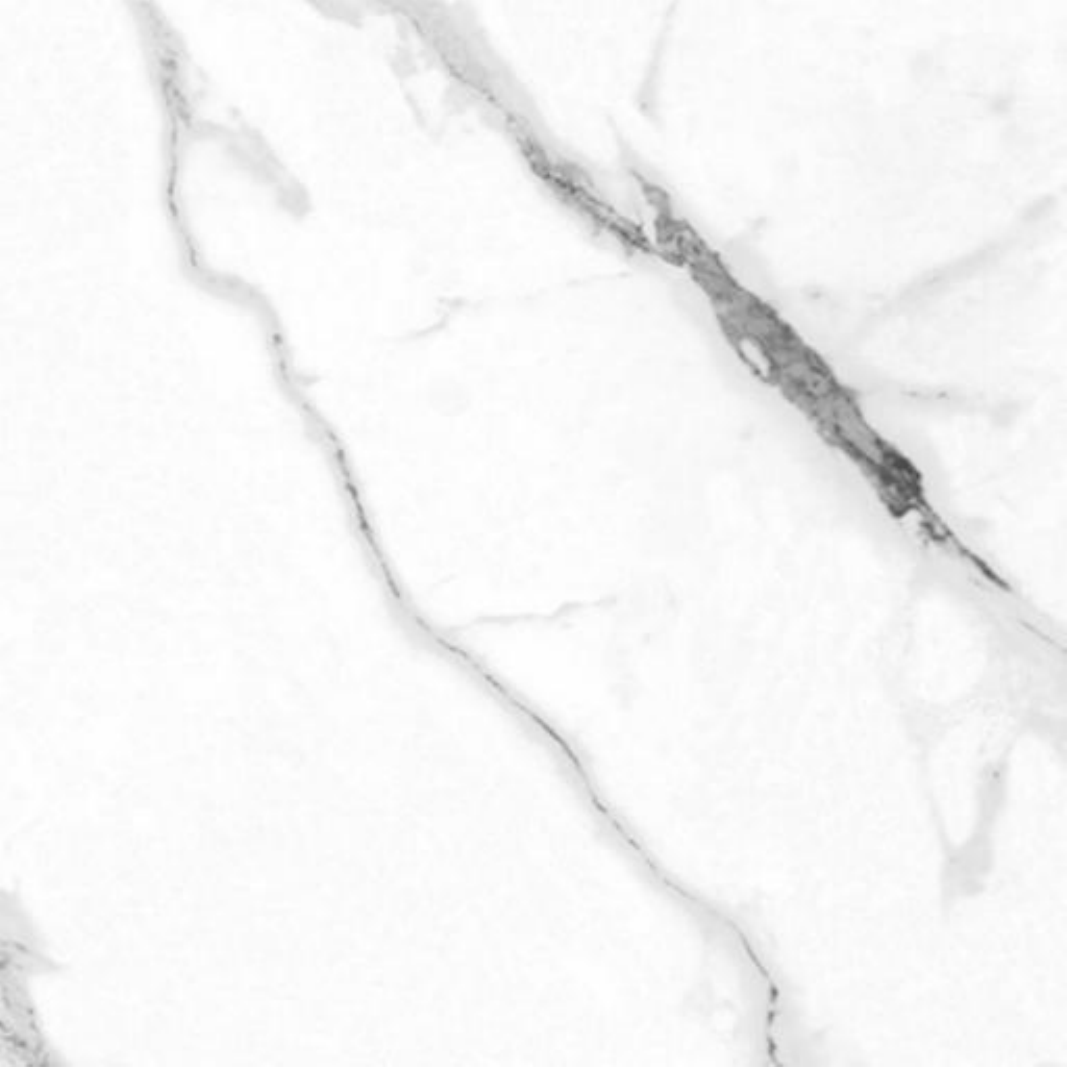}};
        \end{scope}
        \draw[rounded corners=5pt, RubineRed, very thick, dashed] (0,0) rectangle (\stycompression\textwidth,\stycompression\textwidth);
        \end{tikzpicture}}
    \end{overpic}&
    \begin{overpic}[width=\imgwidth\textwidth]{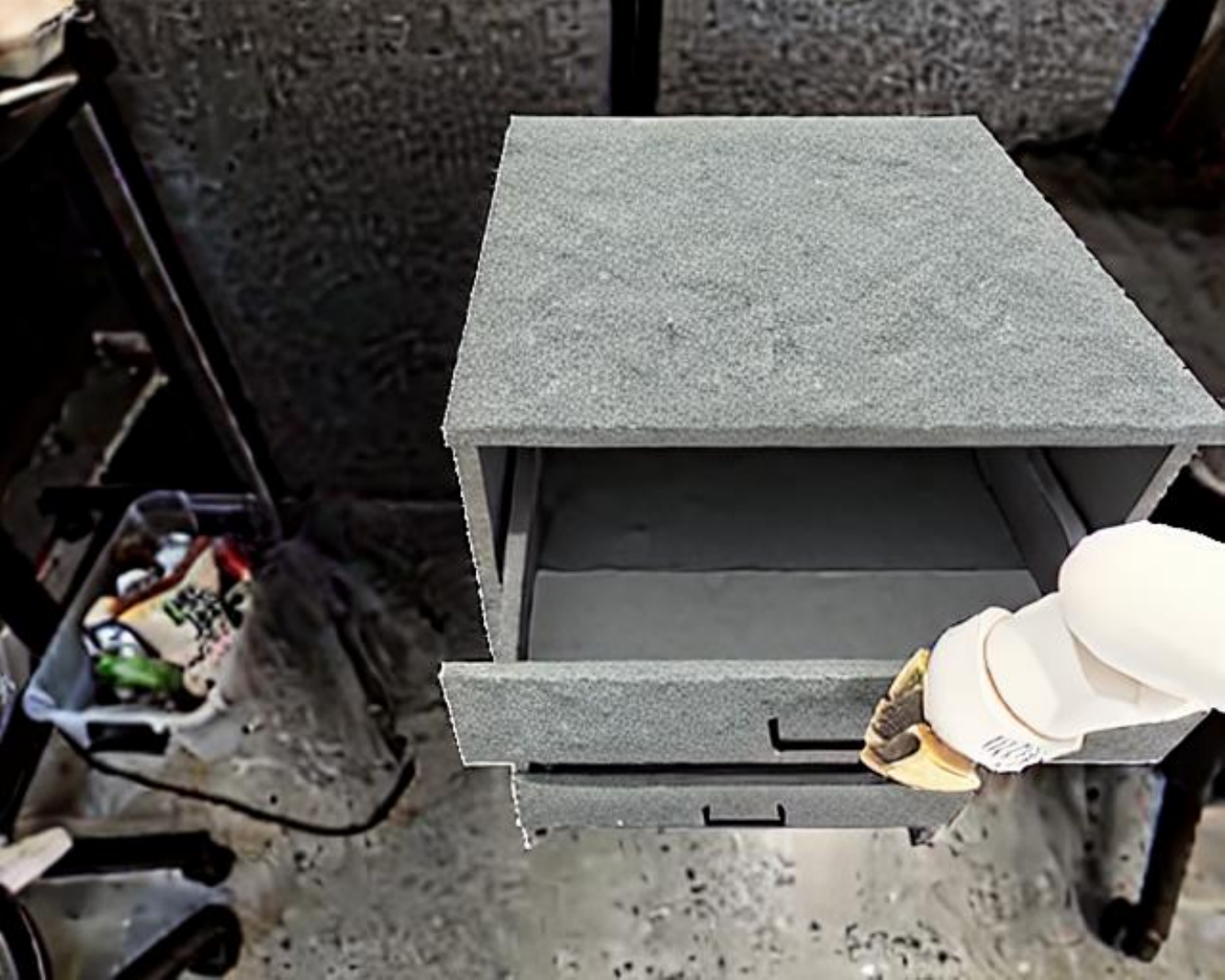}
        \put(1, 50){
         \begin{tikzpicture}
          \begin{scope}
           \clip[rounded corners=5pt] (0,0) rectangle (\stycompression\textwidth, \stycompression\textwidth);
           \node[anchor=north east, inner sep=0pt] at (\stycompression\textwidth,\stycompression\textwidth) {\includegraphics[width=\stycompression\textwidth]{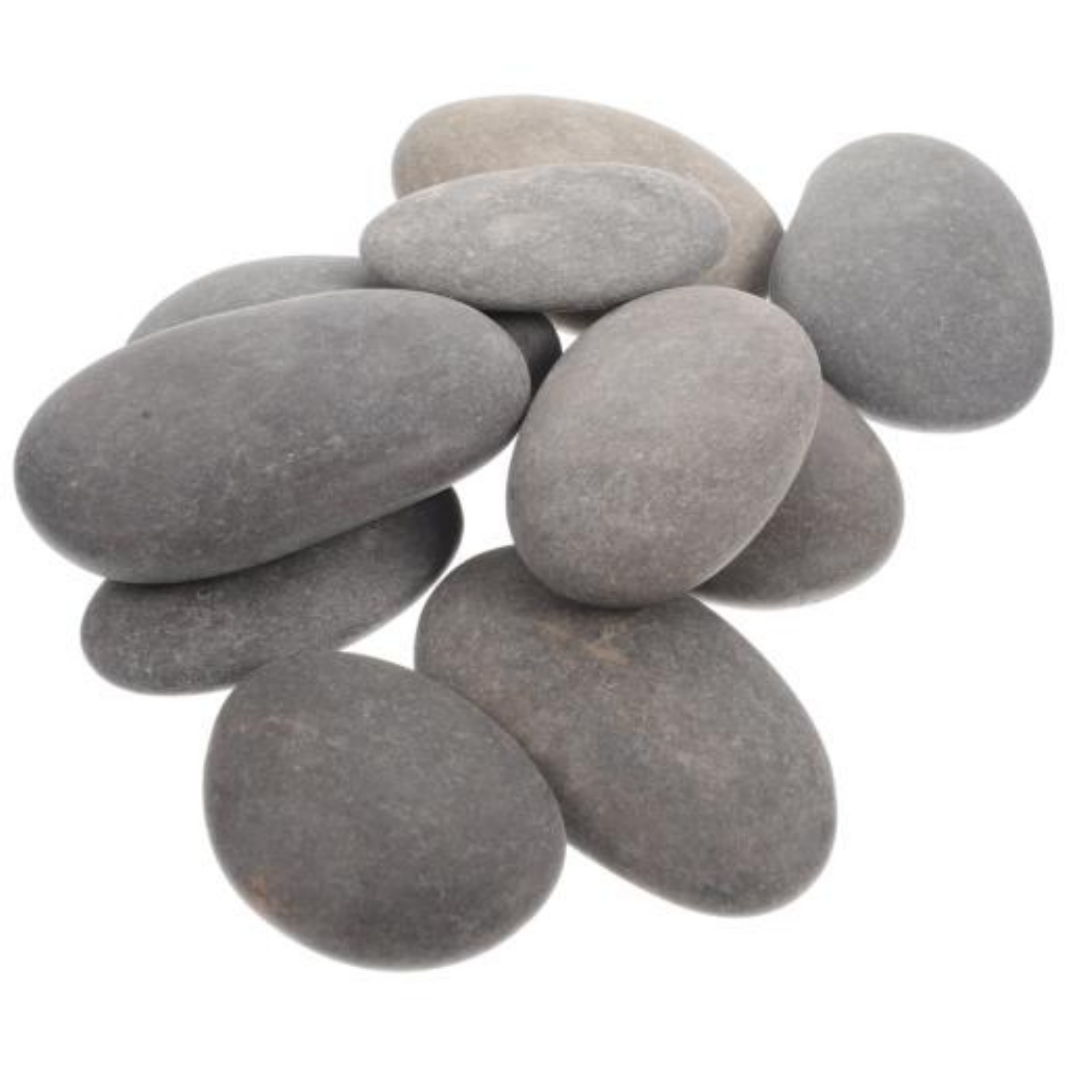}};
        \end{scope}
        \draw[rounded corners=5pt, RubineRed, very thick, dashed] (0,0) rectangle (\stycompression\textwidth,\stycompression\textwidth);
        \end{tikzpicture}}
    \end{overpic}&
    \begin{overpic}[width=\imgwidth\textwidth]{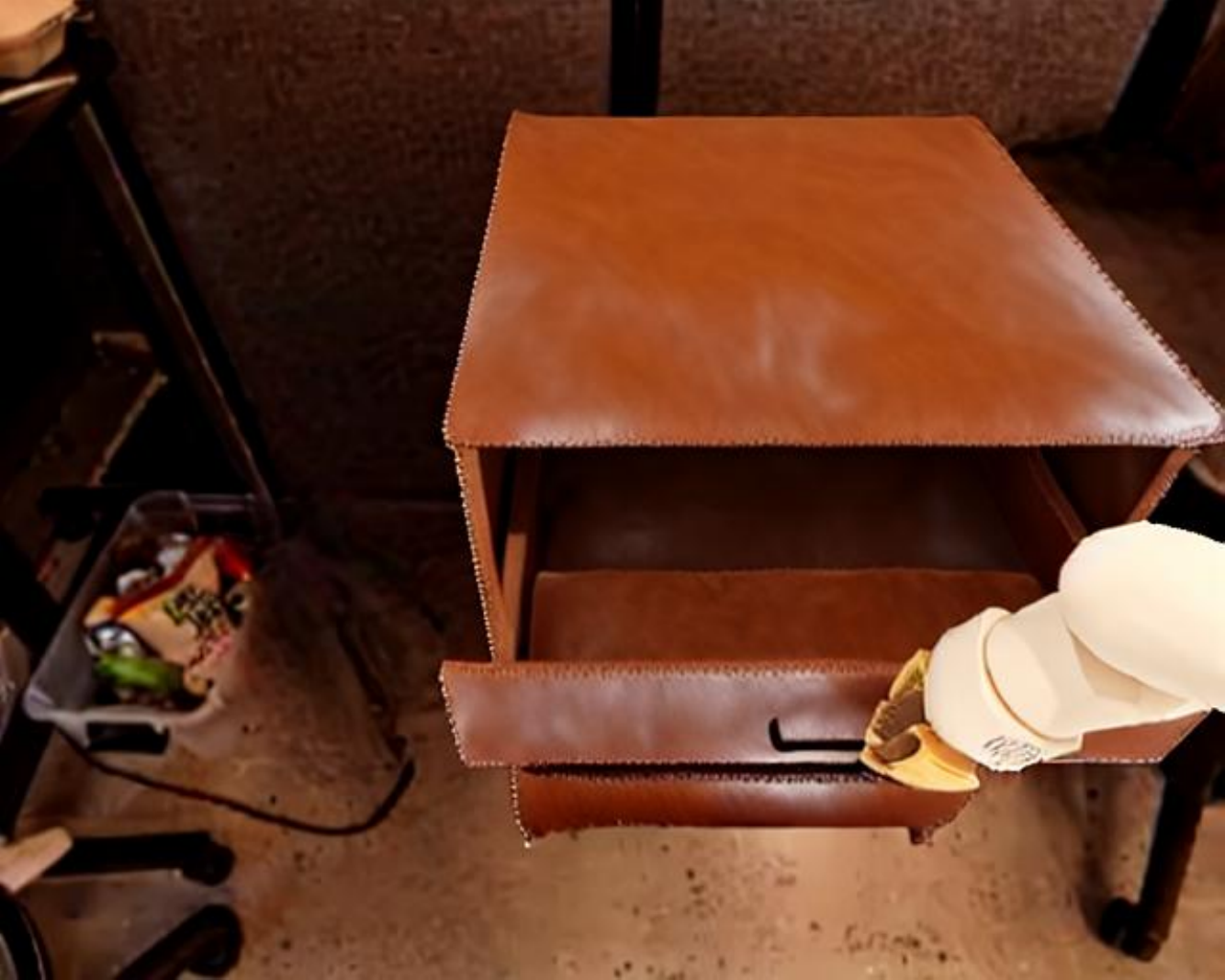}
        \put(1, 50){
         \begin{tikzpicture}
          \begin{scope}
           \clip[rounded corners=5pt] (0,0) rectangle (\stycompression\textwidth, \stycompression\textwidth);
           \node[anchor=north east, inner sep=0pt] at (\stycompression\textwidth,\stycompression\textwidth) {\includegraphics[width=\stycompression\textwidth]{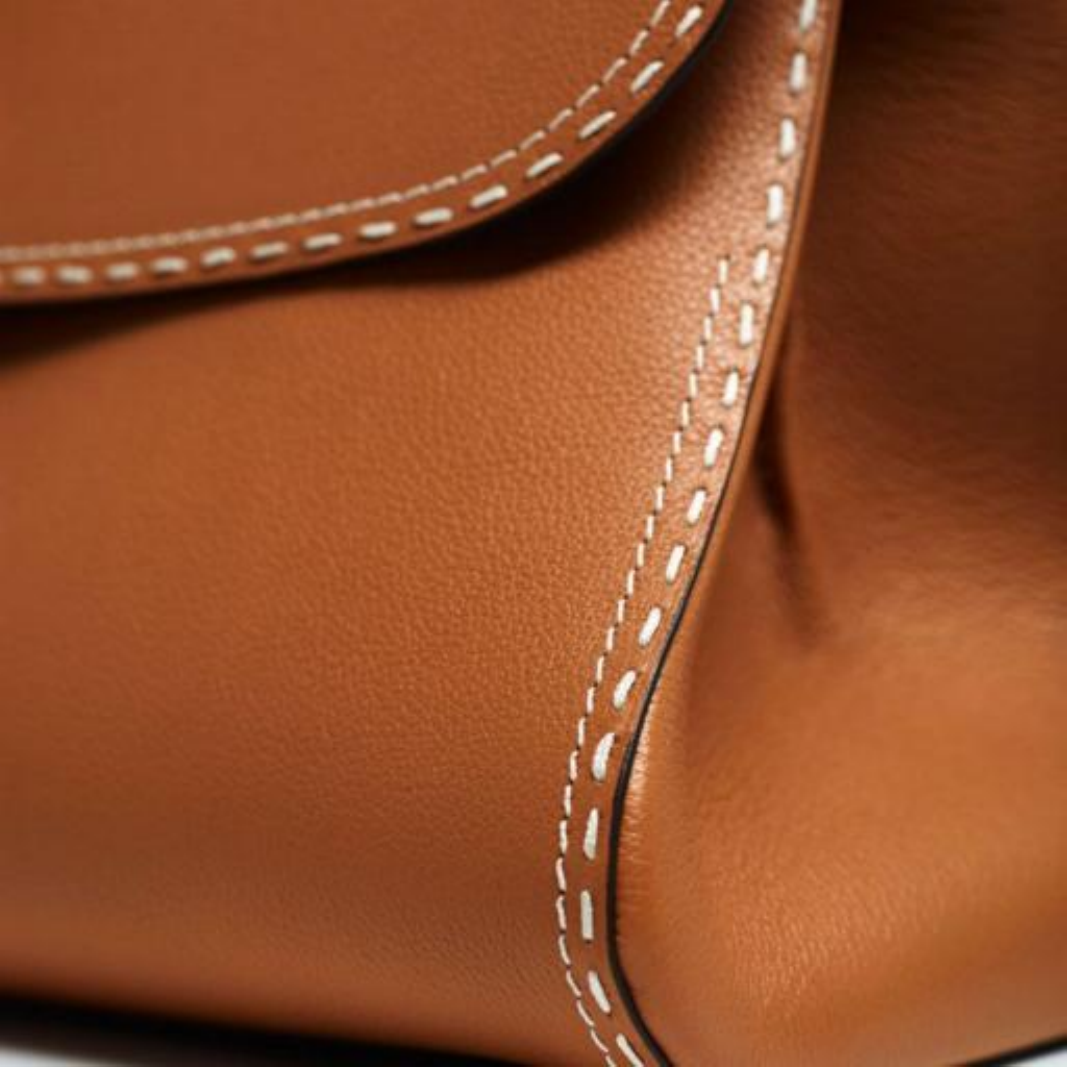}};
        \end{scope}
        \draw[rounded corners=5pt, RubineRed, very thick, dashed] (0,0) rectangle (\stycompression\textwidth,\stycompression\textwidth);
        \end{tikzpicture}}
    \end{overpic} \\ [0.5em]

    \includegraphics[width=\imgwidth\textwidth]{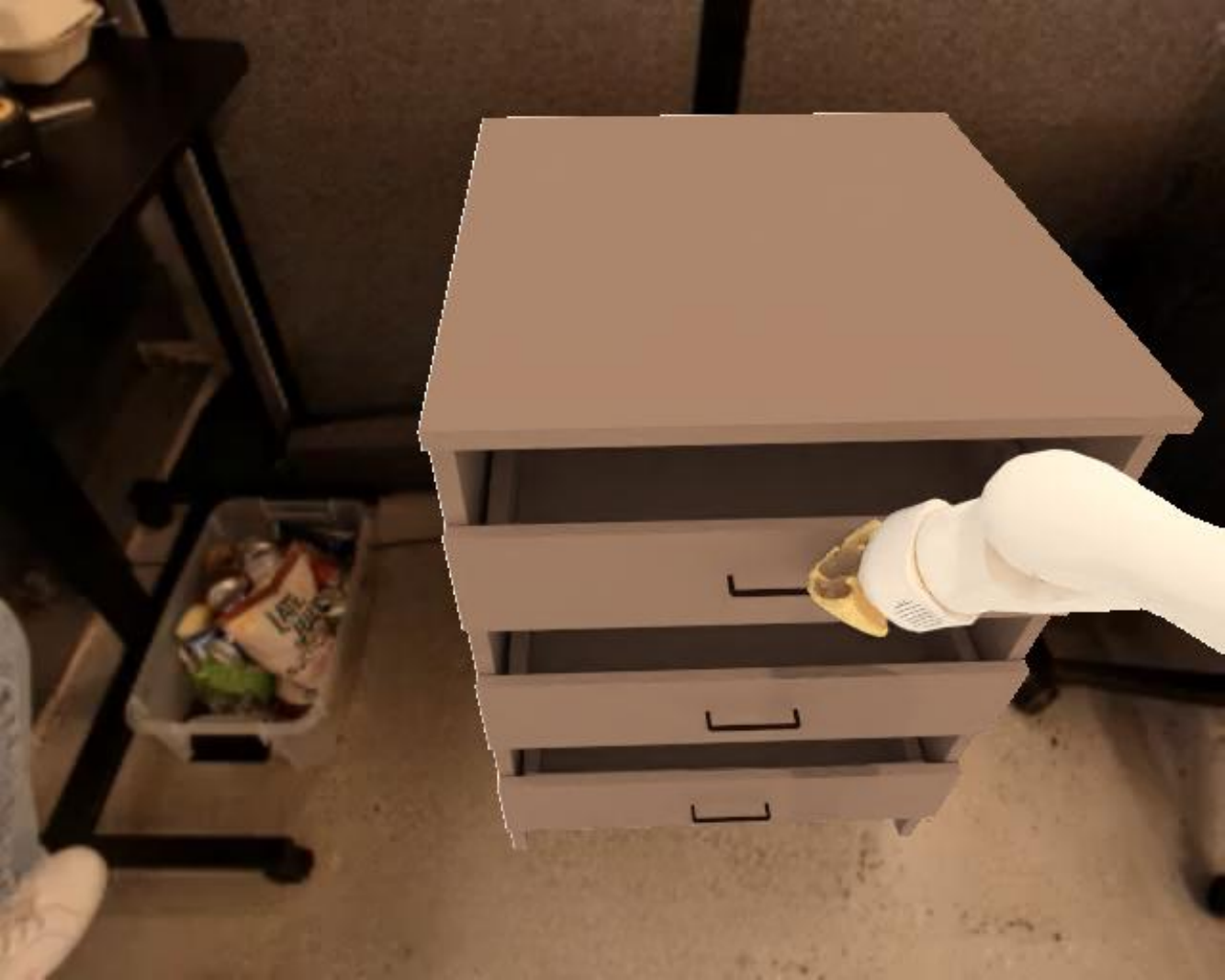} &
    \begin{overpic}[width=\imgwidth\textwidth]{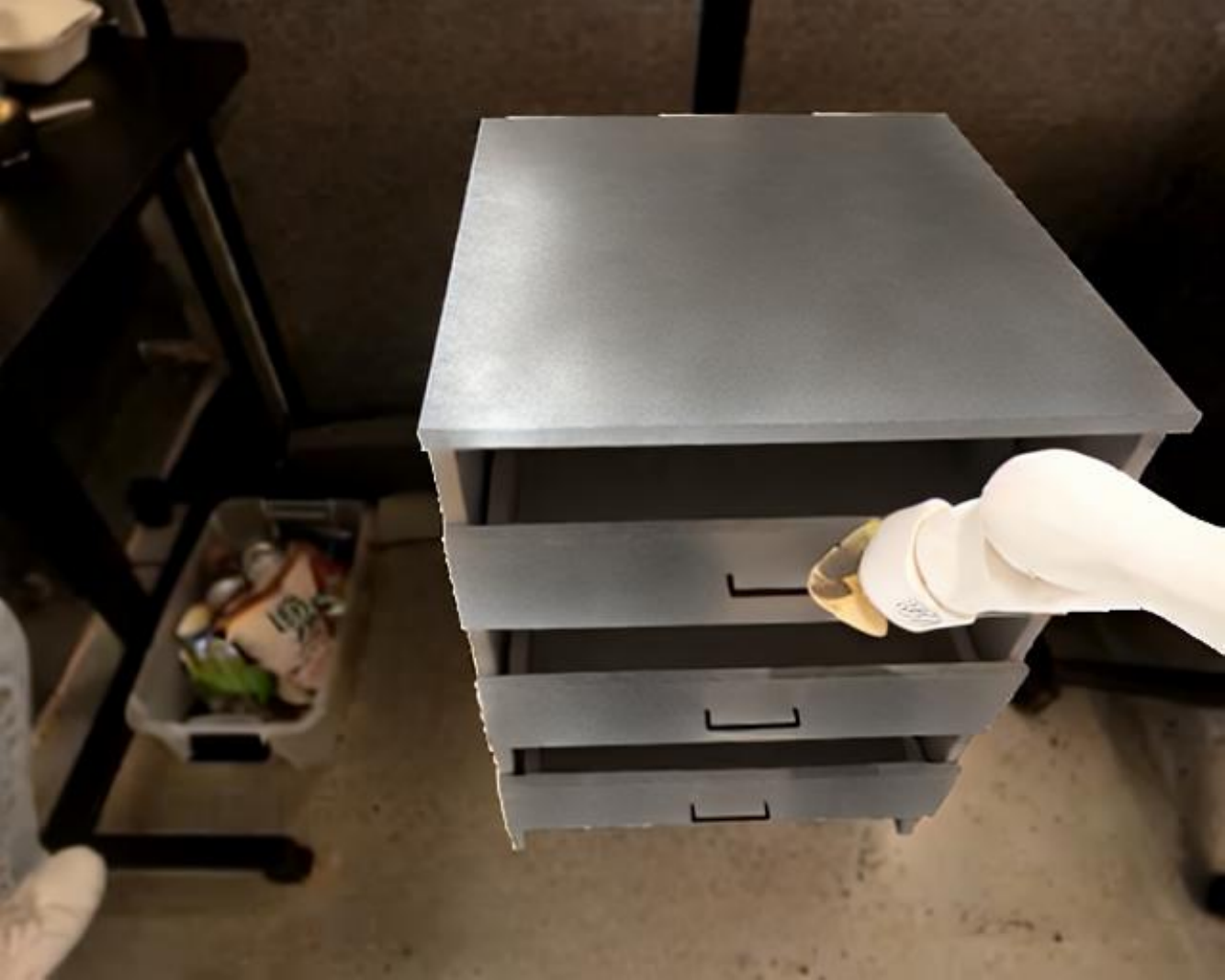}
        \put(1, 50){
         \begin{tikzpicture}
          \begin{scope}
           \clip[rounded corners=5pt] (0,0) rectangle (\stycompression\textwidth, \stycompression\textwidth);
           \node[anchor=north east, inner sep=0pt] at (\stycompression\textwidth,\stycompression\textwidth) {\includegraphics[width=\stycompression\textwidth]{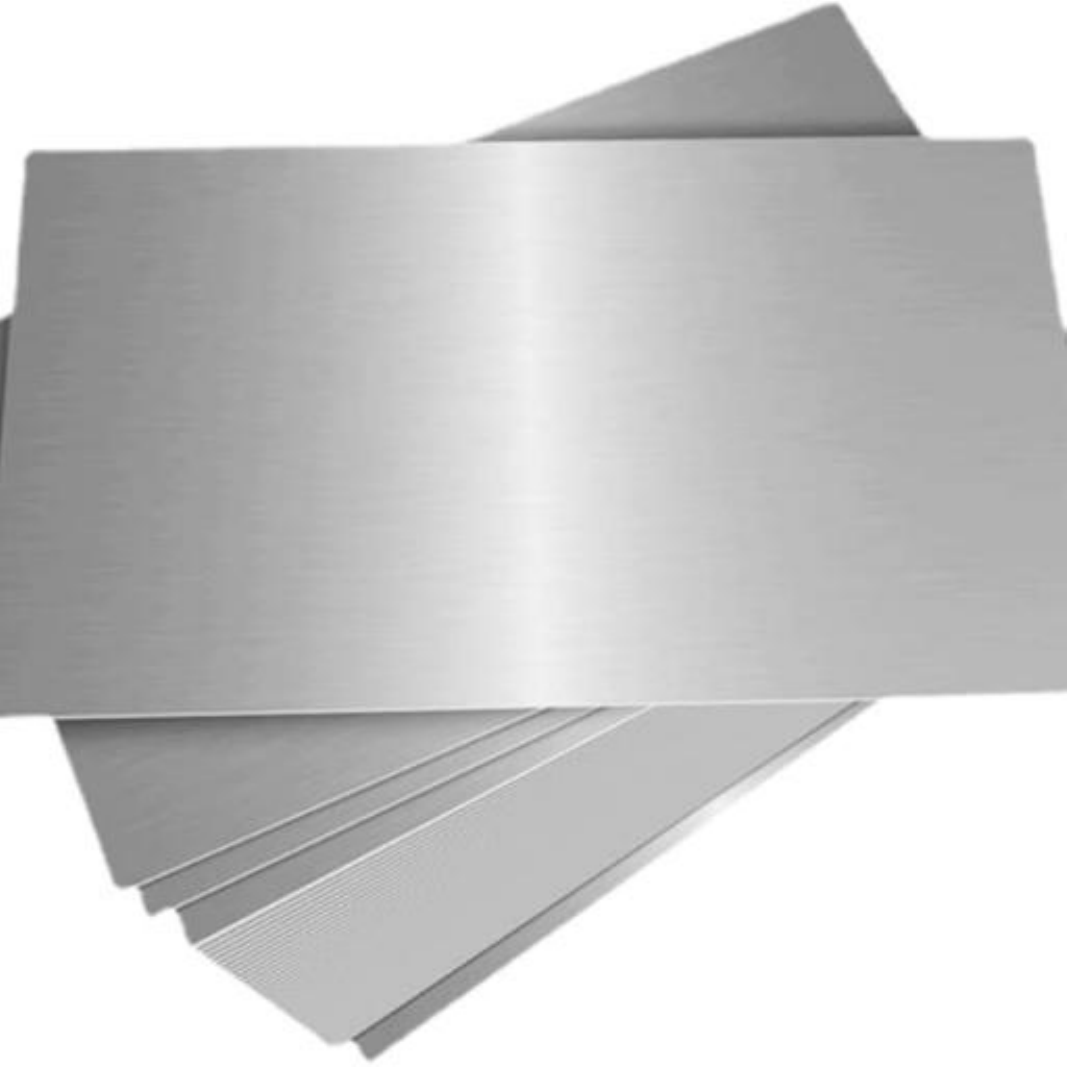}};
        \end{scope}
        \draw[rounded corners=5pt, RubineRed, very thick, dashed] (0,0) rectangle (\stycompression\textwidth,\stycompression\textwidth);
        \end{tikzpicture}}
    \end{overpic}&
    \begin{overpic}[width=\imgwidth\textwidth]{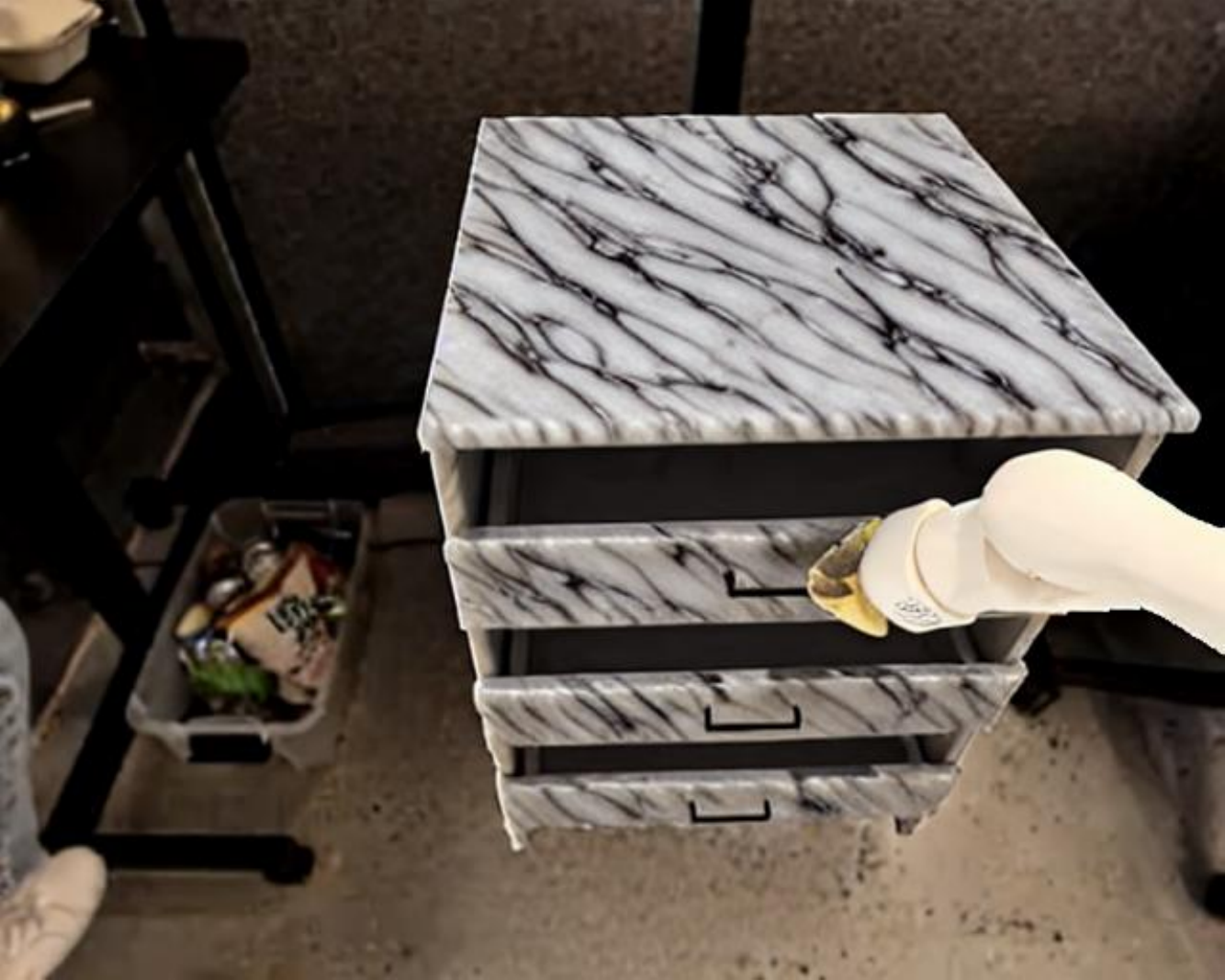}
        \put(1, 50){
         \begin{tikzpicture}
          \begin{scope}
           \clip[rounded corners=5pt] (0,0) rectangle (\stycompression\textwidth, \stycompression\textwidth);
           \node[anchor=north east, inner sep=0pt] at (\stycompression\textwidth,\stycompression\textwidth) {\includegraphics[width=\stycompression\textwidth]{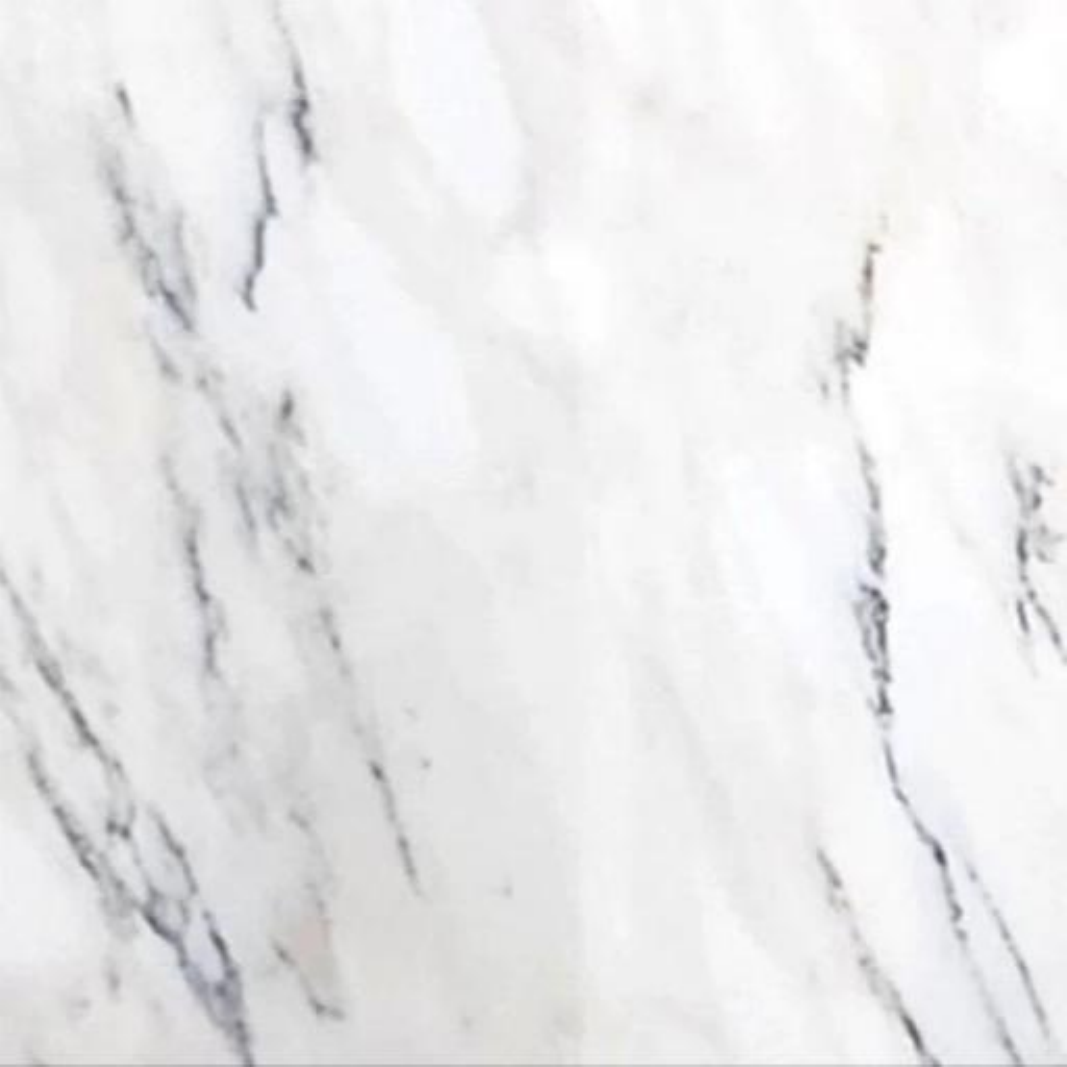}};
        \end{scope}
        \draw[rounded corners=5pt, RubineRed, very thick, dashed] (0,0) rectangle (\stycompression\textwidth,\stycompression\textwidth);
        \end{tikzpicture}}
    \end{overpic}&
    \begin{overpic}[width=\imgwidth\textwidth]{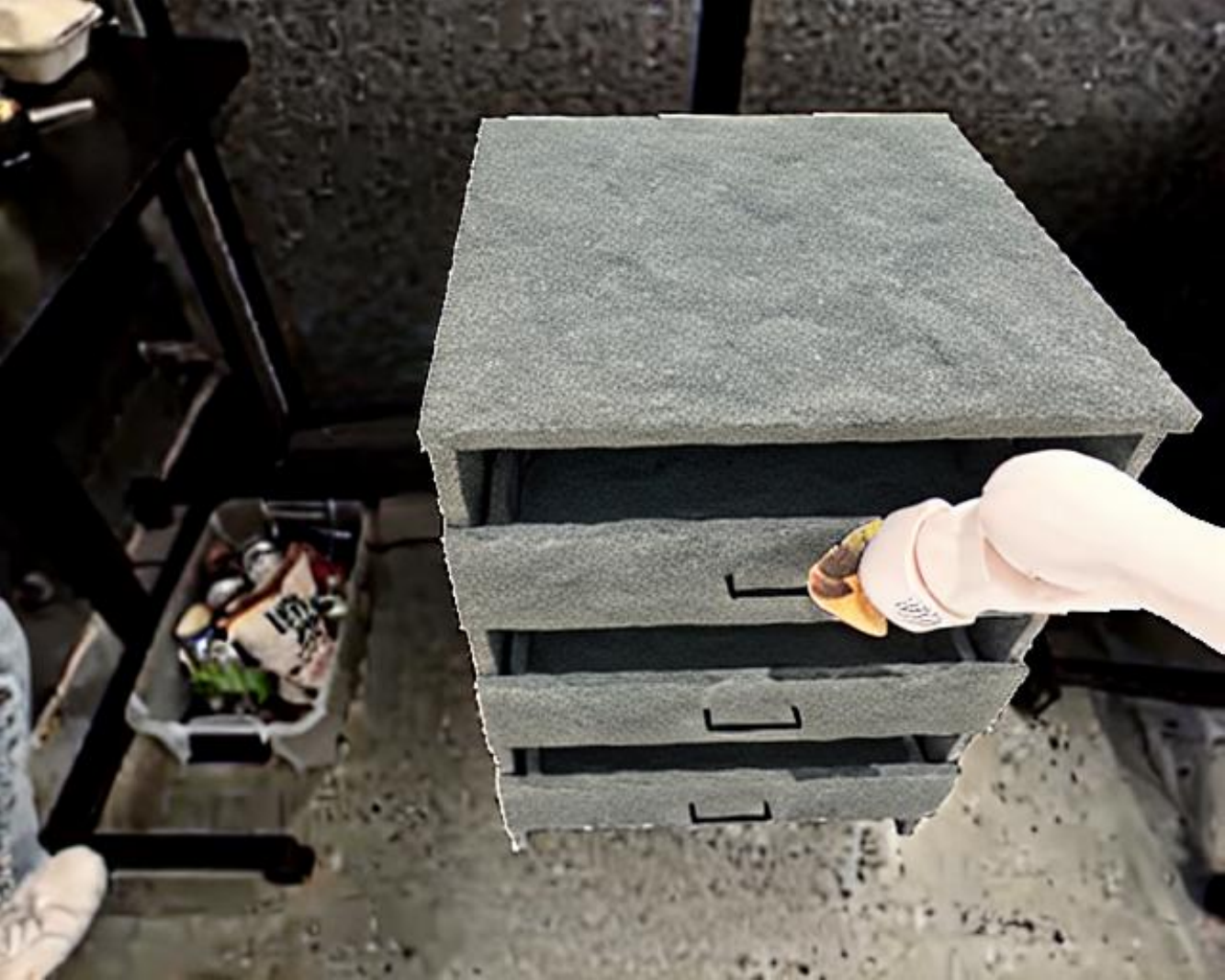}
        \put(1, 50){
         \begin{tikzpicture}
          \begin{scope}
           \clip[rounded corners=5pt] (0,0) rectangle (\stycompression\textwidth, \stycompression\textwidth);
           \node[anchor=north east, inner sep=0pt] at (\stycompression\textwidth,\stycompression\textwidth) {\includegraphics[width=\stycompression\textwidth]{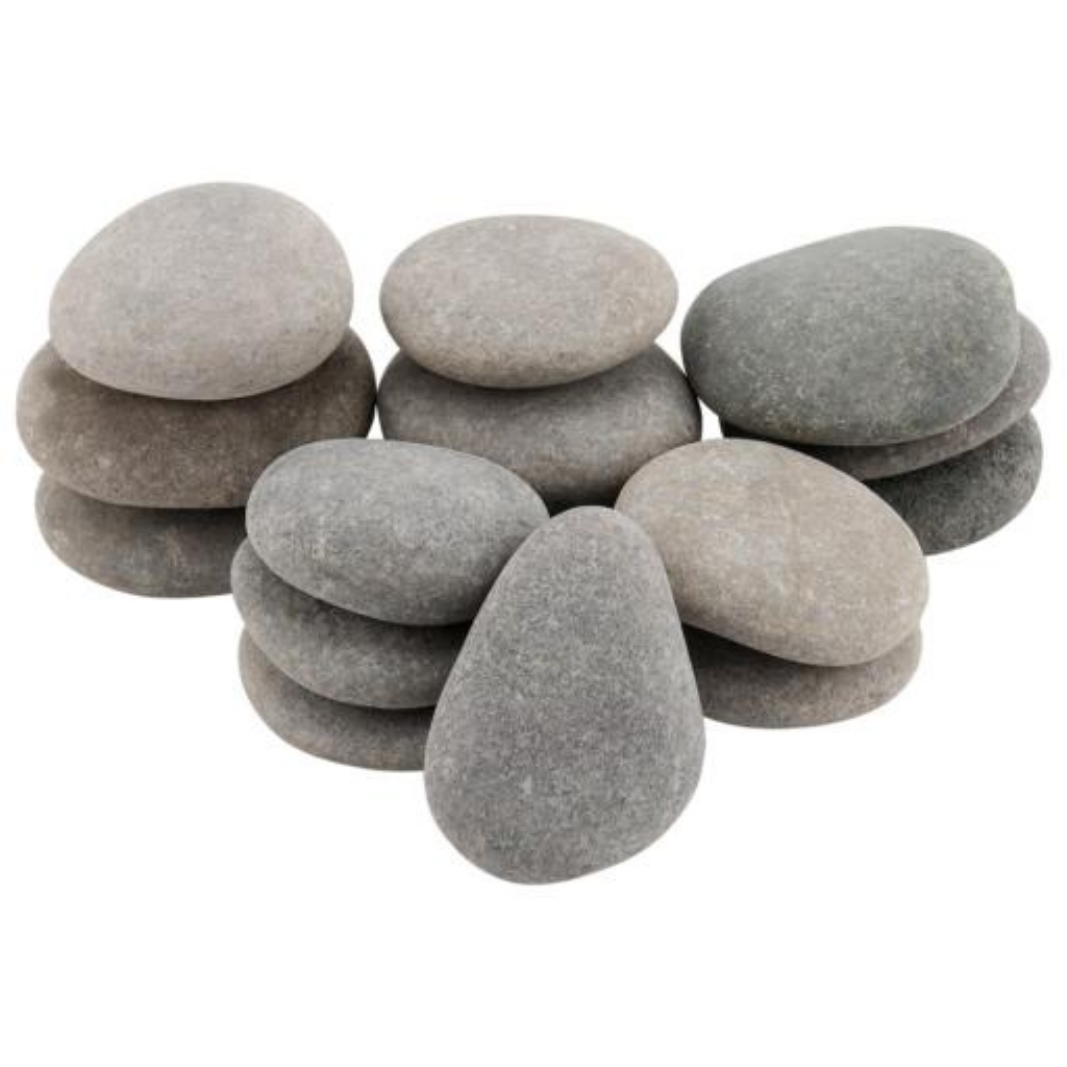}};
        \end{scope}
        \draw[rounded corners=5pt, RubineRed, very thick, dashed] (0,0) rectangle (\stycompression\textwidth,\stycompression\textwidth);
        \end{tikzpicture}}
    \end{overpic}&
       \begin{overpic}[width=\imgwidth\textwidth]{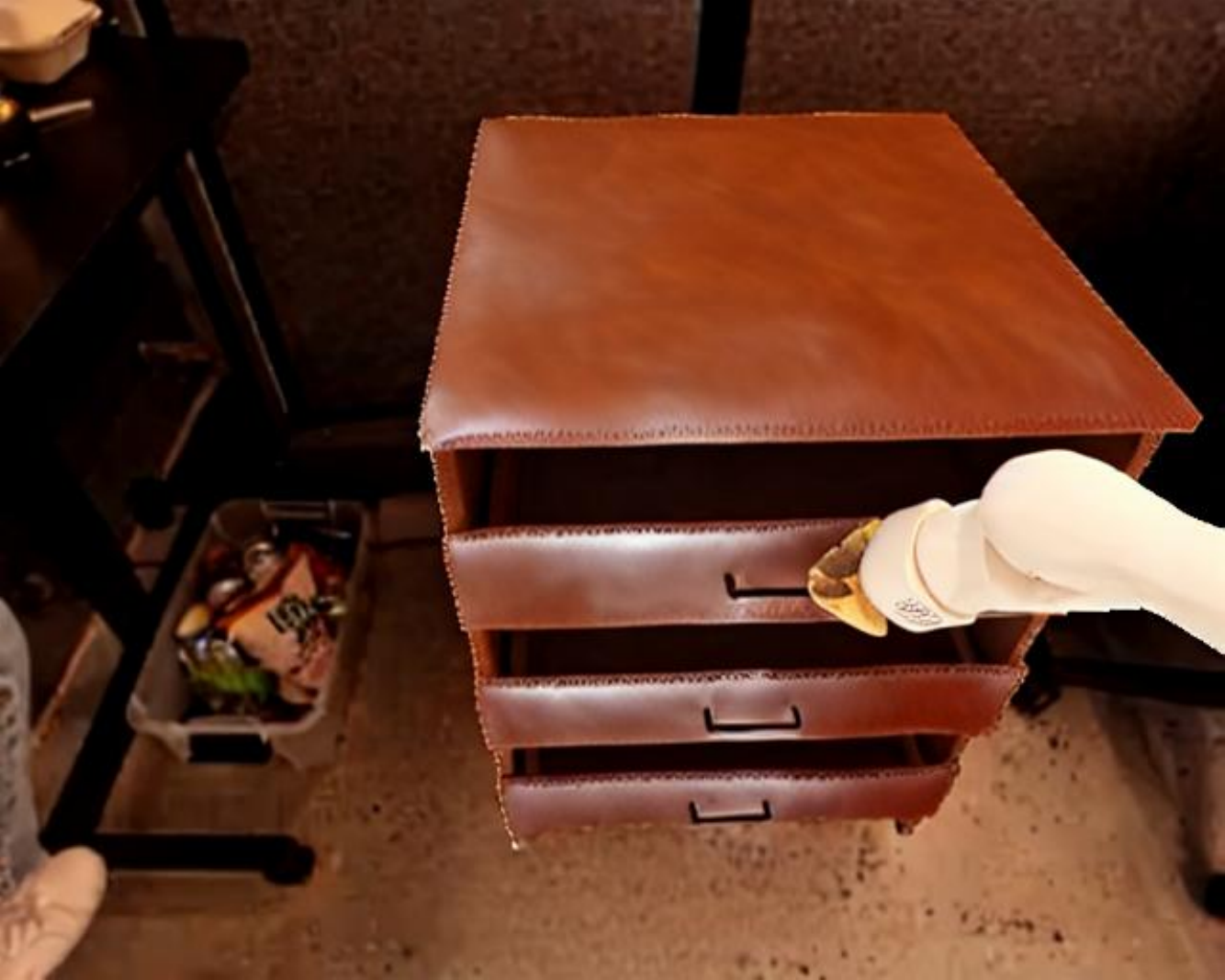}
        \put(1, 50){
         \begin{tikzpicture}
          \begin{scope}
           \clip[rounded corners=5pt] (0,0) rectangle (\stycompression\textwidth, \stycompression\textwidth);
           \node[anchor=north east, inner sep=0pt] at (\stycompression\textwidth,\stycompression\textwidth) {\includegraphics[width=\stycompression\textwidth]{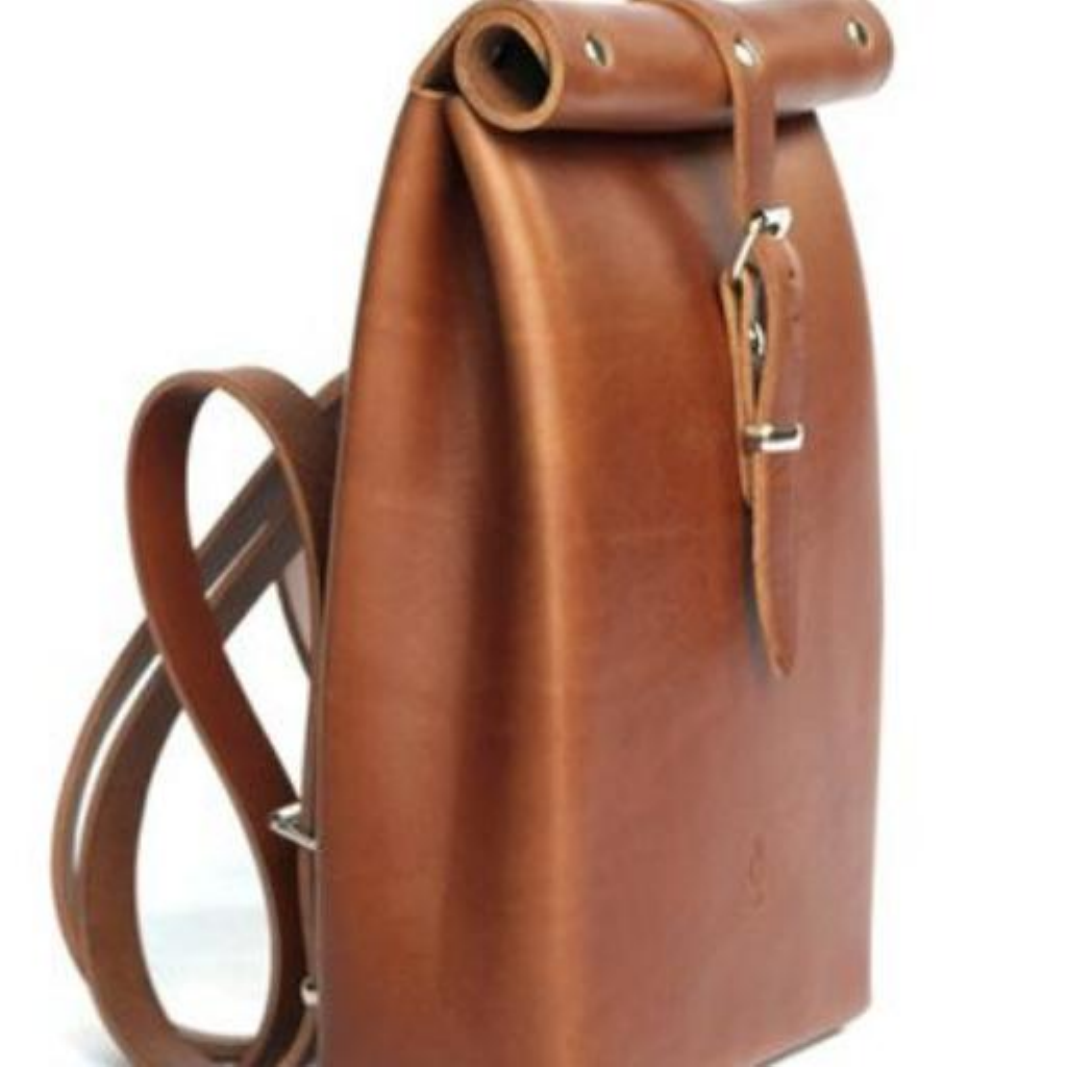}};
        \end{scope}
        \draw[rounded corners=5pt, RubineRed, very thick, dashed] (0,0) rectangle (\stycompression\textwidth,\stycompression\textwidth);
        \end{tikzpicture}}
    \end{overpic} \\

\end{tabular}
\caption{Examples of a simulated scene edited by our method, showcasing enhanced realism compared to the original image across various styles. When training the diffusion model across the diverse styles, we use the same text prompt format: \textit{``make the cupboard out of [X]"}.}

\label{apdx_rtx_row}
\end{figure}

\begin{figure}[htbp]
    \centering
    \newcommand{\imgwidth}{0.125\textwidth} 
    \newcommand{\imgwidthh}{0.25\textwidth} 
    \newcommand{\smallimgheight}{0.95\textwidth}
    \begin{tabular}{cccc} 
        Reference style & Synthetic element & Small real element & Large real element\\[.25em]
        \begin{minipage}[c][0.5\textwidth][c]{\imgwidth} 
            \centering
            \begin{tikzpicture}
                \clip[rounded corners=5pt] (0,0) rectangle (\smallimgheight,\smallimgheight);
                \node[anchor=south west, inner sep=0pt] at (0,0) {%
                    \includegraphics[width=\smallimgheight,height=\smallimgheight]{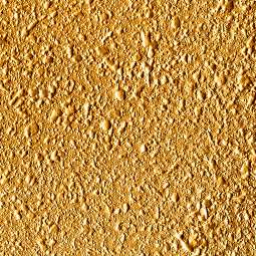}%
                };
            \end{tikzpicture} \\
            \smallskip 
            \begin{tikzpicture}
                \clip[rounded corners=5pt] (0,0) rectangle (\smallimgheight,\smallimgheight);
                \node[anchor=south west, inner sep=0pt] at (0,0) {%
                    \includegraphics[width=\smallimgheight,height=\smallimgheight]{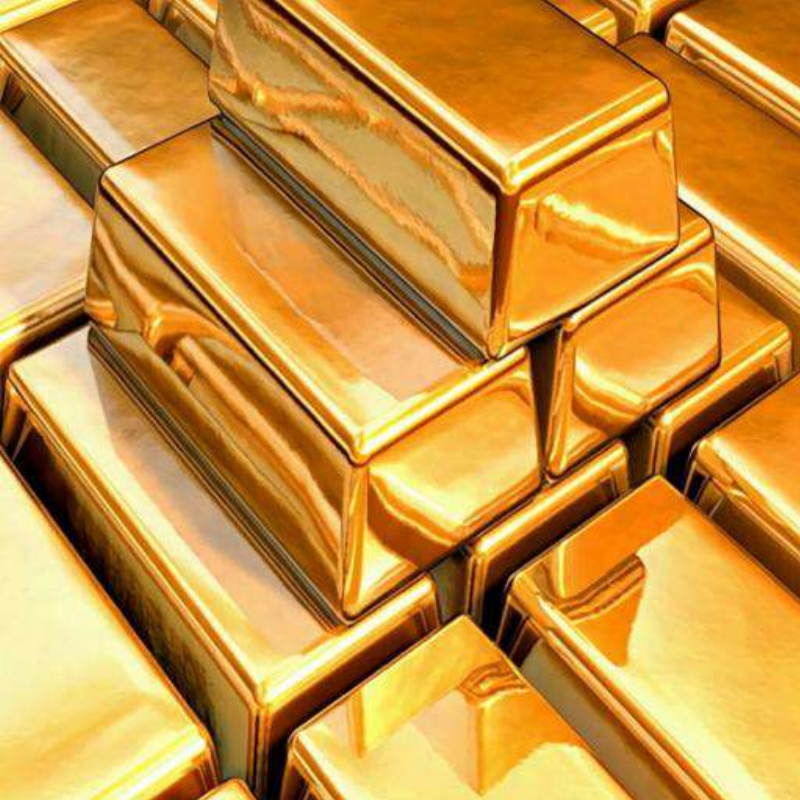}%
                };
            \end{tikzpicture} \\ [.5em]
            \smallskip 
            \begin{tikzpicture}
                \clip[rounded corners=5pt] (0,0) rectangle (\smallimgheight,\smallimgheight);
                \node[anchor=south west, inner sep=0pt] at (0,0) {%
                    \includegraphics[width=\smallimgheight,height=\smallimgheight]{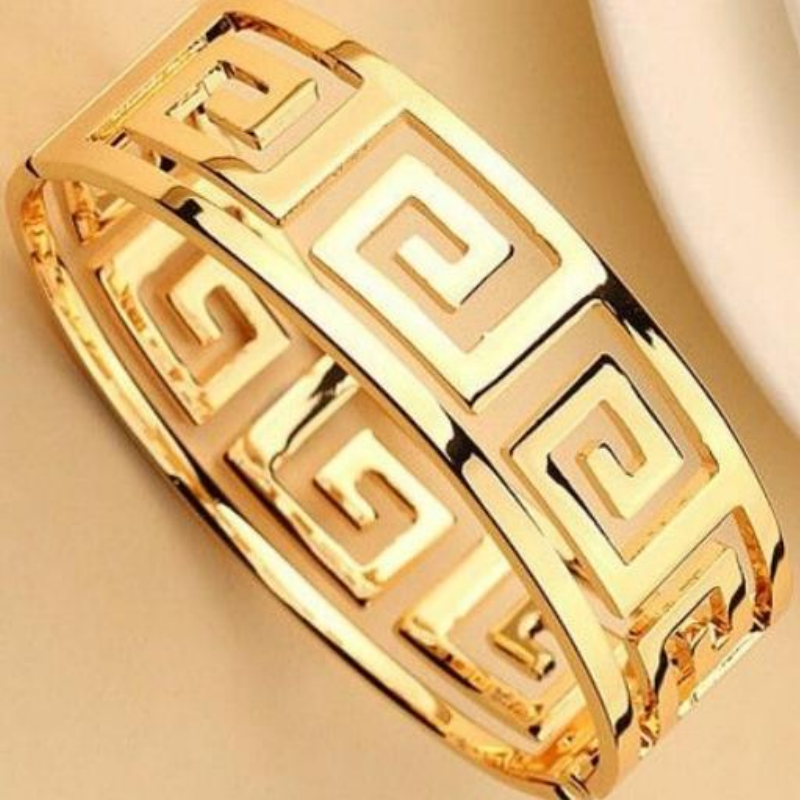}%
                };
            \end{tikzpicture} \\ 
            \smallskip 
            \begin{tikzpicture}
                \clip[rounded corners=5pt] (0,0) rectangle (\smallimgheight,\smallimgheight);
                \node[anchor=south west, inner sep=0pt] at (0,0) {%
                    \includegraphics[width=\smallimgheight,height=\smallimgheight]{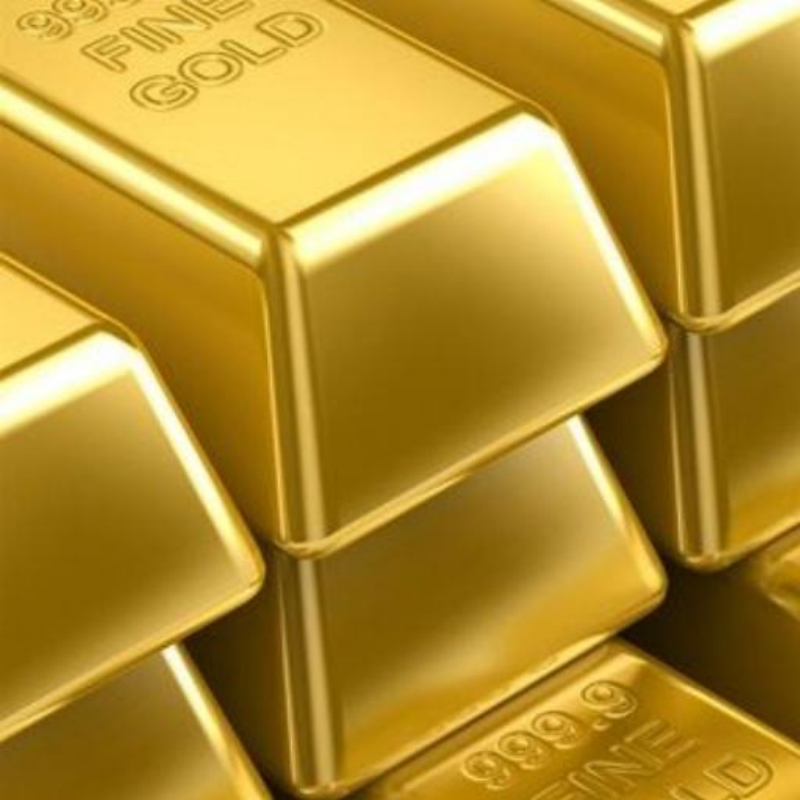}%
                };
            \end{tikzpicture}
        \end{minipage} &

        \begin{minipage}[c][0.5\textwidth][c]{\imgwidthh}
            \centering
            \includegraphics[width=\textwidth,height=\textwidth]{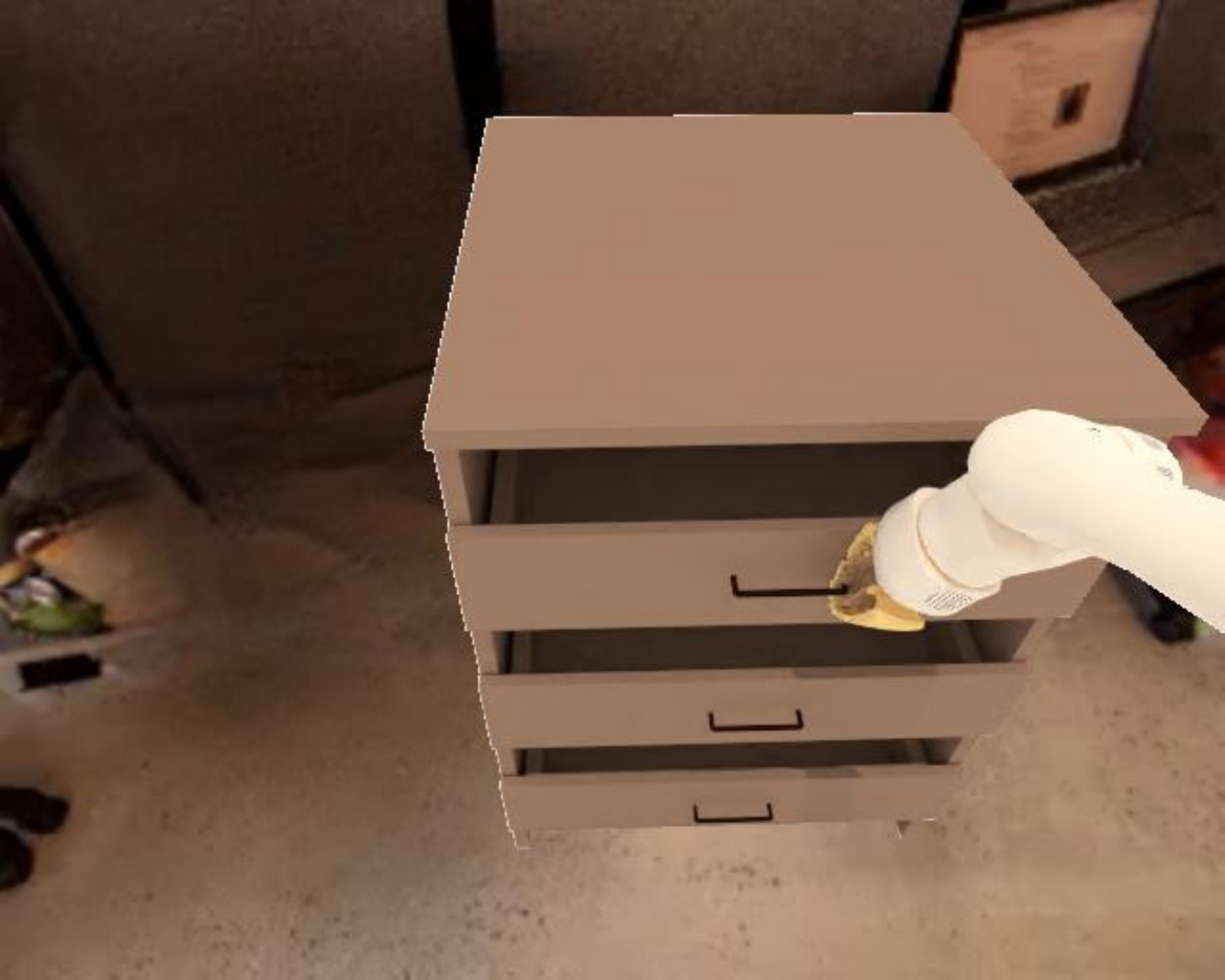} \\ [.5em]
            \includegraphics[width=\textwidth,height=\textwidth]{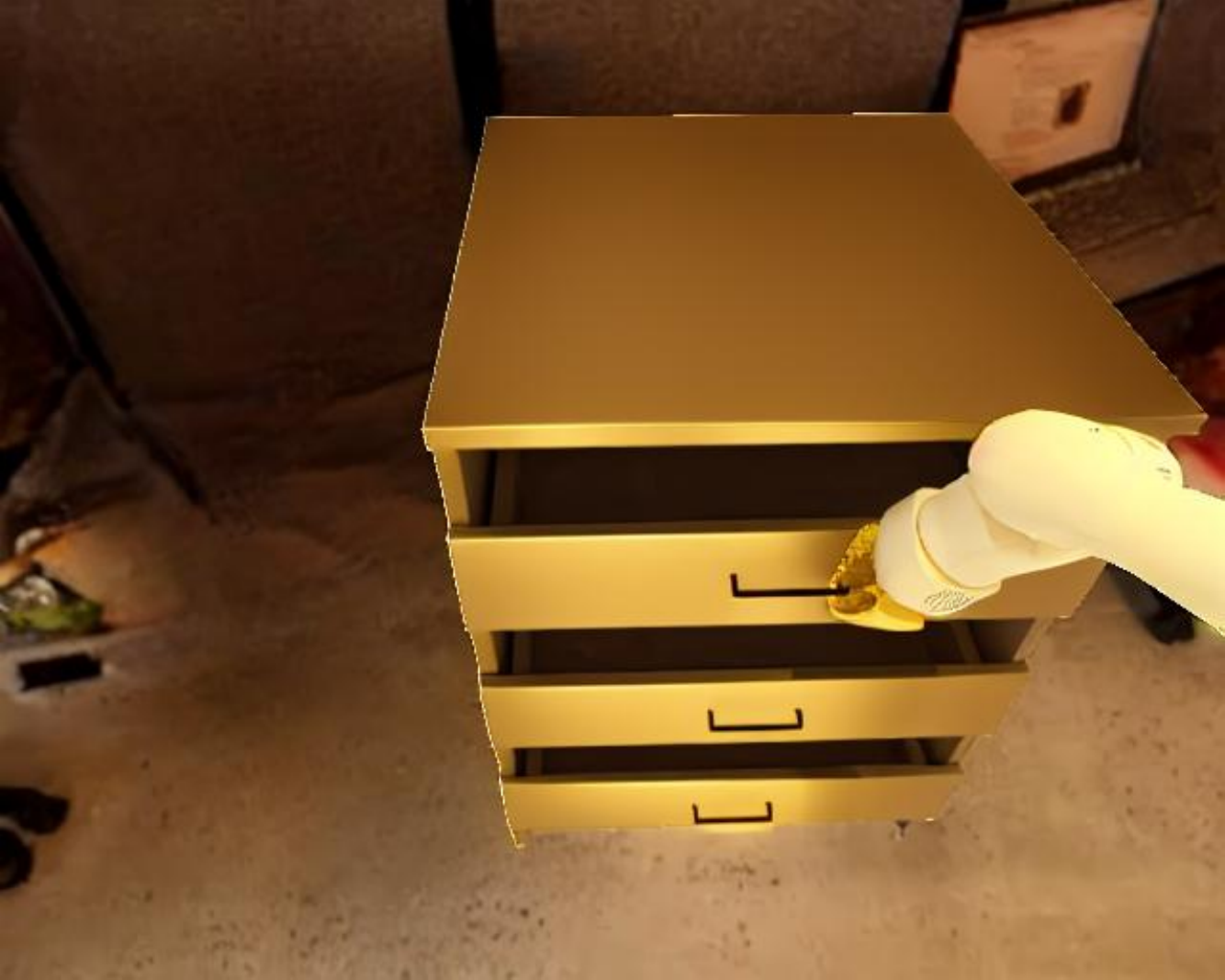}
        \end{minipage} &

        \begin{minipage}[c][0.5\textwidth][c]{\imgwidthh}
            \centering
            \includegraphics[width=\textwidth,height=\textwidth]{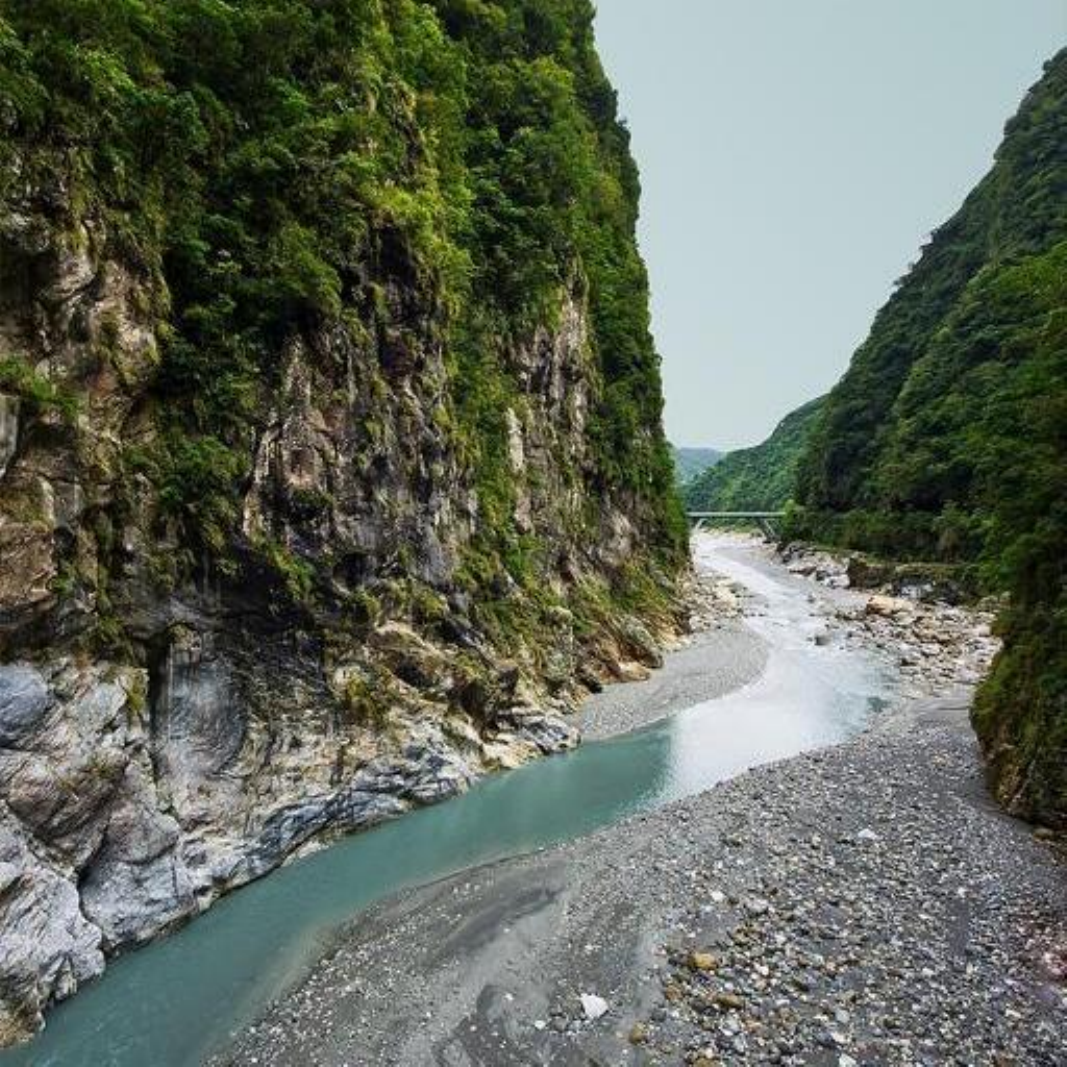} \\ [.5em]
            \includegraphics[width=\textwidth,height=\textwidth]{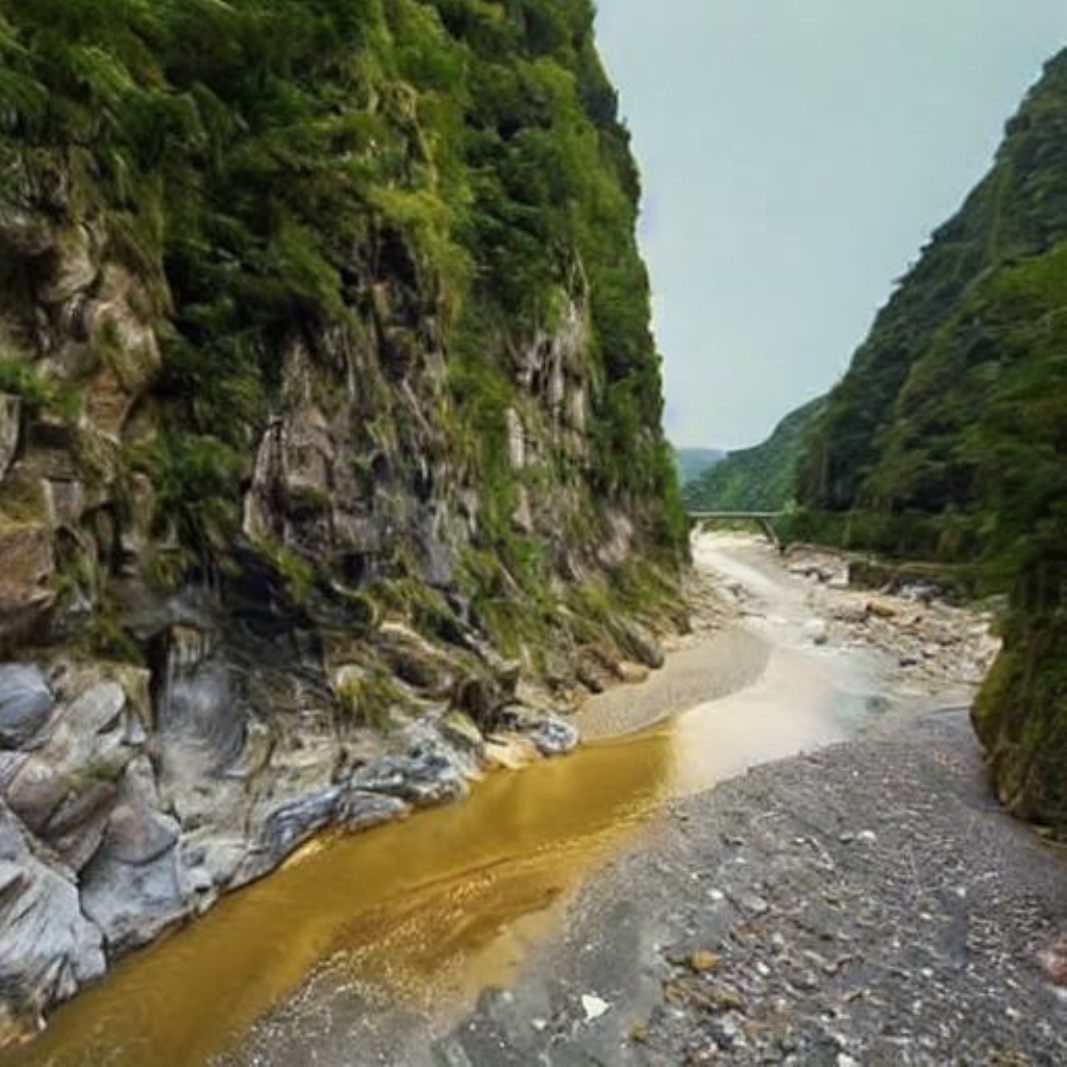}
        \end{minipage} &

        \begin{minipage}[c][0.5\textwidth][c]{\imgwidthh}
            \centering
            \includegraphics[width=\textwidth,height=\textwidth]{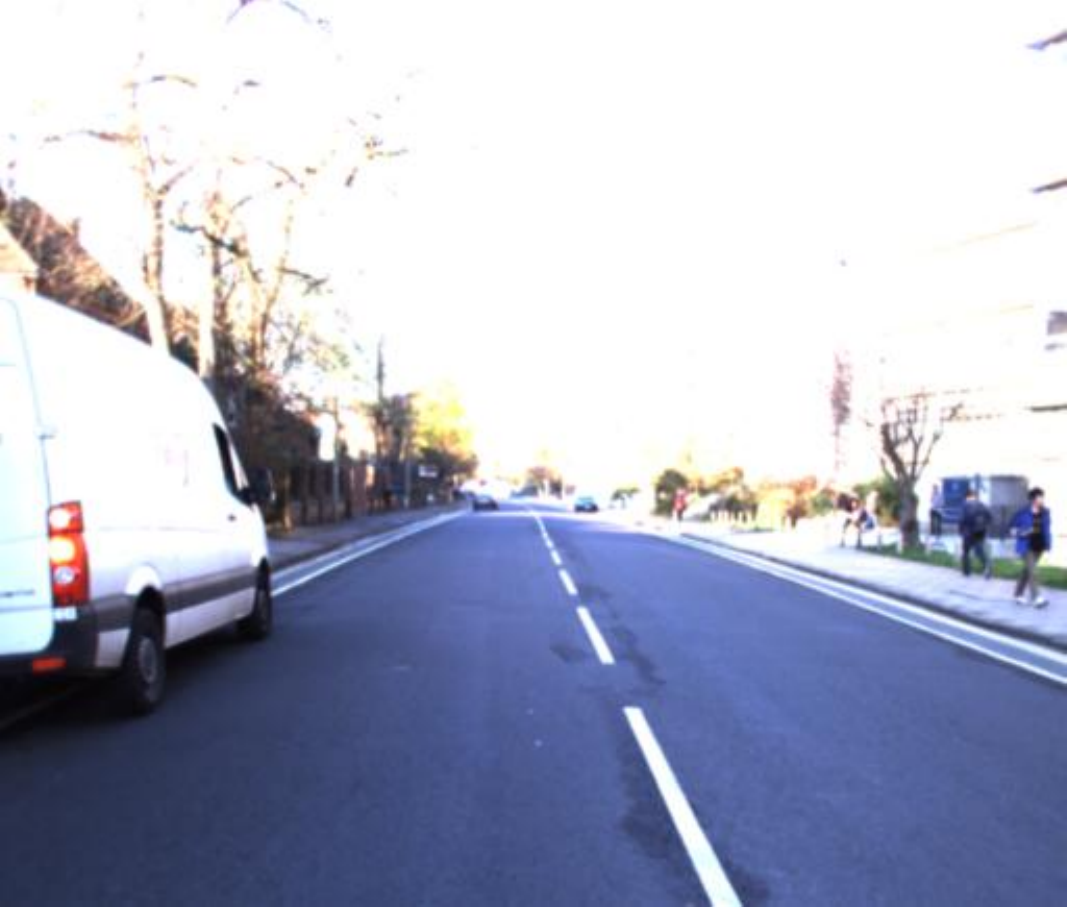} \\ [.5em]
            \includegraphics[width=\textwidth,height=\textwidth]{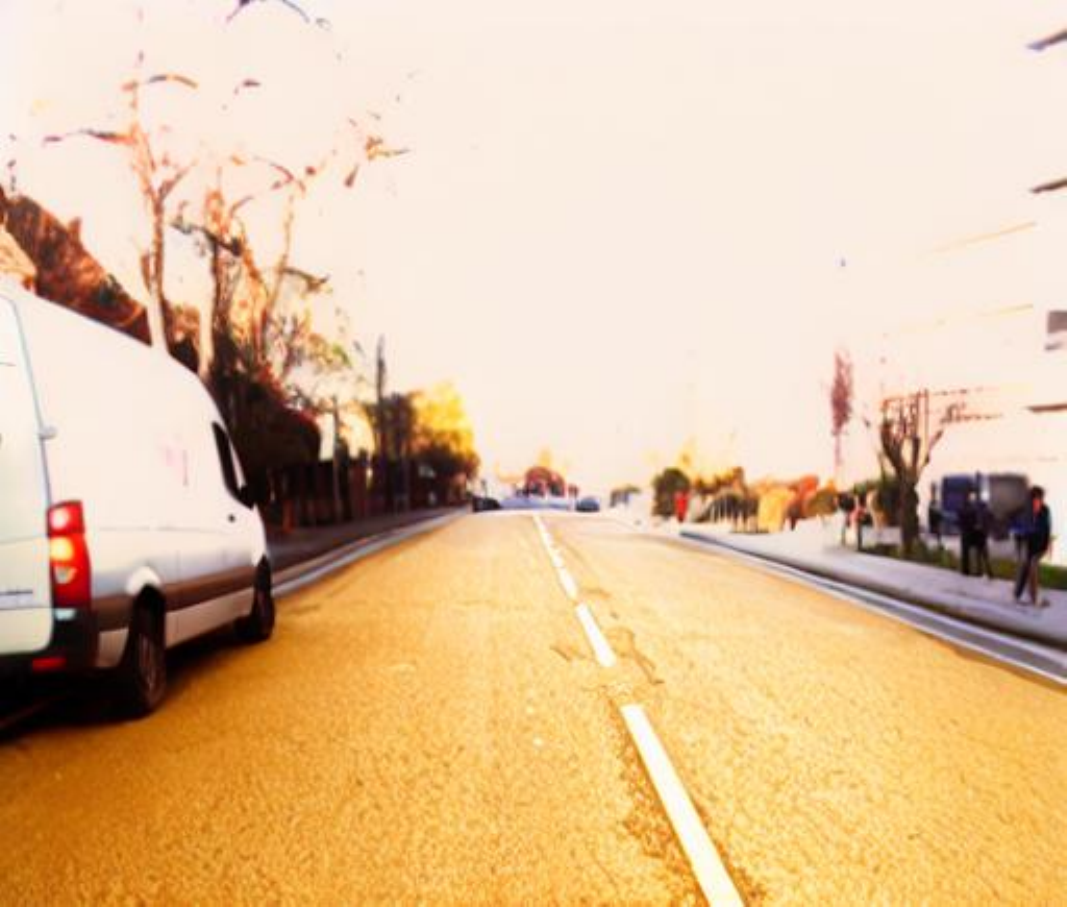}
        \end{minipage}
    \end{tabular}
    \caption{The visual prompts can be precisely replicated across scenes with diverse contexts and layouts, demonstrating the flexibility of our approach. We showcase variants of input images, including synthetic and real-world scenes, small and large elements, and various materials, to illustrate the model's ability to generalize effectively. Notably, our method achieves this versatility by leveraging only a few visual exemplars during training, ensuring robust performance across a wide range of inputs.}
    \label{appdx_gold_multi}
\end{figure}

\begin{figure*}[h]
\centering
\newcommand{\imgwidth}{0.23}
\setlength{\tabcolsep}{2pt} 
\renewcommand{\arraystretch}{1} 
\newcommand{\textspace}{0.7em}
\newcommand{\myhspace}{0.5em} 
\newcommand{\stycompression}{0.05}
\begin{tabular}{cccc} 
    Input & Edit \hspace{\myhspace} & \hspace{\myhspace} Input & Edit\\ [0.3em]
    \begin{overpic}[width=\imgwidth\textwidth]{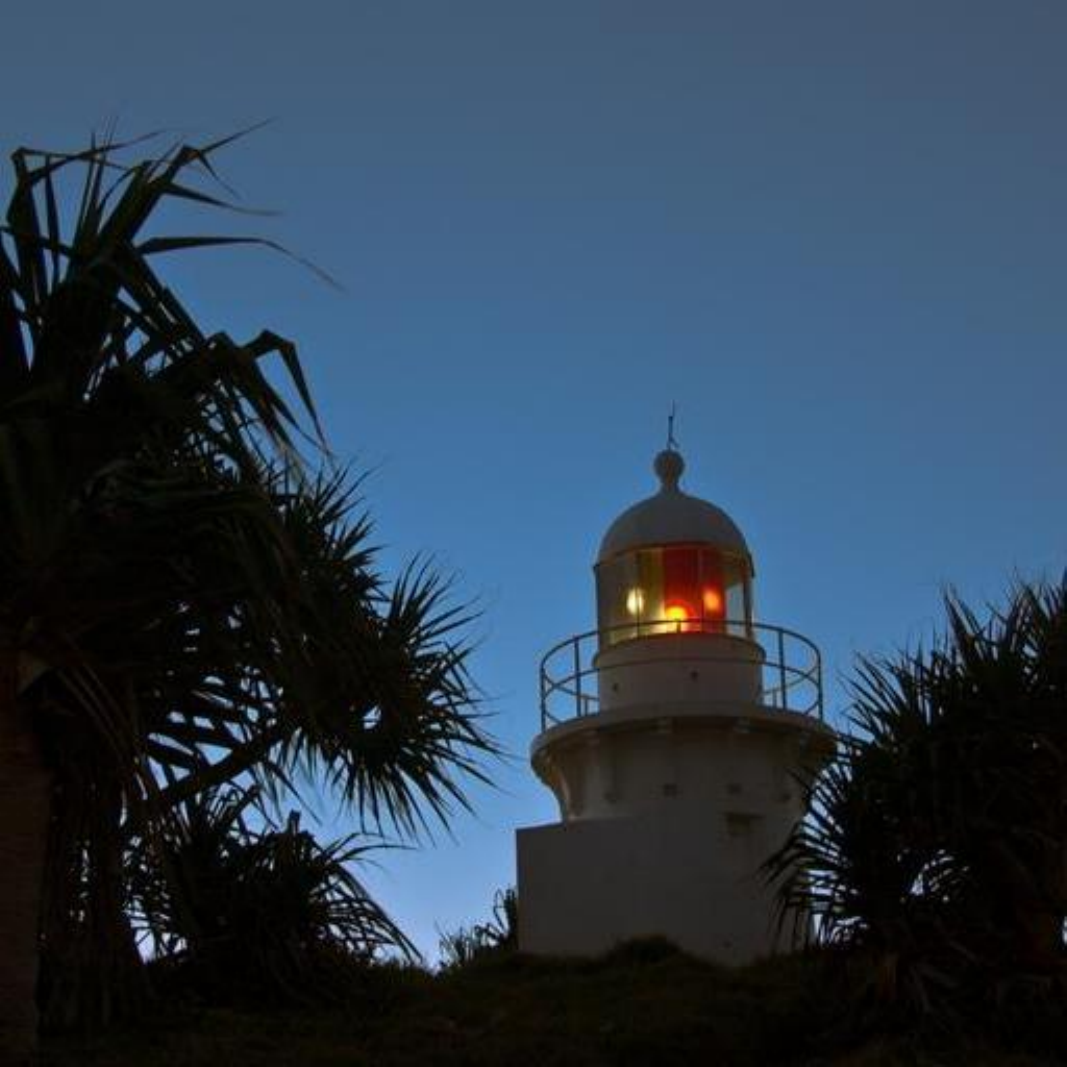}
    \put(1, 74){
     \begin{tikzpicture}
      \begin{scope}
       \clip[rounded corners=5pt] (0,0) rectangle (\stycompression\textwidth, \stycompression\textwidth);
       \node[anchor=north east, inner sep=0pt] at (\stycompression\textwidth,\stycompression\textwidth) {\includegraphics[width=\stycompression\textwidth]{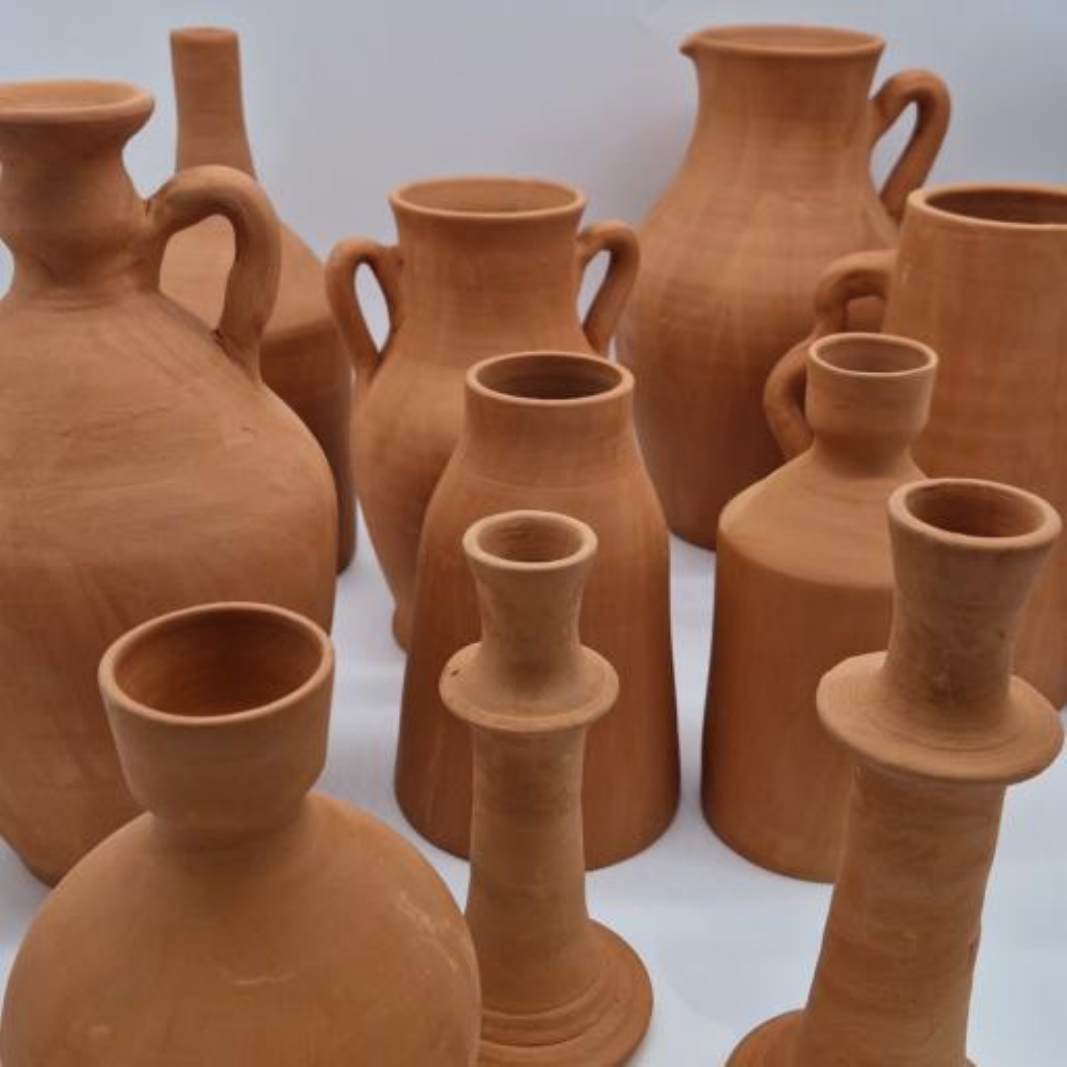}};
    \end{scope}
    \draw[rounded corners=5pt, RubineRed, very thick, dashed] (0,0) rectangle (\stycompression\textwidth,\stycompression\textwidth);
    \end{tikzpicture}}
    \end{overpic}&
    \includegraphics[width=\imgwidth\textwidth]{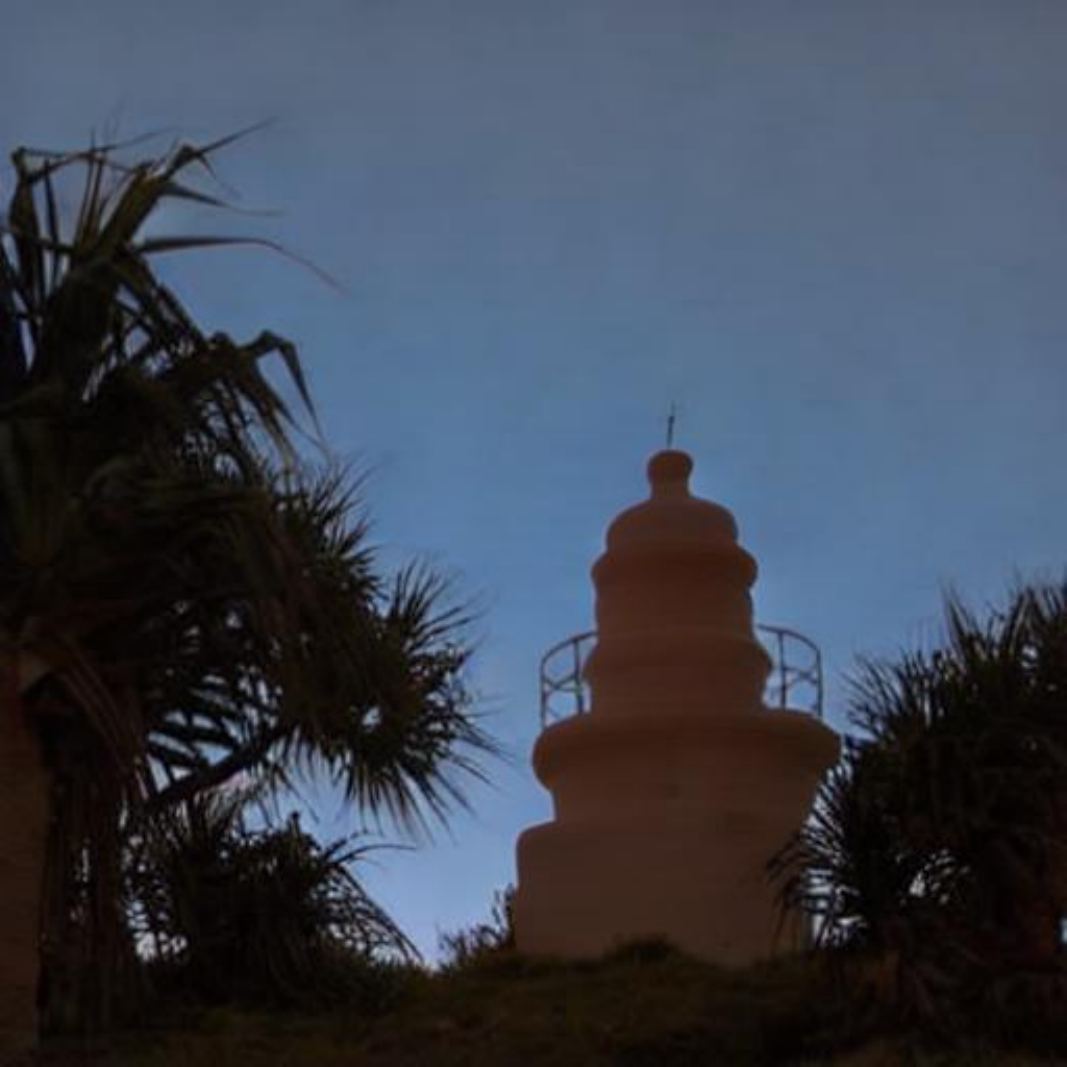} \hspace{\myhspace} &\hspace{\myhspace}
    \begin{overpic}[width=\imgwidth\textwidth]{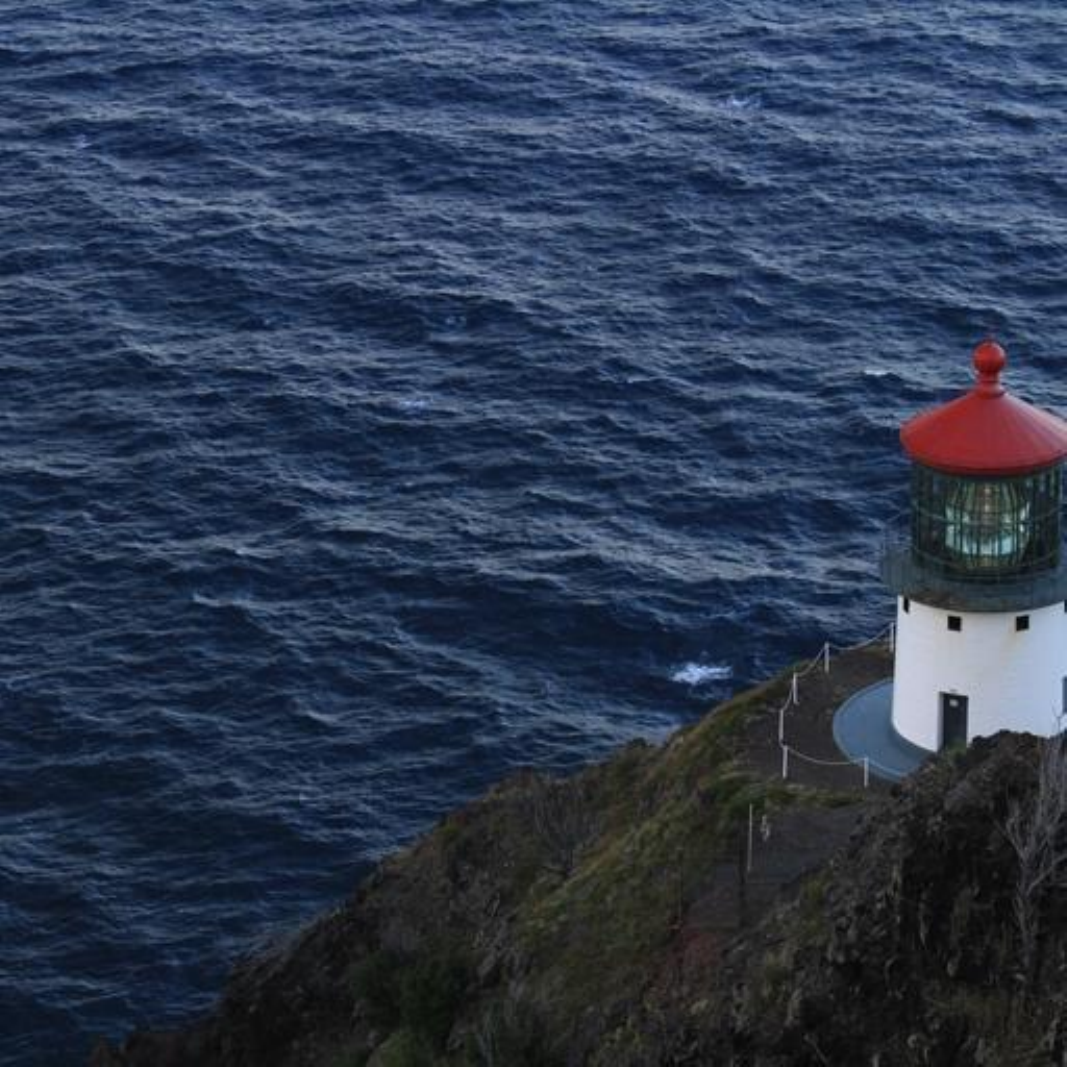}
    \put(1, 74){
     \begin{tikzpicture}
      \begin{scope}
       \clip[rounded corners=5pt] (0,0) rectangle (\stycompression\textwidth, \stycompression\textwidth);
       \node[anchor=north east, inner sep=0pt] at (\stycompression\textwidth,\stycompression\textwidth) {\includegraphics[width=\stycompression\textwidth]{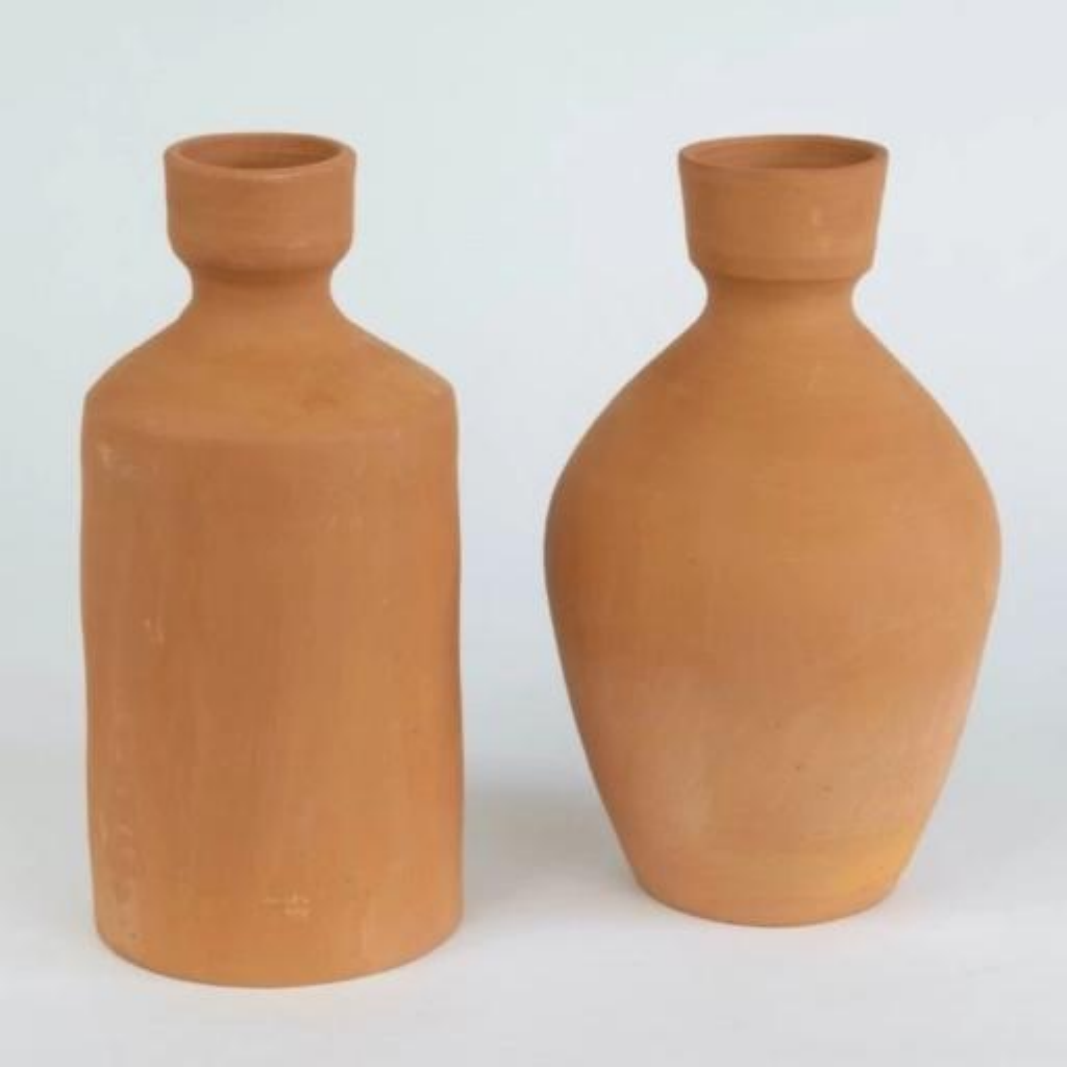}};
    \end{scope}
    \draw[rounded corners=5pt, RubineRed, very thick, dashed] (0,0) rectangle (\stycompression\textwidth,\stycompression\textwidth);
    \end{tikzpicture}}
    \end{overpic} &
    \includegraphics[width=\imgwidth\textwidth]{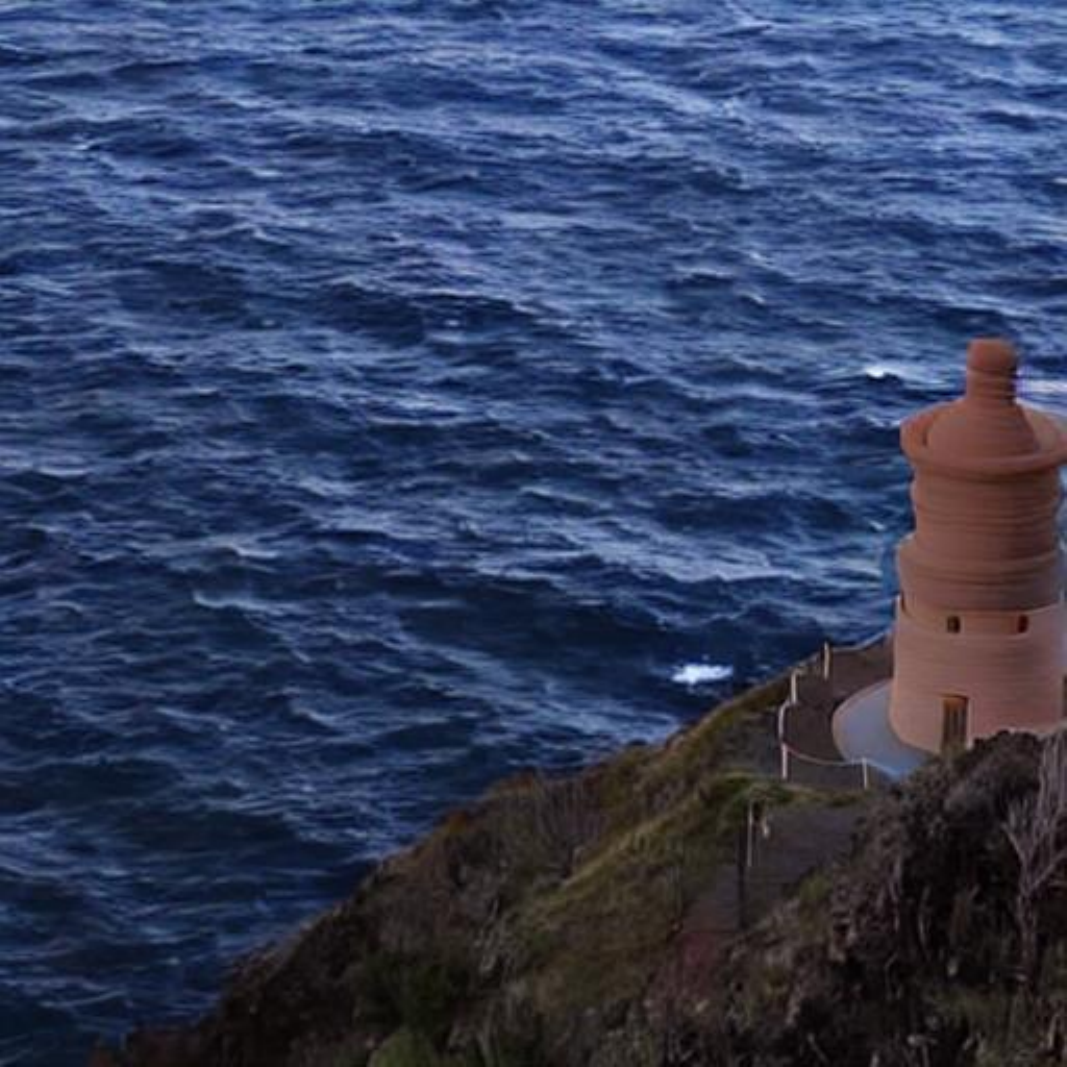}\\
    \multicolumn{4}{c}{``Change the lighthouse into terracotta"}\\[\textspace]

    \begin{overpic}[width=\imgwidth\textwidth]{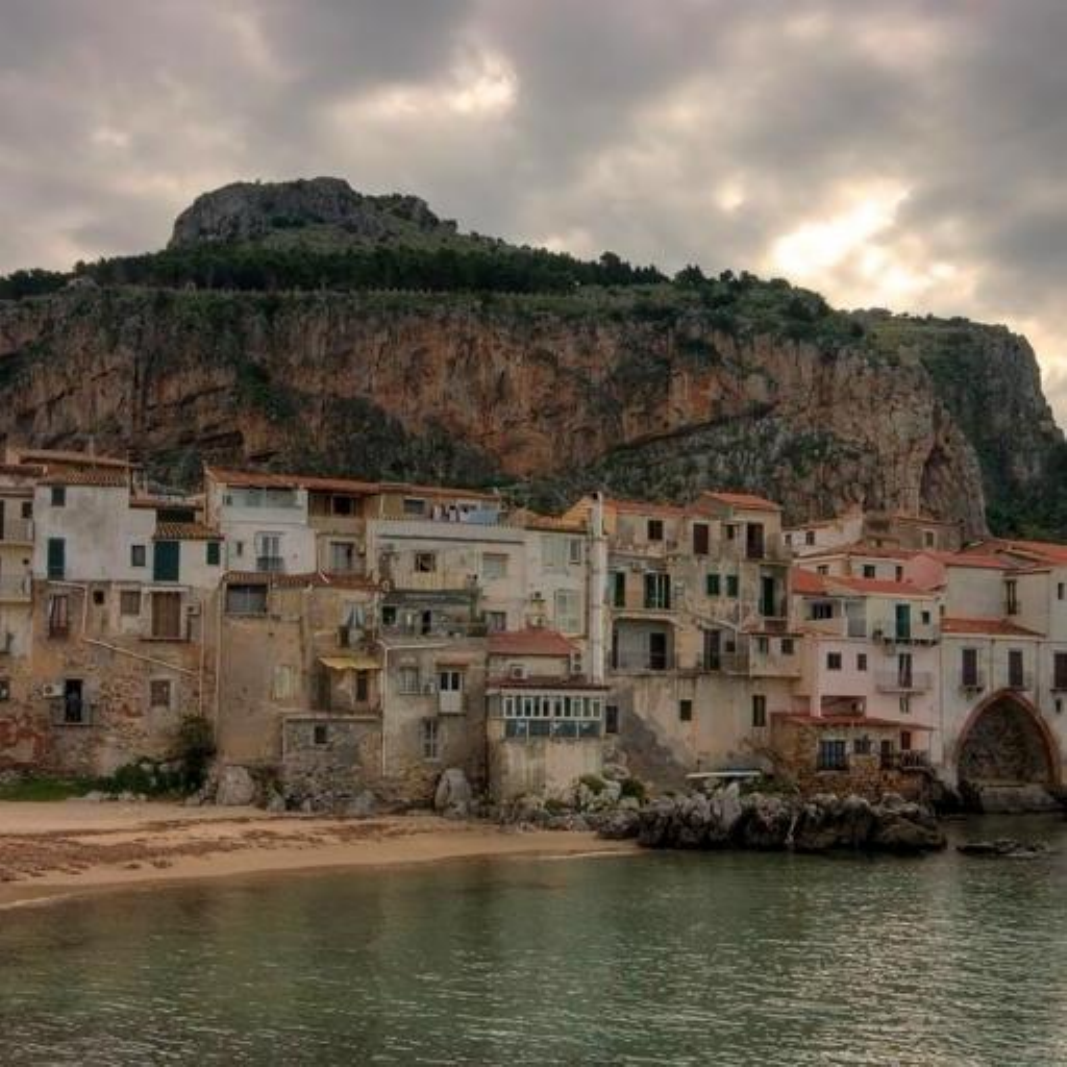}
    \put(1, 74){
     \begin{tikzpicture}
      \begin{scope}
       \clip[rounded corners=5pt] (0,0) rectangle (\stycompression\textwidth, \stycompression\textwidth);
       \node[anchor=north east, inner sep=0pt] at (\stycompression\textwidth,\stycompression\textwidth) {\includegraphics[width=\stycompression\textwidth]{images/qualitative/all_pdf_512/tgt/gold/gold0..pdf}};
    \end{scope}
    \draw[rounded corners=5pt, RubineRed, very thick, dashed] (0,0) rectangle (\stycompression\textwidth,\stycompression\textwidth);
    \end{tikzpicture}}
    \end{overpic} &
    \includegraphics[width=\imgwidth\textwidth]{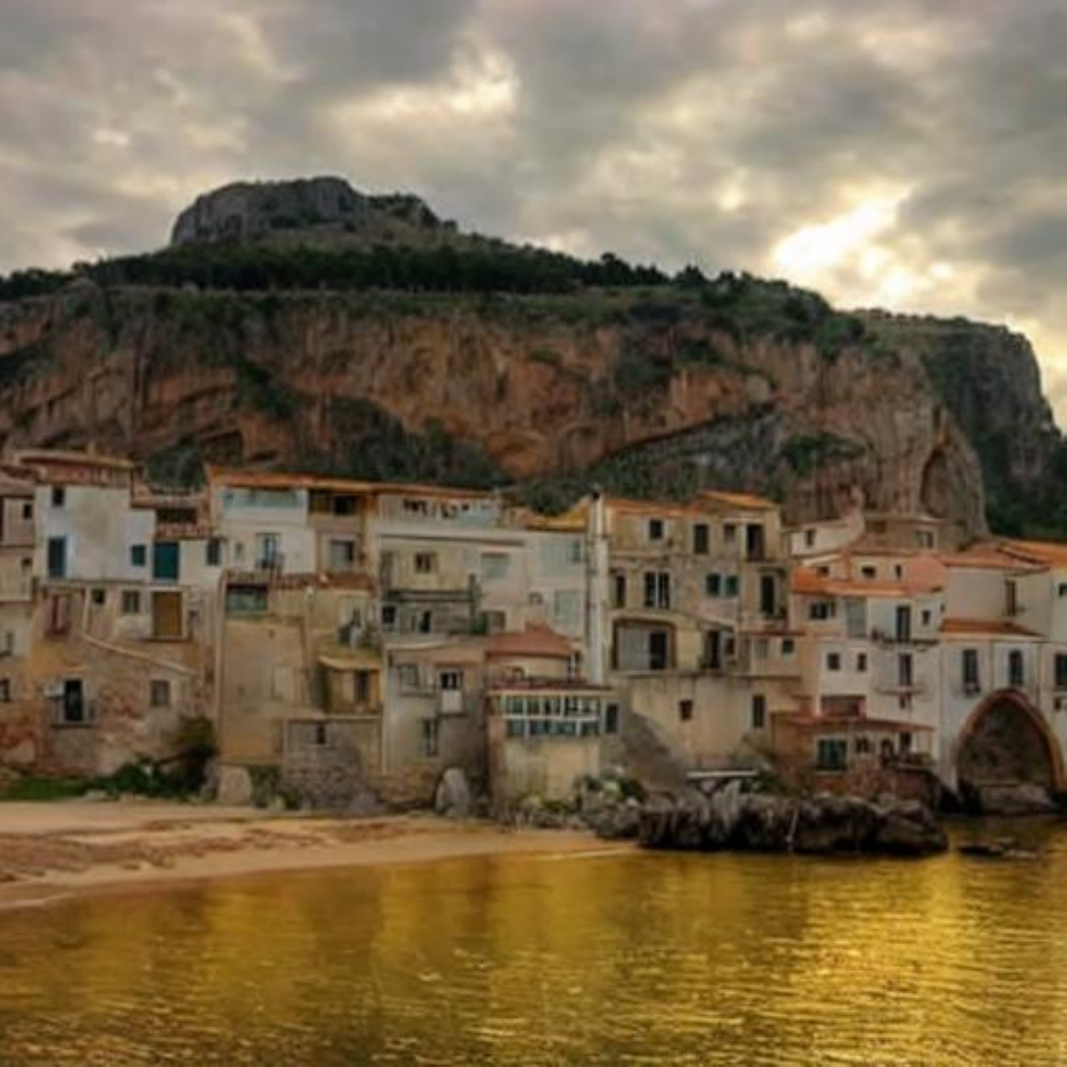} \hspace{\myhspace} &\hspace{\myhspace
    }
    \begin{overpic}[width=\imgwidth\textwidth]{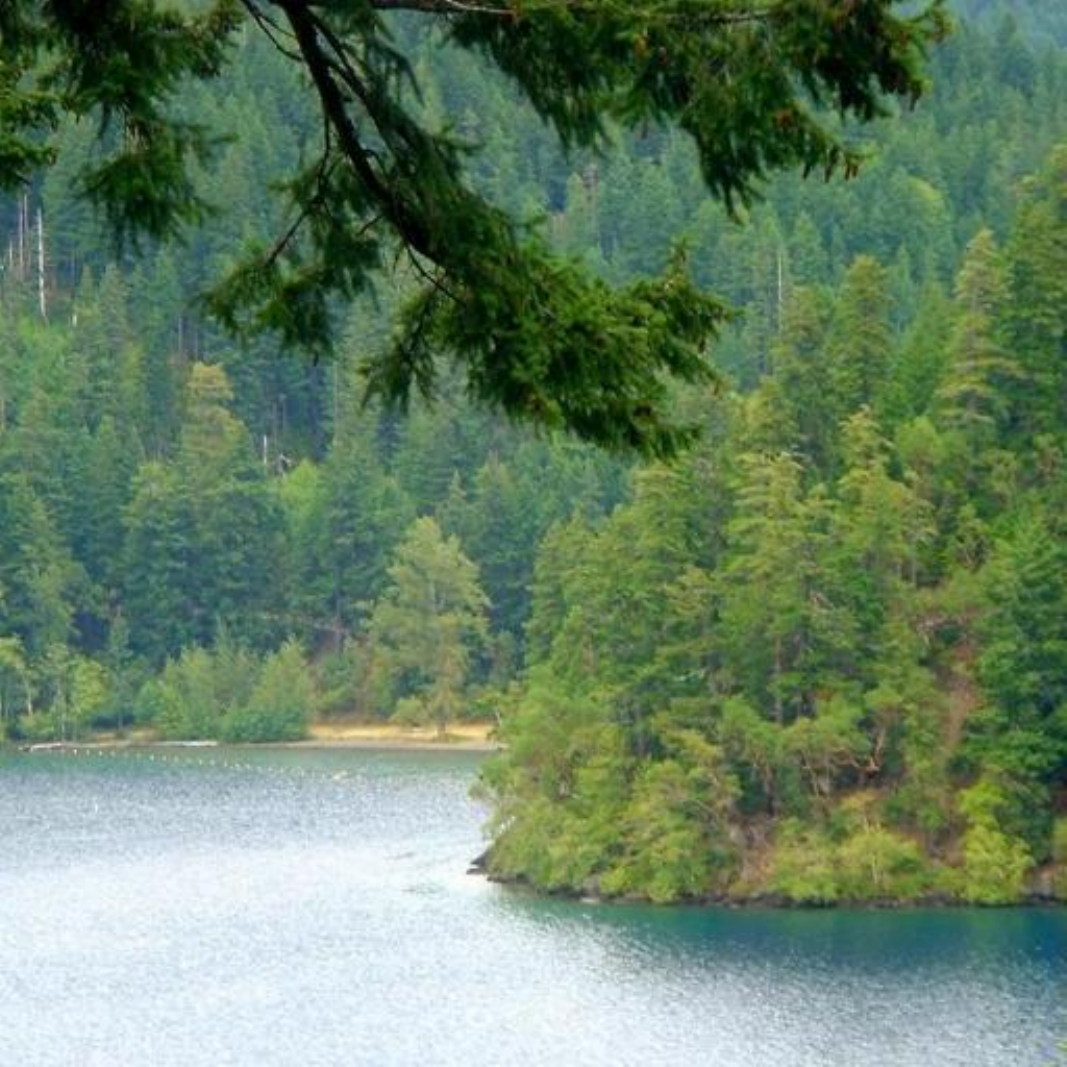}
    \put(1, 74){
     \begin{tikzpicture}
      \begin{scope}
       \clip[rounded corners=5pt] (0,0) rectangle (\stycompression\textwidth, \stycompression\textwidth);
       \node[anchor=north east, inner sep=0pt] at (\stycompression\textwidth,\stycompression\textwidth) {\includegraphics[width=\stycompression\textwidth]{images/qualitative/all_pdf_512/tgt/gold/gold_extra.pdf}};
    \end{scope}
    \draw[rounded corners=5pt, RubineRed, very thick, dashed] (0,0) rectangle (\stycompression\textwidth,\stycompression\textwidth);
    \end{tikzpicture}}
    \end{overpic}&
    \includegraphics[width=\imgwidth\textwidth]{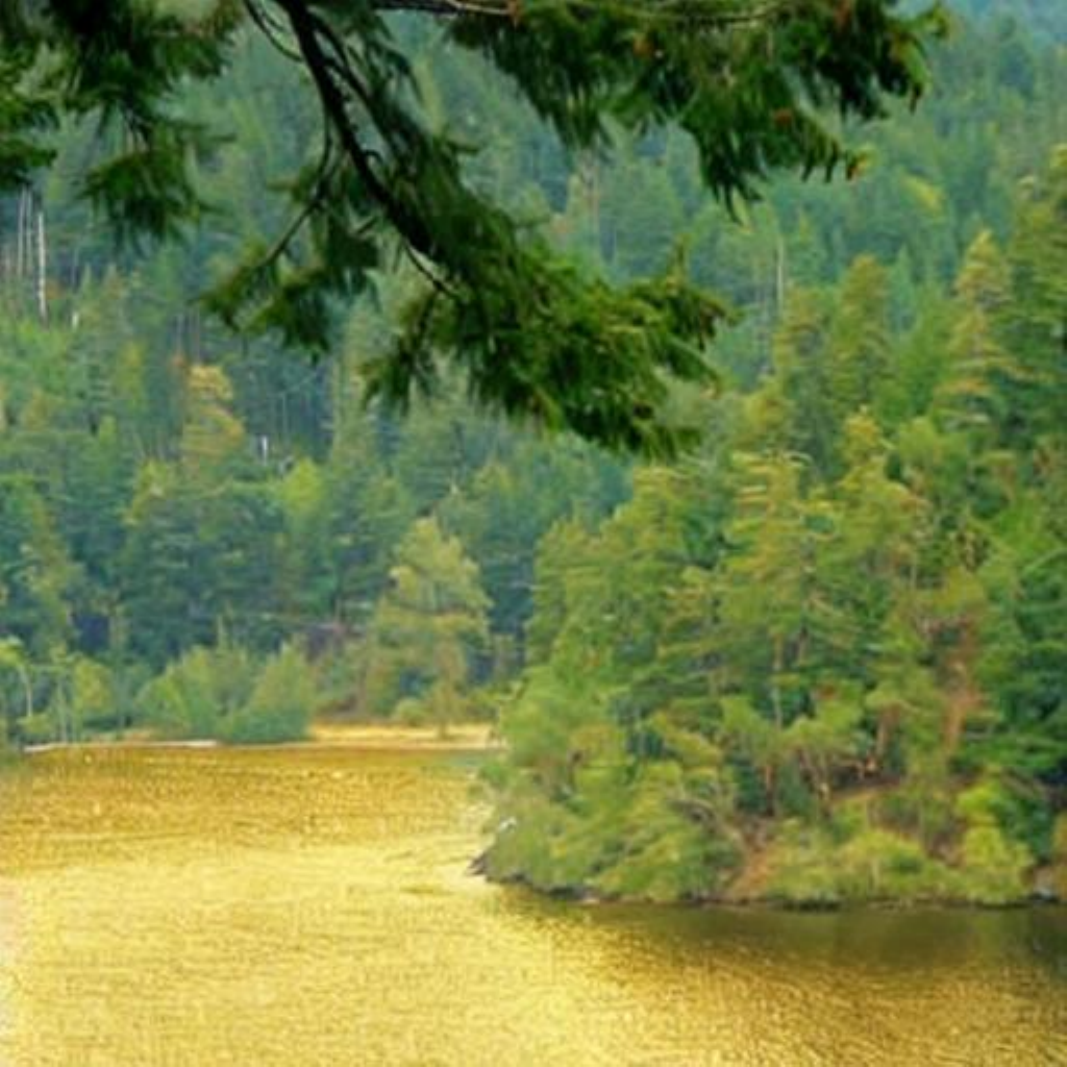}\\
    \multicolumn{4}{c}{``Change the water for gold"}\\[\textspace]

    \begin{overpic}[width=\imgwidth\textwidth]{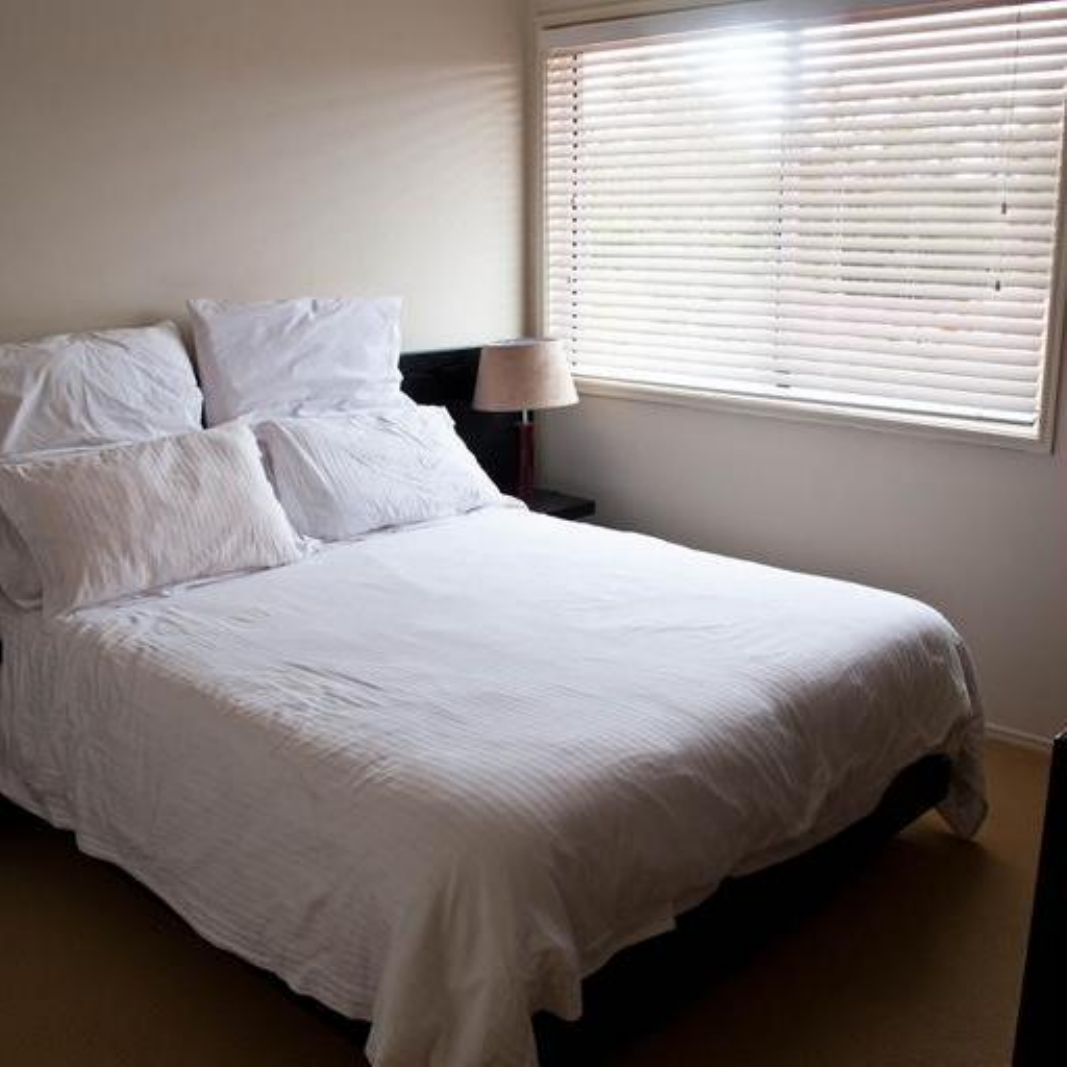}
    \put(1, 74){
     \begin{tikzpicture}
      \begin{scope}
       \clip[rounded corners=5pt] (0,0) rectangle (\stycompression\textwidth, \stycompression\textwidth);
       \node[anchor=north east, inner sep=0pt] at (\stycompression\textwidth,\stycompression\textwidth) {\includegraphics[width=\stycompression\textwidth]{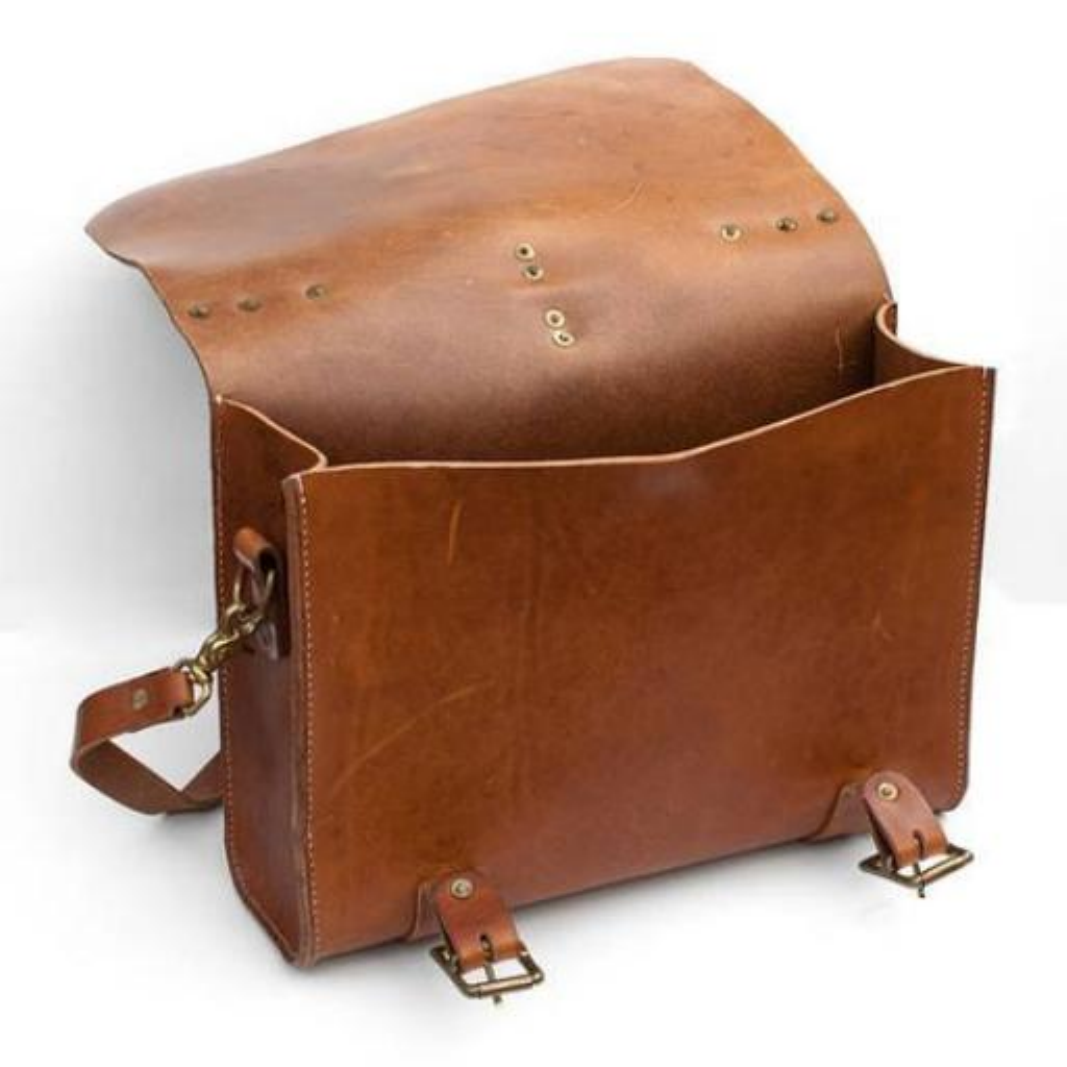}};
    \end{scope}
    \draw[rounded corners=5pt, RubineRed, very thick, dashed] (0,0) rectangle (\stycompression\textwidth,\stycompression\textwidth);
    \end{tikzpicture}}
    \end{overpic}&
    \includegraphics[width=\imgwidth\textwidth]{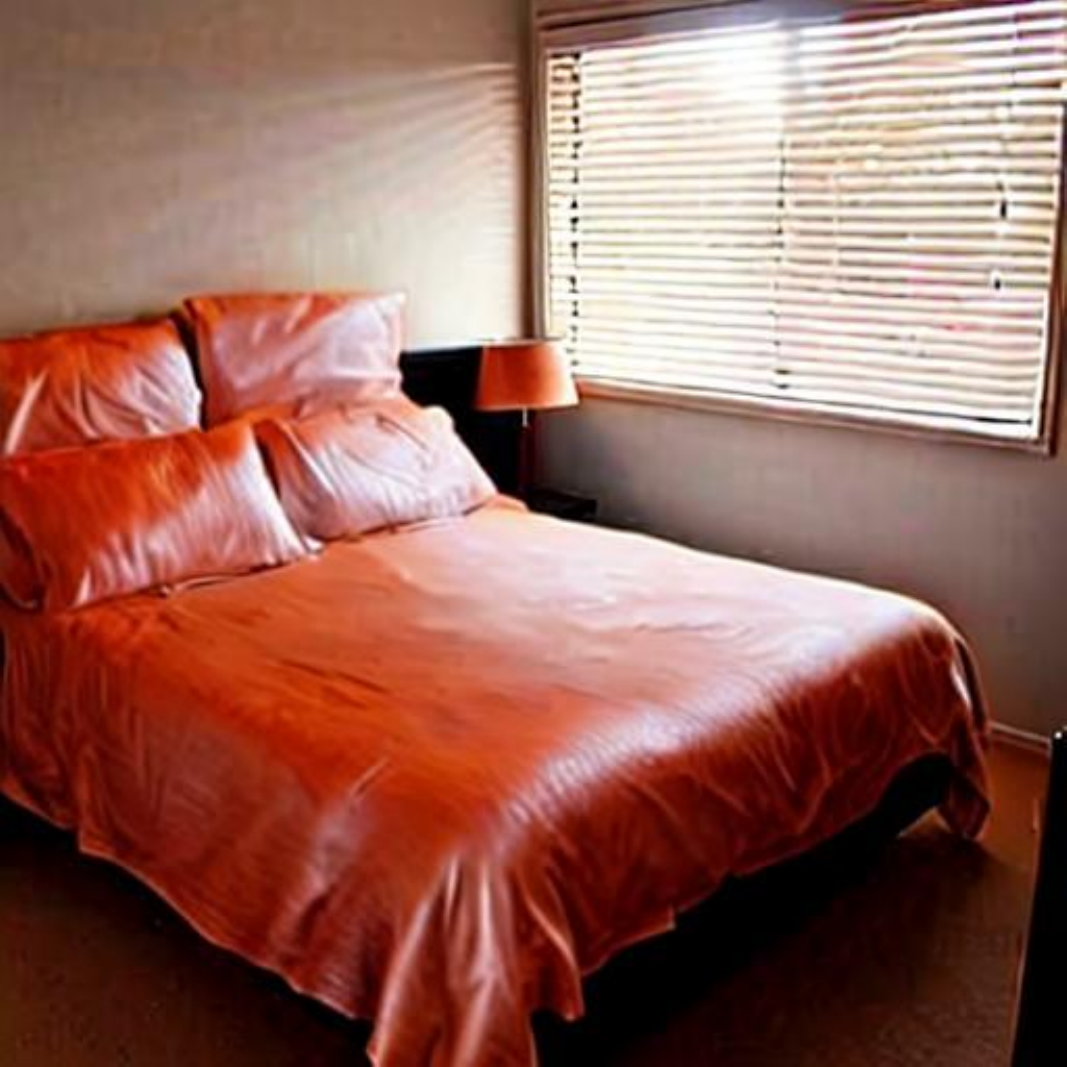} \hspace{\myhspace} &\hspace{\myhspace}
    \begin{overpic}[width=\imgwidth\textwidth]{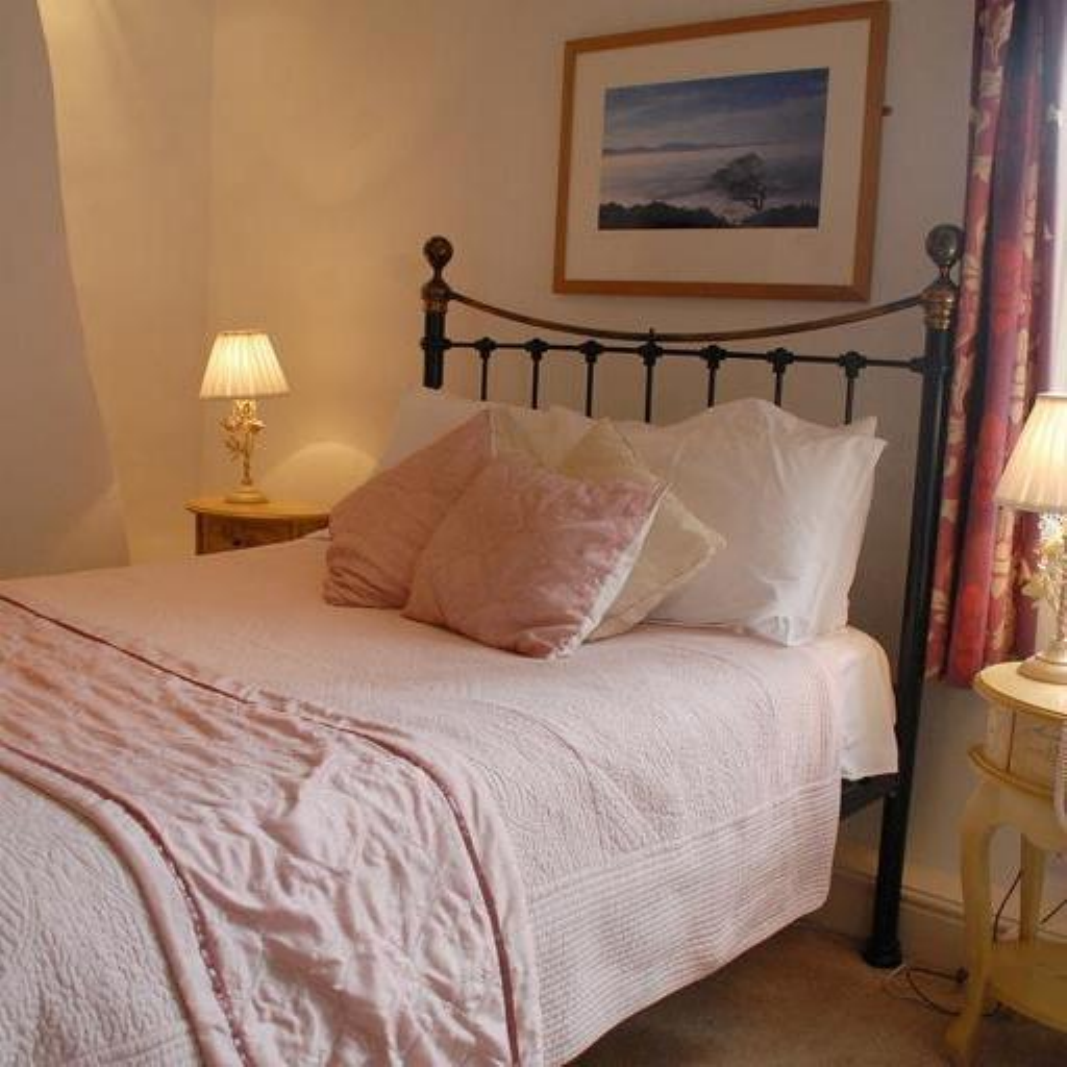}
    \put(1, 74){
     \begin{tikzpicture}
      \begin{scope}
       \clip[rounded corners=5pt] (0,0) rectangle (\stycompression\textwidth, \stycompression\textwidth);
       \node[anchor=north east, inner sep=0pt] at (\stycompression\textwidth,\stycompression\textwidth) {\includegraphics[width=\stycompression\textwidth]{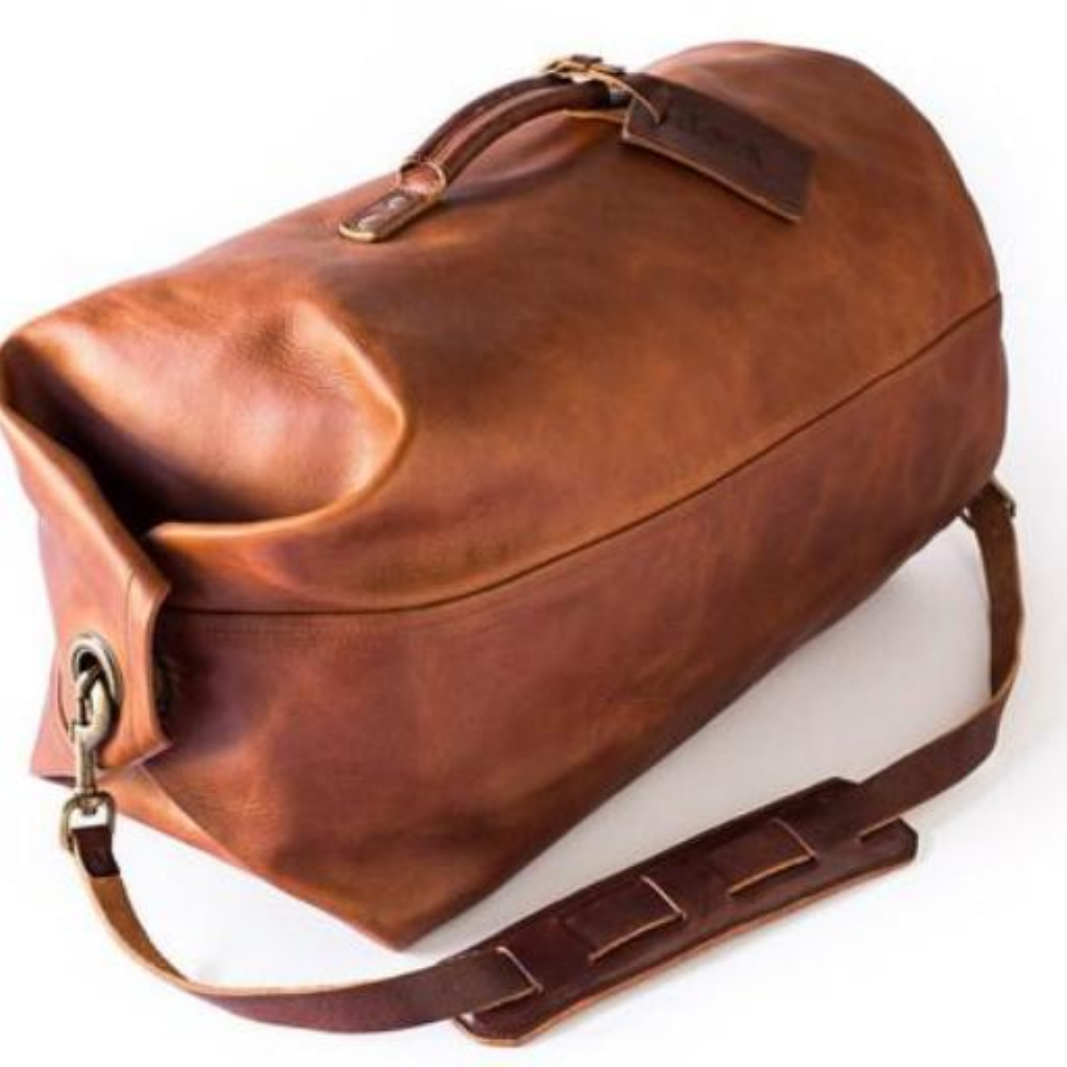}};
    \end{scope}
    \draw[rounded corners=5pt, RubineRed, very thick, dashed] (0,0) rectangle (\stycompression\textwidth,\stycompression\textwidth);
    \end{tikzpicture}}
    \end{overpic}&
    \includegraphics[width=\imgwidth\textwidth]{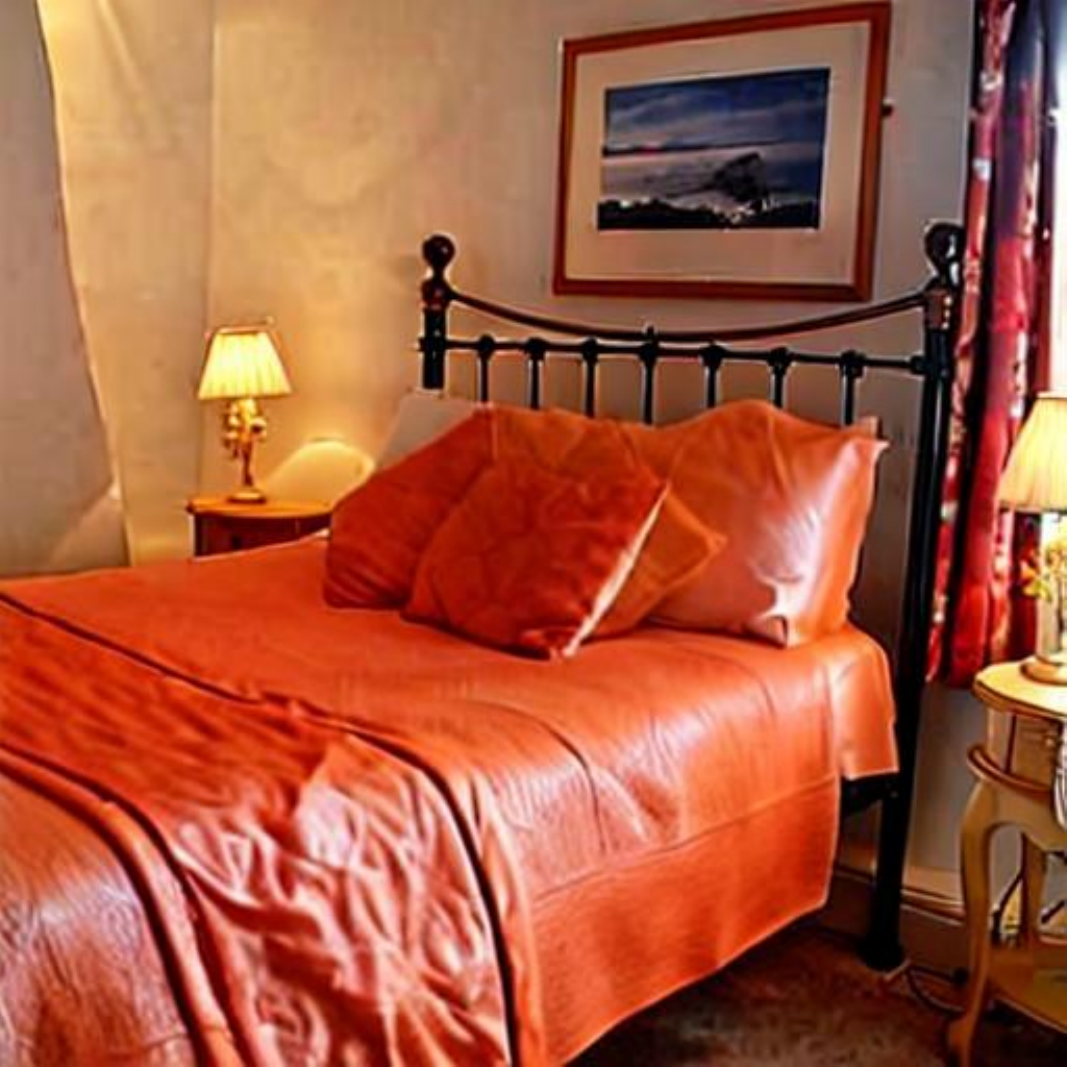}\\
    \multicolumn{4}{c}{``Make the bed out of leather"}\\[\textspace]

    \begin{overpic}[width=\imgwidth\textwidth]{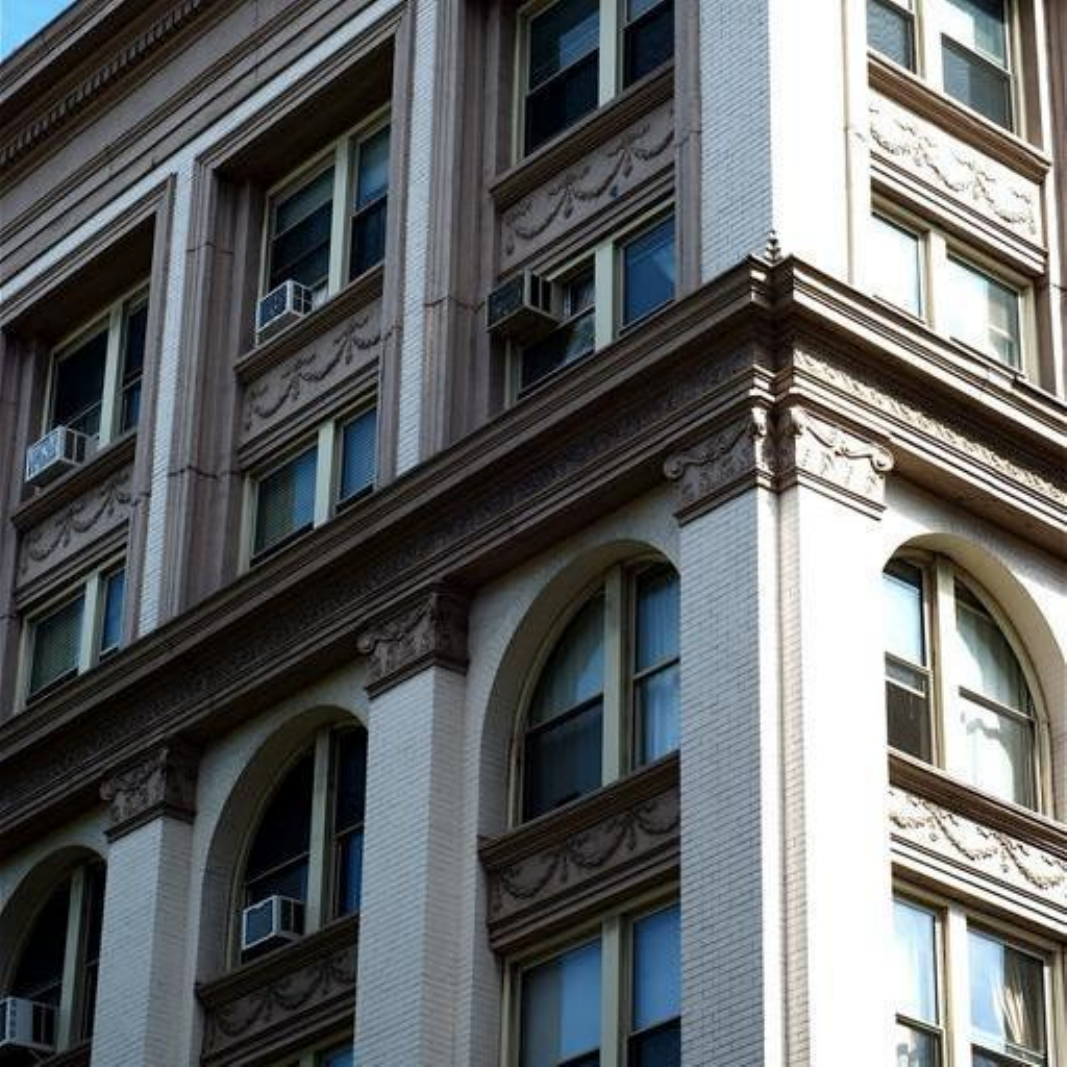}
    \put(1, 74){
     \begin{tikzpicture}
      \begin{scope}
       \clip[rounded corners=5pt] (0,0) rectangle (\stycompression\textwidth, \stycompression\textwidth);
       \node[anchor=north east, inner sep=0pt] at (\stycompression\textwidth,\stycompression\textwidth) {\includegraphics[width=\stycompression\textwidth]{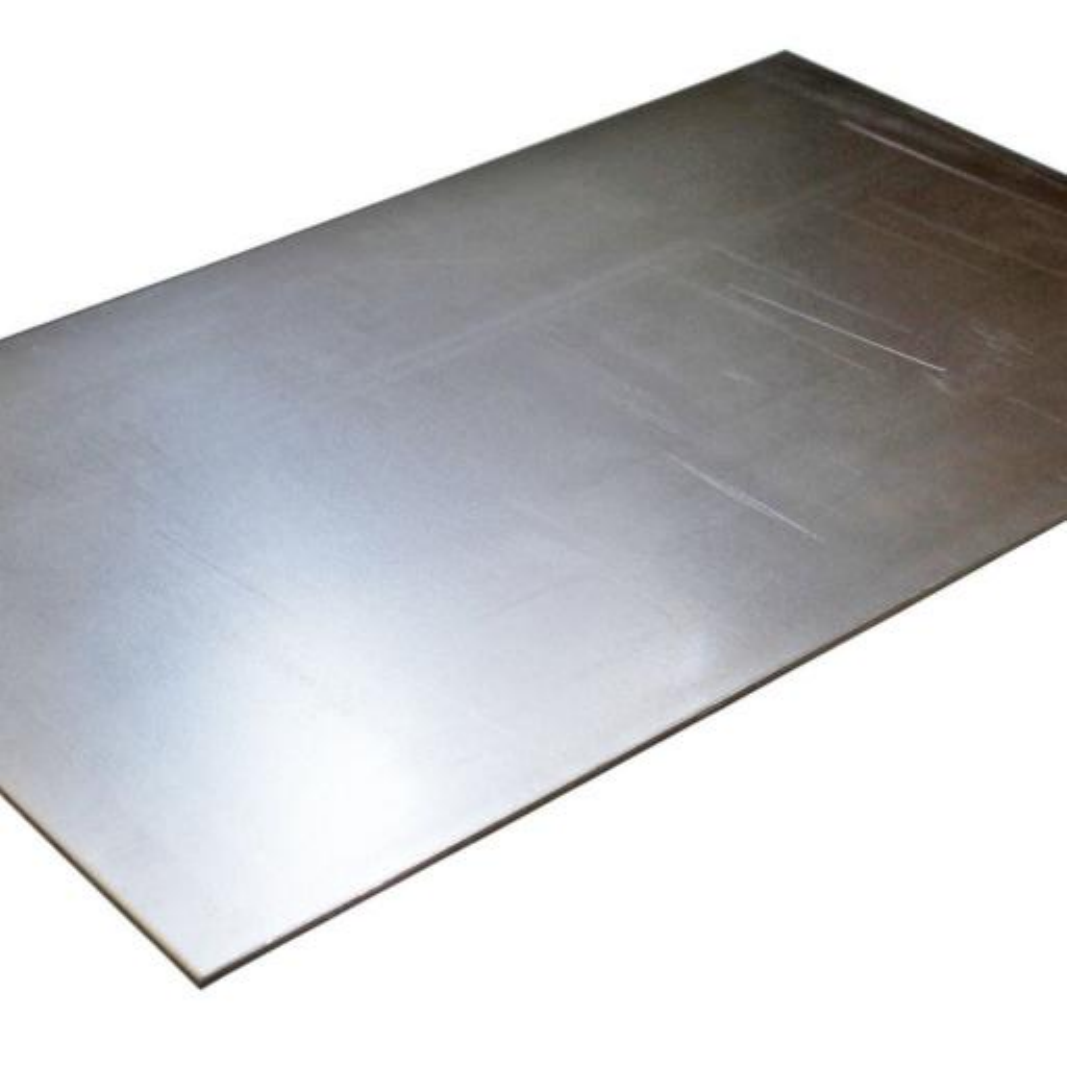}};
    \end{scope}
    \draw[rounded corners=5pt, RubineRed, very thick, dashed] (0,0) rectangle (\stycompression\textwidth,\stycompression\textwidth);
    \end{tikzpicture}}
    \end{overpic}&
    \includegraphics[width=\imgwidth\textwidth]{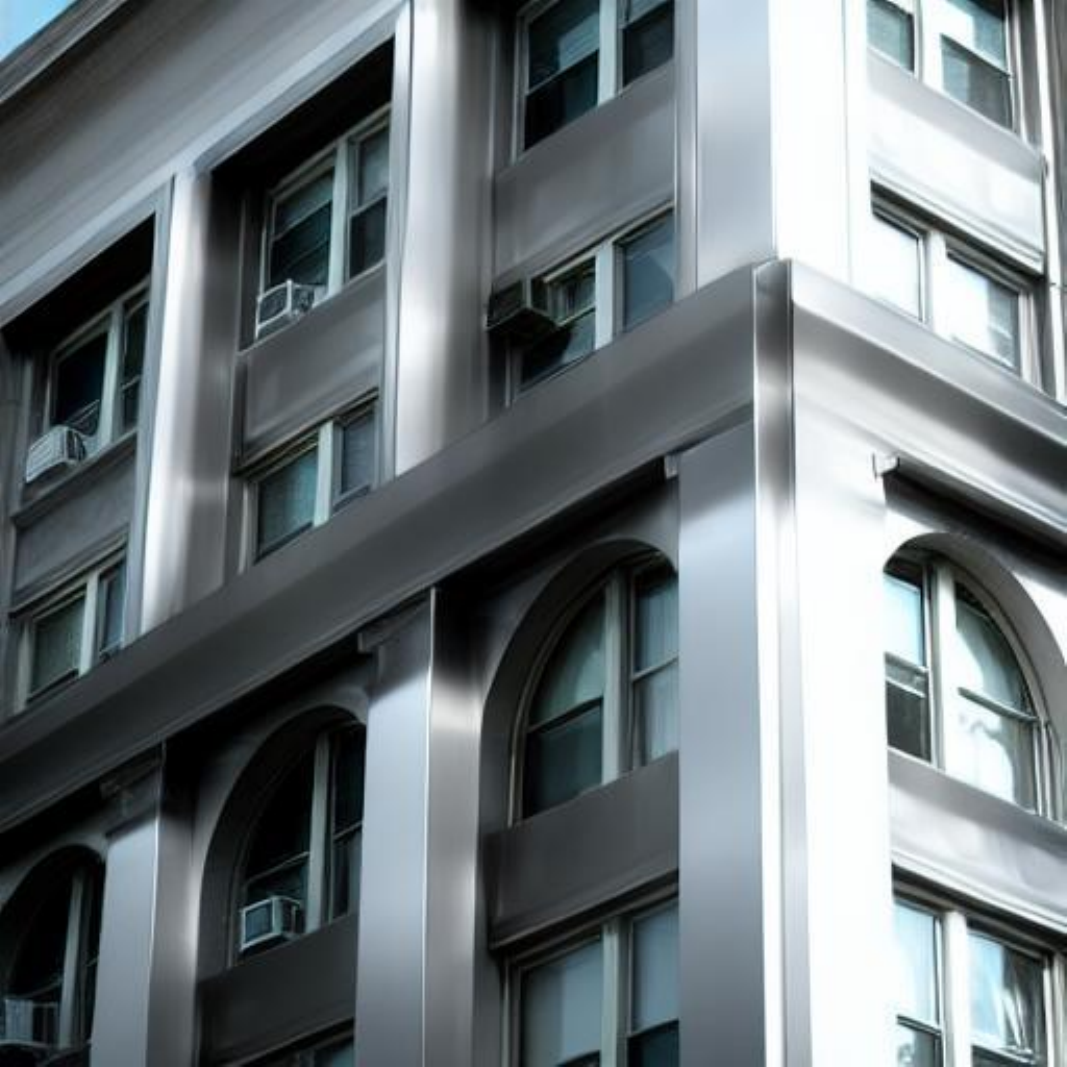} \hspace{\myhspace} & \hspace{\myhspace}
    \begin{overpic}[width=\imgwidth\textwidth]{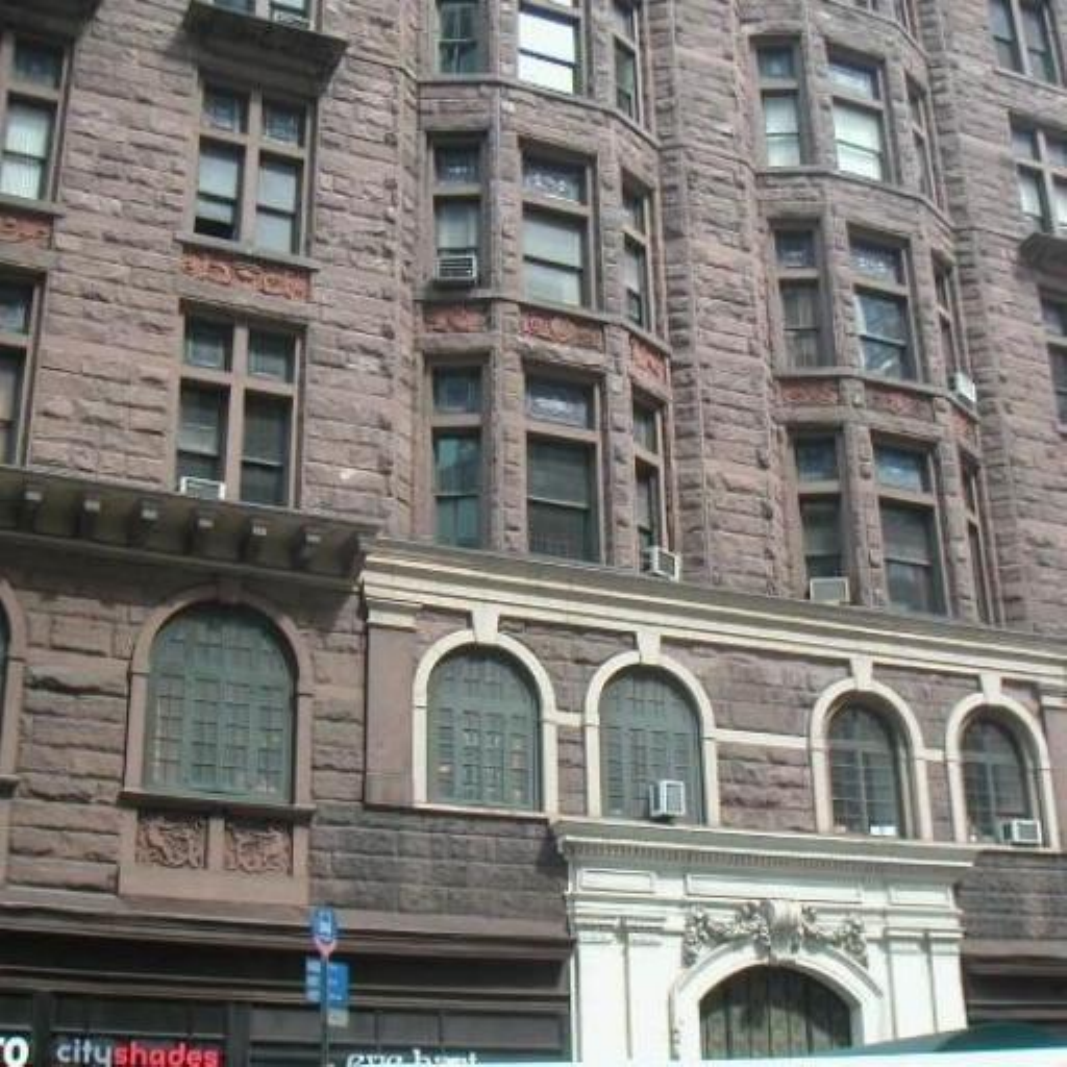}
    \put(1, 74){
     \begin{tikzpicture}
      \begin{scope}
       \clip[rounded corners=5pt] (0,0) rectangle (\stycompression\textwidth, \stycompression\textwidth);
       \node[anchor=north east, inner sep=0pt] at (\stycompression\textwidth,\stycompression\textwidth) {\includegraphics[width=\stycompression\textwidth]{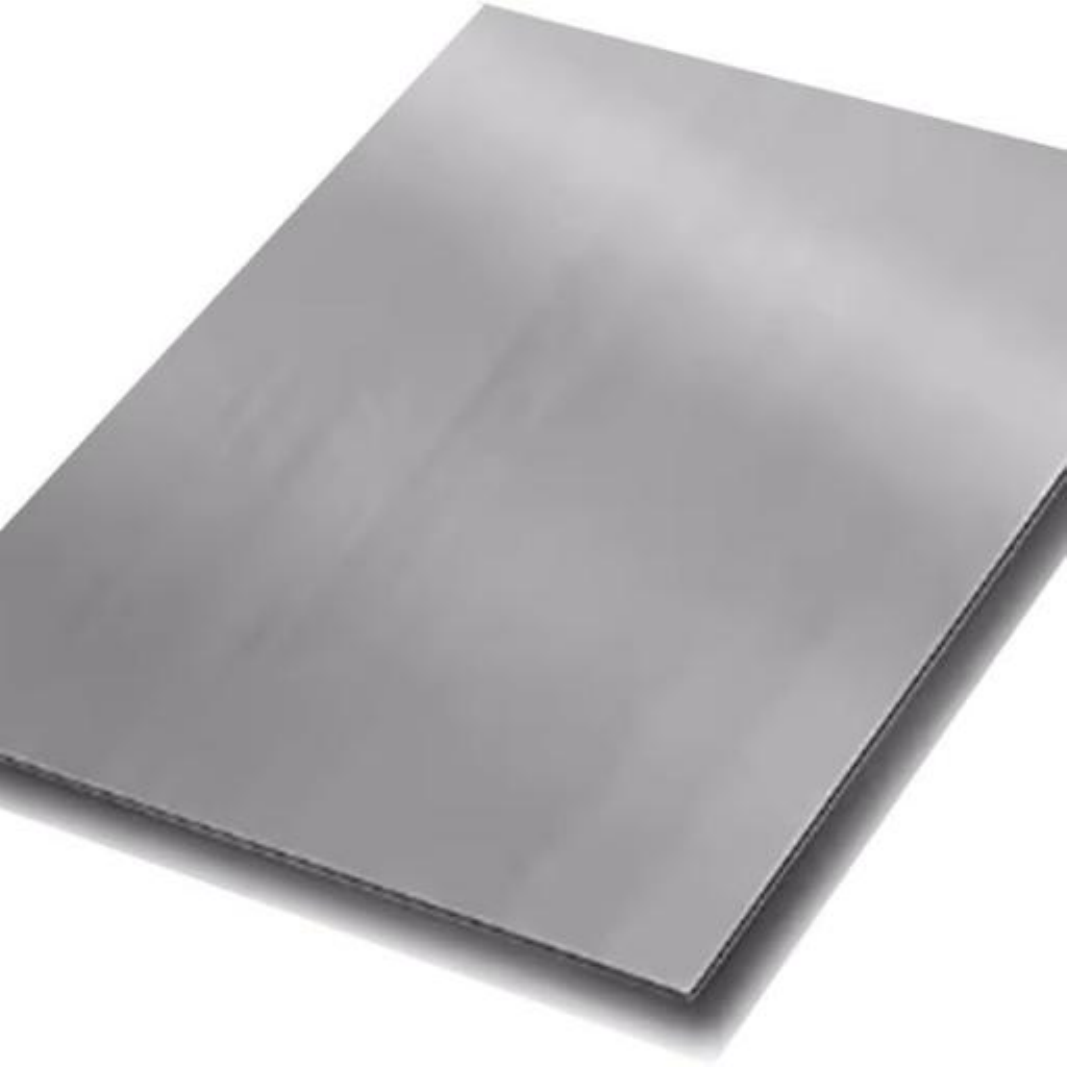}};
    \end{scope}
    \draw[rounded corners=5pt, RubineRed, very thick, dashed] (0,0) rectangle (\stycompression\textwidth,\stycompression\textwidth);
    \end{tikzpicture}}
    \end{overpic}&
    \includegraphics[width=\imgwidth\textwidth]{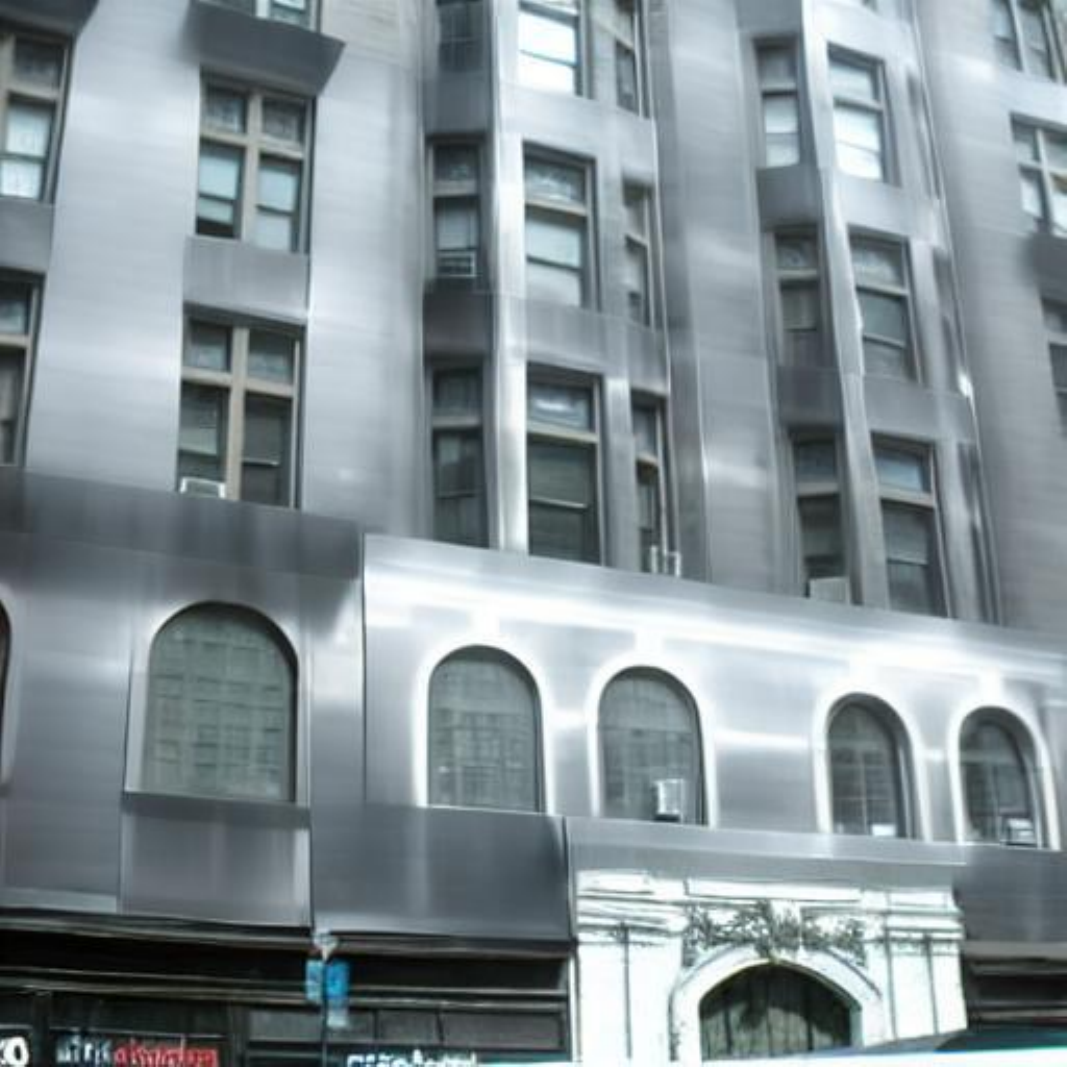}\\
    \multicolumn{4}{c}{``Turn the building into steel"}\\[\textspace]

\end{tabular}
\caption{Diverse examples of realistic edits generated by our method, demonstrating precise structural preservation and semantic alignment across various scenes and styles. We show the input images, as well as both the text and visual prompts used to generate these edits.}
\label{appdx_soloresults1}
\end{figure*}

\begin{figure*}[h]
\centering
\newcommand{\imgwidth}{0.23}
\setlength{\tabcolsep}{2pt} 
\renewcommand{\arraystretch}{1} 
\newcommand{\textspace}{0.7em}
\newcommand{\myhspace}{0.5em} 
\newcommand{\stycompression}{0.05}
\begin{tabular}{cccc} 
    Input & Edit \hspace{\myhspace} & \hspace{\myhspace} Input & Edit\\ [0.3em]

    \begin{overpic}[width=\imgwidth\textwidth]{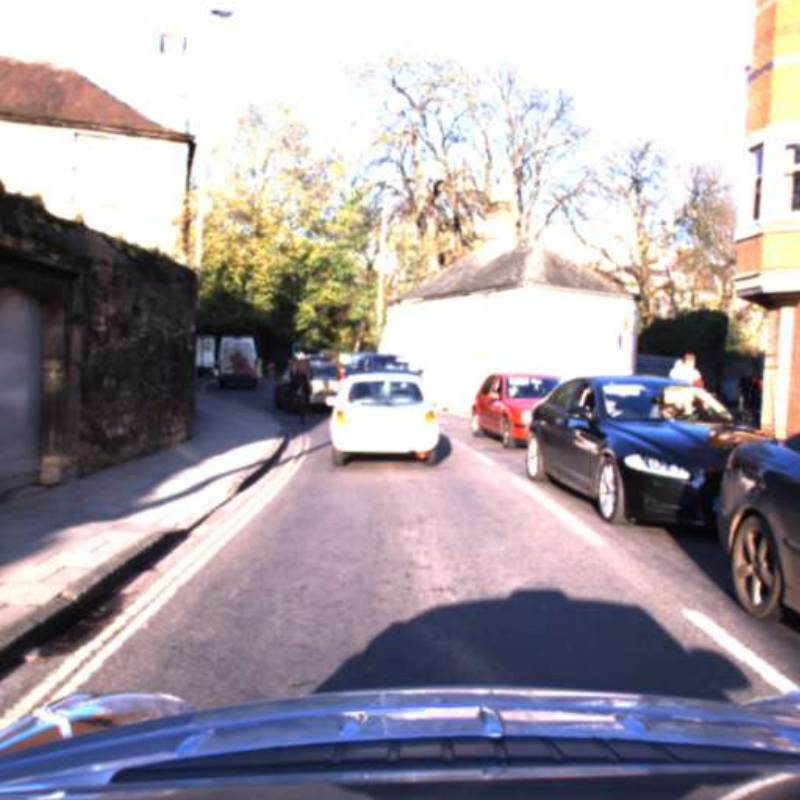}
    \put(1, 74){
     \begin{tikzpicture}
      \begin{scope}
       \clip[rounded corners=5pt] (0,0) rectangle (\stycompression\textwidth, \stycompression\textwidth);
       \node[anchor=north east, inner sep=0pt] at (\stycompression\textwidth,\stycompression\textwidth) {\includegraphics[width=\stycompression\textwidth]{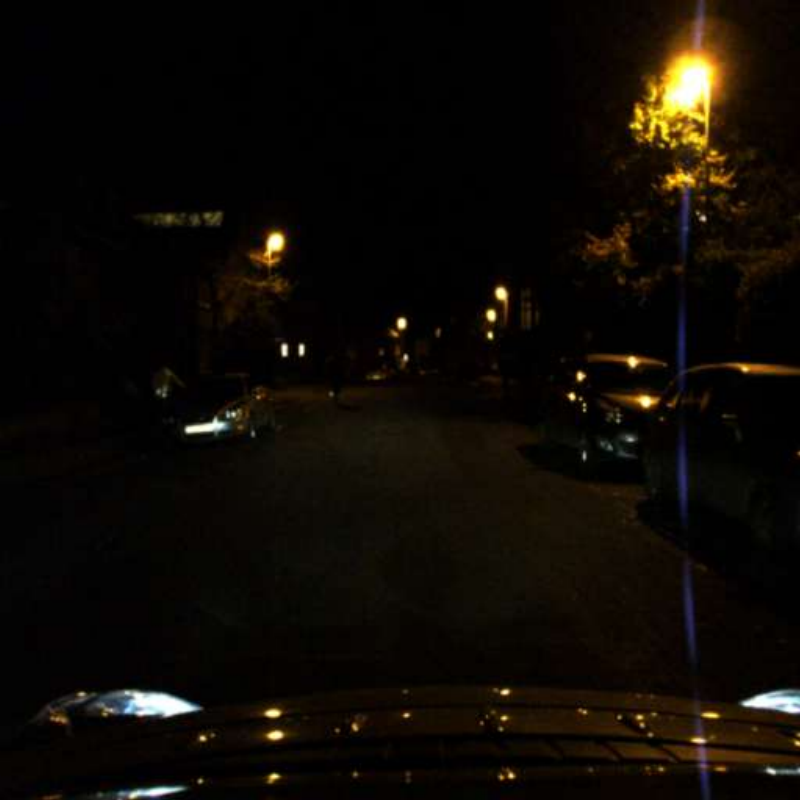}};
    \end{scope}
    \draw[rounded corners=5pt, RubineRed, very thick, dashed] (0,0) rectangle (\stycompression\textwidth,\stycompression\textwidth);
    \end{tikzpicture}}
    \end{overpic}&
    \includegraphics[width=\imgwidth\textwidth]{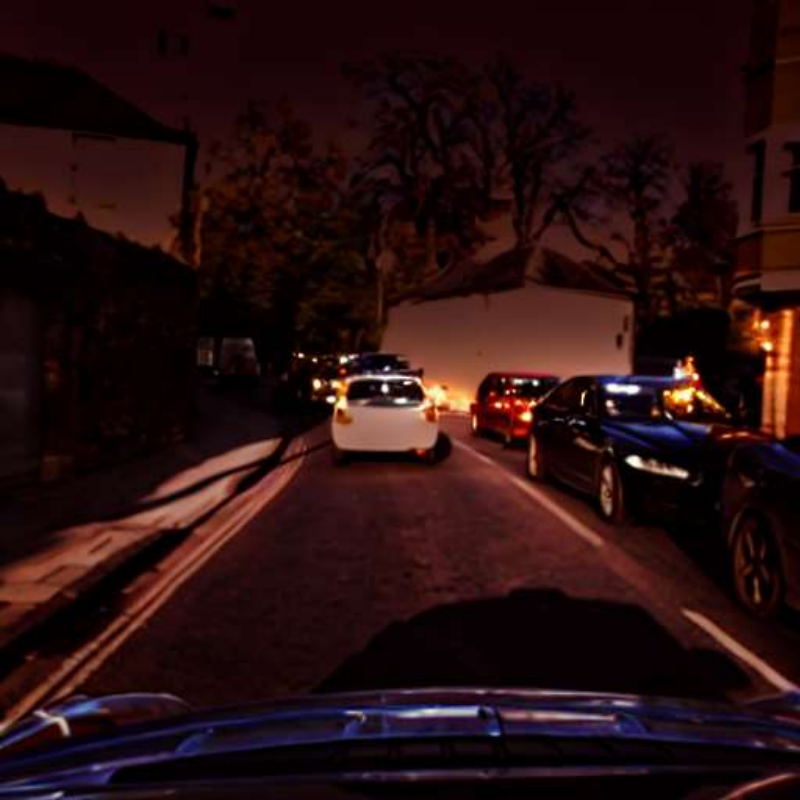} \hspace{\myhspace} &\hspace{\myhspace}
    \begin{overpic}[width=\imgwidth\textwidth]{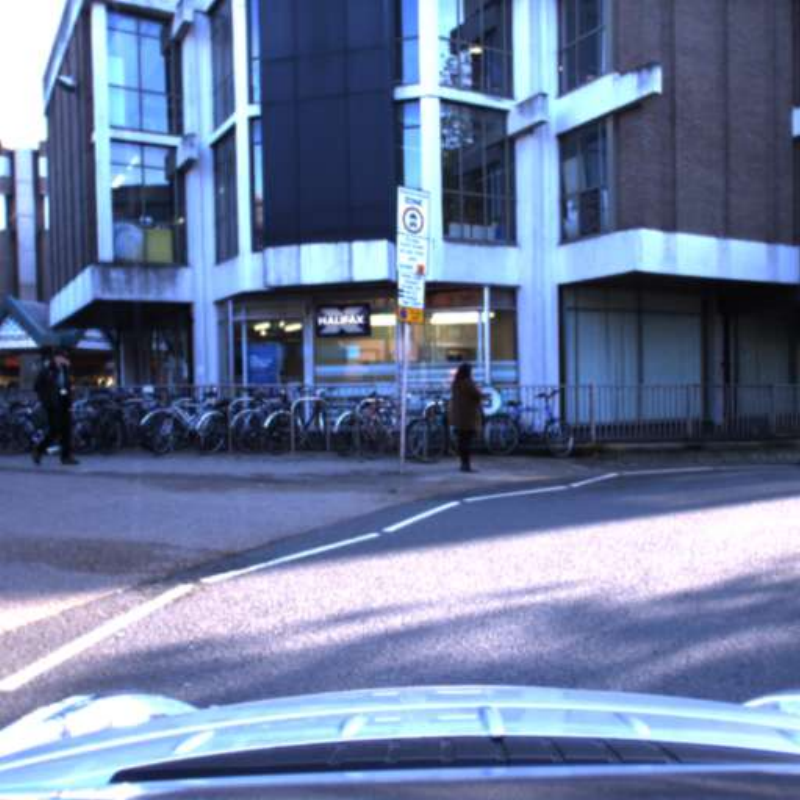}
    \put(1, 74){
     \begin{tikzpicture}
      \begin{scope}
       \clip[rounded corners=5pt] (0,0) rectangle (\stycompression\textwidth, \stycompression\textwidth);
       \node[anchor=north east, inner sep=0pt] at (\stycompression\textwidth,\stycompression\textwidth) {\includegraphics[width=\stycompression\textwidth]{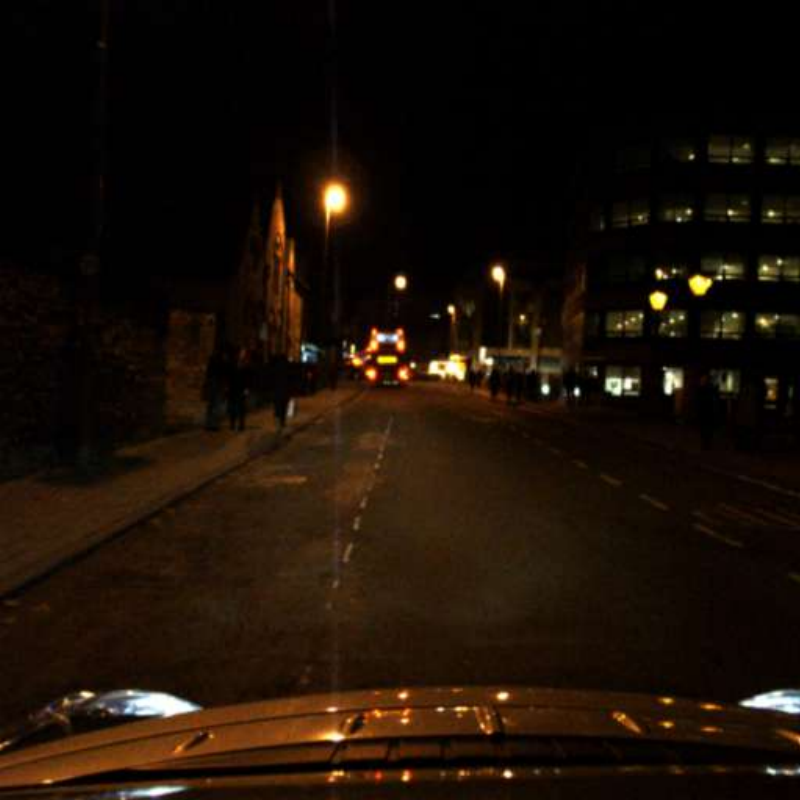}};
    \end{scope}
    \draw[rounded corners=5pt, RubineRed, very thick, dashed] (0,0) rectangle (\stycompression\textwidth,\stycompression\textwidth);
    \end{tikzpicture}}
    \end{overpic}&
    \includegraphics[width=\imgwidth\textwidth]{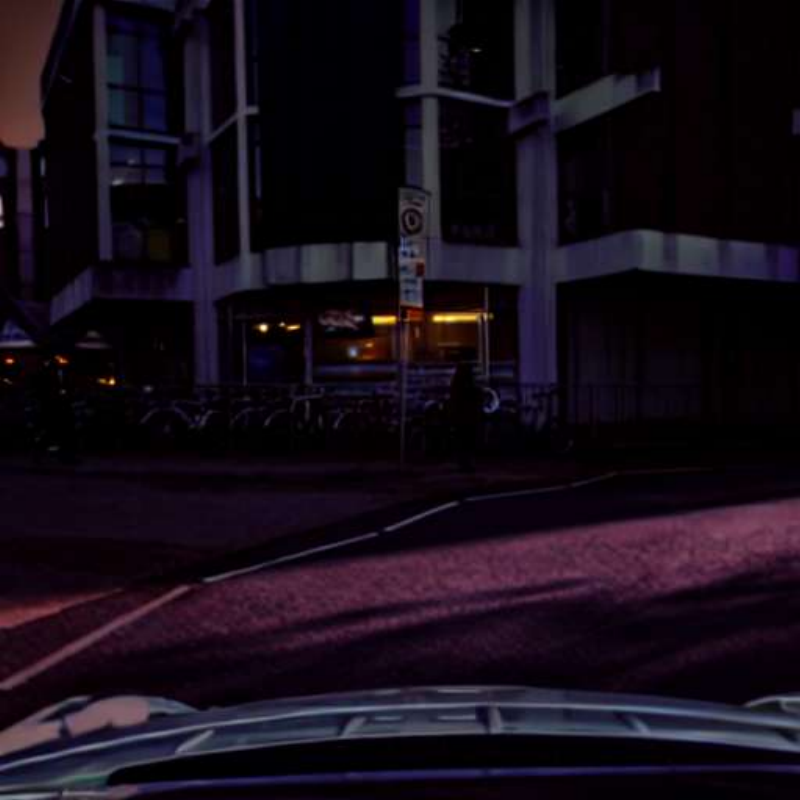}\\
    \multicolumn{4}{c}{``Change the time to nighttime"}\\[\textspace]

    \begin{overpic}[width=\imgwidth\textwidth]{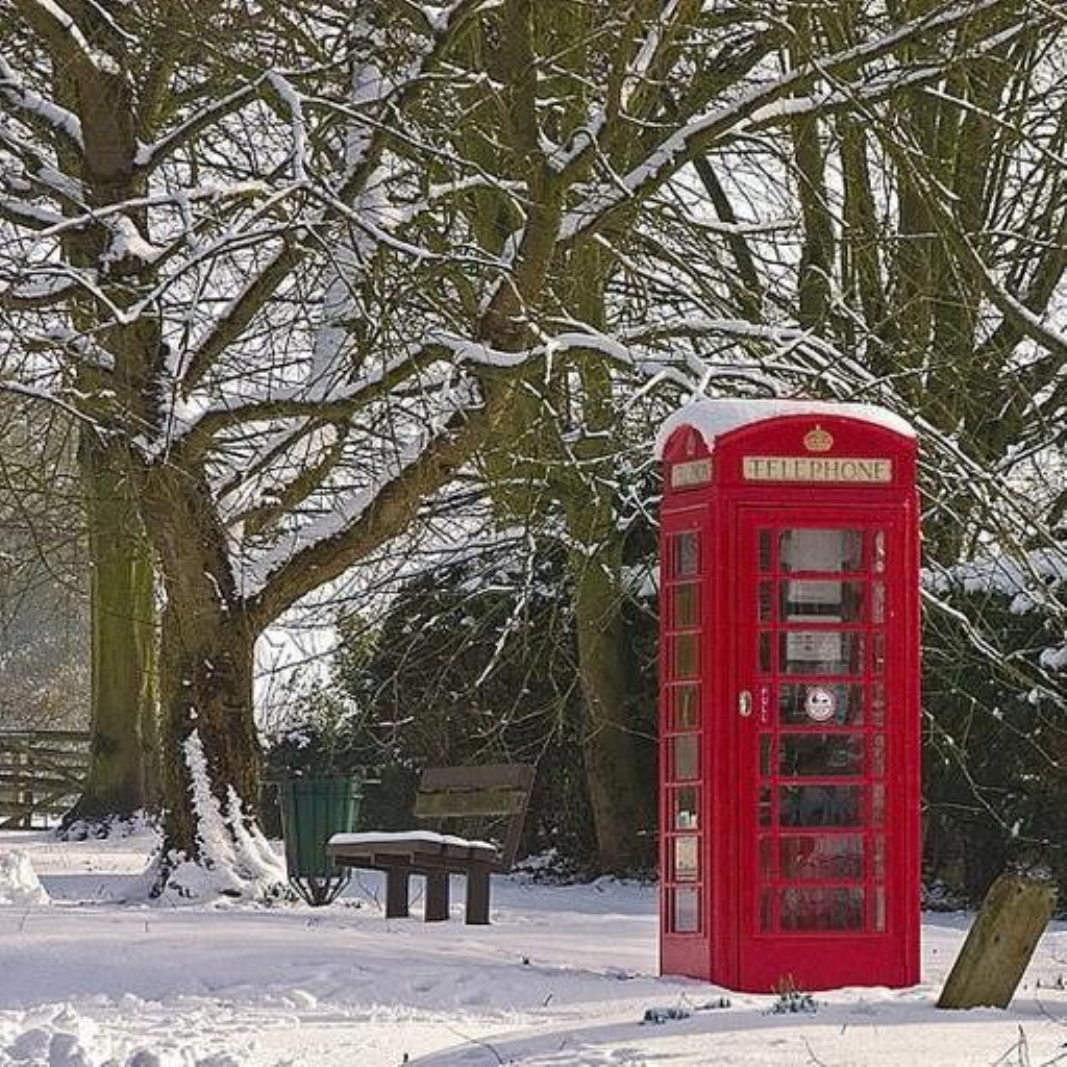}
    \put(1, 74){
     \begin{tikzpicture}
      \begin{scope}
       \clip[rounded corners=5pt] (0,0) rectangle (\stycompression\textwidth, \stycompression\textwidth);
       \node[anchor=north east, inner sep=0pt] at (\stycompression\textwidth,\stycompression\textwidth) {\includegraphics[width=\stycompression\textwidth]{images/qualitative/all_pdf_512/tgt/stone/stone5.pdf}};
    \end{scope}
    \draw[rounded corners=5pt, RubineRed, very thick, dashed] (0,0) rectangle (\stycompression\textwidth,\stycompression\textwidth);
    \end{tikzpicture}}
    \end{overpic}&
    \includegraphics[width=\imgwidth\textwidth]{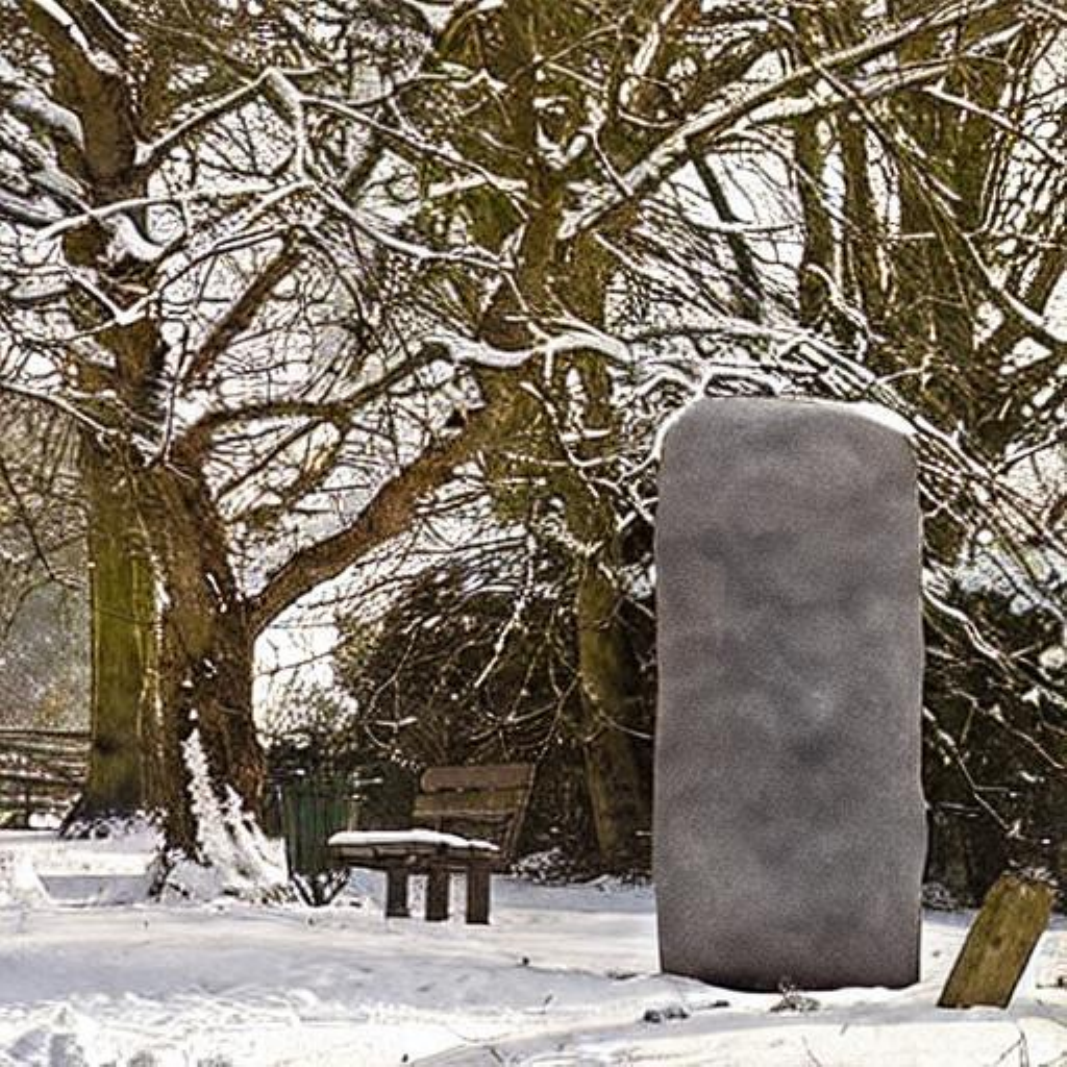} \hspace{\myhspace} &\hspace{\myhspace}
    \begin{overpic}[width=\imgwidth\textwidth]{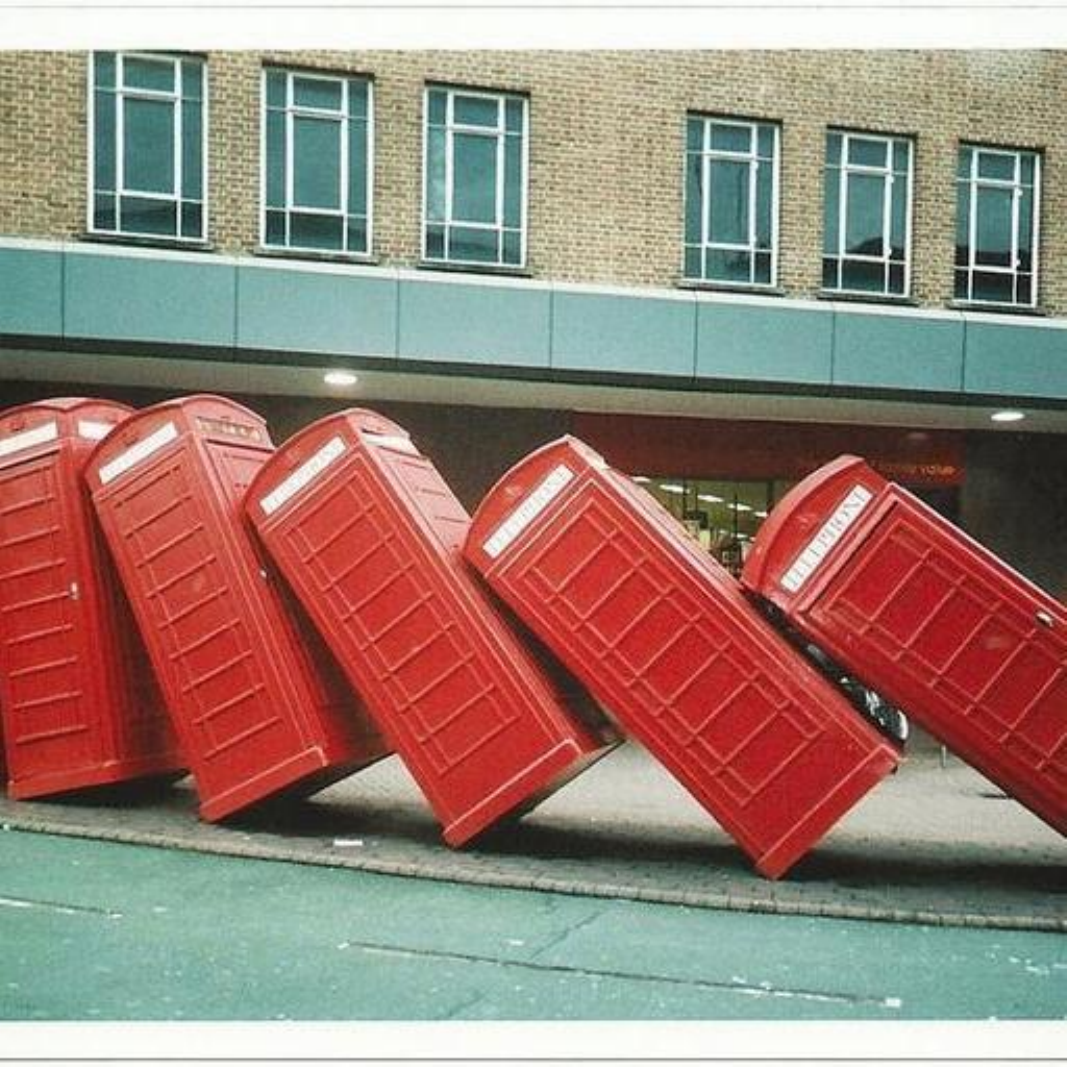}
    \put(1, 74){
     \begin{tikzpicture}
      \begin{scope}
       \clip[rounded corners=5pt] (0,0) rectangle (\stycompression\textwidth, \stycompression\textwidth);
       \node[anchor=north east, inner sep=0pt] at (\stycompression\textwidth,\stycompression\textwidth) {\includegraphics[width=\stycompression\textwidth]{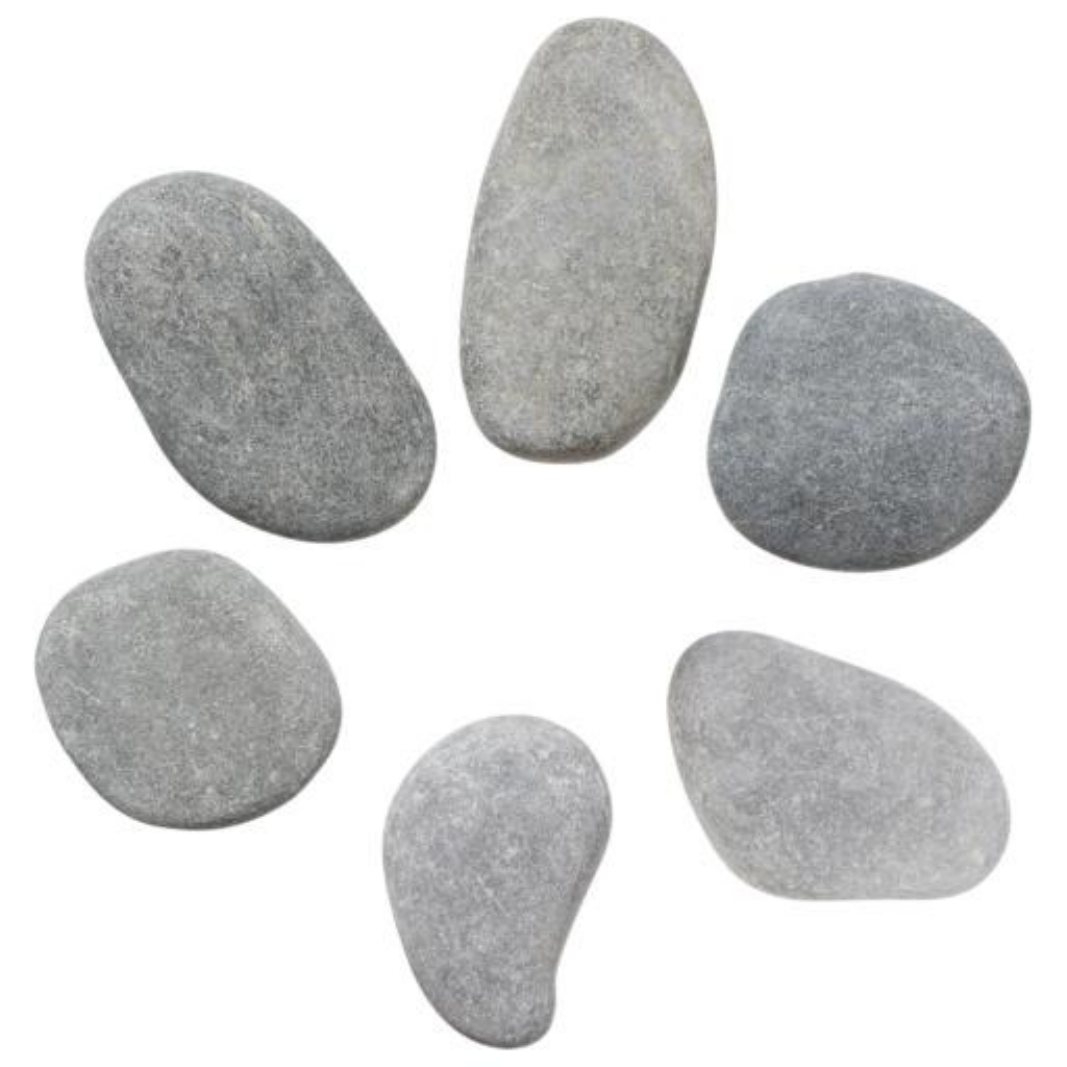}};
    \end{scope}
    \draw[rounded corners=5pt, RubineRed, very thick, dashed] (0,0) rectangle (\stycompression\textwidth,\stycompression\textwidth);
    \end{tikzpicture}}
    \end{overpic}&
    \includegraphics[width=\imgwidth\textwidth]{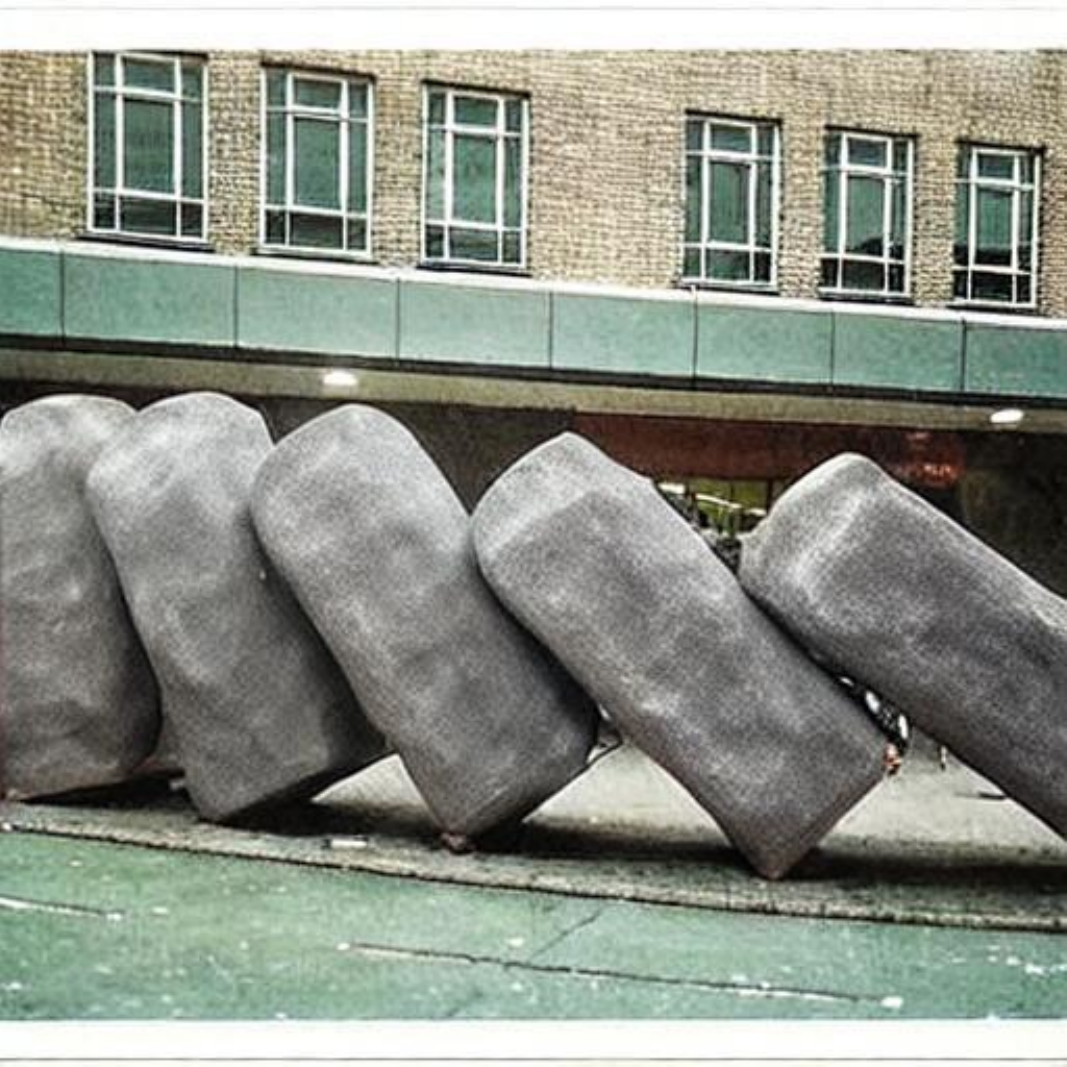}\\
    \multicolumn{4}{c}{``Transform the booth into stone"}\\[\textspace]

    \begin{overpic}[width=\imgwidth\textwidth]{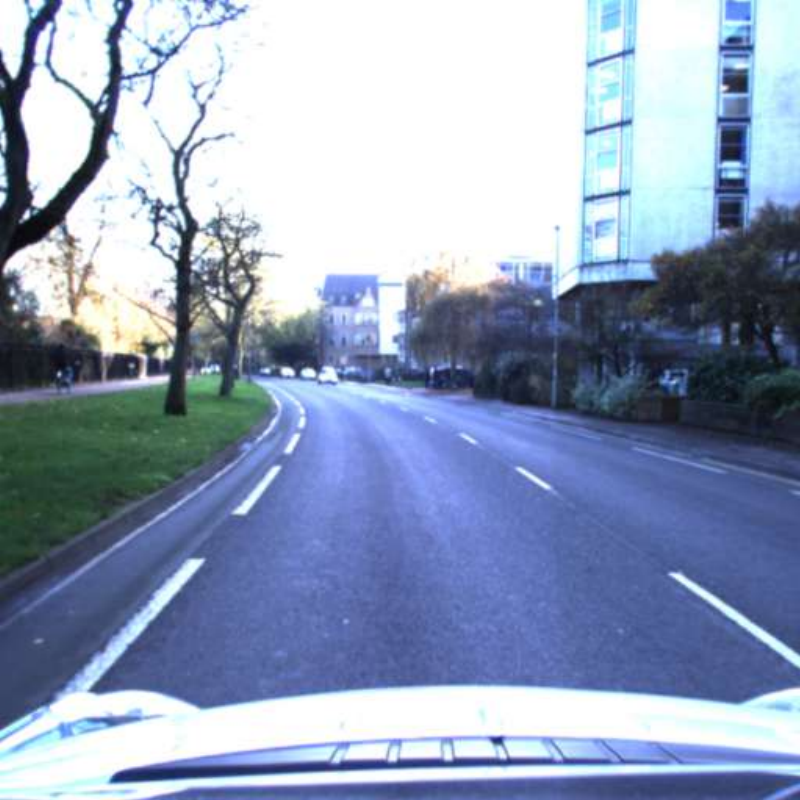}
    \put(1, 74){
     \begin{tikzpicture}
      \begin{scope}
       \clip[rounded corners=5pt] (0,0) rectangle (\stycompression\textwidth, \stycompression\textwidth);
       \node[anchor=north east, inner sep=0pt] at (\stycompression\textwidth,\stycompression\textwidth) {\includegraphics[width=\stycompression\textwidth]{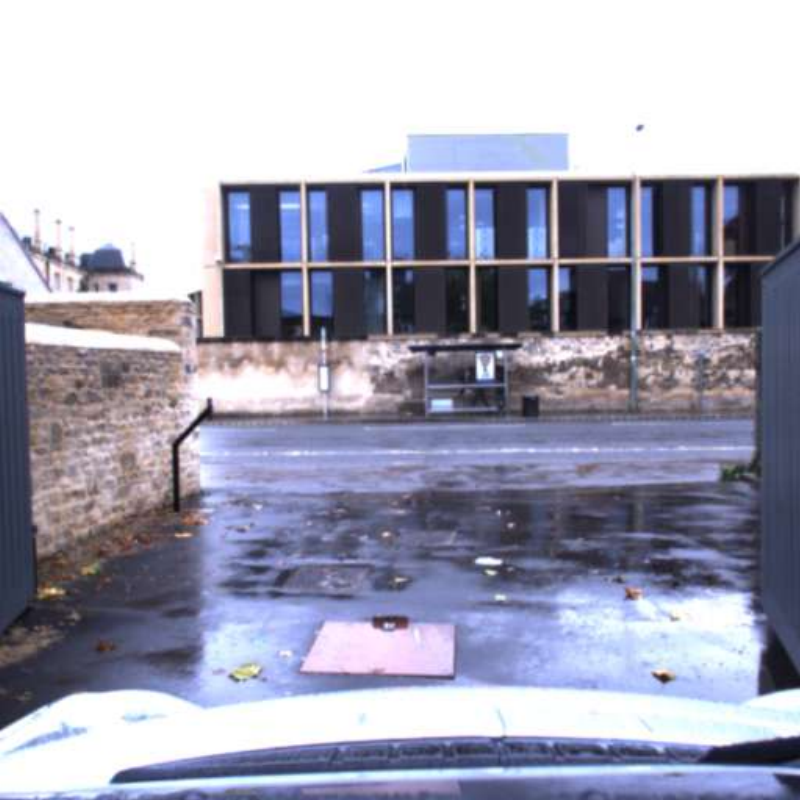}};
    \end{scope}
    \draw[rounded corners=5pt, RubineRed, very thick, dashed] (0,0) rectangle (\stycompression\textwidth,\stycompression\textwidth);
    \end{tikzpicture}}
    \end{overpic}&
    \includegraphics[width=\imgwidth\textwidth]{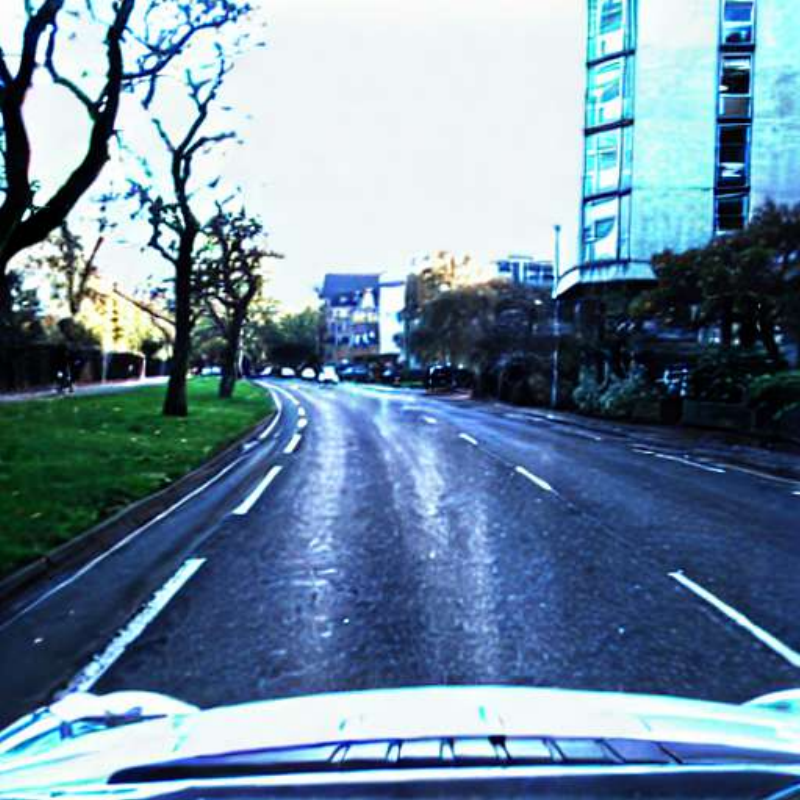} \hspace{\myhspace} &\hspace{\myhspace}
    \begin{overpic}[width=\imgwidth\textwidth]{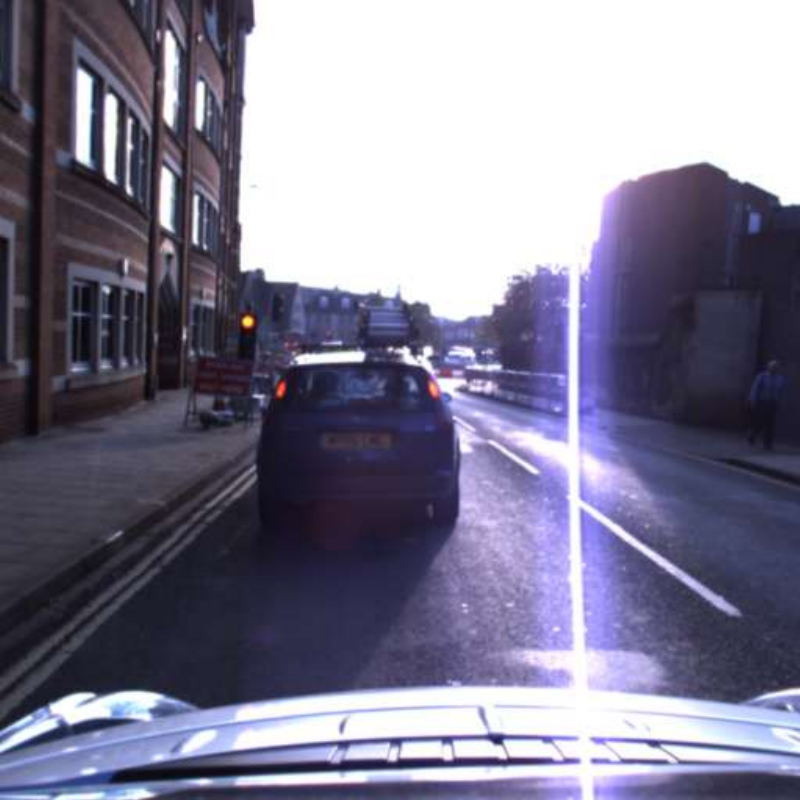}
    \put(1, 74){
     \begin{tikzpicture}
      \begin{scope}
       \clip[rounded corners=5pt] (0,0) rectangle (\stycompression\textwidth, \stycompression\textwidth);
       \node[anchor=north east, inner sep=0pt] at (\stycompression\textwidth,\stycompression\textwidth) {\includegraphics[width=\stycompression\textwidth]{images/qualitative/all_pdf_512/tgt/snow/1422953597799692.pdf}};
    \end{scope}
    \draw[rounded corners=5pt, RubineRed, very thick, dashed] (0,0) rectangle (\stycompression\textwidth,\stycompression\textwidth);
    \end{tikzpicture}}
    \end{overpic}&
    \includegraphics[width=\imgwidth\textwidth]{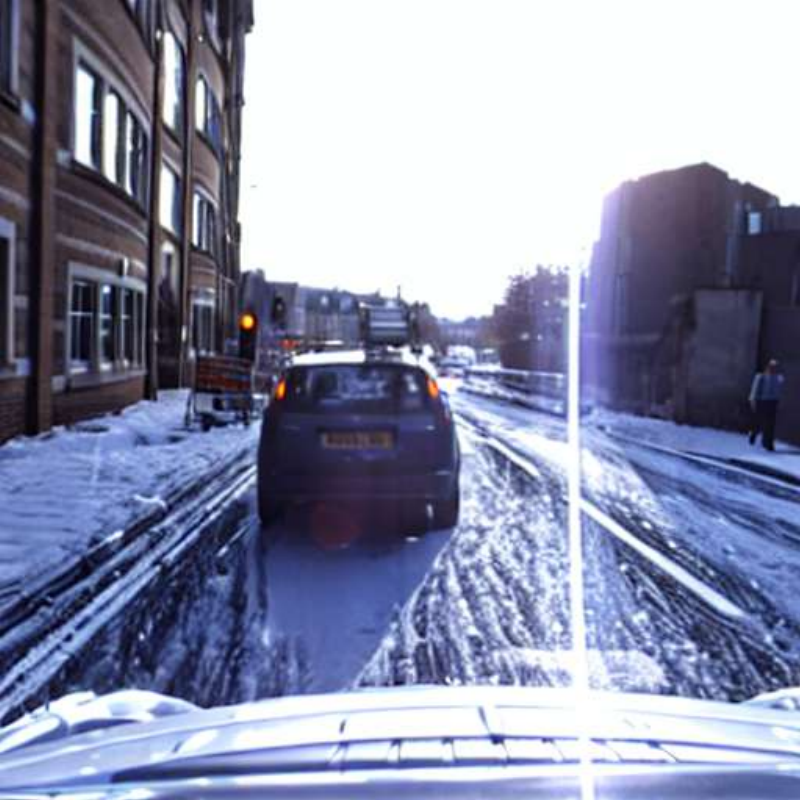} \\
    \multicolumn{2}{c}{``Add rain on the road" \hspace{\myhspace}} &
    \multicolumn{2}{c}{\hspace{\myhspace} ``Add snow on the road"} \\[\textspace]
\end{tabular}
\caption{Diverse examples of realistic edits generated by our method, demonstrating precise structural preservation and semantic alignment across various scenes and styles. We show the input images, as well as both the text and visual prompts used to generate these edits}
\label{appdx_soloresults2}
\end{figure*}

This section of the appendix provides supplementary qualitative results, including a comparative evaluation against baselines (Figures \ref{appdx_baseline1} and \ref{appdx_baseline2}), demonstrating SPIE's superior performance in structural preservation, semantic alignment, and realism. 
We also provided a extended visualization of the impact of different visual prompts on generated samples (Figure \ref{appdx_real_conditioning_qualitative}), showcasing how SPIE captures semantic nuances beyond text prompts. 
Additionally, we present examples of simulated scenes edited with enhanced realism (Figure \ref{apdx_rtx_row}), and highlight the flexibility of our approach in replicating visual prompts across diverse scenes (Figure \ref{appdx_gold_multi}). 
Finally, we display an extensive array of realistic edits generated by our method (Figures \ref{appdx_soloresults1} and \ref{appdx_soloresults2}), illustrating precise structural preservation and semantic alignment across various scenes and styles.


\clearpage

\end{document}